\documentclass[epsfig,a4paper,12pt,titlepage]{book}
\usepackage{epsfig}
\usepackage{fancyhdr}
\usepackage{subcaption}
\usepackage{plain}
\usepackage{fancyhdr}
\usepackage{amsmath}
\usepackage{algorithm}
\usepackage[noend]{algpseudocode}
\usepackage{amsmath}
\usepackage{slashbox}
\usepackage{lineno,hyperref}
\usepackage{amsmath}
\usepackage{multirow}
\usepackage{subcaption}
\usepackage{relsize}
\usepackage[intoc]{nomencl}

\include{pythonlisting}
\usepackage{algorithm}
\usepackage[noend]{algpseudocode}

\usepackage{graphicx}

\usepackage[toc,page]{appendix}

\DeclareMathOperator*{\argmin}{arg\,min}
\DeclareMathOperator*{\argmax}{arg\,max}
\makeatletter
\renewcommand\part{%
  \if@openright
    \cleardoublepage
  \else
    \clearpage
  \fi
  \thispagestyle{empty}%
  \if@twocolumn
    \onecolumn
    \@tempswatrue
  \else
    \@tempswafalse
  \fi
  \null\vfil
  \secdef\@part\@spart}
\makeatother

\newcommand{\clearemptydoublepage}{\newpage{\pagestyle{empty}\cleardoublepage}}

\makeindex
\makenomenclature

\begin{document}
\pagestyle{plain}

\newpage
\clearemptydoublepage
\thispagestyle{empty}
\begin{center}

\begin{figure}[h!]
  \centerline{\psfig{file=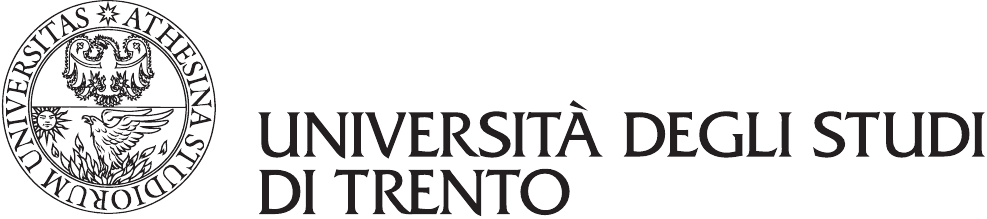,width=0.4\textwidth}}
\end{figure}

\hrulefill

DEPARTMENT OF INFORMATION ENGINEERING AND COMPUTER SCIENCE\\
\textbf{ICT International Doctoral School}\\

\vspace{1 cm} 
\Huge\textsc{Deep Learning \\ for \\ Distant Speech Recognition\\} \Large

\vspace{0.3 cm}

\begin{center}
\begin{tabular}{l}
\LARGE{Mirco Ravanelli}\\
\end{tabular}
\end{center}
\vspace{1 cm} 
\begin{flushleft}
\begin{tabular}{ll}
\multicolumn{2}{l}{\large Advisor:}\\
 & \large Maurizio Omologo\\
 & \large Fondazione Bruno Kessler\\
\end{tabular}
\end{flushleft}

\begin{flushleft}
\begin{tabular}{ll}
\end{tabular}
\end{flushleft}

\hrulefill

\normalsize
December $2017$
\end{center}

\newpage
\clearemptydoublepage
\thispagestyle{empty}
\large

\nomenclature{CASA}{Computational Auditory Scene Analysis}
\nomenclature{ASA}{Auditory Scene Analysis}
\nomenclature{ANN}{Artificial Neural Network}
\nomenclature{AI}{Artificial Intelligence}
\nomenclature{GMM}{Gaussian Mixture Model}
\nomenclature{HMM}{Hidden Markov Model}
\nomenclature{DNN}{Deep Neural Network}
\nomenclature{FF-DNN}{Feed-Forward Deep Neural Network}
\nomenclature{ResNET}{Residual Neural Network}
\nomenclature{CTC}{Connectionist Temporal Classification}
\nomenclature{RBM}{Restricted Boltzmann Machines}
\nomenclature{H-DNN}{Hierarchical Deep Neural Network}
\nomenclature{MLP}{Multi-Layer Perceptron}
\nomenclature{RNN}{Recurrent Neural Network}
\nomenclature{LSTM}{Long Short Term Memory}
\nomenclature{GRU}{Gated Recurrent Units}
\nomenclature{M-GRU}{Minimal Gated Recurrent Units}
\nomenclature{Li-GRU}{Light Gated Recurrent Units}
\nomenclature{CW}{Context Window}
\nomenclature{ACW}{Asymmetric Context Window}
\nomenclature{SCW}{Symmetric Context Window}
\nomenclature{ASR}{Automatic Speech Recognition}
\nomenclature{DSR}{Distant Speech Recognition}

\nomenclature{MFCC}{Mel Frequency Cepstral Coefficients}
\nomenclature{FBANK}{Filter Bank}
\nomenclature{fMLLR}{Feature space Maximum Likelihood Linear Regression}
\nomenclature{PER}{Phone Error Rate}
\nomenclature{WER}{Word Error Rate}
\nomenclature{FA}{Frame Accuracy}

\nomenclature{CT}{Close-Talk}
\nomenclature{PT}{Pre-Training}

\nomenclature{MLP}{Maximum Likelihood Estimation}

\nomenclature{NLL}{Negative Log-Likelihood}
\nomenclature{CE}{Cross-Entropy}
\nomenclature{MSE}{Mean Squared Error}

\nomenclature{GD}{Gradient Descend}
\nomenclature{SGD}{Stochastic Gradient Descend}

\nomenclature{BN}{Batch Normalization}

\nomenclature{Tanh}{Hyperbolic Tangent}

\nomenclature{ReLU}{Rectified Linear Units}
\nomenclature{CNN}{Convolutional Neural Network}
\nomenclature{TDNN}{Time Delay Neural Networks}

\nomenclature{CIFAR}{Canadian Institute for Advanced Research}

\nomenclature{GAN}{Generative Adversarial Network}

\nomenclature{ToF}{Time of Flight}

\nomenclature{SNR}{Signal-to-Noise Ratio}

\nomenclature{LPC}{Linear Predictive Coding}
\nomenclature{DICIT}{Distant-talking Interfaces for Control Interactive TV}
\nomenclature{MAP}{Maximum a Posteriori}
\nomenclature{AMI}{Advanced metering infrastructure}
\nomenclature{AMIDA}{Augmented Multi-party Interaction with Distance Access}
\nomenclature{PLP}{Perceptual linear prediction}
\nomenclature{CMVN}{Cepstral Mean and Variance Normalisation}
\nomenclature{VTLN}{Vocal Tract Length Normalization}
\nomenclature{LDA}{Linear Discriminative Analysis}
\nomenclature{HLDA}{Heteroscedastic Linear Discriminative Analysis}
\nomenclature{PCA}{Principal Component Analysis}
\nomenclature{PoV}{Probability of Voicing}
\nomenclature{PDT}{Phonetic Decision Tree}
\nomenclature{LM}{Language Model}
\nomenclature{AM}{Acoustic Model}
\nomenclature{WFST}{Weighted Finite State Transducers}
\nomenclature{MMI}{maximum mutual information}
\nomenclature{MPE}{minimum phone error}
\nomenclature{MBR}{minimum Bayes risk}
\nomenclature{ROVER}{Recognition Output Voting Error Reduction}
\nomenclature{TDOA}{Time Difference of Arrival}
\nomenclature{AEC}{Acoustic Echo Cancellation}
\nomenclature{LMS}{Least Mean Square}
\nomenclature{SBSS}{Semi-Blind Source Separation}
\nomenclature{DCASE}{Detection and Classification of Acoustic Scenes and Events}
\nomenclature{SVM}{Support Vector Machines}
\nomenclature{UBM}{Universal Background Model}
\nomenclature{MLS}{Maximum Length Sequence}
\nomenclature{LSS}{Linear Sine Sweep} 
\nomenclature{ESS}{Exponential Sine Sweep}
\nomenclature{FBK}{Fondazione Bruno Kessler}
\nomenclature{IM}{Image Method}
\nomenclature{LJ}{Lehmann-Johansson algorithm}
\nomenclature{CD}{Context Dependent}
\nomenclature{CI}{Context Independent}
\nomenclature{NN}{Neural Network}
\nomenclature{IWSLT}{International Workshop on Spoken Language Translation}
\nomenclature{ADAM}{Adaptive Moment Estimation}
\nomenclature{NIST}{National Institute of Standards and Technology}
\nomenclature{LIMABEAM}{Likelihood Maximization Beamforming}

\nomenclature{DIRHA}{Distant-speech Interaction for Robust Home Applications}

\nomenclature{WSJ}{Wall Street Journal}

\nomenclature{IR}{Impulse Response}
\nomenclature{DRR}{Direct to Reverberant Ratio}

\nomenclature{GCF}{Global Coherence Field}
\nomenclature{GCC-PHAT}{Generalized Cross Correlation with Phase Transform}

\chapter*{Acknowledgements}
\thispagestyle{empty}
\vspace{0cm}

This achievement would not have been possible without the help and support of many people. Firstly, I would like to express a special appreciation  to my advisor  Maurizio Omologo, that has been an exceptional mentor for me. I would like to thank him for encouraging my research and for allowing me to grow as a research scientist.  Besides my advisor, I would also like to thank all the members of the SHINE research unit for their scientific and technical support in the last years.

Special thanks to Prof. Yoshua Bengio,  for hosting and advising me during my stay at the MILA lab. It was really a pleasure working  with him and with the other top deep learning scientists of his amazing lab. In particular, I would like to mention Philemon Brakel, co-author of several papers I published during the last year,  for his advice, support, and helpful discussions. 

I would like to thank Prof. Laface and Prof. Squartini for reviewing this thesis, Prof. Sebe for being a committee member,  as well as the FBK and UNITN administration staff for their availability and professionalism. 

This PhD would not have been possible without my family. I would like to thank my wife Serena.  Her encouragement,  patience, and love were fundamental to face both beautiful and bad life moments. Last but not the least, I would like to express my sincere gratitude to my parents. They did multiple sacrifices to provide me a top-level education and their constant support over the years was priceless. 

\clearpage

\newpage\null\thispagestyle{empty}\newpage

{\bf \huge Abstract}

\vspace{1cm}

\noindent
\emph{Deep learning is an emerging technology that is considered one of the most promising directions for reaching higher levels of artificial intelligence.  
}
\emph{Among  the other achievements, building computers that understand speech represents a crucial leap towards intelligent machines. 
} 

\emph{Despite the great efforts of the past decades, however, a natural and robust human-machine speech interaction still appears to be out of reach, especially when users interact with a distant microphone in noisy and reverberant environments.  The latter disturbances severely hamper the intelligibility of a speech signal, making  Distant Speech Recognition (DSR) one of the major open challenges in the field.}

\emph{This thesis addresses the latter scenario and proposes some novel techniques, architectures, and algorithms to improve the robustness of distant-talking acoustic models. We first elaborate on methodologies for realistic data contamination, with a particular emphasis on DNN training with simulated data.} \emph{We then investigate on approaches for better exploiting speech contexts, proposing some original methodologies for both feed-forward and recurrent neural networks.
} \emph{Lastly, inspired by the idea that cooperation across different DNNs could be the key for counteracting the harmful effects of noise and reverberation, we propose a novel deep learning paradigm called ``network of deep neural networks". 
} 

\emph{The analysis of the original concepts were based on extensive experimental validations conducted on both real and simulated data, considering different corpora, microphone configurations, environments, noisy conditions, and ASR tasks.}


\vspace{0.5cm}
\noindent
{\bf Keywords}

\noindent
Deep Learning, Distant Speech Recognition, Deep Neural Networks.

\clearemptydoublepage

\pagenumbering{roman}
\tableofcontents
\clearemptydoublepage
\printnomenclature
\clearemptydoublepage
\pagestyle{fancy}

\chapter*{Publication List}
This thesis is based on the following publications:

\begin{enumerate}

    \item \textbf{M. Ravanelli}, P. Brakel, M. Omologo, Y. Bengio,  ``Light Gated Recurrent Units for Speech Recognition", in IEEE Transactions on Emerging Topics in Computational Intelligence (to appear).

    
	
	\item \textbf{M. Ravanelli}, P. Brakel, M. Omologo, Y. Bengio, ``A network of deep neural networks for distant speech recognition", in Proceedings of ICASSP 2017 (best IBM student paper award).

    \item \textbf{M. Ravanelli}, P. Brakel, M. Omologo, Y. Bengio, ``Improving Gated Recurrent Units by Revising Gated Recurrent Units", in Proceedings of Interspeech 2017.
    
    \item \textbf{M. Ravanelli}, P. Brakel, M. Omologo, Y. Bengio, ``Batch-normalized joint training for DNN-based distant speech recognition", in Proceedings of STL 2016.
    
    \item \textbf{M. Ravanelli}, P. Svaizer, M. Omologo, ``Realistic Multi-Microphone Data Simulation for Distant Speech Recognition",  in Proceedings of Interspeech 2016.

    \item  M. Matassoni, \textbf{M.Ravanelli}, S. Jalalvand, A. Brutti, ``The FBK system for the CHiME-4 challenge", in Proceedings of the CHiME 4 challenge.
    
    \item \textbf{M. Ravanelli}, M. Omologo, ``Contaminated speech training methods for robust DNN-HMM distant speech recognition", in Proceedings of  INTERSPEECH 2015.

    \item \textbf{M. Ravanelli}, L. Cristoforetti, R. Gretter, M. Pellin, A. Sosi, M. Omologo, ``The DIRHA-English corpus and related tasks for distant-speech recognition in domestic environments", in Proceedings of ASRU 2015.

    \item E. Zwyssig, \textbf{M. Ravanelli}, P. Svaizer, M. Omologo, ``A multi-channel corpus for distant-speech interaction in presence of known Interferences", in Proceedings of  ICASSP 2015.

   \item \textbf{M. Ravanelli}, B. Elizalde, J. Bernd, G. Friedland,  ``Insights into Audio-Based Multimedia Event Classification with Neural Networks", in Proceedings of ACM-MMCOMMONS.
   
   \item \textbf{M. Ravanelli}, M. Omologo, ``On the selection of the impulse responses for distant-speech recognition based on contaminated speech training", in Proceedings of  INTERSPEECH 2014.

   \item L. Cristoforetti, \textbf{M. Ravanelli}, M. Omologo, A. Sosi, A. Abad, M. Hagmueller, P. Maragos, ``The DIRHA simulated corpus",  in Proceedings of LREC 2014.

   \item M. Matassoni, R. Astudillo, A. Katsamanis, \textbf{M. Ravanelli}, ``The DIRHA-GRID corpus: baseline and tools for multi-room distant speech recognition using distributed microphones", in Proceedings of  INTERSPEECH 2014.

  \item  A. Brutti, \textbf{M. Ravanelli}, M. Omologo, ``SASLODOM: Speech Activity detection and Speaker LOcalization in DOMestic environments", in Proceedings of Evalita 2014.

 \item A. Brutti, \textbf{M. Ravanelli}, P. Svaizer, M. Omologo, ``A speech event detection and localization task for multiroom environments",  in Proceedings of HSCMA 2014.

 \item \textbf{M. Ravanelli}, V.H.  Do, A. Janin, ``TANDEM-Bottleneck Feature Combination using Hierarchical Deep Neural Networks",  in Proceedings of ISCSLP 2014.

\item \textbf{M. Ravanelli}, B. Elizalde,  K. Ni, G. Friedland, ``Audio Concept Classification with Hierarchical Deep Neural Networks",  in Proceeding of  EUSIPCO 2014. 
 
\item B. Elizalde, \textbf{M. Ravanelli}, K. Ni, D. Borth, G. Friedland , ``Audio-Concept Features and Hidden Markov Models for Multimedia Event Detection",  in Proceedings of SLAM 2014.

\end{enumerate}

\chapter{Introduction}
\pagenumbering{arabic}
\label{cha:intro}

The human voice is the most natural way to communicate. Building computers that understand speech thus represents a crucial step towards easy-to-use human-machine interfaces \cite{roberto}. According to analysts, the global speech recognition market is estimated to be in order of \$6 billions, with a tremendous growth expected in the next few years. As a matter of fact, voice-based interfaces are rapidly gaining a central role in our everyday lives. 
Current application areas include web-search, intelligent personal assistants, home automation, automotive, healthcare, financial transactions, consumer electronics, just to name a few.
The growing interest in speech recognition is also witnessed by the remarkable investments of the most important tech companies, that have recently committed huge resources in the field. As a result, systems like Siri, Cortana, and Google Voice, gained considerable popularity and are currently used by millions of people worldwide.


The widespread diffusion of such commercial software, however, has given rise to the mistaken belief that speech recognition is a mostly solved problem. Actually, thanks to deep learning, modern speech recognizers achieve an unprecedented performance level. Nevertheless, despite the optimism and excitement surrounding ASR technologies, the road towards a natural and flexible human-machine interaction is still long and full of scientific challenges. At the time of writing, some prominent open issues regard, for instance, the ability of dealing with very large vocabularies, better managing multiple languages with low resources, and properly modeling spontaneous speech. The  development of real-time, small footprint and privacy-preserving systems also represents a crucial need.
Another important problem concerns robustness against variability factors, including speaker, accent and channel variabilities. 

This thesis considers the latter issue, addressing in particular the robustness against noise and reverberation, which we believe is one of the major open challenges in ASR. These disturbances typically arise when the speaker interacts with a distant microphone, making DSR a topic of fundamental interest for the research community. The research reported in this thesis is focused on deep learning, that is the natural candidate for improving DSR robustness, thanks to a more detailed modeling of the underlying properties of speech and acoustic environments. 

The next section introduces deep learning, while the problem of distant speech recognition is summarized in Sec. \ref{sec:intro_dsr}. Sec. \ref{sec:motivation} then discusses the motivations and the scope of this research effort. Our contribution is summarized in Sec. \ref{sec:contributions}, while Sec. \ref{sec:outline} finally provides an outline of the thesis.

\section{Deep Learning} \label{sec:intro_dl}
Building intelligent machines has fascinated humanity for centuries \cite{superintellicence}. The field of Artificial Intelligence (AI), however, started to develope relatively recently, when programmable digital computers were conceived. The history of AI has continuously alternated decades of great enthusiasm, investments and expectations, often followed by  ``AI winters", characterized by reduced fundings and excitement towards this field \cite{ai_history}. The rise of deep learning \cite{Goodfellow-et-al-2016-Book}  has recently contributed to renew the interest in AI and has allowed current technology to achieve higher levels of artificial intelligence.
This paradigm has rapidly become a driving factor in academic and industrial research, and it is now being deployed in a wide range of domains, including computer vision (for object recognition, restoring colors in black and white photos, cancer detection, handwriting recognition, video classification), machine translation, as well as in natural language processing (for dialogue systems, question answering, image captioning, automatic writing) and speech recognition \cite{lideng}. Other interesting applications  are recommendation systems, fraud and risk detection, resource planning, predictions on financial markets, automatic game playing, robotics, self-driving cars, just to name a few.  

Deep learning is actually a very general machine learning paradigm that follows a compositionality principle to represent the world around us efficiently. Current deep learning implementation exploits deep neural networks, that are properly trained to progressively discover complex representations starting from simpler ones. This principle can be applied in several practical problems, including the problem of recognizing distant speech, that will be briefly introduced in the following section.

\section{Recognizing Distant Speech} \label{sec:intro_dsr}

Most of the current speech recognizers are based on a close-talking interaction with a microphone-equipped device, such a smartphone, a tablet, a laptop or even a smart watch.
Although this approach usually leads to better performance, it is easy to predict that, in the future, users will prefer to relax the constraint of handling or wearing any device to access speech recognition services. Distant speech recognition could indeed represent the preferred modality for future human-machine communications, especially in some specific contexts, where a distant interaction is more natural, convenient and attractive \cite{dsrbook}. 
For instance, applications such as meeting transcriptions and smart TVs have been studied over the past decade in the context of the AMI/AMIDA \cite{ami} and the DICIT \cite{dicit_1} projects, respectively.
More recently,  speech-based domestic control gained a lot of attention \cite{vacher,isidoros}.
To this end, the EU DIRHA project developed voice-enabled automated home environments based on distant-speech interaction in different languages \cite{lrec,dirha_asru}. The recent success of commercial products like Amazon Echo and Google Home further confirms the great interest towards DSR in a domestic environment. 
Another possible application is distant speech interaction in operating theaters \cite{AISV}, in which  the surgeon can dictate some notes about the operation or can access the medical records of the patient. 
Robotics, finally, represents another emerging application, where users can freely dialog with distant mobile platforms.

Under such hands-free scenarios, multiple issues arise that radically change the speech recognition problem. The distance between speaker and microphone tremendously degrades the intelligibility of the speech, hampering the performance of a speech recognizer. The recorded signal, in fact, not only accounts for the contribution of the desired speech, but also includes background noise, competing speakers, and non-speech acoustic events. Moreover, every acoustic enclosure introduces reverberation, that is originated by multiple delayed reflections on the room surfaces.


The research of the last decades led to an impressive improvement of these technologies \cite{dsrbook}. However, a natural, robust and flexible speech interaction is far from being reached, especially in adverse environments \cite{adverse}. For instance, even the most advanced state-of-the-art systems work relatively well in rather quiet environments, but often fail when facing more challenging acoustic conditions. This further indicates that several research efforts are still required to continue the maturation of this technology.


\section{Motivation and Scope}
\label{sec:motivation}
The main motivation behind this work is the belief that building machines able to naturally interact with humans is of paramount importance for reaching higher levels of artificial intelligence.

Recognizing speech is a very basic task for human beings. This apparent simplicity has contributed to create huge expectations around ASR, highlighting a significant gap between the possibilities offered by current technology and user requirements. A fundamental motivation is to contribute to bridge this gap, allowing future users to use speech technologies without current limitations and constraints.

The scope of this thesis is thus to explore proper techniques for improving the robustness of a DSR system to noise and reverberation with deep learning. More precisely, our goal is to properly revise standard deep learning principles, architectures and algorithms to better address DSR under such adverse conditions.

\section{Contributions}
\label{sec:contributions}
DSR is a very active research field, that offers a huge  literature on techniques, methodologies and algorithms.
In the following, the main findings and contributions of the research activity undertaken in this PhD are summarized:
\begin{itemize}
\item \textbf{Methods for data contamination}.
Data are playing a fundamental role in deep learning. Developing proper solutions for data contamination is thus of great interest for the research community. In the speech recognition field, however, it is still not clear what are the best practices for deriving realistic contaminated data. 
In this thesis, our contribution is the definition of a  methodology for deriving high-quality simulated data for ASR purposes. Several efforts have thus been devoted to the characterization of the reverberation effects of typical acoustic enclosures considering both measured and synthetic impulse responses. An extensive comparison between real and simulated data has shown the effectiveness of the proposed approach. The satisfactory level of realism obtained with our methodology, allowed us to generate some realistic simulated datasets and released them at international level.
Proper techniques for exploiting these high-quality simulations for DNN training have then been proposed. Examples are the study of close-talking labels and close-talking pre-training techniques for distant speech recognition with contaminated data.

\item \textbf{Better exploiting time contexts}. The importance of speech contexts is clear to the research community since several decades. However, past HMM-GMM systems were basically inadequate to manage large time contexts and one of the main reason behind the success of deep learning in speech recognition is the natural ability of DNN to properly manage long-term speech information. This feature is particularly helpful for distant speech recognition, since embedding more information into the system can counteract the remarkable uncertainty originated by noise and reverberation. A contribution of this thesis is the study of proper methodologies and architectures for better exploiting time contexts for DSR. Our studies involved both feed-forward and recurrent neural networks. For the former architectures, we proposed the use of an asymmetric context window to counteract acoustic reverberation. We also studied hierarchical neural networks for processing longer time contexts. Finally, we revise standard Gated Recurrent Units (GRUs) to improve the context management for Recurrent Neural Networks. As a result, the proposed model called Light-GRU (Li-GRU) turned out to be helpful to both improve the ASR performance and to reduce the training time.

\item \textbf{Cooperative networks of deep neural networks}:
In this thesis we focused on DNN cooperation, that we believe is a factor of paramount importance for solving complex problems. More precisely, we believe that a proper cooperation and interaction across the different modules of a DSR system could be an  interesting strategy to fight uncertainty. Based on this vision, we developed a novel paradigm called  ``\emph{networks of deep neural} networks", where speech recognition, speech enhancement and acoustic scene analysis DNNs are jointly trained and progressively learn how to communicate each other to improve their performance. The proposed architecture is unfolded using a strategy similar to that used for RNNs and it is trained with a variation of the back-propagation algorithm called \emph{back-propagation through network}. 
The experimental results highlighted the effectiveness of this approach.

\end{itemize}

\section{Thesis Outline}
\label{sec:outline}
This thesis is organized as follows. 

The state-of-the art in deep learning is summarized in Chapter \ref{cha:dl}. This Chapter will describe the basic algorithms, the most popular architectures  and will provide a brief overview on deep learning history, recent evolution, and future perspective. 

Chapter \ref{cha:dsr} then proposes an overview on the main state-of-the art technologies involved in distant speech recognition. First, the main challenges of DSR are summarized. Then, the basic components of a DSR systems (i.e., speech recognition, speech enhancement and acoustic scene analysis) will be described. 

The innovative contributions of this work are summarized in the remaining three chapters: the methods for contaminated speech training are detailed in Chapter \ref{sec:cont}, our studies on time contexts are illustrated in Chapter \ref{cha_time}, while the network of deep neural network paradigm is described in Chapter \ref{cha:ndnn}. Finally, some conclusions are drawn in Chapter \ref{cha:conclusion}. 

The thesis also includes three appendices describing in detail the adopted corpora, the considered microphone configurations, and the different setups used in the experiments.

\chapter{Deep Learning} \label{cha:dl}

Current computers are able to efficiently solve problems that can be described by a sequence of well-defined formal and logic rules, but have many difficulties in tackling tasks that are hard to formalize as a list of basic operations (e.g., recognize speech or objects in images). A research challenge of AI is to develop machines able to address the latter category of problems.  

Artificial intelligence can be approached in various ways. \textit{Knowledge-based} solutions rely on hard-coding logical rules in a formal language. Another approach, called \textit{machine learning}, aims to directly acquire knowledge from data, without (ideally) any human effort. The performance of machine learning systems strongly depends on the
representation of the input data. For many years the research in the field has focused on proper methods for feature extraction, transformation and selection \cite{nn_1}. In standard pattern recognition approaches \cite{pattern_rec_bishop}, these features were normally hand-designed and later exploited by classification algorithms.

Differently to past approaches, the idea of \textit{representation learning} is to jointly discover not only the mapping from input feature to output, but also the features itself \cite{bengio_rep_learning}.

\begin{figure}[t!]
   \centering
     \includegraphics[width=1.0\textwidth]{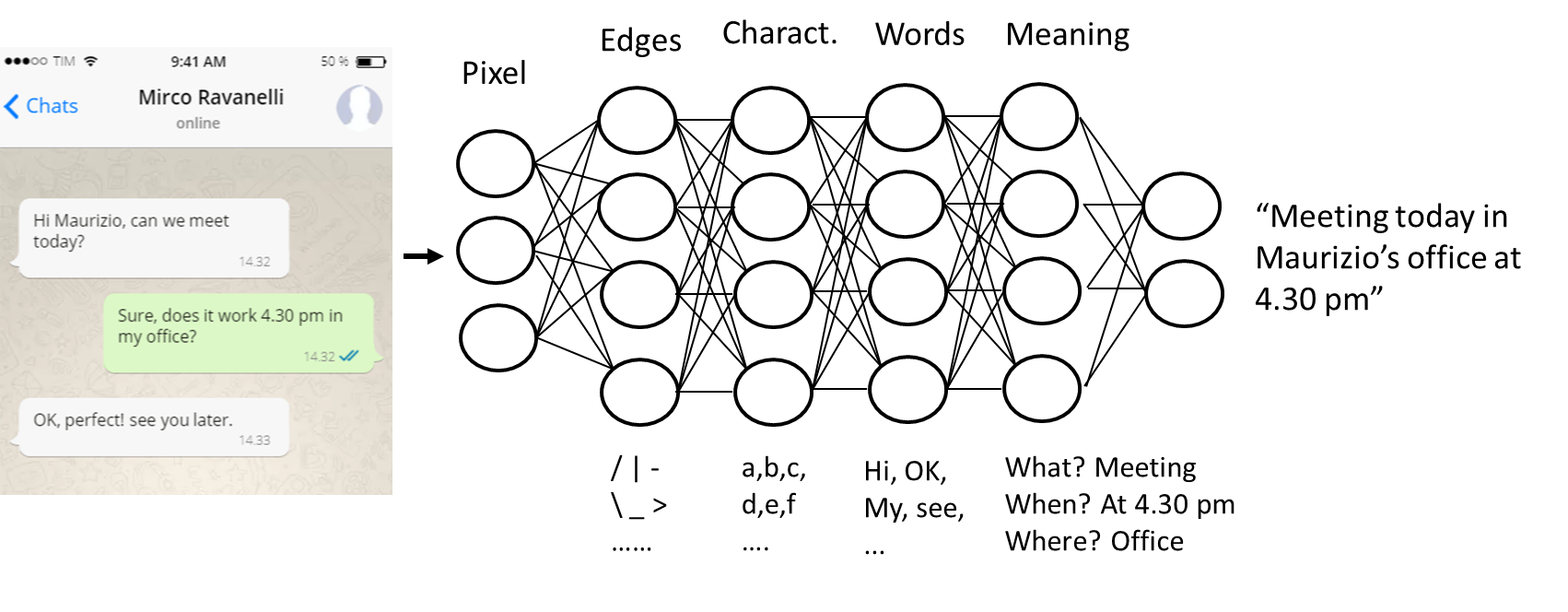}
     \caption{An example of an ideal deep learning system that learns hierarchical representations from low-level to higher level concepts.}
     \label{fig:dl_idea}
\end{figure}

\textit{Deep learning} \cite{Goodfellow-et-al-2016-Book} follows the philosophy of representation learning and aims to progressively discover  complex representations starting from simpler ones. The principle of composition, on the other hand, can be used to describe the world around us efficiently. An example is reported in Figure  \ref{fig:dl_idea}, showing that combination of pixels can describe edges, from combination of edges we can derive characters, words, and, finally, the semantic meaning of a chat.

The deep learning paradigm is currently implemented with Deep Neural Networks (DNNs), that are Artificial Neural Networks (ANNs) based on several hidden layers between input and output. Each layer learns higher-level features that are later processed by the following layer \cite{bengio_ai}. When a suitable high-level representation is reached, a classifier  can perform the final decision.  Modern DNNs provide a very powerful framework for supervised learning: when adding more layers, in fact, a deep network can represent functions of increasing complexity and can potentially reach higher levels of semantic representations. 

ANNs have been object of several research in the past decades \cite{ai_history}. These efforts were extremely important for the research community, since they laid the foundations for the basic learning algorithms \cite{pattern_rec_bishop}. For instance, the back-propagation algorithm was invented in the 60s-70s with the contributions of many scientists \cite{bp_history}. Despite these achievements,  the time was not yet ripe for the explosion of this technology. The current rise of deep learning can be explained with the following motivations: 
\begin{itemize}
\item \textbf{Big Data}; 
A key ingredient for the success of this technology is the availability of large datasets. Current systems, in fact, aim to incorporate considerable knowledge into a machine, requiring  lots of data. More precisely, the effectiveness of DNNs strongly depends on the capacity of the model, that can be increased by adopting deep and wide architectures. The improved network capacity, however, increases the number of parameters, inherently requiring  more data to reliably estimate them. 
Fortunately, the rapid spread of internet and smartphones, allows easy and cheap big-data collections.

\item \textbf{Computational power}; To properly exploit deep models and large datasets, a considerable computational power is required. In the last years, important progresses have been done to develop specialized hardware for deep learning. Modern Graphical Processor Units (GPUs), for instance, are currently used by most of deep learning practitioners to efficiently train complex models. 

\item \textbf{Computationally efficient inference}; DNNs usually require a lot of computational power during the training phase. An interesting aspect is that inference can be computed relatively efficiently, allowing, for instance, the development of real-time speech recognizers or low-latency dialogue systems.

\item \textbf{Powerful priors}; Last but not least, deep learning incorporates reasonable assumptions about the world. The basic assumption is the compositionality principle previously discussed, that efficiently describes the complex world around us as a progressive composition of different elements. 
This assumption acts as a prior knowledge used to defeat curse of dimensionality:  among all the possible functions that are able to explain a dataset, deep learning restricts this selection to a smaller sub-set that satisfies the compositionality constraint. This naturally entails a regularization effect, that allows training deep  architectures.
\end{itemize}


\section{Basic Algorithms} \label{sec:alg}
This section proposes an overview of the main algorithms and techniques used for deep learning. In particular, some general notions about supervised learning are recalled in sub-section \ref{sec:sup}.  The back-propagation algorithm is discussed in sub-section \ref{sec:bp}, while the main optimization techniques are summarized in sub-section \ref{sec:opt}. Regularization methods are finally described in sub-section \ref{sec:reg}. 

\subsection{Supervised Learning} \label{sec:sup}
DNNs are often trained in a supervised fashion \cite{pattern_rec_bishop}. This training modality can be formalized as follows: let's assume to have a set of N training examples \{($x_{1}$,$y_{1}$),...,($x_{i}$,$y_{i}$),...,($x_{N}$,$y_{N}$)\}, where each element is a pair composed of a feature vector $x_{i}$ and its corresponding label $y_{i}$. 
The learning algorithm seeks a function $f:X\rightarrow Y$ that maps the input space X into the output space Y.  Deep learning is a form of parametric machine learning, whose function $f$ depends on a set of trainable parameters $\theta$. For each particular choice of $\theta$, a different mapping function $f$ is obtained. The number of possible  functions that can be represented by the DNN is called \textit{capacity}. 

The goal of supervised learning is to find a function $f$ that is able to ``explain" well the training samples. More formally, this implies finding proper values of $\theta$ able to minimize a certain performance metric:

\begin{equation}
 \hat{\theta}=\argmin_\theta L(Y,f(X,\theta))
 \label{eq:prob}
\end{equation}

The function $L$ is called \textit{loss} (or cost) and, intuitively, should assume low values when the parameters $\theta$ lead to an output well-matching with the reference labels.  

In the context of the Maximum Likelihood Estimation (MLE), the optimization problem can be reformulated in this way:

\begin{equation}
 \hat{\theta}=\argmax_\theta P(Y \lvert X,\theta)
 \label{eq:max_lik}
\end{equation}

where $P(Y \lvert X,\theta)$ is the conditional probability distribution defined at the output of the DNN. To perform a MLE estimation, a popular choice of L is the Negative Log-Likelihood (NLL) or Cross-Entropy (CE). 
In this case, the MLE optimization can be rewritten as:

\begin{equation}
 \hat{\theta}=\argmin_\theta -log(P(Y \lvert X,\theta))
 \label{eq:cross-entropy}
\end{equation}

Another popular choice is the Mean Squared Error (MSE): 

\begin{equation}
 \hat{\theta}=\argmin_\theta || Y-f(X,\theta) ||^2
 \label{eq:cross-entropy}
\end{equation}

Solving the training optimization problem is challenging, especially for DNNs composed of a huge number of parameters. 
The functions $f$ originated by DNNs are, in fact, typically non-linear, making the optimization space highly non-convex. Recently, some theoretical and experimental studies have shown that the main challenge are \textit{saddle points} and not local minima as commonly believed in the past  \cite{saddle_point}.  
Even though these results are still object of an open debate in the research community, they suggest that the parameter space is flat almost everywhere and the (very rare) local minima are almost all also global ones \cite{no_local_minima}.

There are in principle various ways to solve this optimization problem \cite{convex_opt}. A naive solution would be to try all the possible $\theta$ and choose the configuration that minimizes the cost function. This approach is clearly unfeasible, especially for a large number of parameters. Another solution would be to exploit evolutionary optimizations, based on genetic algorithms or particle swarm techniques \cite{genetic_book}. Despite the interesting aspects of these optimizers (e.g, high parallelism, differentiability is not strictly required), such methods require very frequent evaluations of the cost function, that is impractical for DNNs trained on large datasets.
At the time of writing, the most popular choice is gradient-based optimization. The computation of the gradient is described in the following section, while the optimization step  will be discussed in Sec. \ref{sec:opt}.  

\subsection{Back-Propagation Algorithm} \label{sec:bp}

The gradient $\partial L / \partial \theta$ is a very precious information that describes what happens to the cost function $L$ when a little perturbation is applied to the parameters $\theta$. 
If this perturbation causes an improvement of the loss, it could be convenient to do a little step in the direction indicated by the gradient. 

The computation of $\partial L / \partial \theta$ is normally performed with the back-propagation  algorithm \cite{backprop}, that is often misunderstood as meaning the whole learning procedure. Actually, back-propagation is only a method for computing the gradient. 

\begin{figure*}[t!]
   \centering
     \includegraphics[width=1.0\textwidth]{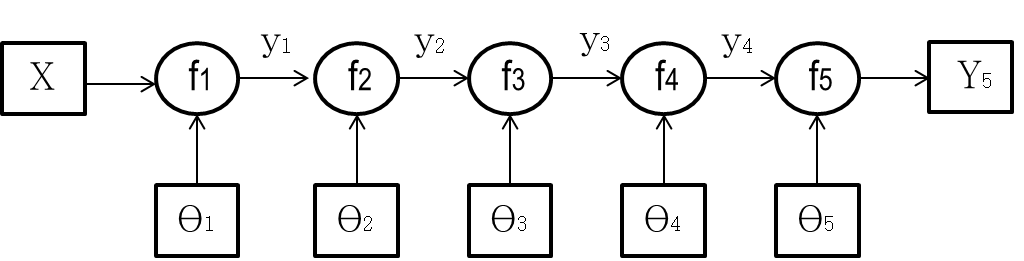}
     \caption{A composite function representing the computations performed in the various layers of a DNN.}
     \label{fig:bp_al}
\end{figure*}

Deriving an analytical expression of $\partial L / \partial \theta$  is rather straightforward for systems that are differentiable almost everywhere. In most of the cases, in fact, the analytical expression of the gradient is a direct application of basic calculus rules \cite{leibniz}. As shown in Figure  \ref{fig:bp_al}, a DNN  can be described as a composite function that performs a chain of computations. Gradients can thus be computed with the chain rule \cite{leibniz}, as reported in the following equations:

\begin{subequations}
\begin{align}
\frac{\partial y_{5}}{\partial \theta_1}&=\frac{\partial y_{1}}{\partial \theta_1} \cdot \boldsymbol{\frac{\partial y_{2}}{\partial y_1} \cdot \frac{\partial y_{3}}{\partial y_2} \cdot \frac{\partial y_{4}}{\partial y_3} \cdot \frac{\partial y_{5}}{\partial y_{4}}} \\
\frac{\partial y_{5}}{\partial \theta_2}&=\frac{\partial y_{2}}{\partial \theta_2} \cdot \boldsymbol{ \frac{\partial y_{3}}{\partial y_2} \cdot \frac{\partial y_{4}}{\partial y_3} \cdot \frac{\partial y_{5}}{\partial y_{4}}} \\
\frac{\partial y_{5}}{\partial \theta_3}&=\frac{\partial y_{3}}{\partial \theta_3} \cdot \boldsymbol{\frac{\partial y_{4}}{\partial y_3} \cdot \frac{\partial y_{5}}{\partial y_{4}}} \\
\frac{\partial y_{5}}{\partial \theta_4}&=\frac{\partial y_{4}}{\partial \theta_4} \cdot \boldsymbol{\frac{\partial y_{5}}{\partial y_{4}}}
\end{align}
\end{subequations}

Beyond the mathematical expression, the intuition behind the chain rule is this: the little change applied to the parameter $\theta_1$ causes a change on the output of function $f_1$. This perturbation will propagate until the end of the chain, eventually causing a perturbation on the final node, that typically computes the cost function L.

Despite the relative simplicity of the gradient expression, its numerical evaluation can be computationally expensive. A naive (but inefficient) solution would be to independently evaluate the gradient expression for each parameter. 
The back-propagation algorithm \cite{backprop} is a simple and inexpensive procedure that performs the various operations in a specific order. From the previous equations, one can notice that there are several shared computations. It would be, for instance, very convenient to start from the last equation, store the results in bold, and reuse them to compute the following gradients.

The back-propagation algorithm is based on a dynamic programming approach and is summarized by the following step:

\begin{enumerate}
\item \textbf{Forward Propagation}; propagate the input features to the output and store the activations $y$.
\item \textbf{Compute Cost}; evaluate the cost function $L$ at end of the chain.
\item \textbf{Back Propagation}; compute the gradient from the last element of the chain to the first one.
\end{enumerate}

When propagating gradients through long computational chains, however, some issues might arise \cite{pascanau}. The chain rule, in fact, implies several gradient multiplications that can lead to \textit{vanishing} or, less often, to \textit{exploding gradients} \cite{Bengio94}.   

Exploding gradients can effectively be tackled with simple clipping strategies \cite{pascanau}, while vanishing gradient is more critical and might impair training very deep neural networks. As will be discussed in the following part of this chapter, special architectures are often needed to properly address this issue.

\subsection{Optimization} \label{sec:opt}
Once computed,  the gradient has to be  exploited  by an optimizer to progressively derive better parameters.

The most popular optimization algorithm is Gradient Descend (GD), that updates $\theta$ according to the following equation:

\begin{equation}
 \theta=\theta - \eta \frac{\partial L}{\partial \theta}
 \label{eq:sgd}
\end{equation}

The parameters are updated in the direction pointed by the gradient,  with a step size determined by the learning rate $\eta$ (the minus is due to the minimization of the loss). Note that gradient-based optimization does not provide any guarantee on global optimality. In such  complex high-dimensional spaces, a crucial role is thus played by a proper initialization of the $\theta$ and by a suitable choice of loss and activation functions.

Depending on how many data are used to compute the gradient, some variants of GD can be defined. When the gradient is computed on the full training dataset, the optimizer is called batch-GD. A popular alternative is Stochastic Gradient Descend (SGD) that splits the dataset into several smaller chunks (called mini-batches) and update the parameters more frequently, with well-known benefits in terms of both accuracy and training convergence.  

Standard SGD, however, has trouble navigating on areas where the surface curves much more steeply in one dimension rather than in another. To mitigate this issue, a momentum is often applied to the update equations for accelerating the training convergence \cite{momentum}.

Another issue is that the same learning rate applies to all  the parameters. This stiffness can  be critical, since each $\theta$ has its own characteristics, possibly requiring independent learning rates. For instance, if the absolute value of the gradient is very large in a specific dimension, it means that we are in a very steep area and would be more prudent to do a little step using a small learning rate. On the contrary, when the gradient is small, a higher learning rate can be used. 
Following this philosophy, several variations of SGD, such as Adagrad \cite{adagrad}, Adadelta \cite{adadelta}, RMSprop, Adam \cite{adam},  have been recently proposed (see the reference papers for more details). These solutions, in general, rescale $\eta$ by gradient-history metrics,  resulting in a faster and more robust optimization.

\subsection{Regularization} \label{sec:reg}
A crucial challenge in machine learning is the ability to perform well on previously unobserved inputs. This ability is called \textit{generalization} \cite{Goodfellow-et-al-2016-Book}.
Possible causes of poor generalization are \textit{underfitting} and \textit{overfitting}, that are two central issues when training DNNs. Underfitting occurs when the model is not able to reach a sufficiently low error on the training set. This might arise when the DNN has not enough capacity.  Overfitting occurs when the gap between training and test errors is too large. Differently to undefitting, this might happen when the model has excessive capacity.
The strategies to counteract overfitting are known as \textit{regularization}, and normally consist of methods for reducing the network capacity based on prior knowledge. From this point of view, even deep learning itself can be regarded as a regularization technique, due to the prior knowledge naturally embedded in the compositionality principle. 

The great importance of these methodologies has made regularization one of the major research directions in the field. 
In the following, the most popular regularizers are described.

\subsubsection{$L^2$ Regularization} \label{sec:l2}
Many regularization approaches are based on limiting the model capacity by adding a parameter norm penalty to the cost function. These approaches tend to penalize too complex solutions, following the philosophy of the \textit{Occam's razor} principle: ``among competing hypotheses leading to the same performance, the simpler one should be selected".  The most common method following this approach is the $L^2$ regularization, that add a penalty to the loss function L:

\begin{equation}
  \tilde{L}= L + \alpha ||\theta||^2
 \label{eq:l2}
\end{equation}
where $\alpha$ is an hyperparameter that weights the  contribution of the
regularization term.
The insight behind this kind of regularization is that  solutions characterized by higher norms of the the trainable parameters are more complex and are more likely to overfit the training dataset.

\subsubsection{Dropout}
An effective way to improve generalization is to combine several different models \cite{ensamble}. If each classifier has been trained separately, it might have learned different aspects of the data distribution and their mistakes are likely to be complementary. Combining them helps produce a stronger model, that is hopefully less prone to overfitting.
Even though these methods are very effective, a major limitation lies in the considerable computational efforts needed to train and test different DNNs. Dropout \cite{dropout} is an effective regularization method that provides an inexpensive
approximation to training and evaluating an exponential number of neural networks. 

The key idea is to randomly drop neurons during training with a probability called dropout rate $\rho$. This way, a subnetwork is sampled from an exponential number of smaller DNNs for each training sample.
At test time, the whole network is used (i.e., the DNN with all the neurons active), but the activations are scaled down by $\rho$. 
 
This ensemble learning approach significantly reduces overfitting and gives major performance improvements. Many variants of dropout have been proposed in the literature \cite{fast_drop,drop_distillation}. For instance, in \cite{drop_connect} the regularizer is applied to the weight connections rather than on neurons, while in \cite{drop_asru,Gal2016} dropout is extended to recurrent neural networks.

\subsubsection{Data Augmentation}
The best way to achieve generalization would be to train the model with more data. However, in practice, the amount of data is limited and the collection of large annotated corpora is very expensive. A possible alternative is to artificially process the available data, in order to generate novel training samples. For a computer vision application, one can for instance, rescale, rotate, or shift the available images to generate novel samples. For speech recognition, one way is to apply proper algorithms able to modify pitch, formants, and other speaker characteristics.  Data augmentation can also be regarded as a way for adding prior knowledge to a model, since  we exploit the information that the new samples do not change the class label when are processed.

\subsubsection{Other Regularizers}
Other approaches have been proposed for counteracting overfitting.
A popular way is to add random noise  during learning. Several methods have been proposed in the literature, proposing to add it at gradient, weight, input, output  or hidden activation levels \cite{adding_noise1,adding_noise2,adding_noise3}. 

An alternative consists in adopting a semi-supervised approach, where unsupervised data can be exploited as prior knowledge to improve generalization. From this point of view, the pre-training approach based on Restricted Boltzman Machines (RBM) \cite{rbm1} or autoencoders \cite{autoencoder} can be regarded as a form of regularization. 
Multi-task learning \cite{multi_task} (i.e., building DNN solving multiple correlated tasks) can also be considered as a sort of regularization, since it encourages the DNN to discover very general features at the first hidden layers.

Finally, one the most popular and simple approach for regularization is early stopping \cite{early_stopping}. The idea is to periodically monitoring the performance of the DNN on held-out data and stop the training algorithm when this performance begins to deteriorate.

\subsection{Hyperparameter selection} \label{sec:hyp}
Most deep learning algorithms are based on some hyperparameters that must be properly set to ensure a good performance. The most important ones describe the structure of the network (e.g., number of hidden layers, number of hidden units per layer), determine the optimization characteristics (e.g., learning rate), and specify the behaviour of the regularizer (e.g., weight decay or dropout rate).
Properly setting the hyperpameters is rather difficult, mainly because a new model should be trained for each new setting. Moreover, several hyperparameters can be correlated each other, making an independent optimization of them  usually not viable.  

There are two basic approaches to derive them: choosing them manually and choosing them automatically. The manual selection requires a considerable experience and familiarity with the addressed task as well as a precise knowledge of the role of each specific hyperparameter.
This is possible for well-explored machine learning tasks, where a detailed literature suggesting reasonable settings is available.

When the manual approach is not feasible, a possible alternative is to automatically select the hyperparameters with grid search.
Grid search simply tries all the combinations over a specified range of values. Although this search can be easily parallelized, its computational expense is exponential in the number of hyper-parameters. 
A straightforward alternative is to sample them randomly \cite{random_search}. This approach is still very easy to parallelize, and is generally faster than grid search. 

\section{Main Architectures} \label{sec:arch}
This section summarizes the main architectures used in deep learning. In particular, sub-sec. \ref{sec:act} discusses the main type of neurons. 
Sub-sec. \ref{sec:ff} describes feed-forward neural networks, sub-sec. \ref{sec:bn_intro} introduces batch normalization, while sub-sec. \ref{sec:rnn} describes Recurrent Neural Networks (RNNs). Finally, some architectural variations recently proposed in the literature are summarized in sub-sec. \ref{sec:other_dnn}.

\subsection{Neuron Activations} \label{sec:act}

The computations performed by the DNNs are a sequence of linear operations followed by non-linear ones. More precisely, the output of a hidden layer $h_{i+1}$ composed of $n$ neurons and fed by $m$ input features $h_{i}$, can be represented as follows:
\begin{equation}
h_{i+1}=g(\underbrace{W h_{i}+b}_{a})
\label{eq:hidden}
\end{equation}
where $W$ is the weight matrix ($n \times m$), b is the bias vector of $n$ element and $g$ is the activation function. The linear transformation performed before applying $g$ is called affine transformation $a$. The choice of $g$ is particularly important, and several research efforts have been devoted to study proper activation functions. The most popular are:
\begin{itemize}
\item \textbf{Linear};  The simplest neurons are obtained by directly taking the affine transformation without any non-linearity. These neurons are called linear and can be used, for instance, for predicting real numbers  in the output layer (regression problem). 

\item \textbf{Sigmoid};  For several decades, the most popular activation was the  logistic sigmoid:
\begin{equation}
\sigma(x)=\frac{1}{1+exp(-x)}=\frac{exp(x)}{1+exp(x)}
\label{eq:softmax}
\end{equation}
The main advantage is that such activations are bounded between 0 and 1. However, the use of sigmoid activations in modern feedforward networks is now discouraged,  since they are characterized by close-to-zero gradients across most of their domain. In fact, $\sigma(x)$ saturates when
their input is very positive or very negative, slowing down the training.

\item \textbf{Hyperbolic Tangent}; This kind of non-linearity is a rescaled version of the sigmoid function: 
\begin{equation}
tanh(x)=\frac{exp(x)-exp(-x)}{exp(x)+exp(-x)}=2\sigma(2x)-1
\label{eq:softmax}
\end{equation}

It suffers from the main disadvantages of  sigmoid units, but in general it provides better performance. This is due to the fact that tanh is symmetric around zero, making this activation less prone to saturation in the last layers.

\item \textbf{Rectified Linear Units}; This kind of activation is currently the most popular choice in modern feed-forward neural networks \cite{relu_jarret,relu_speech}:
\begin{equation}
ReLU(x)=max(0,x)
\label{eq:softmax}
\end{equation}
The main insight behind the use of these activations is that they are rather similar to linear units. Linear units, as we have outlined before, lead to a convex optimization space, that is very easy to optimize. ReLUs inherit, at least in part, the benefits of the latter activations. More precisely, the
derivatives of a ReLU remain large whenever the unit is active.
When initializing the parameters of the affine transformation, a good
practice is to set the bias $b$ to a small positive value (e.g., 0.1),  fostering the unit to be active during the first part of the training. This allows the derivatives to pass unaltered through all the non-linearities without vanishing gradient problems. The main issue is that the unboundedness  of ReLU can cause large activations with possible problems of numerical stability. Several modifications to standard ReLU have been proposed in the literature, including leaky ReLU \cite{leaky_relu}, parametric ReLU \cite{param_relu} and Maxout units \cite{maxout}.

\begin{figure}
    \centering
    \begin{subfigure}[b]{0.35\textwidth}
        \includegraphics[width=\textwidth]{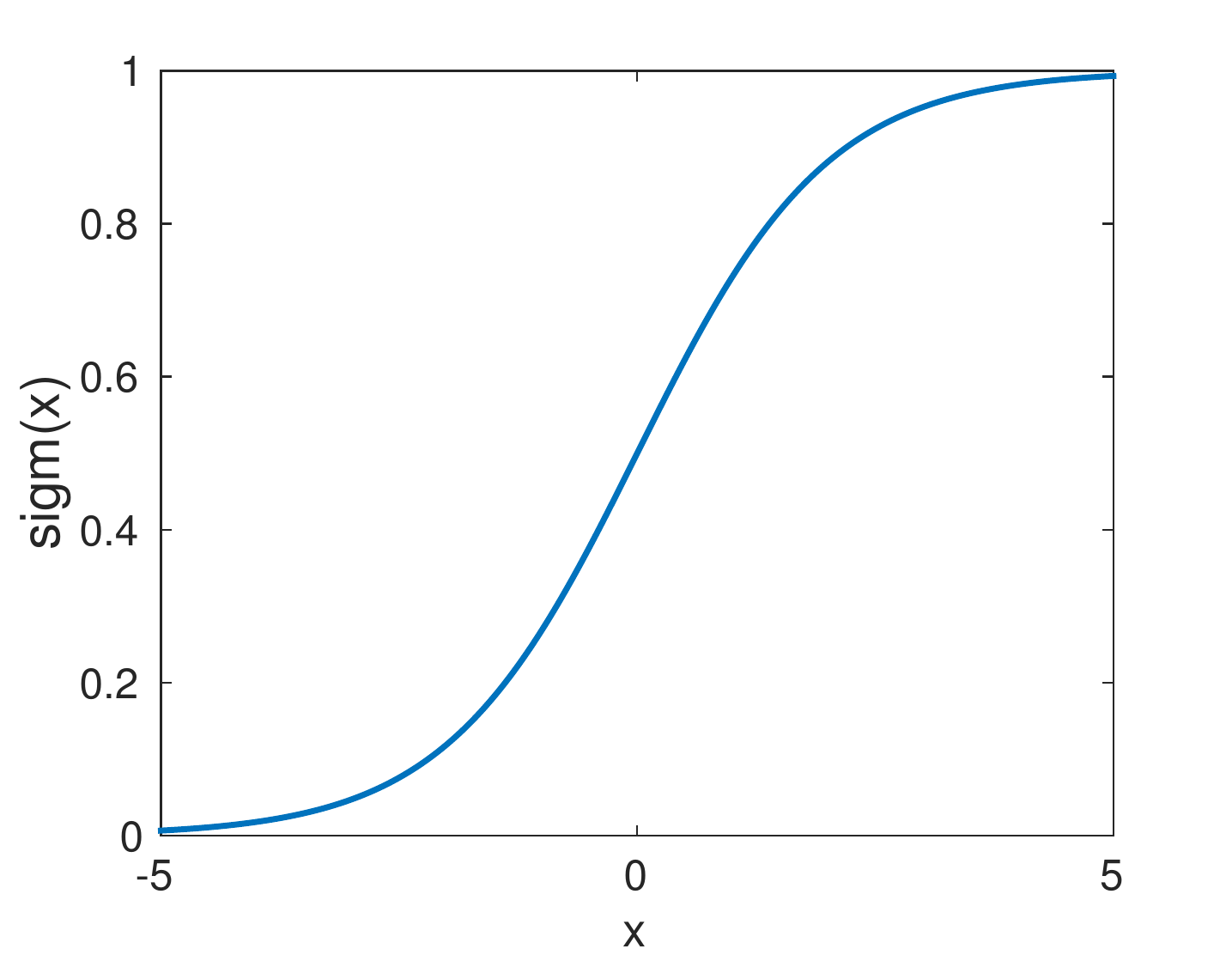}
        \caption{Sigmoid}
        \label{fig:sigm}
    \end{subfigure}
    ~ 
    \begin{subfigure}[b]{0.35\textwidth}
        \includegraphics[width=\textwidth]{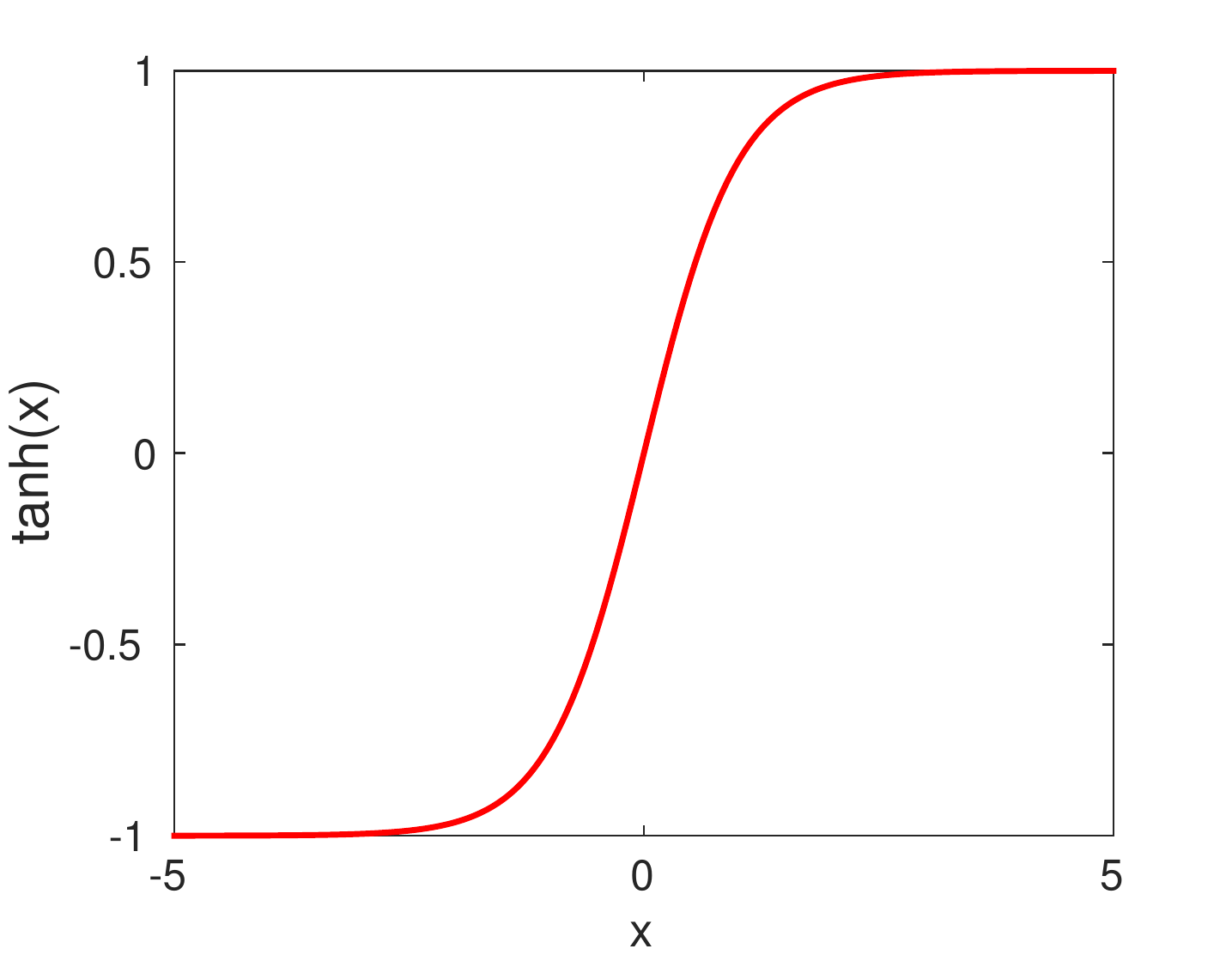}
        \caption{Tanh}
        \label{fig:tanh}
    \end{subfigure}
    ~ 
    \begin{subfigure}[b]{0.35\textwidth}
        \includegraphics[width=\textwidth]{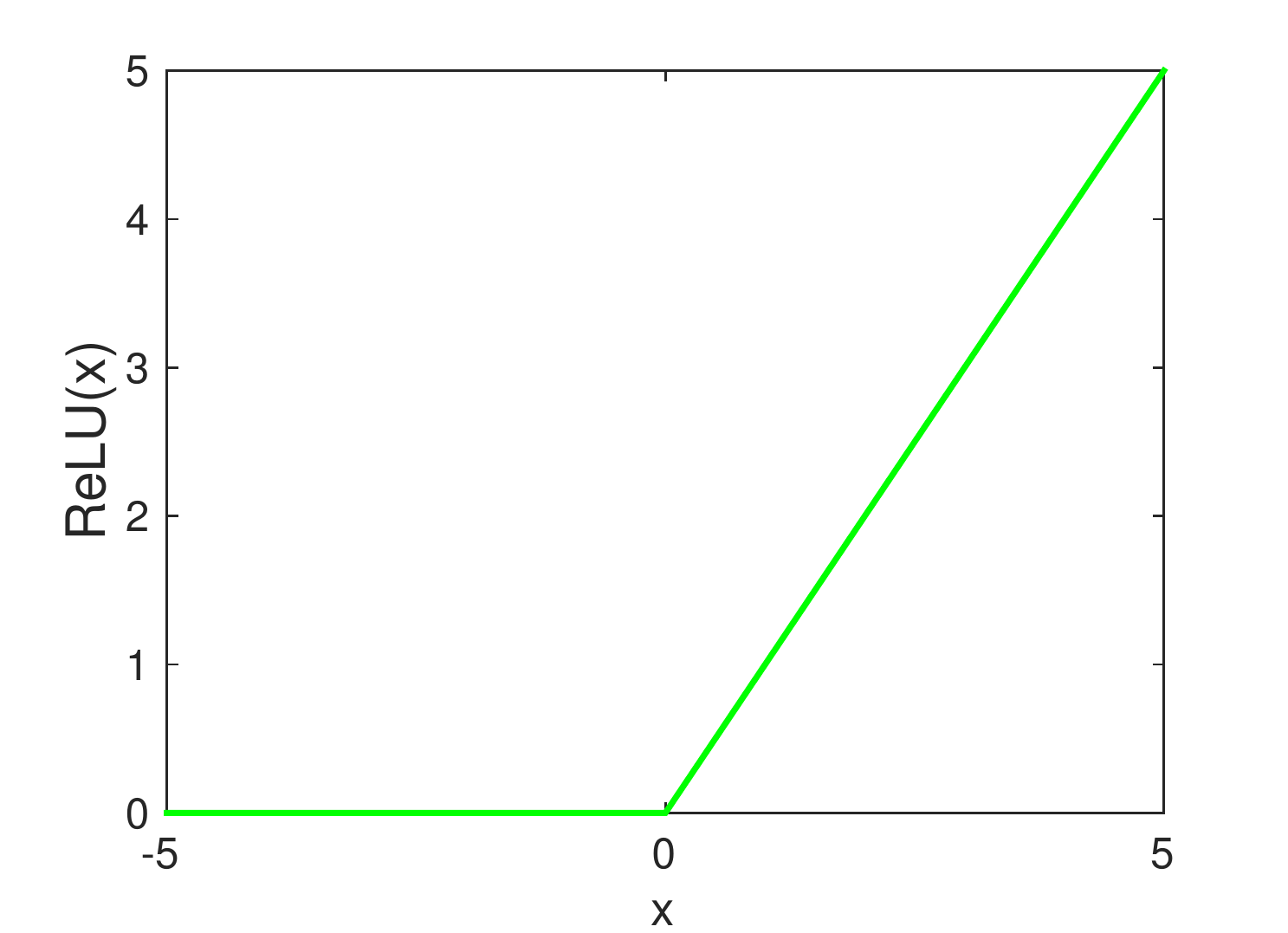}
        \caption{ReLU}
        \label{fig:relu}
    \end{subfigure}
    \caption{Main activation functions used in deep neural networks.}\label{fig:act}
\end{figure}

\item \textbf{Softmax}; Softmax neurons are often used in the output layer to  estimate a probability distribution over a set of $n$ alternatives. To represent probabilities, each output neuron must assume values between 0 and 1, and the sum of all of them must be  1.  Formally, the softmax of the i-th neuron is defined as follows:
\begin{equation}
softmax(a_{i})=\frac{exp(a_{i})}{\sum_{j=1}^n exp(a_{j})}
\label{eq:softmax}
\end{equation}
Softmax creates a competition across the neurons through  its normalization of the affine transformation $a$.  This normalization can be done, in principle, with many other functions, including the linear one. The choice of the exponential function is done because it couples well with the maximum log likelihood optimization carried out to train DNNs. The log-softmax, in fact, can be written as follows:

\begin{equation}
log\big(softmax(a_{i})\big)=a_{i} - log\big(\sum_{j=1}^n exp(a_{j})\big)  
\label{eq:logsoftmax}
\end{equation}

The MLE optimization would try to push up the activation of correct neurons, while penalizing the other ones, in particular the most active incorrect predictions.

\end{itemize}

When training DNNs, a crucial aspect is the parameter initialization, that has to be carefully designed according to the specific choice of the activation function. A crucial requirement is to break the symmetry (i.e., avoiding setting all the parameters with the same initial value). In fact, same weights will compute the same gradients with identical updates. 
A popular approach is to randomly initialize the parameters with small random real values. One problem with this method is that the variance at the output of each neuron grows with the number of input weights. As proposed in \cite{xavier}, one solution is to normalize the initial weight variance by the number of neurons.

\begin{figure*}[t!]
   \centering
     \includegraphics[width=0.9\textwidth]{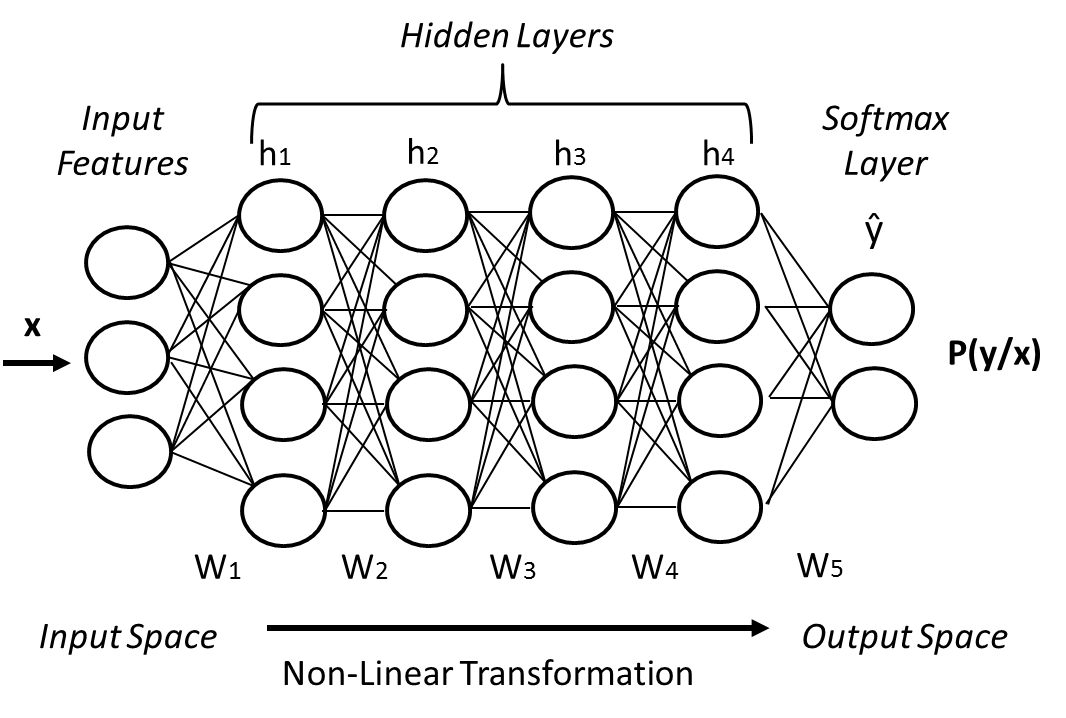}
     \caption{A Multi-Layer Perceptron (MLP) composed of four hidden layers.}
     \label{fig:DNN}
\end{figure*}

\subsection{Feed-forward Neural Networks} \label{sec:ff}
According to their architecture, DNNs can be classified into two main categories: feed-forward and recurrent neural networks.
In the context of Feed-Forward DNNs (FF-DNN), the prediction $P(y/x)$ is just a function of the current input $x$ and of the parameters $\theta$:
\begin{equation}
y=f(x,\theta)  
\label{eq:ff}
\end{equation}

These models are called feed-forward because the information flows from the input to the output without any feedback connection. An example of feed-forward neural network composed of four hidden layers is shown in Figure  \ref{fig:DNN}. 
The figure depicts a popular architecture called fully-connected DNN or Multi-Layer Perceptron (MLP), where all the neurons are connected with the ones of the following layer. 

\begin{figure*}[t!]
   \centering
     \includegraphics[width=1.0\textwidth]{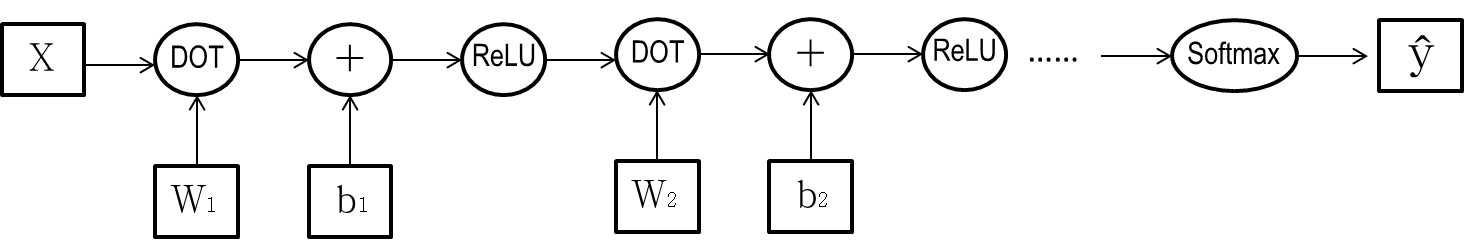}
     \caption{A computational graph representing the chain of computation for a feed-forward MLP.}
     \label{fig:graph}
\end{figure*}

Figure   \ref{fig:graph} shows an alternative representation of the network, using the so-called computational graphs, that are flexible tools for formalizing sequences of computations.  There are several ways of formalizing computational graphs. In the figure, for instance, the variables (that can be scalars, matrices or tensors) are represented by squares, while the operations are represented by circles. This approach can be extended to describe all the DNN computations, including the operations performed for gradient and cost function  estimation.

Interesting, the universal approximation theorem states that a MLP with at least one hidden layer can approximate any continuous  function\footnote{The theorem is actually restricted to continuous functions on compact sets that operates a mapping from a finite-dimensional space to another.}  with any desired non-zero error, provided that the network is composed of  enough hidden neurons \cite{universal_approx}. This theorem seems in contrast with the deep learning principle, which suggests to adopt DNNs composed of several hidden layers.
However, the universal approximation theorem only guarantees that the function exists, but does not provide any insight on how to find it. There are thus no guarantees that the training algorithm will be able to learn that function. Moreover, the theorem does not provide any way for properly choosing this transformation such that it generalizes to points not included in the training set. On the other hand, another theorem  (called ``\textit{no free lunch}” theorem \cite{no_free_lunch}) says that there is no universally superior machine learning algorithm. 

An alternative to fully-connected neural neural networks are Convolutional Neural Networks (CNNs) \cite{cnn_lecun}. Differently to the former ones, CNNs are based on local connectivity, weight sharing and max pooling. 
The combination of these characteristics make CNNs particularly suitable for managing correlations across features. Moreover, the presence of max pooling allows the network to obtain shift-invariant properties. 
CNNs are inspired by biological studies of the visual cortex and  are extremely successful in practical applications, especially in computer vision \cite{dnn_4}. CNNs have also been applied in speech recognition, achieving interesting performance thanks to some degree of invariance to small shifts of speech features along the frequency axis, that resulted important to deal with speaker and environment variations \cite{cnn1}. 

\subsection{Batch normalization} \label{sec:bn_intro}
Training DNNs is complicated by the fact that the distribution of each layer's inputs changes during training, as the parameters of the previous layers change.
This problem, known as \textit{internal covariate shift}, slows down the training of deep neural networks. 
Batch normalization \cite{batchnorm}, that has been recently proposed in the deep learning community, addresses this issue by  normalizing the mean and the variance of each layer's pre-activation for each training mini-batch. 
It has been long known that the network training converges faster if its inputs are properly normalized \cite{yann} and, in such a way, batch normalization extends this normalization to all the layers of the architecture. More precisely, batch normalization is defined as follows:

\begin{equation}
BN(a)=\gamma
\label{eq:bn} \frac{a-\mu_b}{\sqrt[]{\sigma_b^2+\epsilon}}+\beta
\end{equation}
where $a$ is the neuron pre-activation (i.e., the output before applying the non-linearity), $\mu_b$ and $\sigma_b$ are the mean and standard deviations computed for each minibatch and  $\epsilon$ is constant introduced for numerical stability. The variables $\gamma$ and $\beta$ are trainable scaling and shifting parameters, introduced to allow each neuron of the network to have an explicit control of mean and variance statistics. The computations involved for batch normalization are fully differentiable, and it is possible to back-propagate through it.

Batch normalization resulted particularly helpful to achieve regularization,  to significantly speed-up the convergence of the training phase as well as to improve the overall accuracy of a DNN. The regularization effect is due to the fact that mean and variance normalizations are performed on each mini-batch rather than on the entire dataset. This approximated estimation of the normalization statistics introduces a sort of noise in the learning process, that resulted helpful to counteract overfitting \cite{cesar,initbn}. Moreover, batch normalization inherently reduces the capacity of the DNN: among all the possible functions able to explain the training data, only the subset that respect the normalization constraint can be chosen.

\subsection{Recurrent Neural Networks} \label{sec:rnn}
Recurrent Neural Networks (RNN) are architectures suitable for processing sequences \cite{rnn_overview}. The elements of a sequence are, in most of the cases, not independent. This means that, in general, the emission of a particular output might depend on the surrounding elements or even on the full-history. To properly model the sequence evolution, the presence of memory to keep track of past or future elements is thus of fundamental importance. The memory can be implemented using feedback connections, that introduce the concept of \textit{state}. In particular, RNNs are characterized by the following equation:
\begin{equation}
h_{t}=f(x_{t},h_{t-1},\theta)  
\label{eq:rnn}
\end{equation}
where $h_{t}$ and $h_{t-1}$ are the current and the previous states, respectively. 
Due to this recurrence, the current state actually depends  on all the previous ones.

The simplest form of RNN is the so-called \textit{vanilla} RNN, that is described by the following equation:

\begin{subequations}
\begin{align}
&h_{t}=tanh(W x_{t} + U h_{t-1} + b) \\
\end{align}
\end{subequations}

In this case, the parameters $\theta$ are the weight matrix W (feed-forward connections), the matrix U (recurrent weights) and the vector b (bias).

\begin{figure*}[t!]
   \centering
     \includegraphics[width=1.0\textwidth]{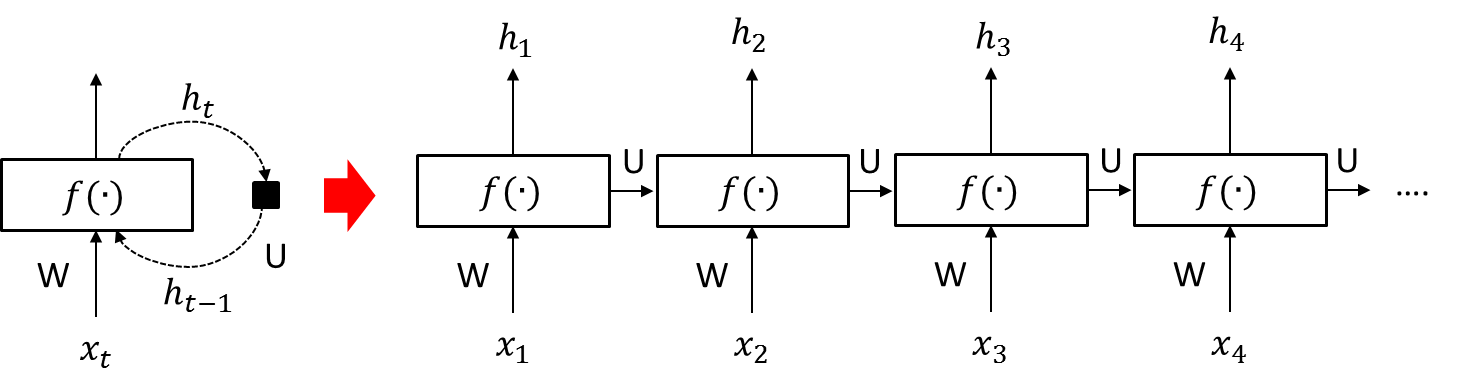}
     \caption{An example of RNN unfolding procedure.}
     \label{fig:RNN_unfolding}
\end{figure*}

To train RNNs, an unfolding of the network is necessary. This operation, that is graphically represented in Figure  \ref{fig:RNN_unfolding}, can be carried out as highlighted in the following equation:

\begin{subequations}
\begin{align}
&h_{3}=f(x_3,h_{2},\theta)  \\
&h_{3}=f(x_3,f(x_2,h_{1},\theta),\theta) \\
&h_{3}=f(x_3,f(x_2,f(x_1,h_{0},\theta),\theta),\theta) 
\end{align}
\end{subequations}

After unfolding, the RNN can be treated as feed-forward neural network, that is very deep along the time axes. Therefore, the same algorithms adopted for training feed-forward neural networks can be used. Sometimes, the back-propagation algorithm in the context of recurrent neural networks is called back-propagation through time, to emphasize that the gradient is propagated through the time axes. 
An important aspect of RNNs is that the parameters are shared across the time steps, making model generalization more easy.

Back-propagating the gradient through many time steps, however, can be complicated by vanishing and exploding gradients, that might impair learning long-term dependencies \cite{Bengio94}.
As we have seen in Sec. \ref{sec:bp}, exploding gradients can effectively be tackled with simple clipping strategies \cite{pascanau}, while vanishing gradient requires special architectures to be properly addressed. A common approach relies on the so-called gated RNNs, whose core idea is to introduce a gating mechanism for better controlling the flow of the information through the various time-steps. Within this family of architectures, vanishing gradient issues are mitigated by creating effective ``shortcuts", in which the gradients can  bypass multiple temporal steps.

The most popular gated RNNs are LSTMs \cite{lstm}, that often achieve state-of-the-art performance in several machine learning tasks, including speech recognition \cite{graves,lstm_speech,baidu,dnn_se3,joint6,chime4_paper}.
LSTMs rely on a network design consisting of memory cells that are controlled by forget, input, and output gates (see \cite{lstm} for the full list of equations describing this model).
Despite their effectiveness, such a sophisticated gating mechanism might result in an overly complex model. On the other hand, computational efficiency is a crucial issue for RNNs and considerable research efforts have recently been devoted to the development of alternative architectures \cite{lstm_odyssey,gru3,lstm_highway}. A noteworthy attempt to simplify LSTMs has recently led to a novel model called Gated Recurrent Unit (GRU) \cite{gru1,gru2}, that is based on just two multiplicative gates. This typology of RNN has been object of several studies in this thesis, as will be discussed in sec. \ref{sec:rnn_context}.


\subsection{Architectural Variations} \label{sec:other_dnn}
During the last years, several novel architectures have been proposed in the deep learning field. Although a complete overview of all the recently-proposed architecture is out of the scope of this thesis, this subsection summarizes the most popular ones.

First of all, architectures based on combinations of convolutional, recurrent and fully-connected layers gained a lot of popularity, especially in the field of speech recognition \cite{conv_lstm_dnn}. For ASR, convolutional layers are used for feature extraction, recurrent layers for temporal processing, and fully-connected layers for the final classification.

Other recently-proposed DNNs extend the shortcut idea to feed-forward DNN for improving the propagation of the gradient across the various hidden layers. With this regard, a popular architecture is called Residual Neural Networks (ResNet) \cite{res_net}. ResNets consider a direct connection through the various hidden layers,  that allow the gradient to flow unchanged. By stacking these layers, the gradient can theoretically pass over all the intermediate layers and reach the bottom one without being diminished.
These networks are called residual because, instead of directly modeling the output distribution, they actually model the difference between the input and output distributions. When several hidden layers are available, the network representations evolve rather slowly from a low-level to a high-level features. It would be thus natural to assume that input and output distributions would be rather similar and could be convenient to model this small difference.

A related idea is exploited in the context of Highway connections \cite{highway,lstm_highway}. 
Highway Networks preserve the shortcuts introduced in ResNets, but augments them with a learnable gate that manages the information flow through the hidden layers. The latter networks actually extend the idea of multiplicative learning gates  to feed-forward DNNs.

\section{Deep Learning Evolution} \label{sec:del_ev}
In the following subsections a brief history, the recent progress as well as the future challenges of deep learning are summarized.

\subsection{Brief History}  \label{sec:dl_history}
To better understand deep learning, it is useful to briefly summarize its history. 
Contrary to common belief, deep learning is not a new research field and has a rather long and troubled history, starting approximatively in the 40s. During the past 80 years, deep learning changes its name (and fortune) several times. There were, indeed, three main historical trends that have characterized the evolution of this technology under different names:  cybernetics (40s-60s), connectionism (80s-90s) and deep learning (2006-).

During 40s, cybernetics was inspired by theories of biological learning. For instance, McCulloch-Pitts \cite{first_neuron} proposed  a first simplified model of a neuron. This neuron  was able to recognize two different categories by testing whether the output (derived from a simple linear transformation)  was positive or negative.  Interestingly, this neuron resembles the ones used nowadays in modern deep learning systems, where the linear transformation processes the inputs of each hidden layer.
In the model proposed by McCulloch-Pitts, the weights W were set by a human operator.
The work by Rosenblatt \cite{rosenblatt}, who led to a neuron called Perceptron,  became the first model that could learn the weights from data in a supervised way. The adaptive linear element (ADALINE) proposed almost simultaneously by Widrow and Hoff \cite{Widrow} was able to exploit a linear neuron to predict a real number (regression problem), using an algorithm very similar to Stochastic Gradient Descend (SGD).

The second evolution of deep learning started with the connectionism approach (80s-90s) \cite{pattern_rec_bishop}. Connectionism was also strongly influenced by the studies on biological neurons and it is based on the  idea that complex non-linear functions can be obtained from the combination of several simple artificial neurons. In effect, biological neurons are cells performing a basic operation and networking them is the key for archiving intelligence. A major accomplishment of the connectionist era was the popularization of the back-propagation algorithm \cite{bp1}. 
During the 90s, important advances were made in modeling time sequences
with recurrent neural networks. In particular,  Hochreiter and Bengio \cite{Bengio94} identified some of the fundamental challenges in modeling long sequences. Some years later, Hochreiter and Schmidhuber (1997) proposed a novel approach for mitigating some of these issues, leading to an architecture called Long Short Term Memory (LSTM) \cite{lstm} that is still very popular to process sequences. The interest towards connectionism  began to wane since mid 90s, in favor of kernel methods \cite{kernel_methods} and graphical models \cite{graphical_model_book}, that achieved good performance in many tasks. During this long deep learning winter, the progress in the field was mainly promoted by the Canadian Institute for Advanced Research (CIFAR), that gathers the research labs led by Geoffrey Hinton the at University of Toronto, Yoshua Bengio at University of Montreal, and Yann LeCun at New York University. 

The new life of neural networks began in 2006, when Geoffrey Hinton showed that neural networks with many hidden layers (called Deep Neural Networks) could be trained using an unsupervised pre-training strategy based on  deep belief networks \cite{rbm1}. This milestone contributed to rekindle interest in deep learning, paving the way for the recent progress in the field that will be discussed in the following sub-section.

\subsection{Recent Progress} \label{sec:dl_recent}
The initial deep learning breakthrough of 2006, seemed to be suggesting that the only possible approach for training deep neural networks was to  initialize them in an unsupervised way. Actually, shortly after Hinton's paper, it was shown  that deep autoencoders were equally able to properly pre-training DNNs \cite{autoencoder}. In the following years, it was gradually more clear that  pre-training step was not strictly necessary to train deep models. In fact, several works showed that simply using ReLU \cite{relu} activations (more effective to fight vanishing gradients)  with a proper weight initialization \cite{xavier}, allows one to avoid any pre-training step.

Moreover, noteworthy progresses have then been done in sequence-to-sequence learning. Before 2014, it was widely believed that this kind of learning would require labeling of each individual element of the sequence. In 2014, the study of attention mechanisms \cite{attention1} has offered a flexible and powerful framework able to overcome previous limitations, revolutionizing  many fields, including machine translation. The same year,  memory-augmented deep neural networks \cite{memory_augmented} have been successfully trained for learning simple programs from examples of desired behaviors (e.g.,  sort lists).  

Important achievements have also been obtained in the context of reinforcement learning \cite{sutton}, where autonomous agents learn to solve tasks with a trial and error strategy. DeepMind showed that a deep reinforcement learning-based agent can learn to play Atari games, reaching
human or super-human performance \cite{atari}. DeepMind also stunned the world when the AlphaGo system was able to defeat the world champion of Go, an ancient Chinese game with a huge number of possible moves (enormously larger than the combinations of chess) \cite{alpha_go}.  

A key element for the success of the reinforcement learning is the competition between agents. This concept is also exploited in Generative Adversarial Networks (GAN) \cite{gan}, proposed by Ian Goodfellow in 2014. The idea is to promote a competition between a network able to generate samples (generator) and another network (discriminator) able to discriminate whether such samples are original or drawn by the DNN. This method has recently  succeeded to generate natural data samples of different nature, including images \cite{gan}, videos \cite{gen_video}, text \cite{gen_text},  and speech \cite{gen_speech}. 

The recent progress is also fostered by the numerous toolkits available to the research community, such as Theano \cite{theano}, TensorFlow \cite{tensorflow}, CNTK \cite{CNTK}, MXNet \cite{MXNet}, Caffe \cite{caffe}, and Torch \cite{torch}, just to name a few.

\subsection{Future Challenges} \label{sec:dl_future}
The rapid rise of deep leaning contributed to spread optimism towards this technology. However, despite of such a great enthusiasm, there are still major scientific challenges to address for really reaching higher levels of artificial intelligence \cite{dl_future}. 

A noteworthy challenge is the effective use of unsupervised data. The recent progress mostly involved supervised learning, that have contributed to achieve state-of-the-art performance in numerous fields. The goal of unsupervised learning is to understand the world around us by observation, which is an ability largely missing in current supervised DNNs. Unsupervised learning can contribute to integrate  a common-sense knowledge, that might lead to a less superficial understanding of the world. Human brain, for instance, infers important evidences  only through observations: kids are able to predict the trajectory of an object falling down even without knowing differential equations or the newton's law. Future unsupervised learning should be able to properly guess properties, correlations and connections across concepts without labeled data. Data annotation requires huge human efforts, and the study of proper techniques to use unlabelled data, could also pave the way to the development of intelligent machines without human efforts. In order to understand the world around us, another possibility is to learn by interactions. It would be thus of great interest the evolution of current reinforcement learning techniques, possibly with a proper integration of the three basic learning a modalities (i.e, supervised, unsupervised, reinforced) in a unified learning framework.



Another challenge concerns the learning efficiency. Current solutions, in fact, require much more information than humans to learn. For instance, humans are able to learn a simple concept (such as a cat) with few examples, while a machines can model it only with hundreds or even thousands of samples. In the future, the development of learning algorithms able to better disentangle and model the factors of variability will be of primary interest.  

Moreover, current deep neural networks are able to solve rather efficiently only single tasks that are generally rather limited and specific (like recognizing faces, classifying sounds, playing Atari). Differently to human brain, current technology is not able to simultaneously solve very different problems at the same time.  It would be thus of great interest the development of more effective multi-task strategies capable of better mimic the human brain.

Another challenge for the future is long-life learning. Current systems are first trained and later tested. The idea of long-life learning is to build a never-ending learning system, that continues to learn from the experience and progressively improves its performance. 

Beyond the other challenges, a major achievement would also be the development of a kind of ``theory of intelligence'' that can better steer the development of intelligent machines. The research in the field, in fact, is now solely based on empirical attempts, that are mostly guided by human intuitions. We can compare the current situation of AI with the first attempts done by the Wright brothers to build a ``flying machine''. Their approach was only based on a trial-and-error strategy, without any notion about the physics of aerodynamics, whose knowledge is clearly very helpful to design modern aircrafts.

\chapter{Distant Speech Recognition} \label{cha:dsr}

Recognizing distant speech is a difficult problem, especially in challenging acoustic environments characterized by significant levels of noise and reverberation. Despite the noteworthy progress of the last years, DSR is still a very active research field, since a natural and flexible speech interaction with a distant machine remains far from being achieved.

\begin{figure*}[t!]
   \centering
     \includegraphics[width=1.0\textwidth]{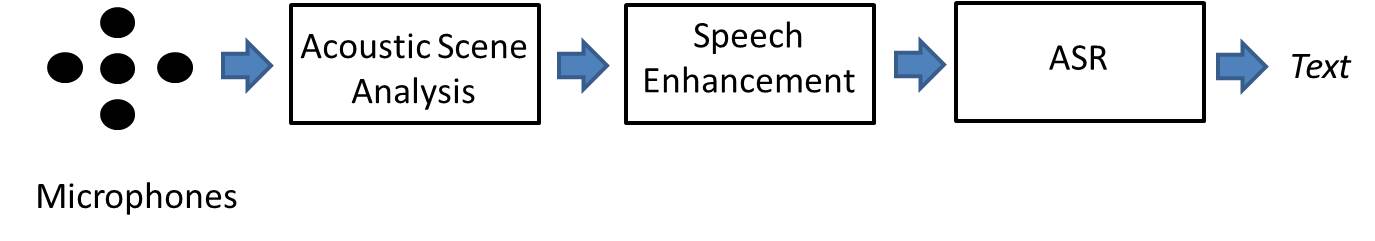}
     \caption{An example of distant-talking interaction with a state-of-the-art speech recognition system.}
     \label{fig:block_diagram}
\end{figure*}

The complexity of this problem normally implies the adoption of complex systems. As shown in Figure  \ref{fig:block_diagram}, state-of-the-art DSR systems are often based on a combination of different modules that have to properly work together \cite{nakatani}. 
The signals recorded by distant microphone arrays can be processed by a front-end \cite{mmprocessing}, that could be composed of both a speech enhancement \cite{speh} and an acoustic scene analysis module. The role of acoustic scene analysis is to provide useful information about the acoustic scenario, while the speech enhancement has the primary goal of improving the quality of the signal recorded by the microphones.  
The resulting enhanced signal feeds a speech recognizer, that tries to identify the sequence of words uttered by the target speaker. 
Note that the scheme of Figure  \ref{fig:block_diagram} represents only one possible implementation of a state-of-the-art DSR system. The order of the components, for instance, might change depending on the specific architecture. For example, the acoustic scene analysis can be performed also after the speech enhancement or, in many cases, can be performed both after and before enhancing the speech.

The following sections will first describe in detail the aforementioned technologies. First, the following section proposes a more detailed description of the main challenges in DSR. After that, a summary of the main state-of-the-art technologies for speech reognition, speech enhancement, and acoustic scene analysis is reported in Sec.  \ref{sec:ASR},  \ref{sec:seh}, and  \ref{sec:asa}, respectively.



\section{Main challenges} \label{sec:challenges}
The signal $y[n]$ recorded by a distant microphone is described by the following equation:

 \begin{equation}
 y[n]=x[n]*h[n]+v[n]
 \label{eq:cont}
 \end{equation}

\begin{figure*}[t!]
   \centering
     \includegraphics[width=0.5\textwidth]{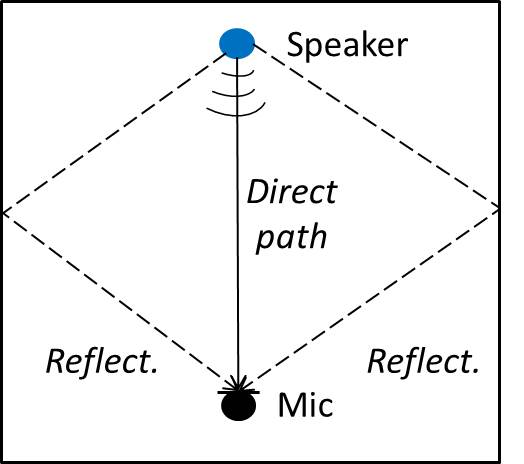}
     \caption{Acoustic reverberation in a typical enclosure.}
     \label{fig:room_rev}
\end{figure*}

The original close-talking speech signal $x[n]$ (i.e, the speech signal before its propagation in the acoustic environment, that is assumed to be a latent variable not directly observed) is  reflected many times by the walls, the floor and the ceiling as well as by the objects within the acoustic environment, as shown in Figure  \ref{fig:room_rev}. 

\begin{figure}[t!]
   \centering

     \includegraphics[width=0.65\textwidth]{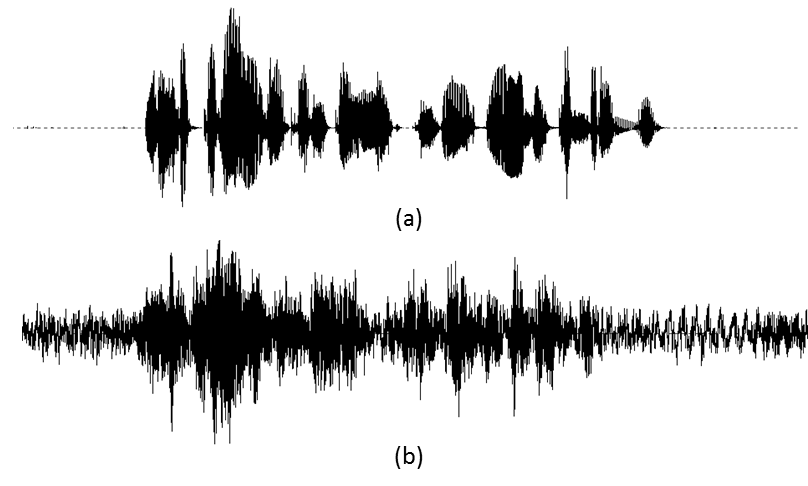}
     \caption{An example of close-talking speech $x[n]$ is depicted in (a), while the same sentence $y[n]$ recorded with a distant microphone in a noisy and reverberant environment is shown in (b).}
     \label{fig:img1}
\end{figure}

Such a multi-path propagation, known as reverberation \cite{kutt}, is represented by a function called impulse response $h[n]$, that is convoluted with $x[n]$. The recorded signal $y[n]$ also includes the contributions $v[n]$ of competitive sources, such as other speakers, telephone ringing, music and other possible interfering background noises. Figure  \ref{fig:img1} shows an example of close-talking speech $x[n]$ with a corresponding distant-talking signal $y[n]$ corrupted by both noise and reverberations, highlighting the deleterious effects of these disturbances. In the following sub-sections, the main characteristics of noise and reverberation in a real DSR application are better discussed.

\subsection{Acoustic reverberation}
\begin{figure}[t!]
\begin{subfigure}{0.50\textwidth}
\includegraphics[scale=0.53,trim={0cm 0cm 0cm 0cm},clip]{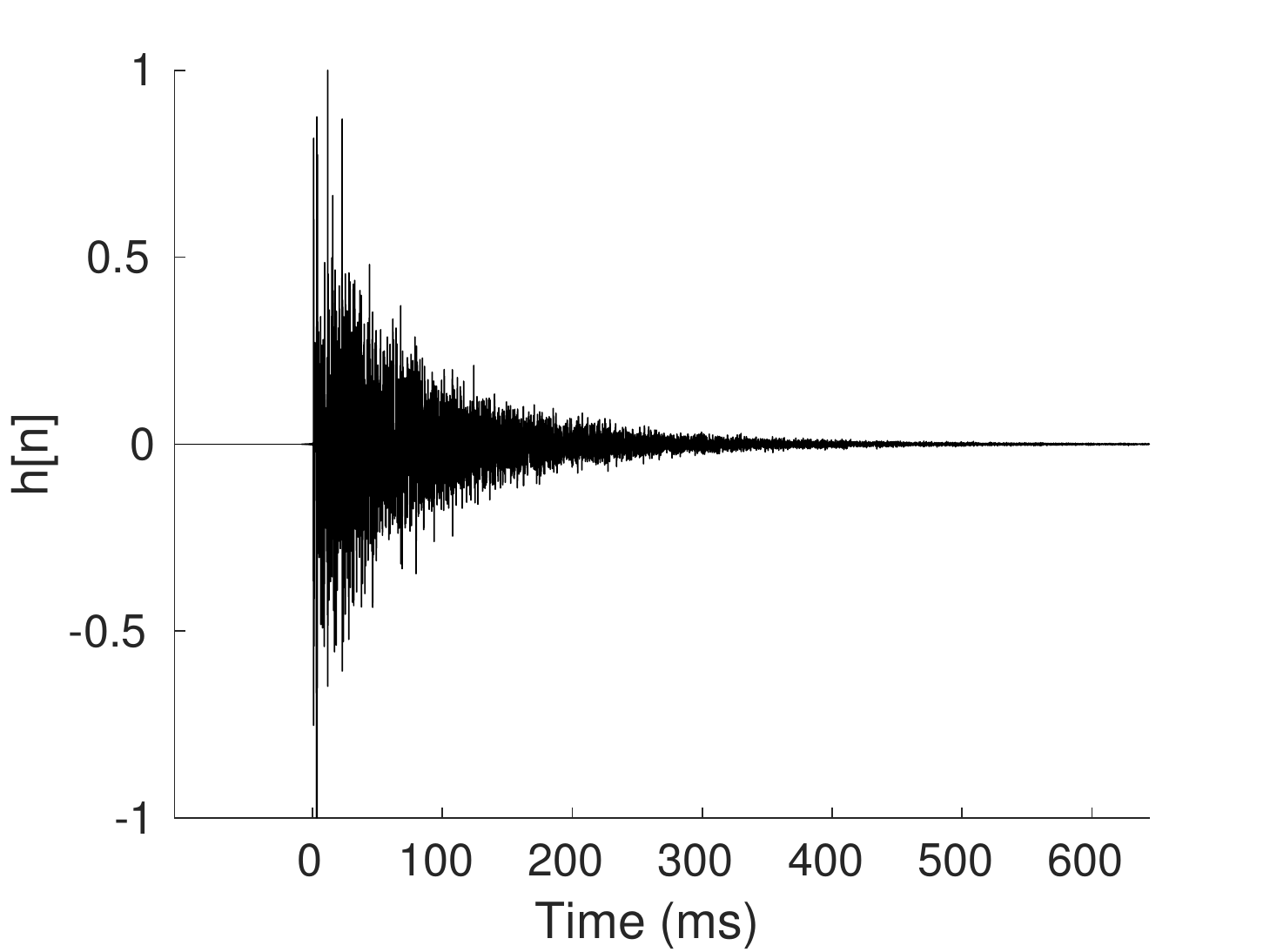}
\caption{Impulse Response $h[n]$}
\label{fig:ir_new}
\end{subfigure} \hspace{0.0\textwidth}
\begin{subfigure}{0.50\textwidth}
\includegraphics[scale=0.53,trim={0cm 0cm 0cm 0cm},clip]{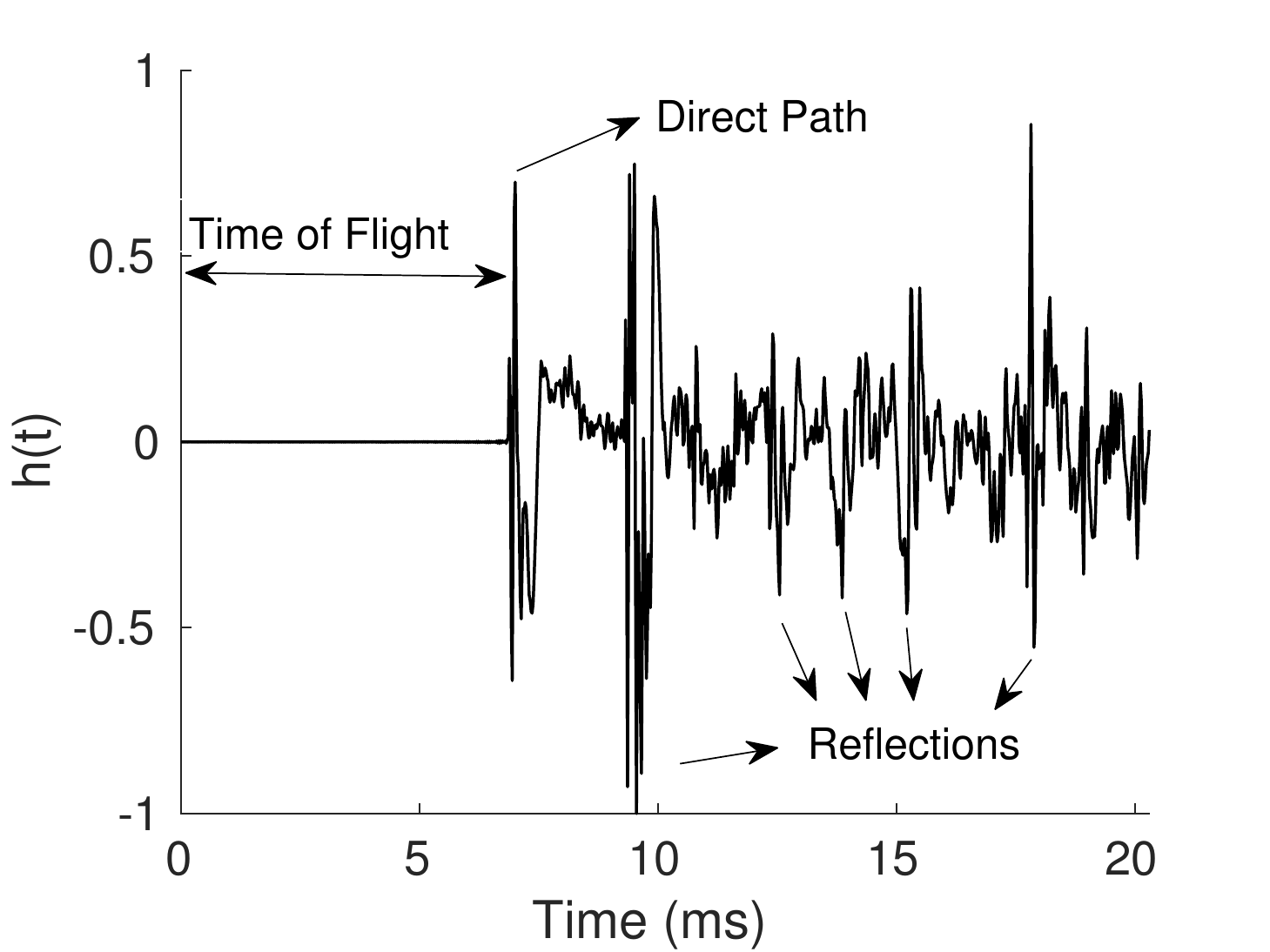}
\caption{Zoom on the first reflections of $h[n]$}
\label{fig:ir_early}
\end{subfigure} \hspace{0.0\textwidth}

\caption{An impulse response measured in a domestic environment with a reverberation time of about 780 ms.}
\label{fig:ir}

\end{figure}

The impulse response $h[n]$ (see Figure  \ref{fig:ir}) can be modeled as a long and causal FIR filter (i.e., $h[n]=0 \quad \forall n<0$), whose taps describe the propagation of the signal in the environment. In particular, if one assumes to deal with a linear time-invariant acoustical transmission system, the impulse response provides a complete description of the changes a sound signal undergoes when it travels from a particular position in space to a given microphone \cite{kutt}.

An impulse response of a typical enclosure is composed of three basic components: the direct sound, early reflections and late reverberation.
The direct sound travels from the sound source to the microphone along a straight line. The time taken by the direct sound to reach the microphone is called Time of Flight (ToF). The early reflections, that arrive at the microphone within approximately 50-100 ms after the direct sound,  are a set of discrete replicas deriving from various reflecting surfaces, such as walls, floor and ceiling. 
Finally, late reflections are based on a dense succession of echoes, where individual contributions can no longer be discriminated. Such late echoes are  more dense in time but characterized by a diminishing energy, due to the attenuation derived from longer paths.
The combination of these three contributions originates the typical exponential decay of the impulse response, that can be appreciated in Figure  \ref{fig:ir_new}.

Acoustic reverberation significantly impairs the intelligibility and the quality of a speech signal, that become more difficult to interpret by both humans and, even to a greater extent, by automatic speech recognizers. Although the global effects of this disturbance can be summarized with a linear filter,  modeling the effects of reverberation in a statistical way or even removing it for the recorded signal is difficult for several reasons. First of all, reverberation introduces long-term effects on the speech signal, originating impulse responses with a large number of filter coefficients, that are difficult to be precisely estimated in a blind fashion. Secondly, the impulse response is a non-stationary function that might change substantially depending on several factors, including the room geometry, the presence of objects in the acoustic environment, position, source/microphone directivity, source/microphone frequency response, temperature, humidity, and air flow, just to name a few.

\begin{figure}[t!]
   \centering
     \includegraphics[width=0.60\textwidth]{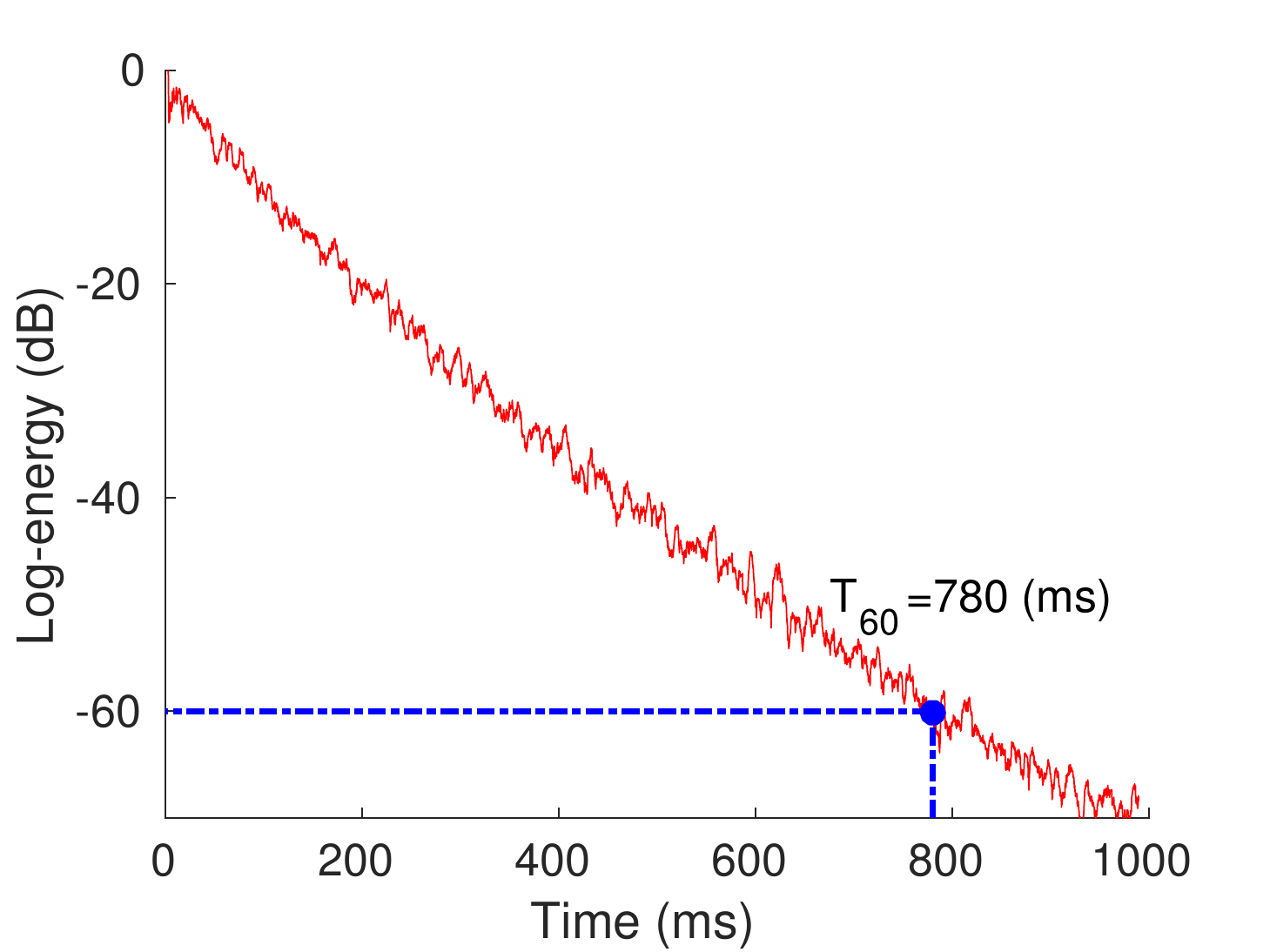}
     \caption{Log-energy decay of $h[n]$}
     \label{fig:ir_en}
\end{figure}

The amount of reverberation characterizing an acoustic enclosure can be estimated with some popular metrics.  One of the most representative measures is the reverberation time $T_{60}$, that is defined as the time required for a sound to decay 60 dB from its initial energy level. Although $T_{60}$ varies significantly depending on the room acoustics, in standard enclosures it typically ranges between 250 ms and 850 ms. 
Figure  \ref{fig:ir_en} shows the log-energy decay obtained with the impulse response of Figure  \ref{fig:ir}. In this case, the energy decay of $h[n]$ indicates that $T_{60}$ is about 780 ms.  Another important measure is the Direct-to-Reverberant Ratio (DRR), that is the energy ratio between the direct and reverberant components of the impulse response. 

A direct measurement of the room characteristics starting from distant-talking speech signals is normally not possible, and several solutions have been proposed in the literature to blindly estimate both $T_{60}$ and DRR \cite{ace}.
The estimation of these parameters can be very useful for DSR systems, since it provides valuable information to speech recognition and enhancement algorithms. For instance, these measures can be used to inform dereverberation algorithms, to select a proper microphone in the environment or to choose a suitable acoustic model, as will  be discussed in the following of the thesis.

\subsection{Additive noise}
The recorded signal $y[n]$ also includes the contributions $v[n]$ of additive interfering  noises (e.g., music, fans, other speakers). 

Additive noise can be classified into different categories.  According to its time evolution, the noise can be continuous (e.g., fans), intermittent (e.g, telephone ringing) or impulsive (e.g, a door knock).  It can also be classified according to its frequency characteristics (white, pink, blue, etc.)  or  according to its statistical properties (gaussian, poisson, etc.). Moreover, the noise can be stationary or non-stationary, depending on the time evolution of its statistical features.  In a realistic environment, $v[n]$ might be composed of  several noise sources of different types that are simultaneously active, making this term very difficult to model or remove from $y[n]$.

The amount of noise is normally measured with the Signal-to-Noise Ratio (SNR). Similarly to reverberation-related measures, several approaches have been proposed to estimate SNR \cite{dsrbook}, and its knowledge can be useful at the different stages of  a distant speech recognition system.

\section{Speech Recognition} \label{sec:ASR}
\begin{figure*}[t!]
   \centering
     \includegraphics[width=1.0\textwidth]{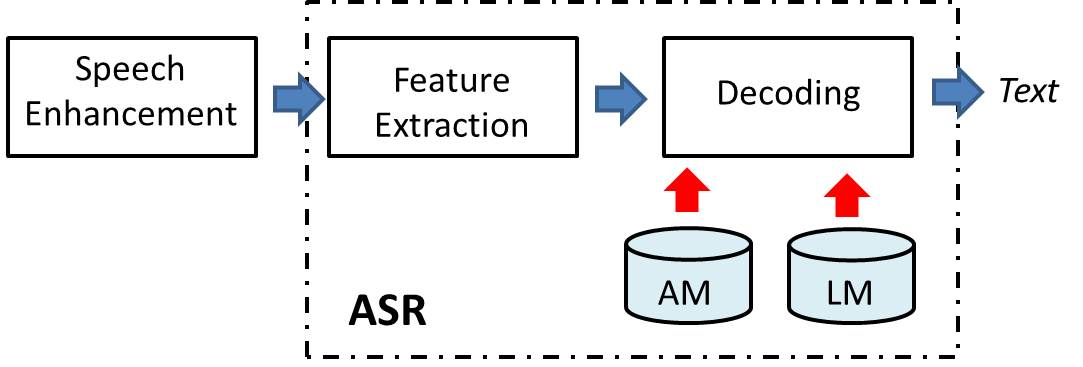}
     \caption{A block diagram highlighting the main components involved in a speech recognizer.}
     \label{fig:ASR}
\end{figure*}

Speech recognition represents the core technology of a DSR system. Its goal is to convert a speech signal (possibly processed by an enhancement module) into the sequence of words uttered by the speaker. As shown in Figure  \ref{fig:ASR}, the speech recognizer is composed of several interconnected modules, that performs feature extraction, acoustic and language modeling as well decoding. 
The following sub-sections will describe these basic components of ASR. The section will then continue with an overview of the most popular techniques for achieving robustness (Sec. \ref{sec:rob}) and with a discussion of modern end-to-end systems (Sec. \ref{sec:e2e}). Finally, a summary of the ASR history is proposed in Sec. \ref{sec:hist_asr}. 

\subsection{Problem Formulation} \label{sec:ASR_pb}
The problem of speech recognition consists in finding the most likely sequence of words $\hat{w}$ given the features $y$ extracted from the signal $y[n]$ \cite{dsrbook}.
More formally, the problem can be formulated using the maximum a posteriori
(MAP) estimate as follows:

\begin{equation}
 \hat{w}=\argmax_w P(w/y)
 \label{eq:eq1}
 \end{equation}
 using the Bayes' theorem, the previous equation can be rewritten as:
 \begin{align}
 \hat{w}&=\argmax_w \frac{P(y/w)\cdot P(w)}{P(y)} \\
        &=\argmax_w P(y/w)\cdot P(w)
 \label{eq:eq2}   
 \end{align}

The likelihood $P(y/w)$ is determined by an acoustic model, while $P(w)$ is determined by a language model. The information provided by these models is integrated in a search graph, that is decoded to estimate the sequence of words uttered by the speaker. 

In the following, the individual components of a speech recognizer will be discussed.

\subsection{Feature Extraction}  \label{sec:fex}
Feature extraction aims to represent the speech waveform $y[n]$ with a reduced set of parameters $y$, while preserving most of the information needed to discriminate the spoken units and the speech characteristics. Although many different features were proposed in the literature, the most popular are the Mel-Cepstral Coefficients (MFCCs) \cite{mfcc}, that attempt to incorporate concepts from the human auditory processing and perception.
MFCCs are commonly derived by processing the speech signal in the following way:

\begin{itemize}
\item Split the signal into frames (usually of 20-25 ms with 10 ms of overlap) using a windowing function (e.g., a Hamming window). 
\item Compute the spectrum of each frame with the Fourier transform.
\item Apply the mel-filterbank to the power spectra. A mel filterbank consists of triangular overlapping windows that are spread over the whole frequency range. To mimic the non-linear human ear perception of sound, these filters are more discriminative at lower frequencies and less discriminative at higher frequencies. 
\item Take the logs of each of filter output. The features processed in this way are called in the literature FBANKs and are  often used as input parameters for DNNs.
\item Take the discrete cosine transform of the FBANK features.
\item Select a subset (usually 13) of the transformed features.
\end{itemize}

\begin{figure}[t!]
   \centering
     \includegraphics[width=1.00\textwidth,trim={4cm 3cm 2.5cm 1cm},clip]{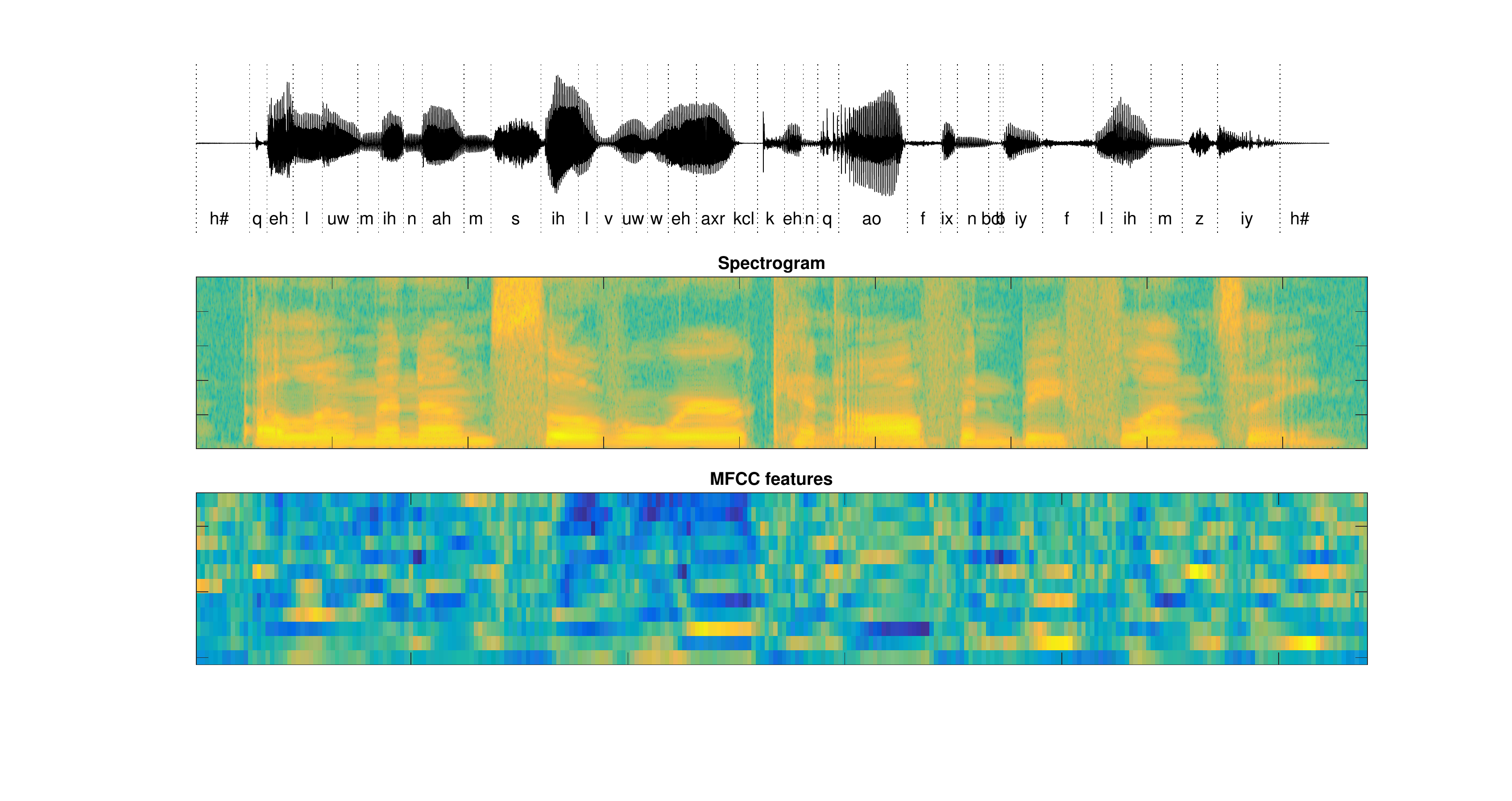}
     \caption{Spectrogram and MFCC features (without log-energy) for a phonetically-rich sentence of the TIMIT dataset.}
     \label{fig:fea}
\end{figure}

An example of MFCC features is reported in Figure  \ref{fig:fea}. Several alternatives have been proposed in the literature \cite{maganti}, including Perceptual linear prediction (PLP) \cite{plp} and biologically-inspired spectro-temporal features such as Gabor \cite{gab1,gab2} or gammatone features \cite{gammatone}. Pitch and Probability of Voicing (PoV) are also often used as additional parameters \cite{IEEEexample:pitch1,IEEEexample:pitch2,kpitch}, that resulted helpful for improving the ASR performance.

Ideally, acoustic features used for ASR should be independent of the specific characteristics of a given speaker. With this goal, several feature normalization techniques such as Cepstral Mean and Variance Normalisation (CMVN), Vocal Tract Length Normalization (VTLN) \cite{vtln} or Feature space Maximum Likelihood Linear Regression (fMLLR) \cite{mllr} have been  proposed. A powerful approach to obtain robust models is to combine standard ASR features with i-vectors, that can be used to inform the ASR system about speaker identity or environmental characteristics \cite{spk_id_ivect}. 
Beyond speaker normalization purposes, a feature transformation can be employed to perform dimensionality reduction or to decorrelate the information. Examples of these transformations are LDA, HLDA \cite{brugnara_hlda} and PCA. More sophisticated solutions are based on non-linear transformations achieved by neural networks using, for instance, TANDEM \cite{IEEEexample:tandem1} and Bottleneck architectures \cite{IEEEexample:bn1}.

As outlined before, several research efforts have been devoted in the past to properly design robust acoustic features for ASR \cite{fea_survey}. Deep learning, however, is drastically changing the way feature extraction is approached \cite{lideng}. In the context of deep learning, in fact, the acoustic features can be automatically learned from data, without the need of human efforts and hand-crafted features. Differently to past ASR systems, the trend is thus to feed the neural network with simple low-level speech representations (such as FBANKs features), leaving the DNN to freely extract higher level parameters.  Some noteworthy attempts have also shown that it is possible to directly feed a neural network with raw speech samples \cite{raw,raw2,joint7}. These recent works have consistently shown that the first layer of a CNN is able to derive a set of spectro-temporal  filters  similar to that used for Gabor feature extraction.

\subsection{Acoustic Model}  \label{sec:am}

The objective of the acoustic model is to provide a statistical representation of the basic sounds making up speech, which is inferred by properly analyzing many hours of recordings with their corresponding text transcriptions. Modern speech recognizers use more than 1000 hours of annotated speech for training the acoustic models \cite{baidu}. As anticipated in the previous sections, state-of-the-art acoustic models are based on Hidden Markov Models (HMMs) \cite{rabiner}. 

\begin{figure}[t!]
   \centering
     \includegraphics[width=1.00\textwidth,trim={0cm 0cm 0cm 0cm},clip]{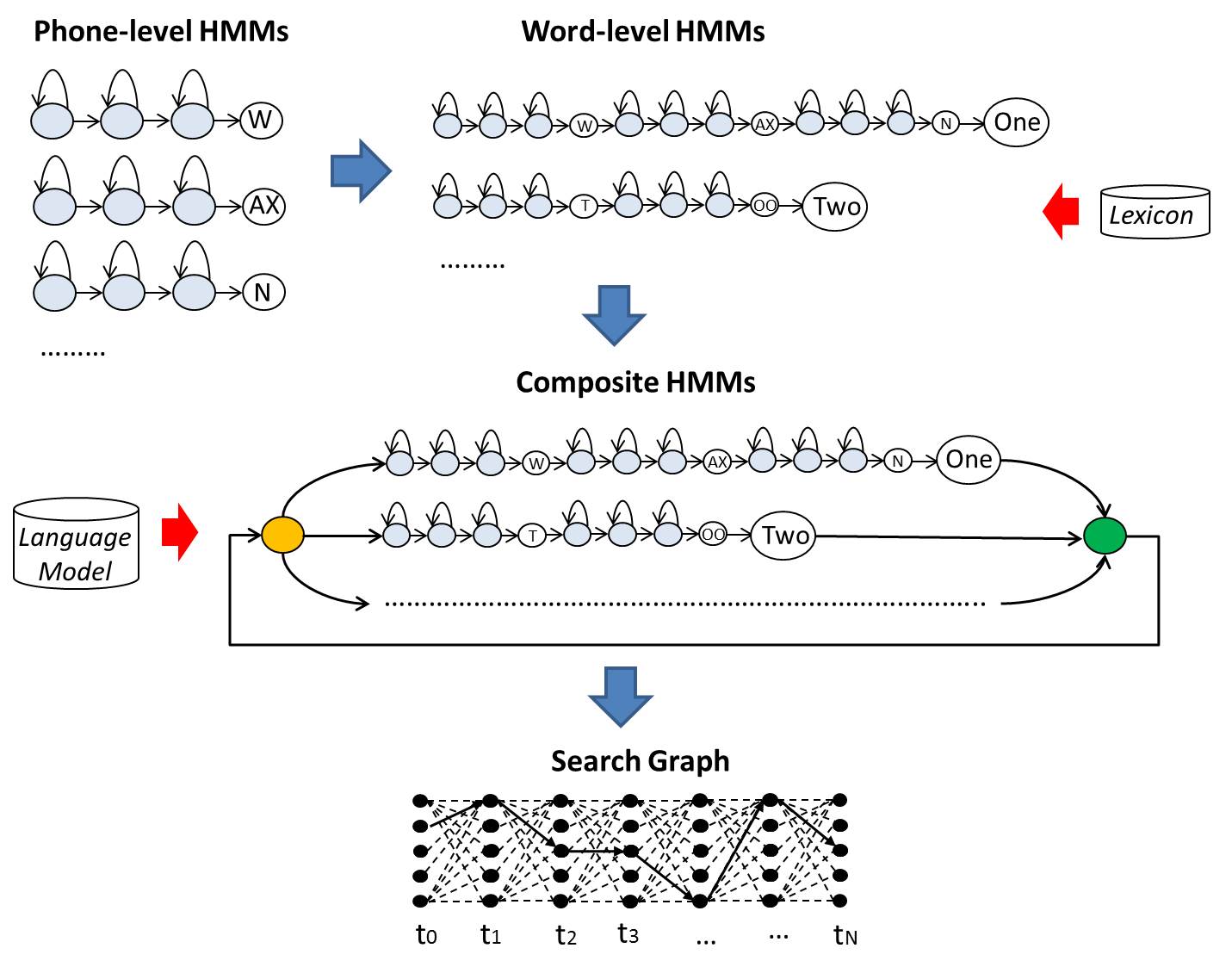}
     \caption{An example of Hidden Markov Models (HMMs) for speech recognition.}
     \label{fig:hmm}
\end{figure}

HMMs are a popular generative models for representing probability distributions  over sequences of observations, that provide a rather straightforward way to model speech. Within the HMM framework, the basic sounds of the speech can be modeled by a set of simple HMMs. As shown in Figure  \ref{fig:hmm}, each basic HMM can describe a particular phone and is generally based on a short sequence of states (normally three or five), that might represent the evolution in time of the considered phoneme. This allows the acoustic model to separately model beginning, central and last parts of each basic sound. Each model is characterized by observation and transition probabilities. The transition probabilities model the duration of each state, while  observation probabilities describe how likely a certain phone unit has been generated by the current HMM state. 


Modern acoustic models actually operate at context-dependent level, where instead of using phones as basic speech units, more articulated fundamental sounds considering also the surrounding phones are adopted. The definition of proper context-dependent phone states is normally based on a Phonetic Decision Tree (PDT), that clusters phones based on a priori
linguistic knowledge and/or exploiting acoustic similarities observed from data \cite{acero_book}.

These  basic HMMs are the building blocks for progressively composing more complex models. As shown in Figure  \ref{fig:hmm}, word-level HMMs can be derived by concatenating phone-level HMMs. This operation requires a lexicon, that converts each word $w$ into a corresponding sequence of phones. Although for some languages (e.g., Italian) this conversion is rather straightforward, for many others (e.g, English) it might be vary tricky and requires sophisticated techniques as well as human efforts to derive an accurate lexicon  \cite{lexicon}.

Word-based HMMs can be, in turn, combined to generate a composite HMM, that can model any sequence of words in the vocabulary thanks to the feedback connections. As will be discussed in Sec. \ref{sec:lm}, the composite HMM can exploit a language model to attribute more weight to the paths leading to likely sequence of words. The obtained HMM is then optimized (by removing, for instance, redundant states) and, starting from it, a search graph is derived to estimate the most likely sequence of words uttered by the speaker, as will be discussed in Sec. \ref{sec:search}.

Depending on the paradigm used for computing the observation probabilities, we can distinguish between GMM or DNN-HMMs. Modern speech recognizers are based on the latter framework and estimate these probabilities with a DNN.  DNN-HMM speech recognizers are often called \textit{hybrid} systems, since they are based on both a generative (HMM) and a discriminative (DNN) model, offering a significant performance gain over GMM-HMM solutions due to the following reasons:
\begin{itemize}
\item \textbf{Better generalization}; DNNs can generalize much better, since they jointly model distributions of different classes. In GMMs, each state is modeled separately with an independent set of GMMs, limiting the possibilities to share correlation across similar states.

\item \textbf{Ability to manage long contexts}; Differently to GMMs, that can effectively handle feature spaces with a reduced dimensionality, DNNs can naturally manage large input spaces. This allows the DNN to account for longer time contexts, that are crucial to improve the system performance. 

\item \textbf{Ability to exploit correlated features}; Within the HMM-GMM framework, diagonal covariance matrices are often considered to train a compact model. However, this assumption forces the system to adopt uncorrelated features (such as MFCCs), since correlation across different dimensions cannot be modeled with simple diagonal covariance matrices. Differently, DNNs are also compatible with correlated inputs and offer the flexibility to effectively exploit features of different nature (such a i-vectors and FBANK features).
\end{itemize}

\begin{algorithm}[t!]
\caption{Kaldi s5 Recipe for DNN Training}
\label{alg}
\begin{algorithmic}[1]
\State Train a GMM-HMM monophone model (\textit{mono})
\State Align the training sentences with the \textit{mono} model
\State Derive a Phonetic Decision Tree and define the context-dependent states $s$
\State Initialize the new model with the mono model
\State Train a GMM-HMM triphone model (\textit{tri1})
\State Align the training sentences with the \textit{tri1} model
\State Initialize the new model with the \textit{tri1} model
\State Train a GMM-HMM triphone model with LDA (\textit{tri2})
\State Align the training sentences with the \textit{tri2} model
\State Initialize the new model with the \textit{tri3} model
\State Train a GMM-HMM triphone model with LDA+fMLLr (\textit{tri3})
\State Align the training sentences with the \textit{tri3} model to obtain DNN labels
\State Train a DNN-HMM
\end{algorithmic}
\end{algorithm}

The training of a GMM-HMM ASR system is carried out with the Baum-Welch algorithm \cite{baum}, that is an example of expectation-maximisation (EM) approach \cite{em}. In the context of modern DNN-HMM systems, instead, the training pipeline is more complex and usually requires the prior training of a GMM-HMM model for deriving the labels needed for DNN training. Such a complexity can be clearly noticed in the standard Kaldi recipe \cite{kaldi} reported in Alg. \ref{alg}, where a large number of incremental steps are needed to train a state-of-the art speech recognizer.



\subsection{Language Model}  \label{sec:lm}
The language model $P(w)$ is learned by means of large text corpora and has the key goal of considering that some sequences of words (e.g., ``open the door'') are more  likely than others (e.g., ``open the dog''). The most popular language models are based on $n$-grams, which gather statistics by counting the occurrences of contiguous sequences of $n$ words.
Typically, this statistics are adjusted by smoothing techniques, such as the Kneser-Ney method \cite{ney}. Depending on the number $n$ of previous words considered to estimate the current one, bigram, trigram, four-gram or even five-grams LMs can be trained.

An alternative to n-gram language models is offered by neural language models \cite{bengio_lm,mikolov}, that learn distributed word representations to counteract the curse of dimensionality problem. Despite the effectiveness of neural language models, their efficient integration into the search algorithm is still an open issue.  Due to this limitation, neural language models can effectively be used only for lattice rescoring after a first decoding step, as discuss more in detail in the following subsection. 

\subsection{Search \& Decoding} \label{sec:search}
Once the acoustic and language models are trained, the decoding procedure allows the speech recognizer to estimate the sequence of words $\hat{w}$ uttered by the speaker.
The information embedded in the acoustic and language models are integrated in a search graph, where each path starting from the initial to the final state represents a particular sequence of words \cite{acero_book}. Among all the possible alternatives, the goal of the decoder is to find the most likely one, that corresponds to the most probable sequence of words. 

A naive solution would be to search for all the possible paths and choose the one with the highest likelihood. This approach, however, is unfeasible especially for large vocabulary ASR.  For a vocabulary of size V words and a sequence of M words, in fact, there are $V^{M}$ alternatives to evaluate. For this reason, search is typically based on the Viterbi algorithm, that is an efficient recursive algorithm based on dynamic programming \cite{viterbi}.  The Viterbi algorithm is able to perform an optimal exhaustive search and only needs to memorize the most probable path at each state, significantly saving both computations and memory. For large vocabulary applications, however, even the straightforward application of  Viterbi search leads to an excessive computational complexity \cite{rabiner}. 

To make decoding computationally tractable, beamsearch methods are  often applied \cite{acero_book}, which define some heuristics to prune non promising paths. Another way to make search more efficient is to save computations by sharing word prefixes inside the search graph. Multi-pass search \cite{gales_book} can also be used for reducing the computational complexity by progressively using  more detailed models. For instance, a first decoding step can be performed with a bigram language model, followed by a second pass based on a more precise trigram LM. In the latter case, the first decoding step outputs a list of the N best alternatives, that can be summarized in a word lattice. The word lattice can be efficiently rescored with a more precise language model, that can also be a neural language model based on recurrent neural networks \cite{mikolov}.  Modern decoders for speech recognition are implemented with Weighted Finite State Transducers (WFST) \cite{mohri}, that offer a flexible and efficient framework to implement the aforementioned decoder functionalities. 

\subsection{Robust ASR}  \label{sec:rob}
The acoustic model is a crucial component of the speech recognizer,  whose accuracy has a significant impact on the final ASR performance. The development of effective acoustic models, however, is a challenging problem, due to the required independence from many variability factors, including speaker accents, speaking rates, hesitations and spontaneous speech. In addition to these issues, a DSR system  should also be robust against noise and reverberation.  As will be discussed in Sec.  \ref{sec:seh}, robustness in adverse acoustic environments can be in part achieved with the speech enhancement module. However, since even very advanced speech enhancement solutions are not able to completely neutralize the unwanted disturbances, robustness is also required to the acoustic model of a speech recognizer.
The most effective DSR systems, in fact, are based on a robust speech enhancement module that attempts to minimize the harmful effect of noise and reverberation, followed by an acoustic model that tries to statistically model the residual disturbance not removed by the front-end \cite{nakatani}.

A number of techniques have been proposed in the literature to derive robust acoustic models. One of the most effective and straightforward approaches consists in training training the ASR system with many data. A popular approach is multi-style training \cite{mstyke,cont2,cont3}, in which the acoustic models are trained with data derived from many  environments characterized by different noisy and reverberant conditions. Since the open availability of large corpora is still an issue, popular approaches are based on transforming existing corpora through data augmentation \cite{dataaug,nakatani,dataaug2,dataaug3} or contaminated speech methods \cite{matassoni,cont2,cont3,revch_full,chime,rav_in14,brutti}, that will be discussed in detail in Chapter \ref{sec:cont}.



Another popular approach is to improve the robustness of a speech recognizer with acoustic model adaptation, that can be supervised or unsupervised depending on the availability of annotations in the adaptation corpus. In the context of GMM-HMM systems, MAP or MLLR \cite{mllr} adaptation gained particular attention in the past. For modern DNNs, various techniques have been proposed \cite{adapt_dnn1}. The simplest solution is to adapt the DNN model by performing some additional retraining iterations. However, when little adaptation data are available, the latter approach can be prone to overfit the adaptation corpus. To mitigate this issue, a possible solution is to add a regularization term in the DNN cost function, such as the  Kullaback-Leiber divergence.  The regularizer forces the adapted distribution to stay close to the original one \cite{adapt_dnn_kl}. Particular attention has also been devoted to unsupervised adaptation. Typically, this kind of adaptation  is carried out  by performing a first decoding step with non-adapted acoustic model, that provides a rough transcriptions of the speech. The obtained transcription is used to  derive  an adapted acoustic model. The effectiveness of this approach depends on the quality of the initial transcription,  that might contain several recognition errors potentially able to impair the effectiveness of the adaptation strategy. An effective approach studied in  \cite{adapt_dnn_falavi} consists in properly filtering the first-step transcriptions according to sentence-level ASR confidence measures.   

Other effective techniques for improving acoustic modeling are the methods for sequence discriminative training, that have the main goal of training the system with metrics like maximum mutual information (MMI),  boosted MMI (BMMI), minimum phone error (MPE) or minimum Bayes risk
(MBR) that have been proved more robust than traditional ones (e.g., log-likelihood, cross-entropy) \cite{acero_book,lideng,sequence_training}.

When multiple recognition systems are available, it could also be convenient to combine the hypothesized word outputs and select the best scoring word sequence. With this goal, a popular technique is ROVER \cite{rover}, that adopts a voting solution implemented with dynamic programming to produce the final combined output.

\subsection{Towards end-to-end ASR} \label{sec:e2e}
As discussed in the previous sections, deep learning has recently contributed to replace GMMs with DNNs inside the HMM framework. 
Following this trend, it is easy to predict that HMMs will be the next element of the ASR pipeline that will be replaced by deep learning. Despite their effectiveness, HMMs have, in fact, some important limitations \cite{hmm_limitations}. First of all, they rely on  the first order Markov assumption, which  states  that  the probability  of  being  in  a  given  state  at  time  $t$  only  depends  on  the  state  at  time  $t-1$. This  is a strong assumption that doesn't represent well speech signals, whose  dependencies  might extend through several states. Moreover, HMMs assume     that    successive    observations    are  independent, while consecutive features are clearly highly correlated in a speech signal. 

Modern RNNs promise to overcome these issues. HMM-free speech recognizers have been, in fact, recently proposed under the name of end-to-end systems.  End-to-end speech recognizers are not only potentially able to overcome the limitations of HMMs, but also aim to avoid any human effort in the design of the speech recognizer, trying to learn everything  in a discriminative way from (large) datasets. 

Popular end-to-end techniques are attention models \cite{attention1} and Connectionist Temporal Classification (CTC) \cite{CTC_graves}.
Attention models are based on an encoder-decoder architecture coupled with an attention mechanism \cite{attention1} that decides which input information to analyze at each decoding step.
CTC \cite{CTC_graves} is based on a DNN predicting symbols from a predefined alphabet (characters, phones, words) to which an extra unit (\textit{blank}) that emits no labels is added.
Similarly to HMMs, dynamic programming is used to sum over all the paths that are possible realizations of the ground-truth label sequence to compute its likelihood and the corresponding gradient with respect to the neural network parameters. This way, CTC allows one to optimize the likelihood of the desired output sequence directly, without the need for an explicit label alignment.

Despite some promising results recently achieved with huge dataset by Baidu \cite{baidu}, the ASR performance of end-to-end systems is generally worse than that currently achievable with state-of-the-art hybrid systems. The former methods, in fact, are relatively young models, and still need to progressively improve to really compete with a more mature DNN-HMM hybrid technology \cite{miao}.

\subsection{Brief History} \label{sec:hist_asr}
The idea of building machines able to recognize speech has fascinated people for long time. The appeal of speech technologies is, after all, also witnessed by the numerous movies, such as ``\textit{2001: A Space Odyssey}” and the ``\textit{Star Wars}" saga,  showing advanced robots or computer naturally interacting with human beings. 

Researchers started addressing the problem of speech recognition in the early 50s. The first attempts were based on template matching approaches, whose core idea is to compare a low-level speech representations with a set of predefined patterns. An example is the pioneering system developed in 1952 by 
Davis, Biddulph, and Balashek at Bell Laboratories \cite{ASR_tm1}. The system was able to recognize isolated digits from a single speaker, using the formant frequencies computed during vowel regions. The obtained formant trajectories served as the “reference pattern” for determining the identity of an unknown digit. Another system using a similar technology was described in \cite{ASR_tm2}, where a speech recognizer able to classify 10 syllables of a single talker was proposed. The work described in \cite{ASR_tm3} was, instead, the first speaker-independent vowel recognizer. 

In the 60s, the concept of adopting non-uniform 
time scale for aligning speech patterns started to gain interest. For instance, Vintsyuk \cite{ASR_dtw} proposed the use of dynamic programming for deriving more robust similarity measures using time alignment between two utterances. This initial work was followed by the studies of Sakoe and Chiba \cite{ASR_dtw2}, that proposed more formal methods, known as dynamic time warping, for speech pattern matching. 

In the early 70s, a significant achievement was the development of Linear Predictive Coding (LPC) \cite{lpc}, that resulted important to derive robust features for recognition performance. Another important milestone was the development of the Hearsay system by Carnegie Mellon University (CMU), that was the first system exploiting beam search for speech recognition.

During 80s, there was a progressive decline of pattern matching methods in favor of more robust statistical approaches \cite{RabinerJuang93}, that were based on a statistical description of the speech signal at both acoustic and linguistic levels. This trend, initially promoted by both Bell Labs and IBM, is well summarized by the Mercer's famous comment: ``\textit{There is no data like more data}", or by the Fred Jelinek's aphorisms : ``\textit{Every time I fire a linguist, the performance of the speech recognizer goes up.}". Hidden Markov Models (HMMs), studied by Baum at the Princeton University in the early 70s \cite{baum}, emerged as the dominant paradigm for statistical ASR. HMMs, which are still used in state-of-the-art speech recognizers, drastically revolutionize ASR due to some important peculiarities \cite{rabiner}, including the availability of efficient training and inference algorithms based on dynamic programming \cite{baum}, the flexibility in merging statistical information deriving from acoustic, lexicon and language models, as well as the robustness in handling acoustic variabilities of speech. 

In the 90s, the progressive evolution of this framework, led to a gradual maturation of the technology \cite{acero_book}, which in some ways lasts to this day. The main achievements were the evolution from discrete to continuous HMMs as well as  the development of context-dependent speech recognizers able to significantly outperform context-dependent HMMs \cite{acero_book}. 

For more than three decades, the dominant approach was based on HMMs coupled with Mixture Gaussian Models (GMMs). 
Starting from 2012, the rise of deep learning started to revolutionized speech recognition \cite{lideng} by replacing GMMs with DNNs \cite{dahl2012context}. DNN-HMMs speech recognizers contributed to significantly outperform previous systems, laying the foundations for a new generation of speech recognizers. Actually, the first attempts to exploit the discriminative power of DNNs in ASR were done in the early 90's by Boulard and Morgan  \cite{IEEEexample:intro7}, following the connectionism trend introduced in the deep learning chapter. 
Such early works were promising but rather premature, since at that time hardware, data and algorithms were not as mature as today.

The progress in ASR, however, was also fostered in last decades by several other factors, including the public release of speech corpora \cite{timit,wsj,sb,apasci} and the development of ASR toolkits \cite{htkbook,kaldi}. The numerous evaluation campaigns promoted by DARPA and NIST, as well as the challenges such as CHiME \cite{chime,chime3} and REVERB \cite{revch_full} were also of fundamental importance to promote ASR progress and to establish common evaluation frameworks across researchers. Important contributions were also given by some speech-related European Projects, such as AMI/AMIDA \cite{ami}, DICIT \cite{dicit_1} and DIRHA. Last but not least, the renewed interest in ASR has recently encouraged huge investments in the field by several big companies (such as Google, Microsoft, IBM, Amazon, Nuance, Baidu, Apple), that in many cases actively contribute to the basic research in the field. 

\section{Speech Enhancement} \label{sec:seh}
Speech enhancement to counteract the effects produced by environmental noise and reverberation was studied for decades, targeting not only speech recognition but also different application fields, ranging from hearing aids to hands-free teleconferencing. Many methods are described in the related literature both for single and for multi-channel input \cite{Benesty_2014,speh}. Common speech enhancement has the primary goal of improving the quality at perceptual level of the recorded signal $y[n]$, trying to minimize the deleterious effects of noises $v[n]$ and reverberation $h[n]$. 
In the following sections, the most popular speech enhancement techniques (e.g., spatial filtering, spectral subtraction, DNN-based speech enhancement) and problems (e.g., dereverberation, source separation, AEC, microphone selection) are discussed.

\subsection{Spatial Filtering} \label{sec:sfil}
One of the most effective multi-microphone technique of speech enhancement is spatial filtering or beamforming \cite{beam}. The goal of these methods is to obtain spatial selectivity (i.e., privilege the areas where a target user is actually speaking), limiting the effects of both noise $v[n]$ and reverberation $h[n]$. 

A straightforward way to perform spatial filtering is provided by the delay-and-sum beamforming, that simply performs a time alignment followed by a sum of the recorded signals. The time alignment, that is necessary since the target signal reaches the microphones at different time instances, is generally achieved by computing delays with TDOA techniques \cite{KnappCarter}. 
A useful tool to analyze the directional properties of a spatial filtering algorithm is the polar pattern, that highlights the sensitivity of the beamformer for all the possible angles from which the sound might arrive.

\begin{figure}[t!]
   \centering
     \includegraphics[width=0.90\textwidth]{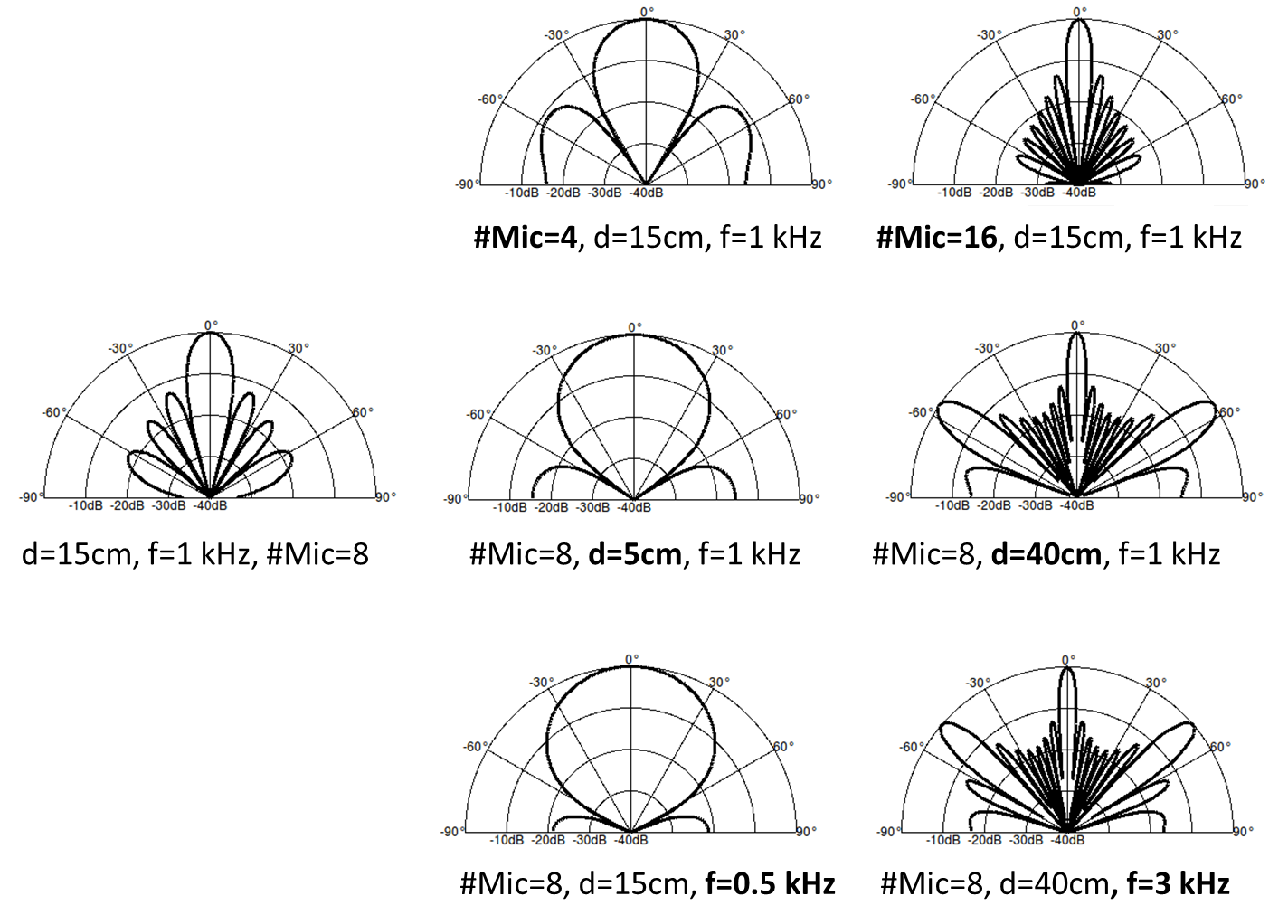}
     \caption{Examples of beampatterns obtained when delay-and-sum beamforming is applied to different linear arrays characterized by different microphone configurations.}
     \label{fig:bp}
\end{figure}

Examples of polar patterns obtained with a delay-and-sum beamforming applied to a linear array are depicted in Figure  \ref{fig:bp}. The figure shows that the number of microphones $\#Mic$, their spacing $d$ as well as the frequency $f$ of the considered signal have an impact on the resulting directivity. In particular, the main lobe tends to become sharper as the number of microphones, the sensor spacing $d$, and the frequency $f$ of the incident signal increase. The microphone spacing $d$, however, should be carefully designed to avoid \textit{spatial aliasing} problems. A linear array, indeed, can detect a plane wave signal at wrong angles if the signal wavelengths are shorter than twice the distance $d$. The spatial aliasing phenomenon causes the appearance of “\textit{grating lobes}” in the delay-and-sum beampattern, that are highlighted in  Figure  \ref{fig:bp} when $d$ and $f$ assume high values. 

Despite its simplicity, delay-and-sum beamforming has some drawbacks. For instance, as shown in Figure  \ref{fig:bp}, it actually generates proper directional patterns only for some specific subbands, and it tends to be less directional  at lower frequency. Moreover, it  introduces spatial aliasing at higher frequency bands.

To mitigate these issues, a more advanced solutions is filter-and-sum beamforming \cite{beam2}, that filters the recorded signals before the time alignment. Another alternative is represented by super-directive beamforming \cite{beam3}, which further enhances the target speech by suppressing the contributions of the noise sources from other directions.

Another limitation of current beamformers is that they usually rely on the assumption that the target speaker is always the dominant source in the environment. Although this can occur in rather quiet environments, such an assumption could not hold anymore in very noisy situations, where the targeted speaker may be weaker than competitive sources.

\subsection{Spectral Subtraction} \label{sec:ss}
Differently from  beamforming, that aims to counteract the effects of both noise and reverberation, some other approaches proposed in the literature try to mitigate just one of these unwanted phenomena. To counteract additive noise, for instance, a popular approach is represented by spectral subtraction algorithms \cite{ss}, that were one of the first algorithms historically proposed for the enhancement of single channel speech.  These techniques reduce the contributions of $v[n]$ by estimating the noise signals when the speaker is not active. The estimated noise spectrum is then subtracted when speech is produced. The main issue concerning these methods is that such techniques are based on the underlying assumption that the additive noise is almost stationary, which is very unrealistic in practical applications. As a result, the processed speech signal is characterized by significant distortions (known as residual or musical noise), that may have an impact on both the signal intelligibility and the speech recognition performance. A difference with beamforming is that most of the spectral subtraction techniques can also operate on a single channel. 

\subsection{DNN-based Speech Enhancement} \label{sec:dnn_se}
The recent rise of deep learning has laid the foundations for the development of   speech enhancement completely based on DNNs. The main difference with the traditional signal processing-based techniques described in previous sub-sections is that the complex non-linear function able to enhance the corrupted speech is not designed by humans, but completely learned from data. 

In the last years, several approaches have been proposed in the literature for dereverberation (i.e., reducing the effect of $h[n]$) \cite{DNN_SE_rev}, denoising (i.e., limiting the effect of $v[n]$) \cite{DNN_SE_noise,DNN_SE_noise2}, and for jointly address both issues \cite{DNN_SE_revnoise}.  The most popular architectures adopted in recent works were feed-forward \cite{dnn_se1,dnn_se2,dnn_se5,dnn_se4}  and recurrent neural networks \cite{dnn_se2,dnn_se6}. The development of DNN-based speech enhancement  has also facilitated an effective integration between this technology with the speech recognition module, as will be better discussed in Chapter \ref{cha:ndnn}. 

Although DNN-based speech enhancement represents a promising research direction, current approaches typically achieve a performance that is still far from the best signal processing-based techniques, especially when addressing both noise and reverberation at the same time. Even though huge corpora reflecting different types of acoustic conditions can be simulated, a  prominent issue concerns the lack of generalization of the former systems, that tend to work well only when test conditions similar to that addressed during training occur. This issue reflects the inherent difficulty to  statistically model the huge variability of noises that might be faced  in a realistic environment. 

\subsection{Speech Dereverberation} \label{sec:derec}
Speech dereverberation  methods are signal processing techniques that aim to limit the harmful effects of acoustic reverberation in a distant-talking speech signal \cite{speechde}. The solutions for reverberation reduction can be divided into many categories (e.g., single vs multiple microphones approaches, feature vs signal domain methods). According to  \cite{speechde}, speech dereverberation algorithms can be categorized depending on whether or not the impulse response $h[n]$ needs to be directly estimated.

The first category consists of techniques, known as blind deconvolution or reverberation cancellation methods \cite{blind_dec}, that are based on an estimate of the impulse response $h[n]$. The estimated impulse response is exploited to reconstruct the anechoic speech $x[n]$ with an inverse-filtering operation. 

The second category gathers approaches known as reverberation suppression methods that do not directly require an explicit estimation of $h[n]$, but exploit the characteristics of the speech signal to mitigate reverberation. The so-called statistical methods (that require the availability of models for speech and noise signals) belong, for instance, to this category. See \cite{speechde} (and the reference therein) for more details about the different approaches proposed in the literature.

In general, speech dereverberation is a particularly challenging problem for several reasons. First, both the speech source and the acoustic channel are unknown, non-stationary and time-varying. As outlined in Sec. \ref{sec:challenges}, in fact, the impulse responses depend on several factors, including source-microphone 3-D positions, orientations and polar patterns. Moreover, although reverberation can be described as a linear FIR filter, the IR is very long and it is difficult to approximate it in a very precise way. Lastly, most of the dereverberation algorithms perform properly in the case of reverberated-only signals, and the presence of additive noise might have a serious impact on the quality of the reconstructed signal.

\subsection{Source Separation} \label{sec:ssep}
Sound Source Separation tackles the problem of segregating a target voice
from background or competing sources. This scenario, often referred to as \textit{``cocktail-party}", is frequent in distant-talking scenarios, where the target speech might be overlapped with some interfering noises. The most popular algorithms of source separation are implemented using independent component analysis \cite{bss} and most of them are completely unsupervised, avoiding the use of any  prior knowledge on the acoustic scene. 
Despite the noteworthy progresses of the last decade and the considerable success of some international challenges  such as CHIME \cite{chime}, a satisfactory separation is still  far from being reached and the enhanced signal can be severely corrupted by artifacts and non-linear distortions. 

\subsection{Acoustic Echo Cancellation} \label{sec:aec}
Although the noises $v[n]$ corrupting the target speech are  in general almost unpredictable, some of these interferences may be directly acquired at their source.  An example is represented by the acoustic signal emitted by a television,  which can be captured before its propagation in the acoustic environment. In this situation, Acoustic Echo Cancellation (AEC) techniques have the specific goal of suppressing the known interference from the signals recorded by the distant microphones.  The most popular approaches achieve this goal by estimating a set of filter parameters with the Least Mean Square (LMS) algorithm. An example is the Subband Acoustic Echo Cancellation (SAEC) \cite{aec}, that approaches the problem using  subband-based solution.  More recently, a Semi-Blind Source Separation (SBSS) paradigm \cite{nesta}, which includes an a priori knowledge of the known interferer as a constraint in the independent component analysis framework, has successfully been proposed.

\subsection{Microphone Selection} \label{sec:msel}
In the case of distributed microphone networks, instead of combining the information of different microphone signals, the signal with better characteristics is obtained through a microphone selection \cite{nadeu}.
These solutions are typically based on distortion measures that aim to rank the channels in a way as close as possible to the unknown performance  of the recognizer.  Several measures have been proposed in the literature.  Examples are scores based on the estimation of the position and the orientation of the speaker \cite{nadeu2}, solutions based on the estimation of the signal to noise ratio (SNR) \cite{nadeu4}, and solutions attempting to estimate some features from the impulse responses, such as the direct-to-reverberant ratio \cite{nadeu3}. More recently, a microphone selection based on the cepstral distance have been proposed \cite{cristina}. The main issue, that make channel selection still a challenging research direction,  concerns the difficulties in devising  distortion measures that are correlated with the ASR performance, especially in real environments characterized by unpredictable noisy and reverberant conditions.

\section{Acoustic Scene Analysis} \label{sec:asa}
Acoustic scene analysis refers to techniques aiming to detect, analyse, and classify the acoustic information diffused in the environment. Acoustic scene analysis is a rather general term that might include a bunch of techniques with several different applications, such as automatic surveillance, smart video conferencing, multi-modal human-computer interaction, hearing-aids technologies \cite{asa_hearing_aids,asa_hearing_aids2}.  This term can also have a different shades of meaning depending on the particular area of research. For instance, Auditory Scene Analysis (ASA) or Computational Auditory Scene Analysis (CASA) \cite{asa_book} typically refer to sound segregation based on perceptually meaningful elements.

Even though several other fields can benefit from an acoustic scene analysis, in this thesis such a technology is only intended to provide additional information to possibly help both the speech recognizer and the speech enhancement systems. To this purpose, acoustic event detection, described in Sec. \ref{sec:aed}, can be used, for instance, to dynamically select an acoustic model more robust against the detected noise. Similarly, the speech enhancement can improve its performance if a hint on the typology of disturbance affecting the recorded signal is provided. Speaker identification techniques, discussed in Sec. \ref{sec:aed}, can be helpful to derive personalized acoustic models, that better match user's voice characteristics. Speaker localization, summarized in  Sec. \ref{sec:sl}, could be instead very precious to guide beamforming algorithms, in order to create spatial selectivity in the direction from where the user is speaking.


\subsection{Acoustic Event Detection and Classification} \label{sec:aed} 

A huge variety of both human and non-human acoustic events can occur in a real situation. Acoustic event detection and classification techniques \cite{aed1} aim at detecting time boundaries of these sounds and categorizing them into classes. 

An important sub-problem consists in  solely detecting the speech activity, distinguishing between speech and non-speech classes only. For close-talking purposes, relatively simple speech activity detection techniques based on energy thresholds or zero-crossing rates work reasonably well. For DSR applications, an automatic segmentation of the distant audio signals into speech and non-speech categories is not only more challenging but also more crucial, since users normally interact with a DSR system in a hands-free modality (i.e., push-to-talk buttons or similar devices are usually not available). Several approaches have been proposed in the literature, including techniques based on periodicity measures \cite{vad2}, statistical models \cite{vad3} or cross-correlation-based approaches \cite{vad4}. Recently, deep learning-based speech activity detection gained particular attention, as witnessed by the numerous works published in the literature \cite{lstm_vad,vad_dnn,vad_dnn2,vad_dnn3,vad_dnn4,vad_dnn5}.

Beyond speech activity detection, a richer classification of the acoustic events can be helpful for improving the DSR performance.
With this purpose, several efforts have been devoted to acoustic event detection in the context of the European project CHIL, with a particular focus on events that can happen in small environments, like lecture and small-meeting rooms \cite{chil_corpus,clear}. The Detection and Classification of Acoustic Scenes and Events (DCASE) and TRECVID Multimedia Event Detection challenges further contribute to fostering the progress in the field.  More recently, Google \cite{google_aed} publicly released a huge dataset that promises to become a benchmark in the field: it is composed of 632 audio event classes and a collection of more that 2 millions human-labeled sound clips drawn from YouTube videos.

Although several solutions have been proposed, the most popular approaches are based on GMM-HMM \cite{aed_hmm}, Support Vector Machines (SVMs) \cite{aed2,aed3} and, more recently, also on DNNs \cite{aed4,ravanelli_eusipco}.

\subsection{Speaker Identification and Verification}  \label{sec:sid} 
Speaker verification consists in validating a user's claimed identity using features extracted from his voice (binary classification). Speaker identification, instead, requires solving a more complex problem, since the system has to explicitly find the right speaker among a set of $N$ alternatives (multi-class classification). 

These technologies can be applied in several fields, including user authentication, surveillance, forensic, security as well as speech recognition.  In a command-and-control application, for instance, speaker verification can be used to enable only a particular user to execute speech commands. Speaker ID can be used to derive personalized acoustic models for each different user, allowing the DSR system to significantly improve its performance.

Research in speaker recognition has  a long history, dated back to more than half a century ago \cite{spk_id_history}. Early approaches were based on template matching \cite{spk_id_pattern_matching,spk_id_pattern_matching2}, progressively followed by methods based on HMMs \cite{spk_id_hmm}, vector quantization \cite{spk_id_vq}, GMMs based  on  Universal Background Model (GMM-UBM) \cite{spk_id_gmm} and SVM \cite{spk_id_svm}. State-of-the-art techniques are based on i-vectors \cite{spk_id_ivect,ivector_laface,ivector_laface2} and DNNs \cite{spk_id_dnn}.


\subsection{Source Localization} \label{sec:sl} 

\begin{figure}[t!]
   \centering
     \includegraphics[width=0.50\textwidth]{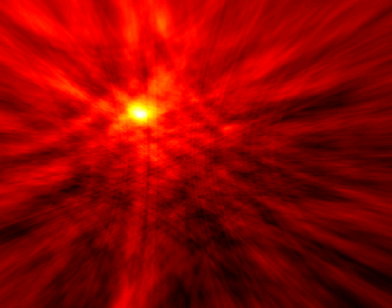}
     \caption{An example of GCF obtained with a circular array composed of six microphones. The position of the speaker is clearly highlighted in the map.}
     \label{fig:gcf}
\end{figure}

Acoustic source localization has the main objective of deriving the spatial coordinates of single or multiple sound sources that are active in a given environment.
These technologies have been deeply investigated and several different approaches are available in the literature. In general, popular algorithms are based on the estimation of the Time Differences Of Arrivals (TDOA) at two or more microphones, from which the source location is inferred by applying geometrical considerations. The Generalized Cross-Correlation Phase Transform (GCC-PHAT)\cite{KnappCarter}, is the most common technique for estimating the TDOA at two microphones and relies on phase information only, that turned out to be much more reliable than magnitude for estimating time delays ~\cite{gcf}.
The TDOA can be exploited to derive acoustic maps, like the Global Coherence Field (GCF)~\cite{DeMori98}, that are particularly effective for accurately localizing acoustic sources as well as to provide important insights on the characteristics of the acoustic environment. An example of a GCF (obtained with a microphone array composed of six microphone) is depicted in Figure  \ref{fig:gcf}, where the position of the speaker is clearly highlighted. Depending on the microphone configuration, speaker position/orientation as well as on the room acoustic, the GCF function can be affected by several artifacts, including the possible presence of the so-called \textit{ghost peaks}, that might, for instance, appear when strong reflections arise.

In the last decade, the research on  speaker localization has led to robust techniques for jointly estimate both speaker position and orientation \cite{brutti2005oriented}. Moreover, techniques for simultaneously localizing multiple acoustic sources \cite{Brutti_EURASIP} and methods for speaker tracking \cite{brutti_tracking} have been explored in the literature. 

In the context of distant-speech recognition, the localization information is typically exploited by the speech enhancement system to steer spatial filtering towards the desired acoustic source, as shown in the speech enhancement section. 

\chapter{Methods for Speech Contamination} \label{sec:cont}
A key ingredient for the success of deep learning is the availability of very large corpora, that can be exploited to train deep neural networks with higher robustness and capacity. Nevertheless, the  open access of so large annotated datasets is still an issue. Huge speech corpora are, in fact, typically collected by big tech companies,  that are reluctant to publicly distribute them.  
This might create a significant gap between industrial and academic research, eventually hindering the progress of the field. 

It is thus of great interest the study of data augmentation approaches \cite{dataaug,nakatani,dataaug2,dataaug3} that can help academic researchers mitigate these limitations. In the field of speech recognition, data augmentation artificially creates new samples by processing available speech sentences. This can be realized, for instance, by perturbing pitch, formants or other speaker characteristics. The process to transform a close-talking signal into a distant-talking one is often called  \textit{data contamination} \cite{matassoni}.  Data contamination is typically carried out by convolving close-talking speech recordings $x[n]$ with impulse responses $h[n]$ and adding some noise sequences $v[n]$, as reported in the following equation:
\begin{equation}
y[n]=x[n] * h[n] + v[n]
\end{equation}

The data generated with this approach are often called \textit{simulated data}. In the literature, some ambiguity between the terms data contamination and multi-style training  \cite{mstyke,cont2,cont3} still exists. In general, the latter  refers to training acoustic models with data coming from different domains, regardless whether they are real or contaminated.

This thesis studied crucial aspects behind the data contamination process. We indeed believe that the realism of simulated corpora and, more importantly, the definition of  common methodologies, algorithms and good practices to generate such data play a crucial role to foster future research in this field and to eventually help researchers better transfer laboratory results into real application scenarios. 
Several ingredients are necessary to generate realistic simulated data, including the quality of both the close-talking recordings and the impulse responses.
Our studies focused in particular on the latter aspect, considering either IRs measured in the real environment, or derived by room acoustic simulators. 
The reference  environment for the experiments reported in this Chapter was a real apartment available for experiments under the \textit{DIRHA project}\footnote{\url{https://dirha.fbk.eu/}}. This apartment, hereinafter referred to as \textit{DIRHA apartment}, was equipped with a microphone network consisting of several microphones. The DIRHA apartment was object of several recording sessions, with the purpose of acquiring real speech datasets and collecting corpora of impulse responses. The reference room for the following experiments was the living-room of the aforementioned apartment, which is characterized by a reverberation time $T_{60}$ of about 750 ms.
See the appendix for a more detailed description of the DIRHA apartment (App. \ref{app:mc-1}, \ref{app:mc-2}). 

The remaining part of this Chapter discusses our main achievements on this topic, summarizing the main findings published in our papers referenced in their respective sections. Sec. \ref{sec:ir_meas} discusses the proposed methodology to generate high-quality simulated data with measured impulse responses. Our approach will then be validated in Sec. \ref{sec:real_vs_sim}, reporting a comparison  between real and simulated data. Sec. \ref{sec:im_method} studies a novel methodology to derive synthetic IRs.
Finally, Sec. \ref{sec:cont_exp} will describe some approaches to better train DNNs with contaminated data.

\section{Measuring IRs with ESS} \label{sec:ir_meas}
IR estimation in an acoustic enclosure is a topic that has been widely discussed in the literature of the last two decades. The early methods were referred to as \emph{Direct}, i.e. methods based on diffusing in the environment a signal of impulsive nature, as a gun shot or a bursting balloon. These approaches were then replaced by \emph{Indirect} ones, characterized by using excitation signals different from the Dirac function, primarily due to the advantage of providing a higher SNR that was not guaranteed by the former one.
In the case of indirect methods, a known excitation signal $s[n]$ of length $L$ is reproduced at a given point (for instance through a loudspeaker), and the corresponding signal $y[n]$ is observed by a microphone placed in another point in space. The related impulse response can be derived by performing a cross-correlation between the known and the recorded signal, as highlighted in the following equation:

\begin{equation}
h[n]=s[n] \star y[n] = \sum_{l=0}^{L-1} s[l]\cdot y[l+n]
\end{equation}

The IR measurement process is affected by environmental noise and non-linearities that may be introduced by instrumentation. In particular, a crucial aspect is the robustness against harmonic distortions, that are non-linear artifacts arising when the loudspeaker does not work in a perfectly linear input-out regime. These artifacts are very common when measuring IRs. In fact, to achieve a high SNR, the monitor should emit a very loud signal. High volumes, however, typically force the loudspeaker to work in a point of the transfer function closer to saturation, where the input-output relation is not anymore perfectly linear, generating harmonic tones that severely degrade the quality of the measurement. 
In the category of indirect methods, some of the most commonly used state-of-the-art techniques are \emph{Maximum Length Sequence} (MLS) \cite{mls}, \emph{Linear Sine Sweep} (LSS) and, more recently, \emph{Exponential Sine Sweep} (ESS) \cite{farina}, that are signals emitted into the acoustic environment with a duration $L$ and a volume $V$. The aforementioned methods  have a different degree of robustness against noise and harmonic distortions, as described in the following:

\begin{figure}[t!]
\centering
\begin{subfigure}[b]{.35\linewidth}
\includegraphics[width=\linewidth]{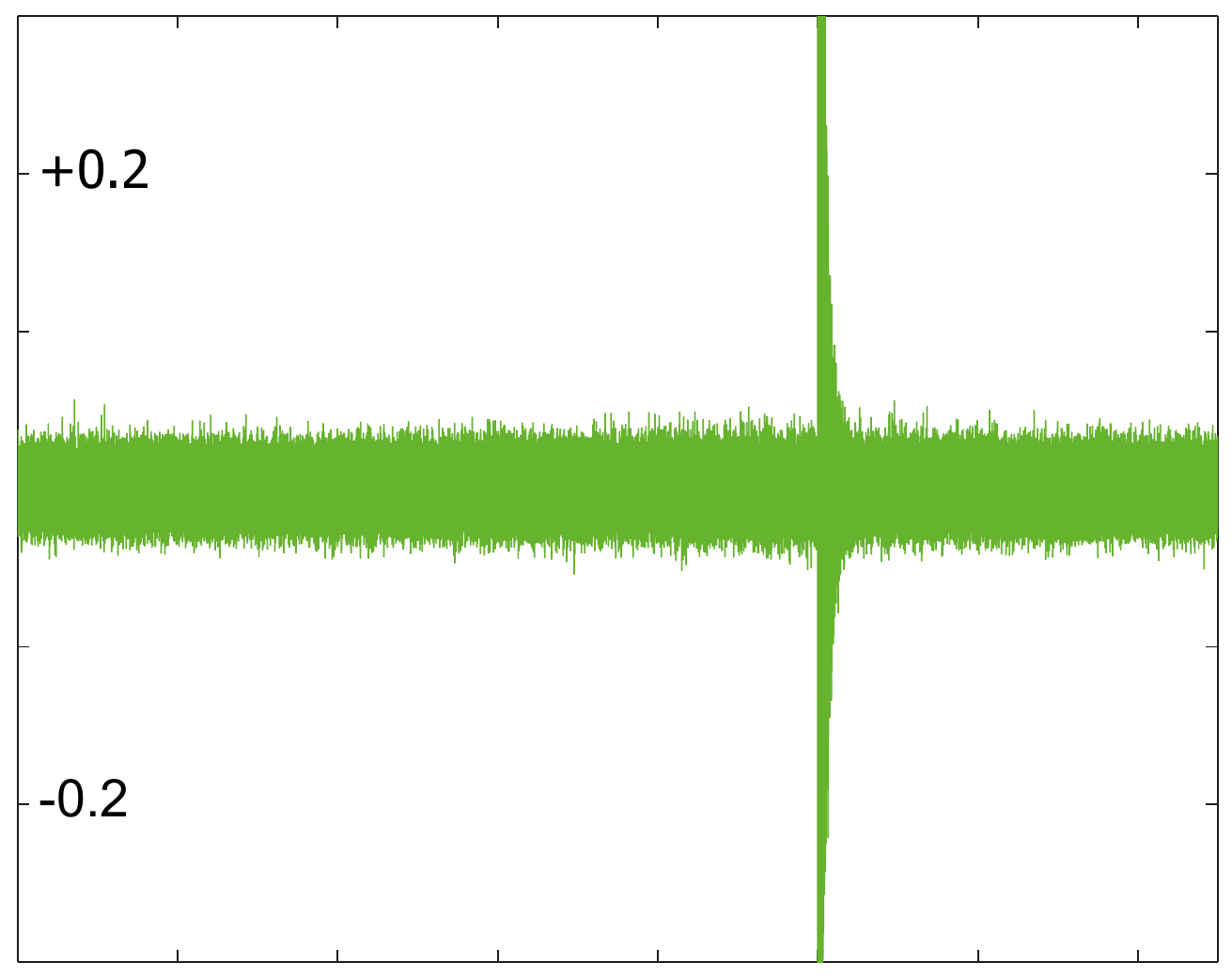}
\caption{IR with MLS.}
\label{fig:MLS}
\end{subfigure}

\begin{subfigure}[b]{.35\linewidth}
\includegraphics[width=\linewidth]{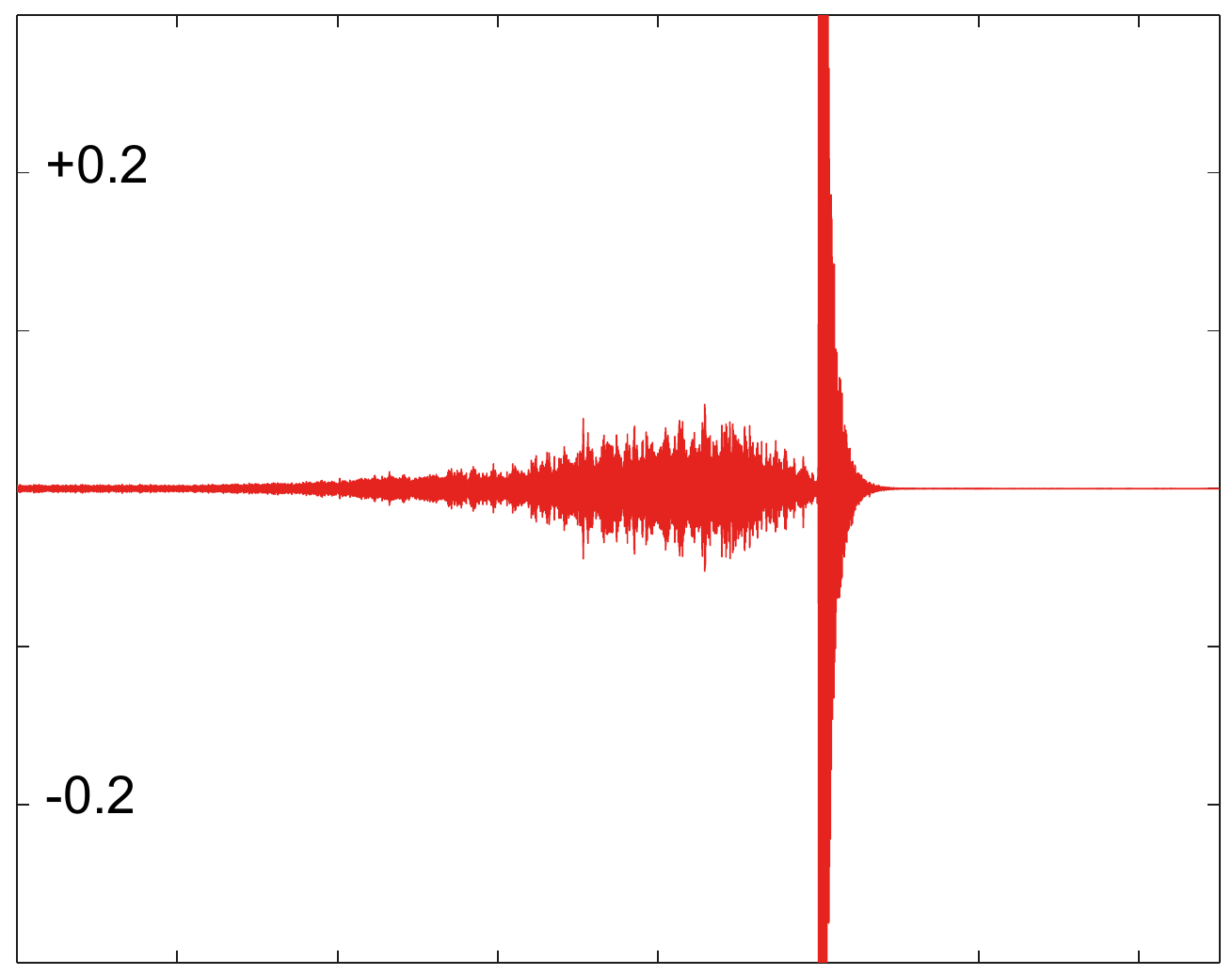}
\caption{IR with LLS.}
\label{fig:LSS}
  
\end{subfigure}
\begin{subfigure}[b]{.35\linewidth}
\includegraphics[width=\linewidth]{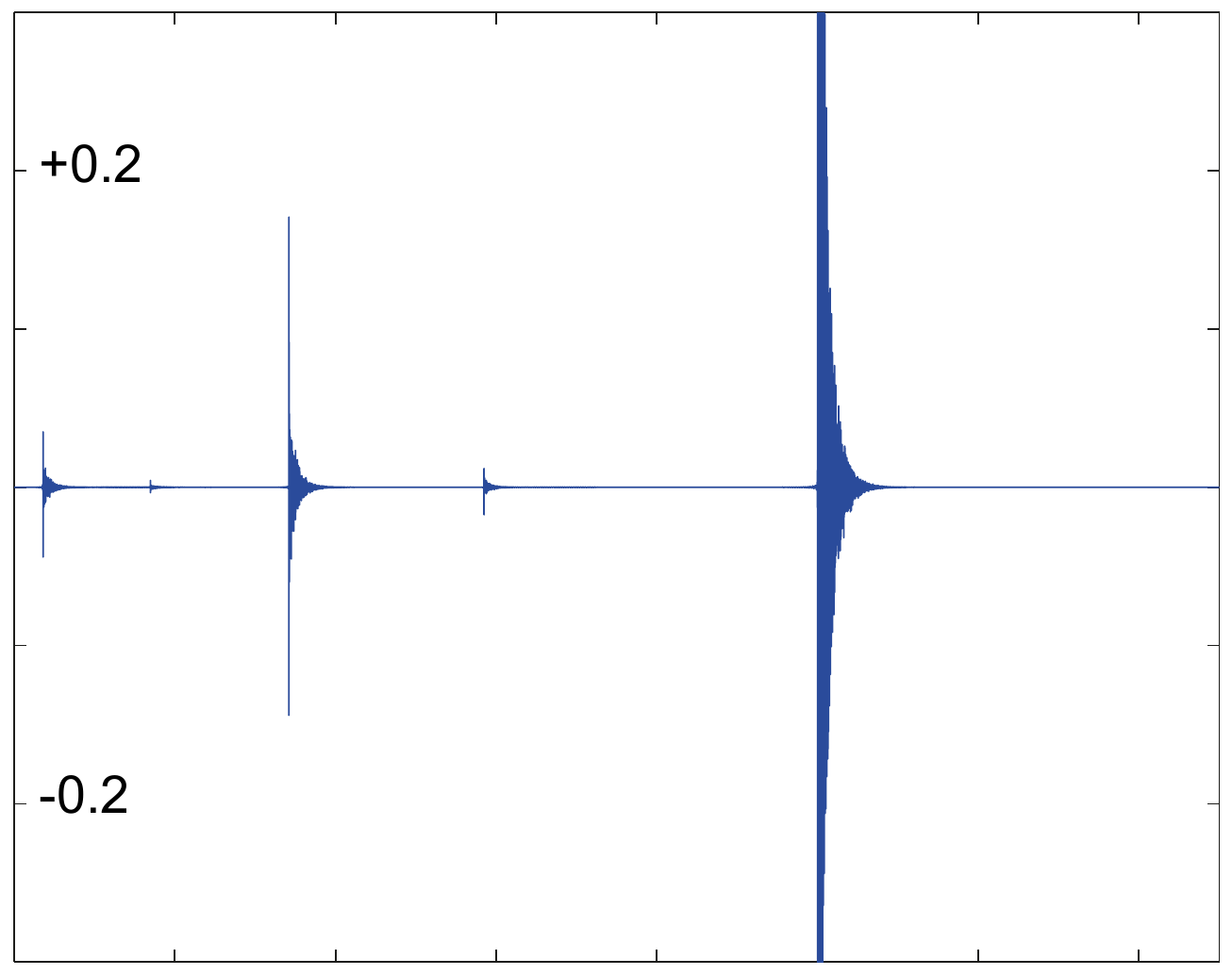}
\caption{IR with ESS.}
\label{fig:ESS}
\end{subfigure}
\caption{Impulse responses measured in an acoustic enclosure with various indirect methods. To increase the SNR, the loudspeaker is set to a high volume, thus introducing non-linear harmonic distortions in the measurement process.}
\label{fig:ir_meas}
\end{figure}

\begin{itemize}
\item \textbf{Maximum Length Sequence (MLS)}. Originally proposed by Schroeder \cite{mls}, the MLS technique provides an indirect measurement of IR based on a finite-length pseudo-random sequence of pulses, that has spectral properties almost equivalent to a pure white noise. Based on this technique, the impulse response is derived by a cross-correlation between the MLS sequence and the microphone signal. As mentioned above, if compared to direct methods, MLS offers a better SNR. It is however sensitive to non-linearities introduced by the measurement system, as shown in Figure  \ref{fig:MLS}.

\item \textbf{Linear Sine Sweep (LLS)}.
LSS \cite{farina} is characterized by an excitation input signal consisting of a sine whose frequency sweeps linearly with time (also referred to as \emph{chirp}). Denoting with $\omega_1$ and $\omega_2$ the initial and final angular frequencies of the sweep, it can be defined as follows:
 \begin{equation}
s[n] =\sin \left( {\it \omega_1}\,n+\frac { \left( {\it \omega_2}-{\it \omega_1} \right) }{L}  \frac{{{n}^{2}}}  {2} \right)
 \end{equation}
As in the case of MLS, the impulse response derives from a cross-correlation between input and output signals. Besides a better SNR than in the MLS case, LSS introduces the advantage of a better (although not perfect) processing of the non-linearities.

\item \textbf{Exponential Sine Sweep (ESS)}.
The ESS technique, introduced by Farina \cite{farina}, is based on an exponential time-growing frequency sweep, as described by the following relationship:
 \begin{equation}
 s[n]= sin \left [ \frac{\omega_1 \cdot L}{ln\left(\frac{\omega_2}{\omega_1}\right)} \left(e^{ \frac{n}{L} \cdot ln\left(\frac{\omega_2}{\omega_1}\right)}-1 \right )  \right ]
 \label{eq:formula_lin}
 \end{equation}
An advantage offered by ESS is the immunity against harmonic distortions. As shown by Figure \ref{fig:ESS}, a perfect separation can be observed between the contributions due to  harmonic distortions, appearing in the left part of the estimated IR, and contributions related to the linear impulse response (i.e., reverberation), observable in its right part.
Another advantage offered by ESS is that its excitation signal spectrum is pink (note that it's white-like for both MLS and LSS), which ensures to have a better SNR at lower frequencies. This is a desirable feature both at perceptual level, since human auditory system is  more discriminative at lower frequencies, and for speech recognition purposes, for which a mel filter-bank is used with higher resolution in the lower part of the frequency axis. Due to this coloration in frequency, ESS impulse responses are characterized by an unnatural dominance of the low-frequency components \cite{farina}. A whitening filter with 3dB/octave should thus be applied before computing the cross-correlation to equalize the computed IRs.
\end{itemize}

In the past, various attempts have been done to compare these impulse response measurement techniques. Our contribution, described in detail in \cite{IRs_paper}, is the comparison of these methods with a specific focus on distant speech recognition. In particular, we compared MLS, LSS and ESS according to some important parameters of the measurement process, including the length $L$ of the emitted signal and its output level $V$.
The training of the DSR system was performed with contaminated versions of the APASCI dataset \cite{apasci}, that is an Italian corpus of phonetically-rich sentences (See App. \ref{app:corpora} for more details about this database).  In order to produce an experimental evidence more directly dependent on the acoustic information, the reference task was a word-loop. The test sentences were command and control utterances recorded in the DIRHA apartment by 11 Italian speakers. See App. \ref{app:es1} for a detailed description of the experimental setup.

 \begin{figure}[t!]
 \centering
   \centering
  \centerline{\includegraphics[width=10cm]{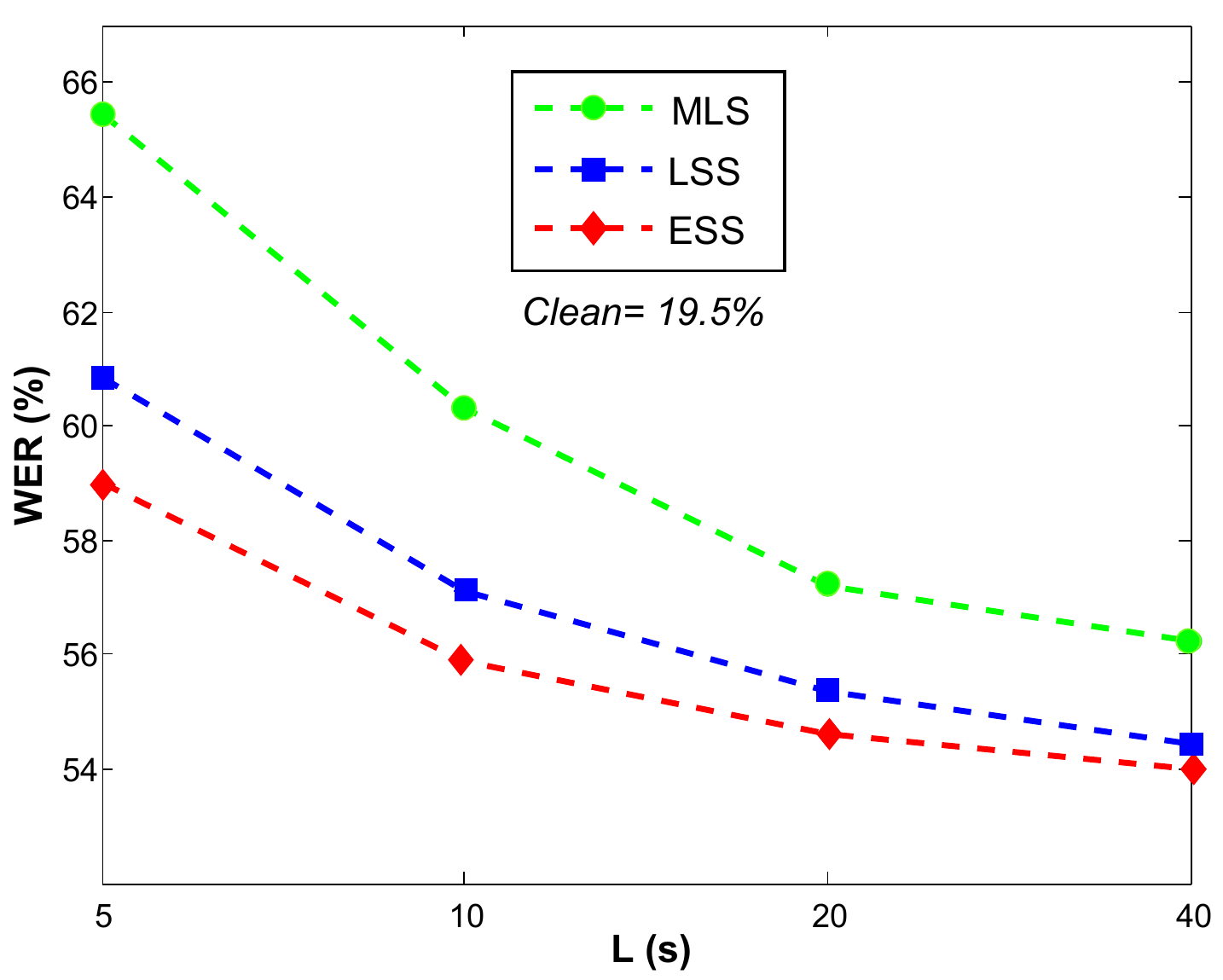}}
 \caption{Performance in terms of WER ($\%$) obtained with a word loop task when varying the length $L$ of the excitation signal using a professional loudspeaker Genelec 8030. }
 \label{fig:lunghezza}
\end{figure}

 \begin{figure}[t!]
 \centering
   \centering
  \centerline{\includegraphics[width=10cm]{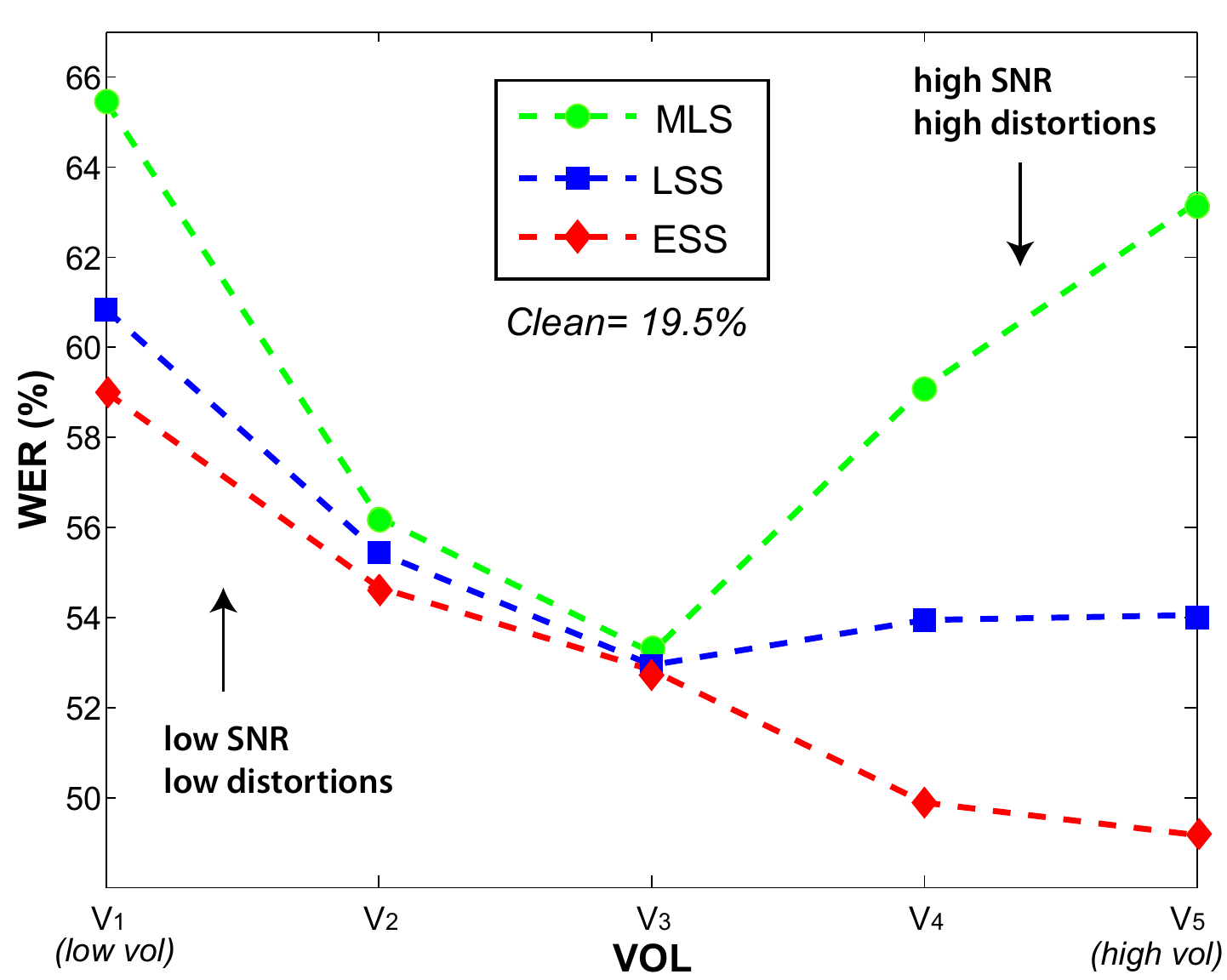}}
\caption{Performance obtained with a word loop task when varying the loudspeaker output level using a professional loudspeaker Genelec 8030.}
 \label{fig:ris_volumi}
\end{figure}

A first set of experiments was conducted to investigate on the impact that excitation length $L$ has on recognition performance. In this case, the impulse responses were derived based on diffusing in the environment the excitation signals with a low amplitude level. 
In Figure \ref{fig:lunghezza}, one can note that an increase of $L$  corresponds to an improved performance for all the investigated techniques, thanks to the improved SNR in the IR measurement process. Results also show that ESS provides the best performance at any excitation signal length. This fact can be due to a better SNR at lower frequencies (pink spectrum), e.g. below 2-3 kHz, typically more critical in speech recognition. MLS and LSS, characterized by a white-like spectrum, do not have this interesting property.
The experimental results also show that MLS has a higher sensitivity to noise than the other two techniques. However, in general, the difference in performance tends to decrease when $L$ increases.

To study the impact of harmonic distortions, a second set of experiments regarded the analysis of recognition performance when impulse response measurements were realized with different dynamics at loudspeaker output level.
From Figure  \ref{fig:ris_volumi} it is worth noting that ESS outperforms the other two techniques. As previously observed, for lower dynamics ($V_{1}$, $V_{2}$), this is due to a better management of SNR.
On the other hand, at higher levels of dynamics ($V_{4}$, $V_{5}$) the best performance provided by ESS is mainly due to a better management of harmonic distortions.

While with MLS and LSS one should choose a trade-off setting in order to have a satisfactory SNR without introducing harmonic distortions, ESS overcomes this trade-off, ensuring a better performance also when the output level of the  loudspeaker increases. This study thus clearly demonstrated the superiority of the ESS technique, that was exploited in the following part of this thesis to derive realistic simulated data.

\section{On the Realism of Simulated Data} \label{sec:real_vs_sim}
Once established a data contamination method, we tried to better validate our approach by performing a more detailed comparison between real and simulated data. 

\begin{figure}[t!]
\centering
\includegraphics[width=0.8\textwidth]{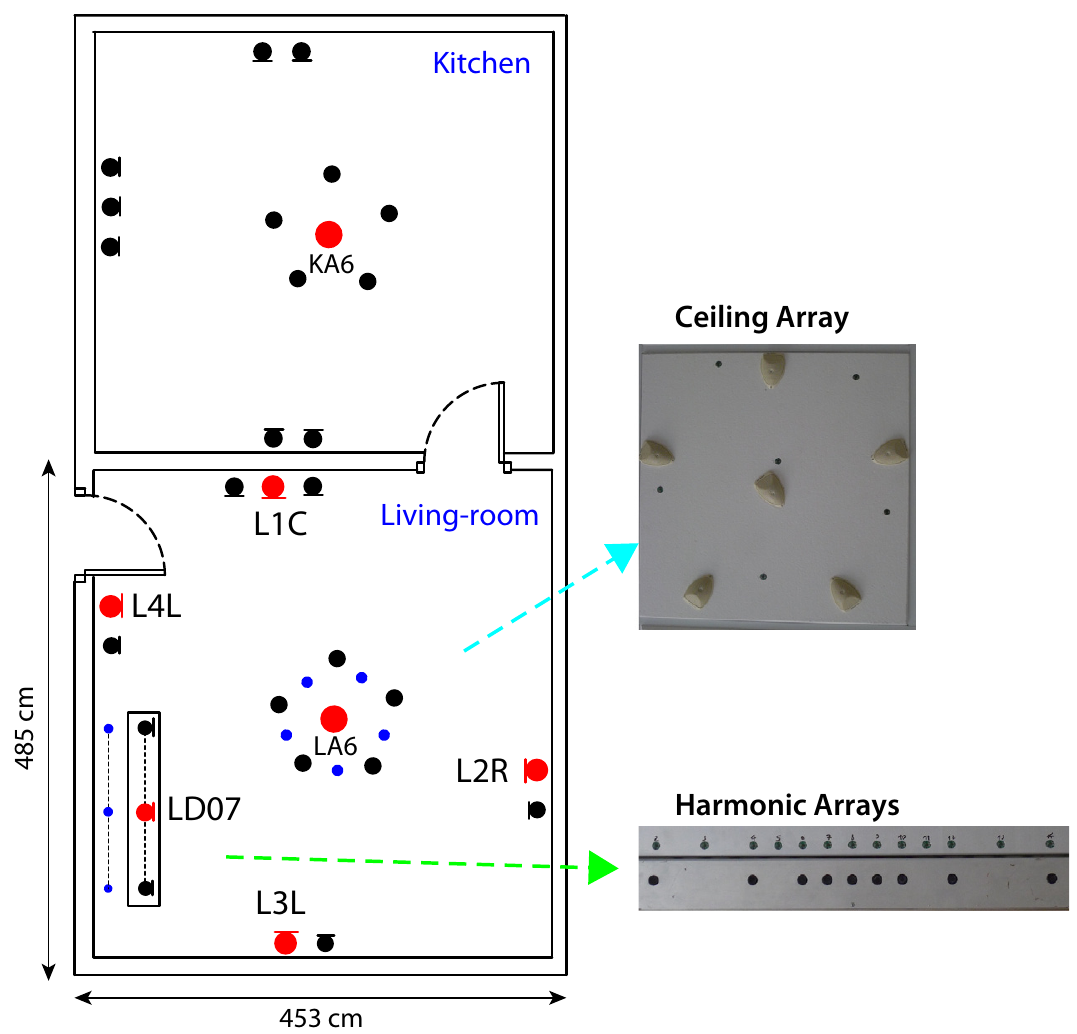}
\caption{An outline of the microphone set-up adopted for the real recordings and for IR measurement. Blue small dots represent digital MEMS microphones, red ones refers to the channels used for most of the experiments, while black ones represent the other available microphones. The right pictures show the ceiling array and the two linear harmonic arrays installed in the living-room.}
\label{fig:dirhaflat_exp}
\end{figure}

The experiments reported in this section summarize our results obtained in \cite{rav_is16,realistic_journal}, 
where we tested the realism of contaminated data under a variety of experimental conditions. The reference scenario was the DIRHA apartment, that was equipped with the  microphone setup  depicted in Figure \ref{fig:dirhaflat_exp} and described in detail in App. \ref{app:mc-2}.

The experiments involved five US speakers that recorded a set of sentences extracted from the wall street journal. To acquire high-quality close-talking material, a first speech acquisition was performed in our recording studio. Using the approach described in the previous section, we then measured several impulse responses in the living-room of the DIRHA apartment  and we generated a simulated dataset. To record real data well-matching with this simulated corpus, we asked the same speakers to utter the same sentences in the same positions of the DIRHA living-room used for measuring the impulse responses. Thanks to this alignment, we were able to perform a comparison as fair as possible between real and simulated data. 
See the Appendix for more detail on the experimental setup (App. \ref{app:es2}). 


The results reported in the first column of Table \ref{tab:res} show the performance obtained when a single distant microphone (i.e., the ``\textit{LA6}'' ceiling microphone depicted in Figure \ref{fig:dirhaflat_exp}) is considered.  The ``$mono$" model  is a  context-independent GMM model (monophones), ``$tri4$" is a context-dependent GMM model based on Speaker Adaptive Training (SAT), while ``$DNN$" represents the considered FF-DNN model.

\begin{figure}
\centering
\includegraphics[width=0.70\textwidth]{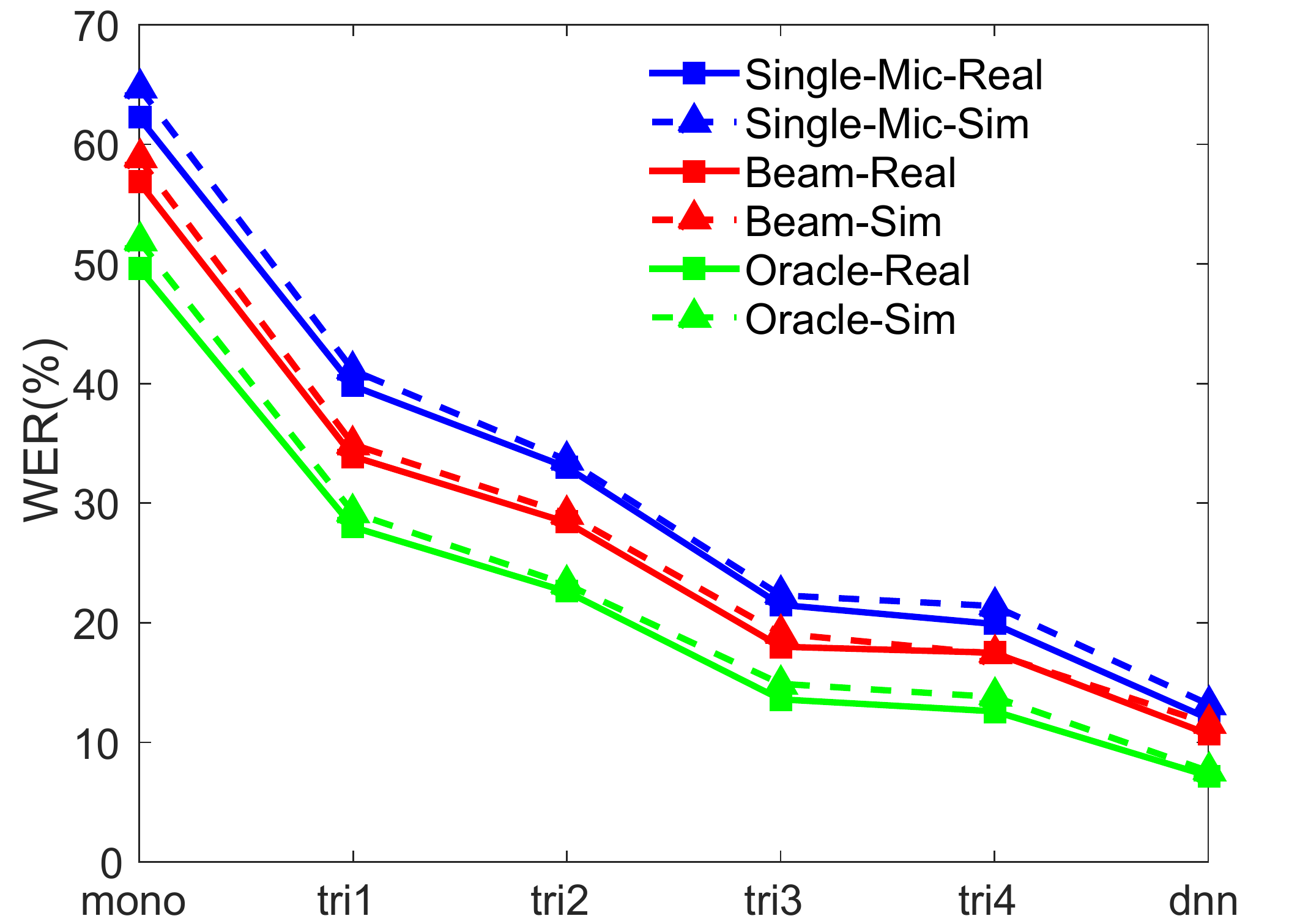}
\caption{Comparison of the performance trends obtained with real and simulated data under a variety of experimental conditions.}
\label{fig:trend1}
\end{figure}

Results clearly highlight that in the case of distant-speech input the ASR performance is dramatically reduced, if compared to a close-talking case (WER=3.7\%). 
The use of robust DNN models trained with contaminated speech material leads, as expected, to a substantial improvement of the WER when compared to other GMM-based systems\footnote{Note that GMM-HMMs systems, whose performance is not anymore competitive with modern DNN-based systems,  are reported here only to analyze the similarity between real and simulated data using different acoustic models.}.
The most interesting result, however, is that a similar performance trend is obtained for both real and simulated data over different acoustic models. This trend can also be appreciated by comparing the continuous (real data) and dashed (sim data) blue lines of Figure  \ref{fig:trend1}. The average relative WER distance between such data-sets computed over the considered acoustic models is about 6\%. We believe that this is a significant result, especially if one takes into account that part of this variability can be attributed, despite our best efforts for aligning simulated and real data, to the fact that in the two recording sessions (i.e, the close-taking recordings to derive simulated data and the real distant speech acquisitions in the apartment) speakers inevitably uttered the same sentence in a different way.

\begin{table*}[t!]
\centering
\small
\tabcolsep=0.20cm
    \begin{tabular}{ | c | c | c | c | c |}
    \hline
    \multirow{2}{*}{} & \multicolumn{2}{ | c | }{\textit{Single Distant Microphone}} & \multicolumn{2}{ | c | }{\textit{D\&S Beamforming}} \\  \cline{2-5}
    & Real Data & Sim Data  & Real Data & Sim Data \\ \hline    
    Mono & 62.2 & 64.7  & 56.8 & 58.8 \\ \hline
    Tri4 & 19.9 & 21.4  & 17.5 & 17.4   \\ \hline    
    DNN & 12.0 & 13.2  & 10.7 & 11.6  \\ \hline    
    \end{tabular}
\caption{WER(\%) obtained in a distant-talking scenario with real and simulated data across different acoustic models and microphone processing.}
\label{tab:res}
\end{table*}

The simulation methodology described in Sec. \ref{sec:ir_meas} can be extended in a very straightforward way to a multi-microphone scenario. It would be thus of crucial importance to ensure that the similar trend between real and simulated data achieved with a single microphone is preserved even when multi-microphone processing is applied to the data.
In the following experiment, we considered the six microphone array placed on the ceiling of the livingroom (see Figure  \ref{fig:dirhaflat_exp}) and applied a delay-and-sum beamforming to both real and simulated data. As described in Sec. \ref{sec:sfil}, the source-microphone delays have been computed with the GCC-PHAT algorithm \cite{gcf}. 

Table \ref{tab:res} and Figure  \ref{fig:trend1} show that beamforming is helpful for improving the system performance (see the red lines). One can also note that, as hoped, a similar performance trend between the datasets is reached when applying beamforming. 
For instance, in the case of real data coupled with FF-DNN acoustic models, delay-and-sum beamforming leads to a relative improvement of about 12\% over the single microphone case, which is similar to the improvement of 13\% obtained with the simulated data. 

Another way to compare real and simulated data in a multi-microphone scenario is to perform a microphone selection. In the following experiment, we considered the six microphones of the DIRHA livingroom depicted as red circles in Figure  \ref{fig:dirhaflat_exp}.
In particular, for each sentence uttered by the speaker, the best WER from the signals acquired by the microphones was considered. This experiment, called \textit{oracle microphone selection}, is reported here only to provide an upper bound of the DSR performance, since real microphone selection techniques are still far from this ideal case \cite{nadeu,cristina}. 
We anyway believe that this test is  particularly interesting because the results might depend on the directional characteristics of the speakers in real and simulated data. Simulated data, in fact, inherit the directivity of the loudspeaker used to measure the impulse responses, and it would be of interest to understand if this aspect affects the spatial realism of our simulations.


Figure  \ref{fig:trend1} shows the WER achieved with the oracle microphone selection (see green lines). Results confirm that the consistency between real and simulated data is largely preserved. This would suggest that simulated data are able to represent  the directional/spatial properties of the speaker with a sufficient level of realism. The experimental results also show that an optimal microphone selection would be particularly helpful to improve the DSR performance.

Thanks to the remarkable level of realism obtained with the proposed approach, several multi-microphone datasets have been developed in the context of the DIRHA project and publicly distributed at international level. The list of corpora based on our realistic data contamination approach is the following:
\begin{itemize}
\item DIRHA Simcorpora \cite{lrec}
\item DIRHA-English \cite{dirha_asru}
\item DIRHA-GRID \cite{dirha_grid}
\item DIRHA-AEC \cite{dirha_icassp}
\end{itemize}

See the Appendix (App. \ref{app:corpora}) for a detailed description of each corpus. 

\section{Directional Image Method} \label{sec:im_method}
When directly measuring an impulse response is not possible, an alternative is to synthetically derive it with a room simulator. 
Concerning this, different methods are described in the literature. The image method (IM) \cite{image},  proposed by Allen and Berkley in 1979, is the most commonly used technique in the speech recognition community. IM refers to the so-called wave equation, and to its frequency domain counterpart called the Helmholtz equation, 
which describes wave propagation in a fluid \cite{kutt}. 

\begin{figure}
\centering
\includegraphics[width=0.90\textwidth]{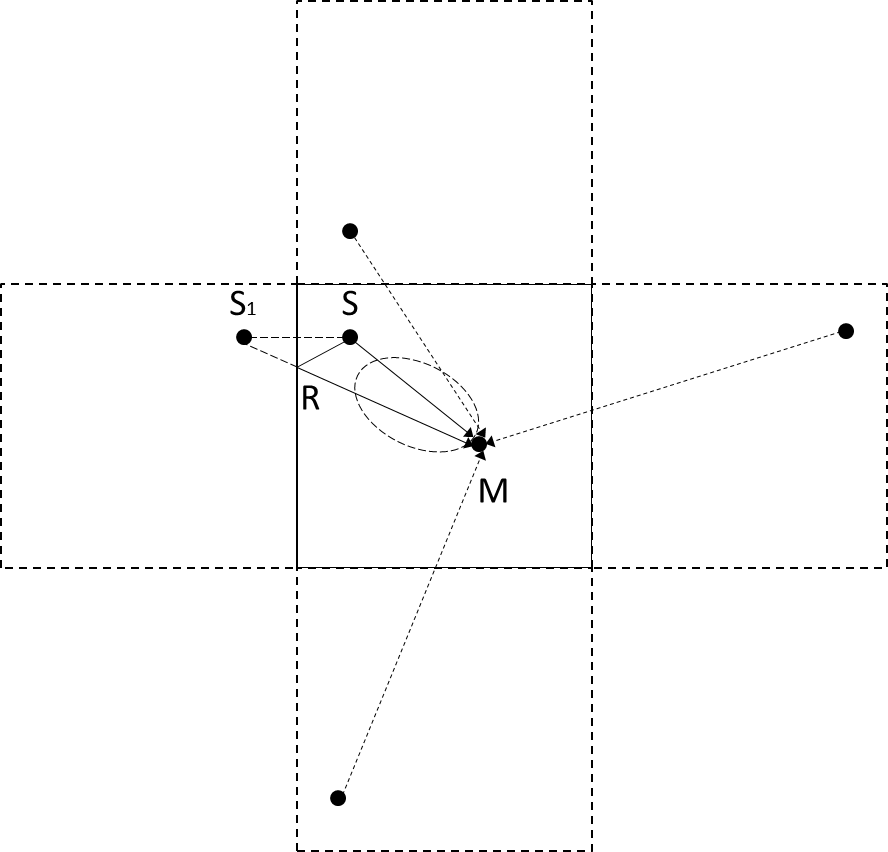}
\caption{Basic principle behind the implementation of the image method.}
\label{fig:im1_ex}
\end{figure}

Fig. \ref{fig:im1_ex} shows the image model used for the 2-D implementation of the standard image method, where a microphone $M$ and a source $S$ are considered. The source, that in the original formulation proposed in \cite{image} is  omidirectional, radiates the acoustic signal in all the directions. To compute the impulse response, the delay $\tau$ of each possible path connecting the microphone with the source should be computed. These delays depend on the length $l$ of each path and on the sound velocity $c$:

\begin{equation}
\tau=\frac{l}{c}
\end{equation}

The length of the direct path can be computed from the known positions of the source and the microphone.  To compute the length of the reflection paths, the IM method exploits an efficient procedure based on the concept of images.  For instance, a first order reflection that follows the path $SR-RM$ is depicted in Fig. \ref{fig:im1_ex}. This reflection can be considered as generated by an equivalent image source $S_{1}$, that is obtained by mirroring the original source $S$ over the reflection wall. The triangle $SS_1R$ is isosceles and therefore the path length $SR+RM$ is the same as $S_1M$. This means that the length of the reflection path corresponds to the distance between the image and the reference microphone. The other first order reflections can be  derived by mirroring the source over the different walls, while higher order reflections can be computed by progressively mirroring the previous images. The attenuation $\alpha$ of each reflection $\delta(t-\tau)$ depends on the path length $l$ and on the number $n$ of reflections involved in the path:

\begin{equation}
\alpha=\frac{\rho^n}{4 \pi l}
\end{equation}

where $\rho$ is the reflection coefficient (ranging from 0 to 1) that described the reflection properties of the environment.

Different algorithms based on IM are described in the literature and some software tools have been made available to the scientific community \cite{Lehmann, Habets}.
In the last decades, several modifications of original Allen-Berkley's algorithm  have been proposed. Some of them consider the extension to 3D room acoustics, a different reflection coefficient for each surface, an implementation in the frequency domain, and the simulation of the microphone polar pattern (e.g., omnidirectional, or cardioid), as described in the literature referenced above. 

In particular, the simulation of directional polar patterns is rather straightforward within the image method framework. As shown in Fig. \ref{fig:im1_ex}, the direction of arrival of the reflection ($RM$) is the same of the image-microphone path $S_1M$. To account directivity it is thus sufficient to multiply the strength of the reflection by a gain factor $D(\theta)$, that depends on the related angle between microphone orientation and reflection.

Beside microphone directivity,  we believe that another fundamental aspect concerns the simulation of sound source directivity, that contributes substantially in characterizing the process of speech propagation in space.  As reported in \cite{Monson}, speech directionality depends not only on the speaker's head pose but also on other factors, as speaker's gender, mouth shape and consequently uttered phonemes. As a matter of fact, the use of an impulse response that derives from an omnidirectional source hypothesis is not sufficient to obtain the required realism, if compared with real speech acquired under equivalent conditions.

    \begin{figure*}
        \centering
        \begin{subfigure}[b]{0.475\textwidth}
            \centering
            \includegraphics[width=\textwidth]{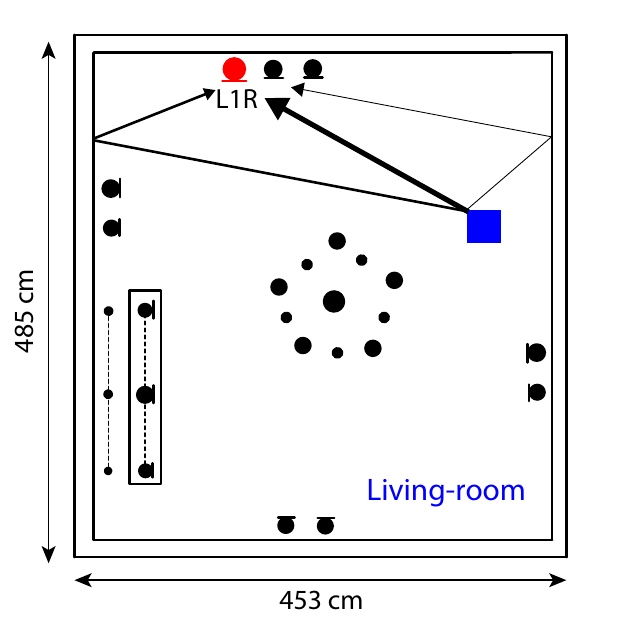}
            \caption[Network2]%
            {{\small Room Geometry.}}    
            \label{fig:dir_room}
        \end{subfigure}
        \hfill
        \begin{subfigure}[b]{0.475\textwidth}  
            \centering 
            \includegraphics[width=\textwidth]{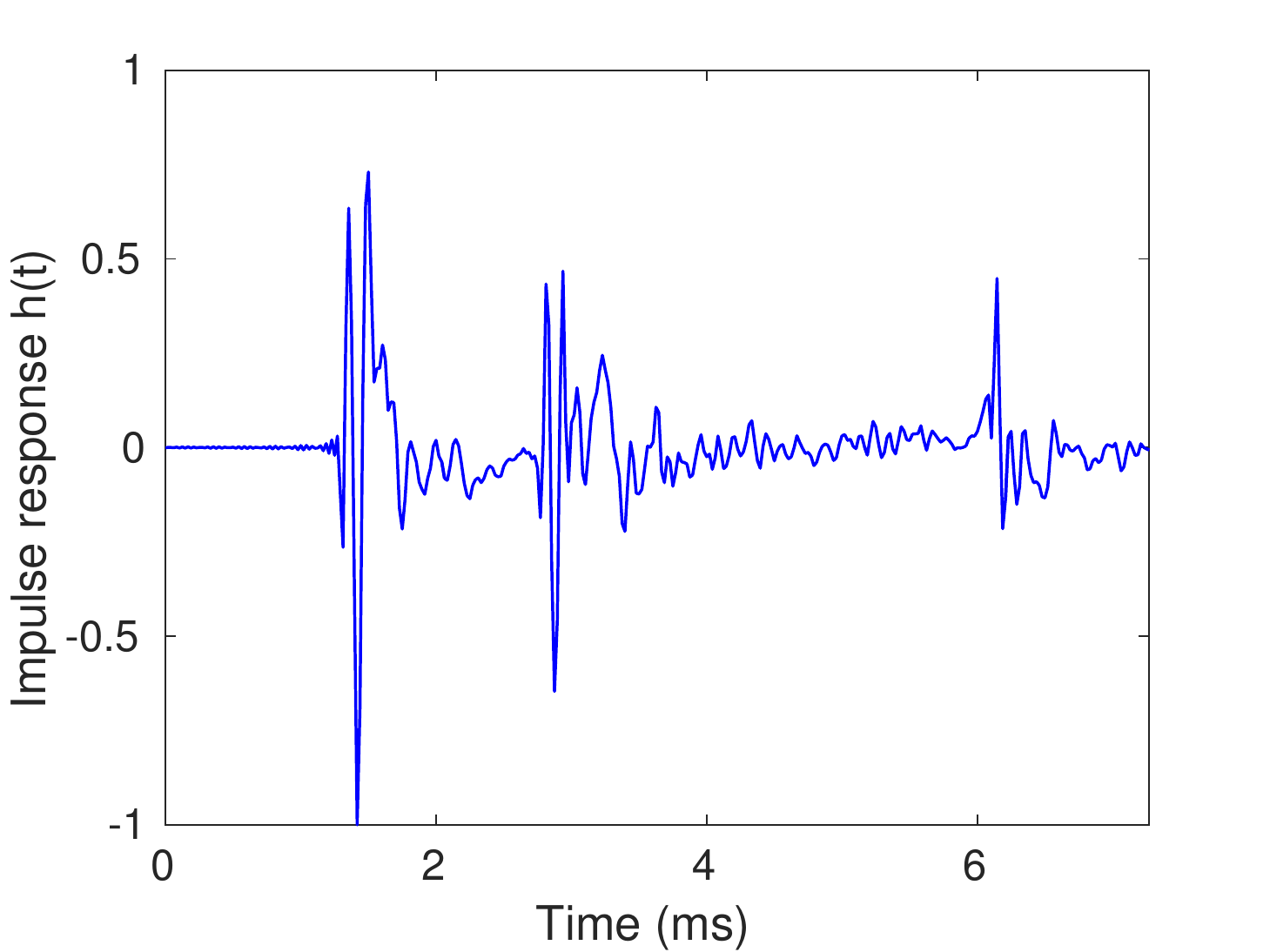}
            \caption[]%
            {{\small Measured impulse response.}}    
            \label{fig:dir_real}
        \end{subfigure}
        \vskip\baselineskip
        \begin{subfigure}[b]{0.475\textwidth}   
            \centering 
            \includegraphics[width=\textwidth]{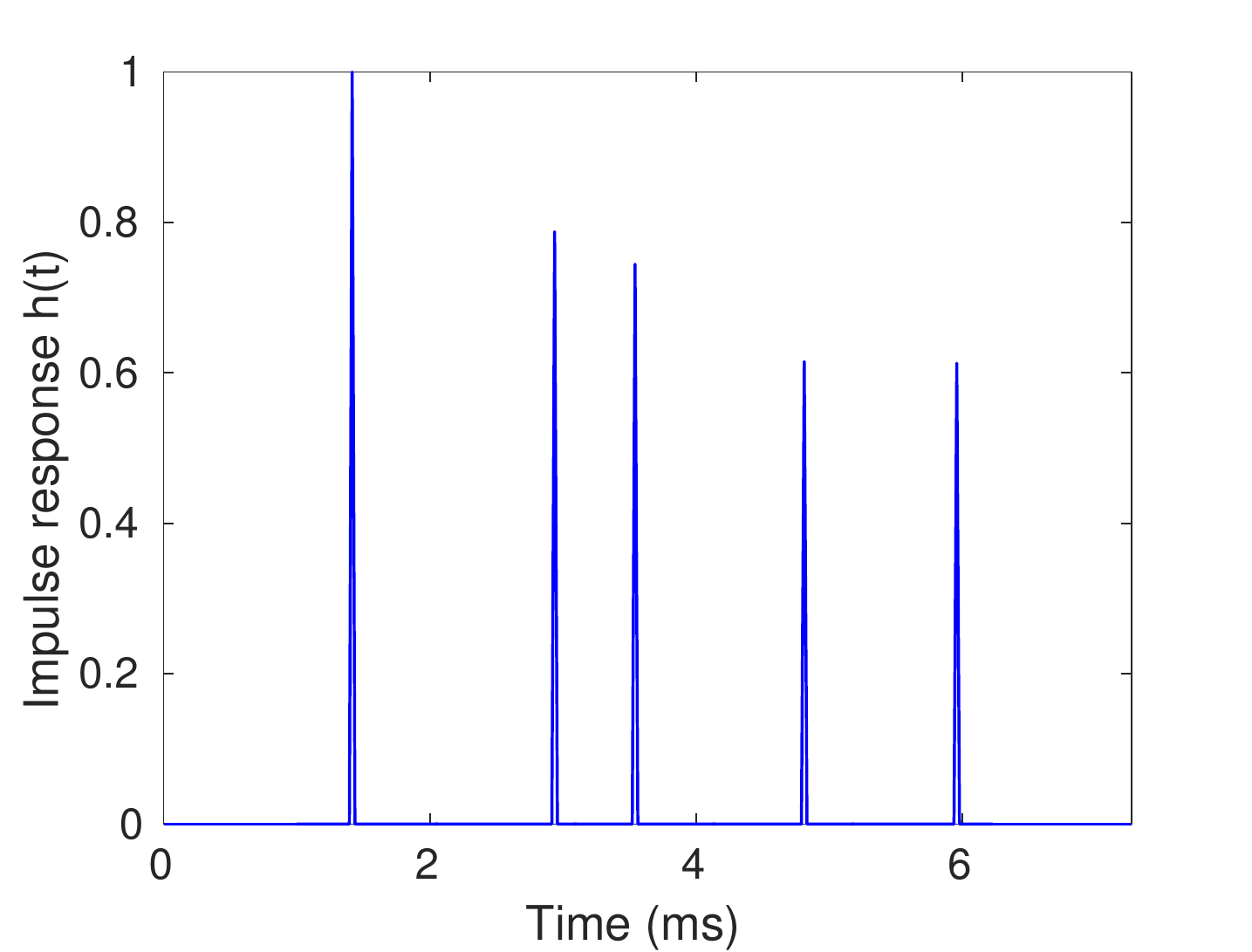}
            \caption[]%
            {{\small Original Omnidirectional IM.}}    
            \label{fig:dir_imomni}
        \end{subfigure}
        \begin{subfigure}[b]{0.475\textwidth}   
            \centering 
            \includegraphics[width=\textwidth]{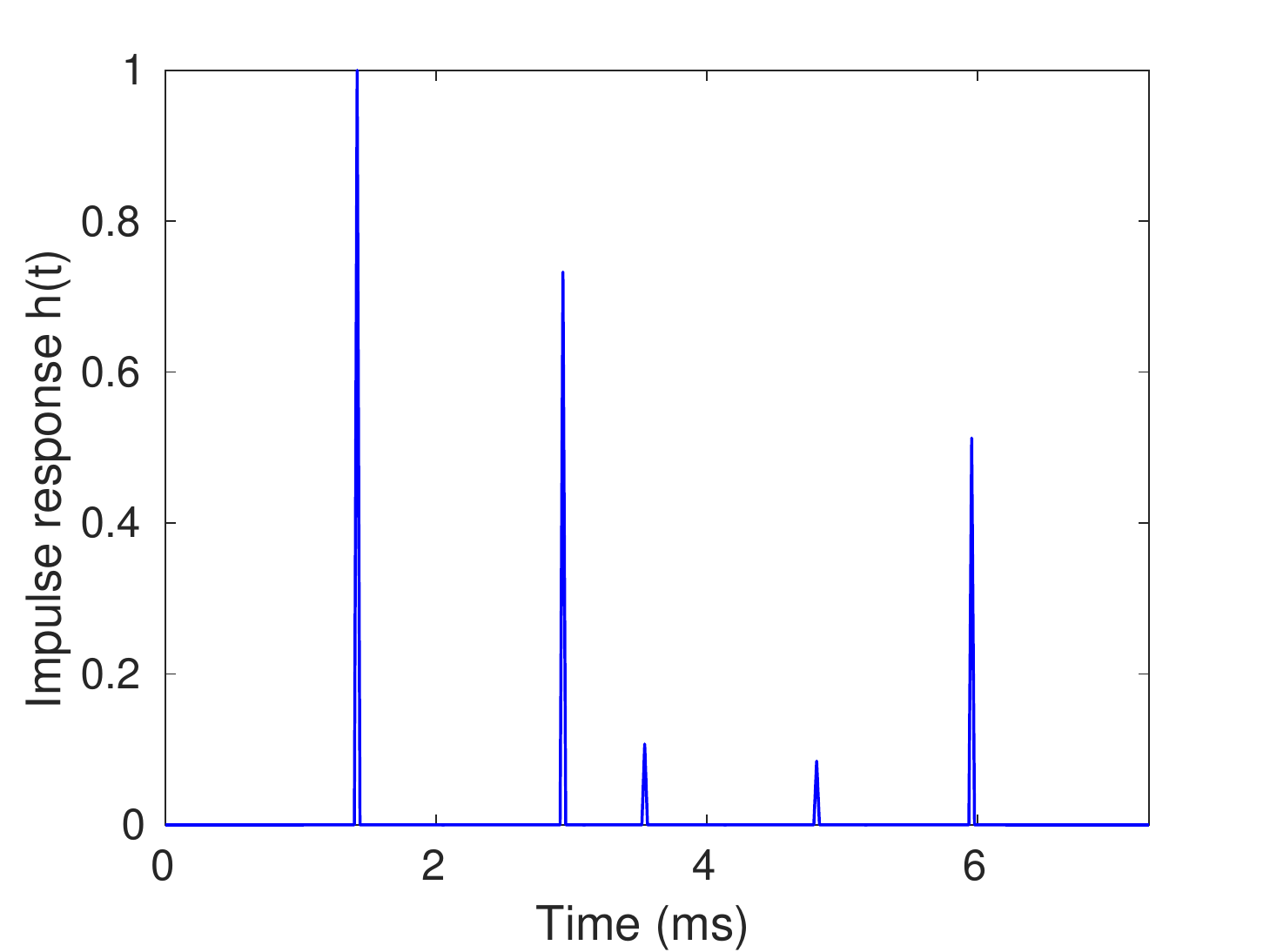}
            \caption[]%
            {{\small Directional IM.}}    
            \label{fig:dir_imdir}
        \end{subfigure}
        \quad

        \caption[ The average and standard deviation of critical parameters ]
        {\small Early reflections obtained from impulse responses measured/simulated in the DIRHA livingroom with the room geometry depicted in (a). The thickness of the highlighted paths is proportional to the energy diffused along that angle by a directional sound source.} 
        \label{fig:di}
    \end{figure*}




  

The importance of a proper directivity matching between real and simulated data has been reiterated in the previous section. 
Differently to  previous experiments, that were based on measured impulse responses, in this section we elaborate on directivity for synthetic IRs. In this work, we used a software available at FBK, which simulates sound source directivity, based on a modification of the original IM algorithm.
This modification simply exploits the principle of reciprocity: instead of mirroring the source, the microphone is mirrored. The various paths connecting the images are then multiplied not only by $\alpha$ but also by the following directivity factor:

\begin{equation}
D(\theta)=\frac{\Big(\frac{1+cos(\theta)}{2}\Big)^p+\epsilon}{1+\epsilon}
\end{equation}

where $\theta$ is the relative angle between source orientation and direction of the reflection, $p$ is a parameter used to tune the source direcitivity factor, and $\epsilon$ is a small constant to avoid paths with zero gain.  Note that there are actually both azimuth $D_{az}$ and elevation $D_{el}$ directivities in the considered 3-D implementation of the image method. The total directivity can be thus obtained as follows: 

\begin{align}
 D_{az}(\theta)=\Big(\frac{1+cos(\theta)}{2}\Big)^p \\
 D_{el}(\phi)=\Big(\frac{1+cos(\phi)}{2}\Big)^q \\
 D_{tot}(\theta,\phi)=\frac{D_{az}(\theta) \cdot D_{el}(\phi)+\epsilon}{1+\epsilon} 
\end{align}

where $p$ and $q$ are the azimuth and elevation directivity factors, respectively.

The experiments reported in the remaining part of this section are based on the experimental evidence emerged in \cite{realistic_journal}. Values of $p=3$, $q=1$, and $\epsilon=0.01$ have been used for the following experiments.  

A qualitative comparison between the considered directional IM and a more standard implementation of the IM is reported in Figure  \ref{fig:di}.
Figure \ref{fig:dir_real} shows the early reflections of an impulse response measured in the DIRHA apartment, while Figure  \ref{fig:dir_imomni} shows a corresponding IR simulated with the standard omnidirectional image method. Finally, Figure  \ref{fig:dir_imdir} represents the directional implementation. Note that the source-microphone position and orientation depicted in Figure  \ref{fig:dir_room}  were considered for all the reported IRs.

The first peak, that is clearly highlighted in all the IRs, is the direct path. The other ones correspond to reflections on the walls, floor and ceiling of the living-room. It is worth noting that the proposed method originates reflections with an amplitude more similar to that encountered in the real case (compare Figure  \ref{fig:dir_real} and Figure  \ref{fig:dir_imdir}). In the omnidirectional case, for instance, the third and fourth reflections  have a significant magnitude. On the other hand, the amplitude of those peaks is close to zero in both the measured and directional IR, thanks to the assumption of source directionality. Properly managing this aspect, thus significantly contributes to generate synthetic IRs that are more similar to the  corresponding measured ones. As a matter of fact, it is also worth noting that a perceptually more natural simulated speech signal is provided by contaminating clean speech with the latter IR.

 \begin{table}[t!]
 \centering
 \tabcolsep=0.12cm
     \begin{tabular}{  | l | c | c | c | c | c | c | c | }
     \cline{1-4}
  IR ID & IM-OMNI & IM-DIR & IR-MEAS \\ \hline
  IR type & \textit{IM-method} & \textit{IM-method} & \textit{Measured} \\ \hline
  IR directivity & \textit{omni} & \textit{directional} & \textit{directional} \\ \hline
  WER(\%) & 13.0 & 12.5 & 12.0 \\ \hline
     \end{tabular}
 \caption{WER(\%) obtained on real data by contaminating the training data with different implementations of the image method algorithm.}
 \label{tab:im_res}
 \end{table}
 
In the following experiment, we extended our study to the speech recognition domain. 
Training was based on contaminated versions of the WSJ dataset, using impulse responses estimated in the reference DIRHA living-room. Test was based on the real recordings described in the previous section. See App. \ref{app:es2} for more details.

Table  \ref{tab:im_res} shows the recognition results obtained with speech material contaminated with different IRs deriving from the considered implementations of the image method. 
As expected, the best performance is achieved with the measured IRs (IR-MEAS). It is interesting to note, however, that the considered directional implementation (IM-DIR) outperforms the omnidirectional version of the image method (13.0\% vs 12.5\%). 


Our results confirmed the importance of simulating source directivity within the IM framework. Taking into account sound source directionality while training the acoustic model indeed implies a better match with real data, that improves the performance of the speech recognizer. We would like to highlight some analogies with what observed in \cite{brutti}, where ASR performance highly correlated with early-to-late reverberation ratio. In general, taking into account sound source directionality while training the acoustic model
implies that a better match with real data will be obtained as far as DRR (and, consequently, other early-to-late reverberation features) is concerned. A correlation between DRR and recognition performance has also been evidenced in another recent work from us \cite{cristina_CSL}. In general, DRR depends on the distance between source and microphone, and it is much more negative when an omnidirectional sound source is simulated, if compared to the case of a directional sound source.

\section{DNN Training with Contaminated Data} \label{sec:cont_exp}
After proposing a methodology for realistic data contamination, we focused on the study of proper approaches to train DNNs with contaminated data. In particular, we tried to exploit an interesting peculiarity of contaminated datasets, i.e. the fact  that the same sentences are available in both a close- and distant-talking conditions. 

The following sections summarize the experiments reported in \cite{Ravanelli-14}. The experimental validation was based on a phone-loop task, in order to achieve an experimental evidence not biased by language information. The reference environment was the  DIRHA livingroom, that was equipped with the microphone setup described in the  Appendix (see App. \ref{app:mc-1}). Training was based on contaminated versions of both APASCI \cite{apasci} and Euronews \cite{gretter} datasets, while test was performed on real and simulated sentences of the DIRHA-phrich dataset. In this work, we considered acoustic conditions of increasing complexity by adopting simulated data with reverberation (Sim-$Rev$) and with both noise and reverberation (Sim-$Rev\&Noise$).
For more details on the experimental setup see the Appendix (App. \ref{app:es3}).

\subsubsection{Close-Talking Labels}
In standard ASR, the labels for DNN training are derived by a forced-alignment of the training corpus over the tied-states. Although some GMM-free solutions have been proposed \cite{gmm-free}, this alignment is typically performed using a standard CD-GMM-HMM system. 
This phase can be very critical because a precise alignment could be difficult to reach, especially in challenging acoustic scenarios characterized by noise and reverberation. As a consequence, the DNN learning process might be more problematic due to a poor supervision.   

In the contaminated speech training framework, however, a more precise supervision can be obtained from the close-talking dataset. Although this is a natural choice for this kind of training modality, our contribution is to better quantify the benefits deriving from this approach. 
The studied methodology requires to train a standard CD-GMM-HMM system with the original clean datasets, and exploit it to generate a precise tied-state forced alignment over the close-talking training corpus, later inherited as supervision for the distant-talking DNN.

Table \ref{tab:test3} reports the results obtained with the standard approach, that is based on labels derived from distant-talking signals, and the proposed close-talking variation. 
Results show that the proposed approach provides a substantial performance improvement, over both real and simulated data. Specifically, a relative improvement of 10\% and 12\% was achieved for APASCI and Euronews, respectively. Results shows that the benefits of this approach holds also under challenging acoustic conditions characterized by noise and reverberation.

\begin{table}[t!]
\centering
\small
\tabcolsep=0.11cm
    \begin{tabular}{ | l | c | c | c | c | }
    \cline{1-5}
    \multirow{2}{*}{\backslashbox{\em{Test}}{\em{Train}}} & \multicolumn{2}{ | c |}{APASCI (6 h)}  & \multicolumn{2}{ | c |}{Euronews (100 h)}  \\ \cline{2-5}
    & Standard & CT-lab & Standard & CT-lab \\ \hline
    Sim-Rev & 37.0 & \bf33.0 & 36.1 & \bf32.1 \\ \hline 
    Real-Rev & 40.2 & \bf35.9 & 39.3 & \bf34.3 \\ \hline 
    Sim-Rev\&Noise & 51.8 & \bf47.3 & 50.1 & \bf46.4 \\ \hline      
    \end{tabular}
\caption{PER(\%) obtained in distant-talking scenarios with the standard and with the proposed technique based on close-talking labels (CT-lab).}
\label{tab:test3}
\end{table}

Besides this significant reduction of the error rates, another interesting aspect is the faster convergence of the learning procedure, due to a better supervision provided to the DNN. In these experiments, 15 epochs were needed to converge with the standard solution, while only 12 were sufficient with the proposed approach, reducing the training time of 20\%.

\begin{table}[t!]
\centering
\tabcolsep=0.15cm
    \begin{tabular}{ | l | c | c | c | c | }
    \cline{1-5}
    \multirow{2}{*}{\backslashbox{\em{Test}}{\em{Train}}} & \multicolumn{2}{ | c |}{APASCI (6 h)}  & \multicolumn{2}{ | c |}{Euronews (100 h)}  \\ \cline{2-5}
    & Stand-PT & CT-PT & Stand-PT & CT-PT \\ \hline
    Sim-Rev & 33.0 & \bf31.4 & 32.1 &  \bf31.2 \\ \hline 
    Real-Rev & 35.9 &  \bf34.7 & 34.3 &  \bf33.0 \\ \hline 
    Sim-Rev\&Noise &  47.3 & \bf46.2 & 46.4 &  \bf45.1 \\ \hline

    \end{tabular}
\caption{PER(\%) obtained in distant-talking scenarios with a standard RBM pre-training (Stand-PT) and with the proposed supervised close-talking pre-training (CT-PT).}
\label{tab:test4}
\end{table}

\subsubsection{Supervised Close-talking Pre-Training} \label{sec:ct2}
A proper DNN initialization is of crucial importance in the context of deep learning, as outlined in Chapter \ref{cha:dl}. One of the first methodologies proposed in the literature consisted to initialize DNNs using unsupervised data. The most popular approach was pre-training based on Restricted Boltzman Machines (RBM), that turned out to discover higher level feature representations starting from low-level ones.

In this work,  we studied a supervised alternative to RBM pretraining, which was able to take advantage of the rich information embedded in the close-talking dataset. In particular, we proposed to use a supervised pre-training method based on the close-talking data. The idea is to train a close-talking DNN and inherit its parameters for initializing the distant-talking DNN. A subsequent fine-tuning phase is then carried out on distant-talking data using a slightly reduced learning rate.  This idea resembles a curriculum learning strategy \cite{curriculum}, since we actually propose to first solve a simpler close-talking problem and address the challenging task of environmental robustness only as a second step.

Results, reported in Table \ref{tab:test4}, clearly show a performance improvement, that is consistent for both APASCI and Euronews datasets. This improvement is obtained  over both real and simulated data and also arises in the challenging scenarios characterized by both noise and reverberation ($rev\&noise$).

This suggests that a close-talk pre-training is a smart way of initializing the DNN, which somehow first learns the speech characteristics and only at a later stage learns how to counteract adverse acoustic conditions.

\section{Summary and Future Work} \label{sec:discussion_ch4}
This Chapter summarized our efforts on speech contamination.
First of all we studied a methodology for measuring high-quality impulse responses with the ESS method. We then validated our contamination approach with an extensive experimental activity carried out on both real and simulated data. Finally, we proposed some original methodologies for training DNNs with our simulated data.

This Chapter has also preliminarily addressed the problem of deriving realistic synthetic impulse responses. We proposed a modified version of the original image method that considers directional acoustic sources. This algorithm is being extended to a frequency domain implementation, that promises to generate more realistic impulse responses. 
Despite our best efforts, current synthetic impulse responses are still far from accurately modeling the actual complexity of a real acoustic environment. For example, typical acoustic enclosures are composed of multiple scattering objects, that can significantly alter the overall impulse response. Even though it is currently not possible to directly address such a level of complexity, we believe that processing synthetic impulse responses with some proper noisy sequences might help to better mimic the presence of multiple object spread in the acoustic enclosure. Some future studies will be focused on this topic.

\chapter{Exploiting Time Contexts} \label{cha_time}
In the previous Chapter, we emphasized the importance of contaminated data for improving the performance of a speech recognizer. Another pivotal aspect to increase DSR robustness is a proper analysis of time contexts. Speech, indeed, is inherently a sequential non-stationary signal that evolves dynamically, making of paramount importance the study of methodologies to manage temporal dynamics. This need  is particularly important for distant speech recognition, where contexts might help counteract the considerable uncertainty introduced by noise and reverberation. 

Actually, the importance of modeling speech contexts is known since long time.
In the case of previous HMM-GMM systems, some attempts were done for better exploiting short-term speech dynamics. For instance, high-order derivatives and LDA/HLDA transformations of concatenated frames represented very popular techniques to broaden time contexts \cite{acero_book}.
However, HMM-GMM systems were largely inadequate for managing longer-term information, due to both their inability to handle high-dimensional input spaces and their limitations in dealing with correlated feature vectors. 
A key reason behind the current success of deep learning is the natural ability of DNNs to better manage large time contexts \cite{dnn_better_context}. 

Despite the progress of the last years,  we believe that it is of great interest to continue the study of techniques for better processing long-term information.

In this thesis we contributed to this area of research considering both feed-forward and recurrent neural networks. The remaining part of this Chapter summarizes the main results reported in the papers we published on this topic \cite{ravanelli15,acw,tb,ravanelli_eusipco,ravanelli_is17,li_GRU}. To better contextualize our contribution, in Sec. \ref{sec:ff_context}  we  first propose a brief introduction of the main techniques adopted to manage short-term information with feed-forward DNNs. Sec. \ref{sec:acw} then reports our study on an asymmetric context window \cite{ravanelli15,acw}. In Sec. \ref{sec:rnn_context} we extend our discussion to the most popular methodologies for embedding long-term information in DNN acoustic models. Finally, Sec \ref{sec:li_gru} summarizes our recent efforts to revise standard Gated Recurrent Units (GRU) \cite{ravanelli_is17,li_GRU}.

\section{Analysing Short-Term Contexts} \label{sec:ff_context}
The most straightforward way to exploit time contexts for feed-forward neural networks is to directly feed the DNN with multiple frames. This standard approach is depicted in Figure  \ref{fig:dnn_context}, where the distant-talking signal $y[n]$ captured by the far microphone is processed by a feature extraction function $f(\cdot)$ that computes a sequence of features frames.
For each frame $\mathbf{y_{k}}$, a DNN is employed to perform frame-level phone predictions. In order to estimate more robust posteriors, the network is not only fed with the current frame $\mathbf{y_{k}}$, but also with some surrounding ones. The set of frames feeding the DNN is often called \textit{context window}. Indicating with $\mathbf{y_{k}}$ the distant-talking feature vector for the $k$-th frame of a speech sentence, the context window is the set of frames defined as follows:

\begin{figure*}[t!]
\centering
  \includegraphics[scale=0.61,trim={0cm 0 0.8cm 0},clip]{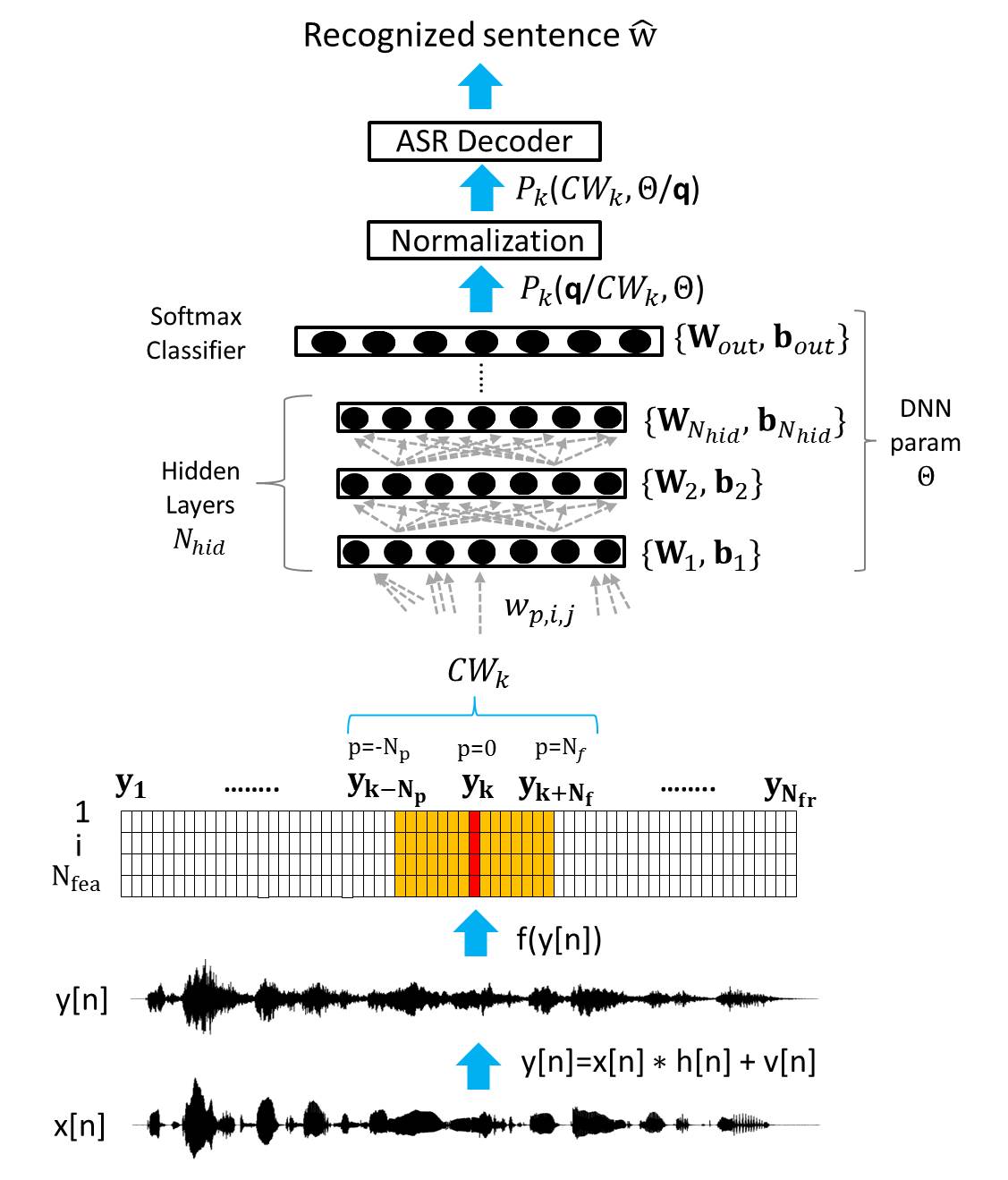}
  \caption{End-to-end pipeline of a HMM-DNN distant speech recognizer. The system is fed with a set features derived from the distant-talking signal $y[n]$.}  
\label{fig:dnn_context}
\end{figure*}

\begin{equation}
CW_{k}=\{ \mathbf{y_{k-N_{p}}},...,\mathbf{y_{k}},...,\mathbf{y_{k+N_{f}}}\} 
\end{equation}
where $N_{p}$ and $N_{f}$ are the number of past and future frames. Although this simple approach works relatively well in most of the cases, the input dimensionality can be too  high when concatenating several frames, possibly causing curse of dimensionality issues.
To mitigate such a concern, LDA \cite{LDA_cw} or DCT \cite{DCT_context}  can be used to perform a dimensionality reduction.

In this thesis we contributed to study a novel context window configuration, that turned out to be suitable for counteracting environmental reverberation. The proposed context window setting, called asymmetric context window, have been described in \cite{ravanelli15,acw}, and is summarized in the following section.


\section{Asymmetric Context Window} \label{sec:acw}
The vast majority of past works on feed-forward DNNs consider symmetric context windows for frame concatenations (i.e., $N_{p}=N_{f}$). Actually, only very few papers adopted asymmetric windows ($N_{p}\neq N_{f}$) in their experimental settings, mainly for real-time applications in close-talking scenarios \cite{online2,small3,tdnn2}. Differently to previous works, we found this approach not only appropriate for low-latency systems, but also better performing than traditional window mechanisms in distant-talking conditions. 
In particular, asymmetric contexts that embeds more past than future frames ($N_{p} > N_{f}$),  turned out to be effective to tackle reverberation.

To account for different balance factors between past and future frames, a useful coefficient $\rho_{cw}$ that will be used in this work is defined as follows:
\begin{equation}
\rho_{cw}(\%)=\frac{{N}_{p}}{N_{p}+N_{f}}\cdot 100
\end{equation}

The asymmetric windows proposed in this work are based on $\rho_{cw}>50\%$.
Under reverberant conditions, the proposed windowing mechanism has proven to be a viable alternative to a more standard symmetric context. The asymmetric window, in fact, feeds the DNN with a more convenient frame configuration which carries, on average,  information that is less redundant and less affected by the correlation effects introduced by reverberation.

Although our work was not the first one proposing asymmetric context windows, our contribution is, to the best of our knowledge, the first attempt to extensively study the role played by the asymmetric context window to counteract reverberation. 

The experimental evidence reported in the following summarizes the main results obtained in \cite{ravanelli15,acw}. 
In particular, a detailed validation has been carried out to show the effectiveness of the investigated method. The following section first provides evidence at signal and feature levels by performing a correlation analysis. In Sec. \ref{sec:acw_inside_dnn} we then analyse the DNN weights and, finally, Sec. \ref{sec:acw_dsr_performance} reports the speech recognition experiments.  

\subsection{Correlation Analysis} \label{sec:acw_corr}
A useful tool to study the redundancy in a reverberated speech signal is the cross-correlation. For instance, we can study the effect of reverberation on a speech signal by analyzing the cross-correlation $R_{xy}$ between the close-talking speech $x[n]$ and the corresponding distant-talking sequence $y[n]$. 
With this purpose,  we can expand the signal $y[n]$, defined in Eq. \ref{eq:cont} of Sec. \ref{sec:challenges}, in this way:

\begin{equation}
y[n] = \sum_{m=0}^{M-1} x[n-m]\cdot h[m]
\end{equation}
Note that, in order to focus our analysis on reverberation effects only, the additive noise $n[n]$ is omitted here.
If we assume $x[n]$ to be a sequence of length L, and $h[n]$ to be an impulse response of length M, the cross-correlation $R_{xy}$ is defined as follows:

\begin{figure}
\begin{subfigure}{0.50\textwidth}
\includegraphics[scale=0.44,trim={0cm 0cm 0cm 0cm},clip]{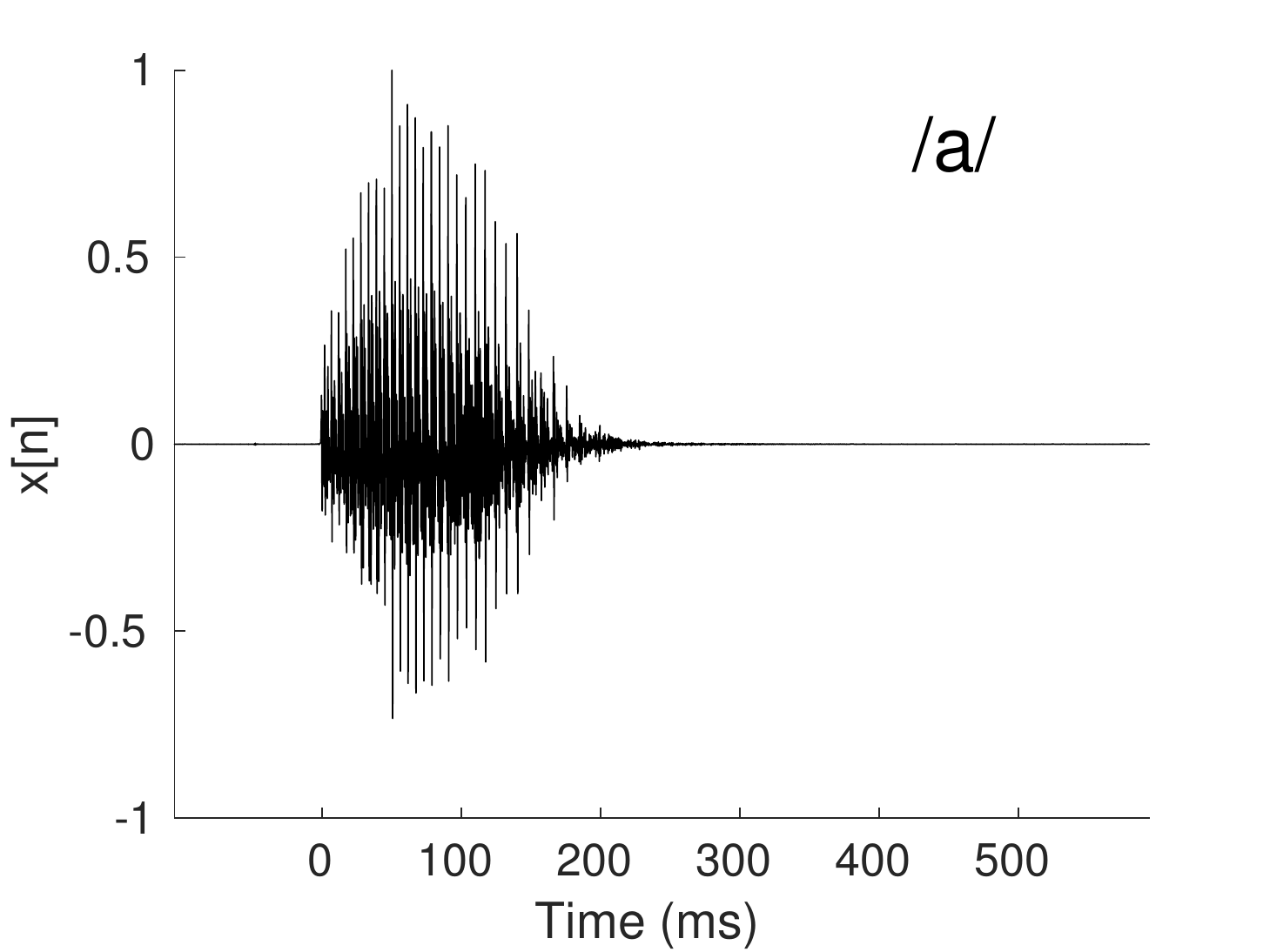}
\caption{Close-talking signal $x[n]$}
\label{fig:a}
\end{subfigure} \hspace{0.0\textwidth}
\begin{subfigure}{0.50\textwidth}
\includegraphics[scale=0.44,trim={0cm 0cm 0cm 0cm}]{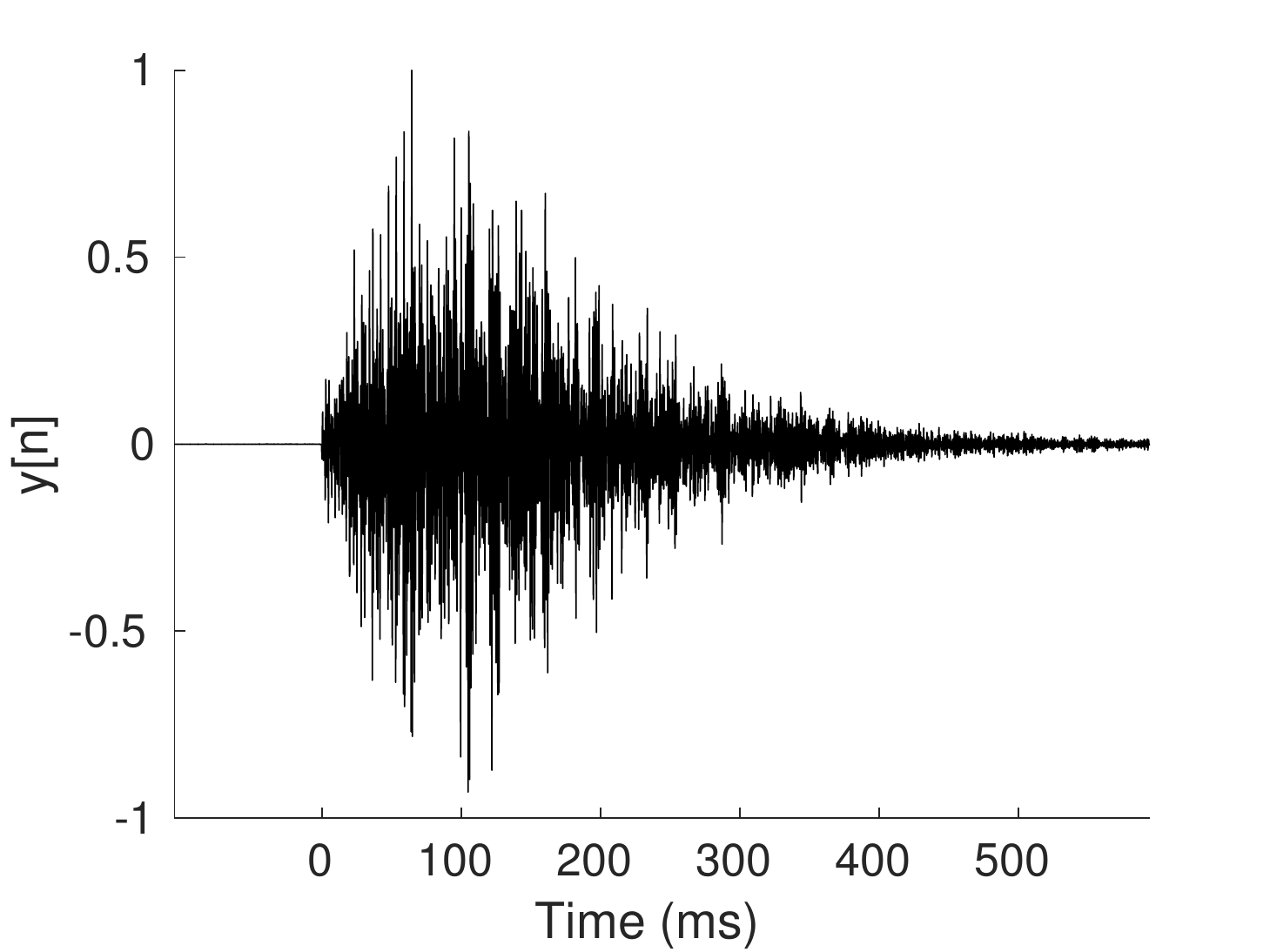}
\caption{Reverberated signal $y[n]$}
\label{fig:a_rev}
\end{subfigure}
\begin{subfigure}{0.50\textwidth}
\includegraphics[scale=0.44,trim={0cm 0cm 0cm 0cm}]{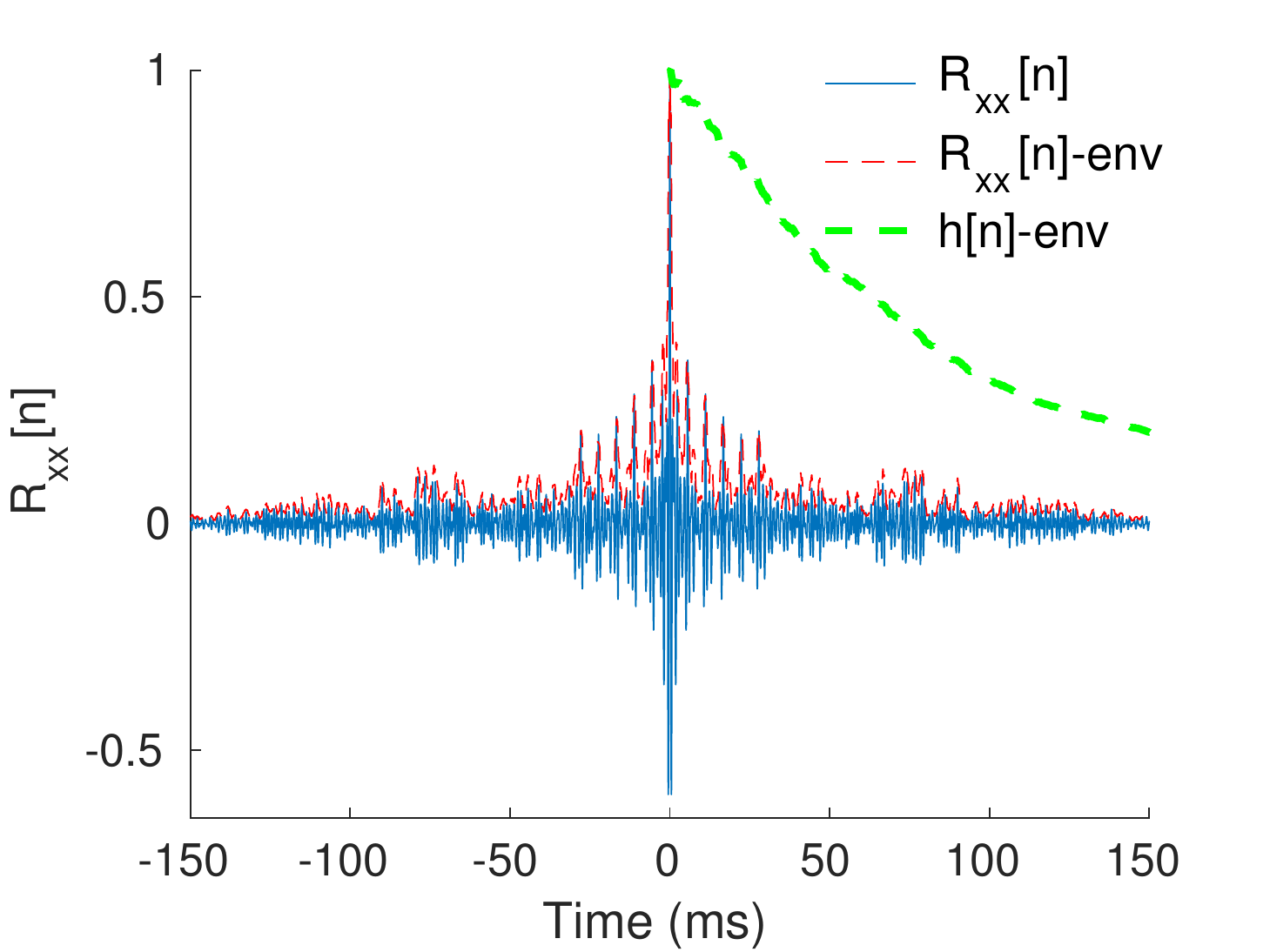}
\caption{Autocorrelation $R_{xx}[n]$}
\label{fig:a_corr}
\end{subfigure}
\begin{subfigure}{0.50\textwidth}
\includegraphics[scale=0.44,trim={0cm 0cm 0cm 0cm}]{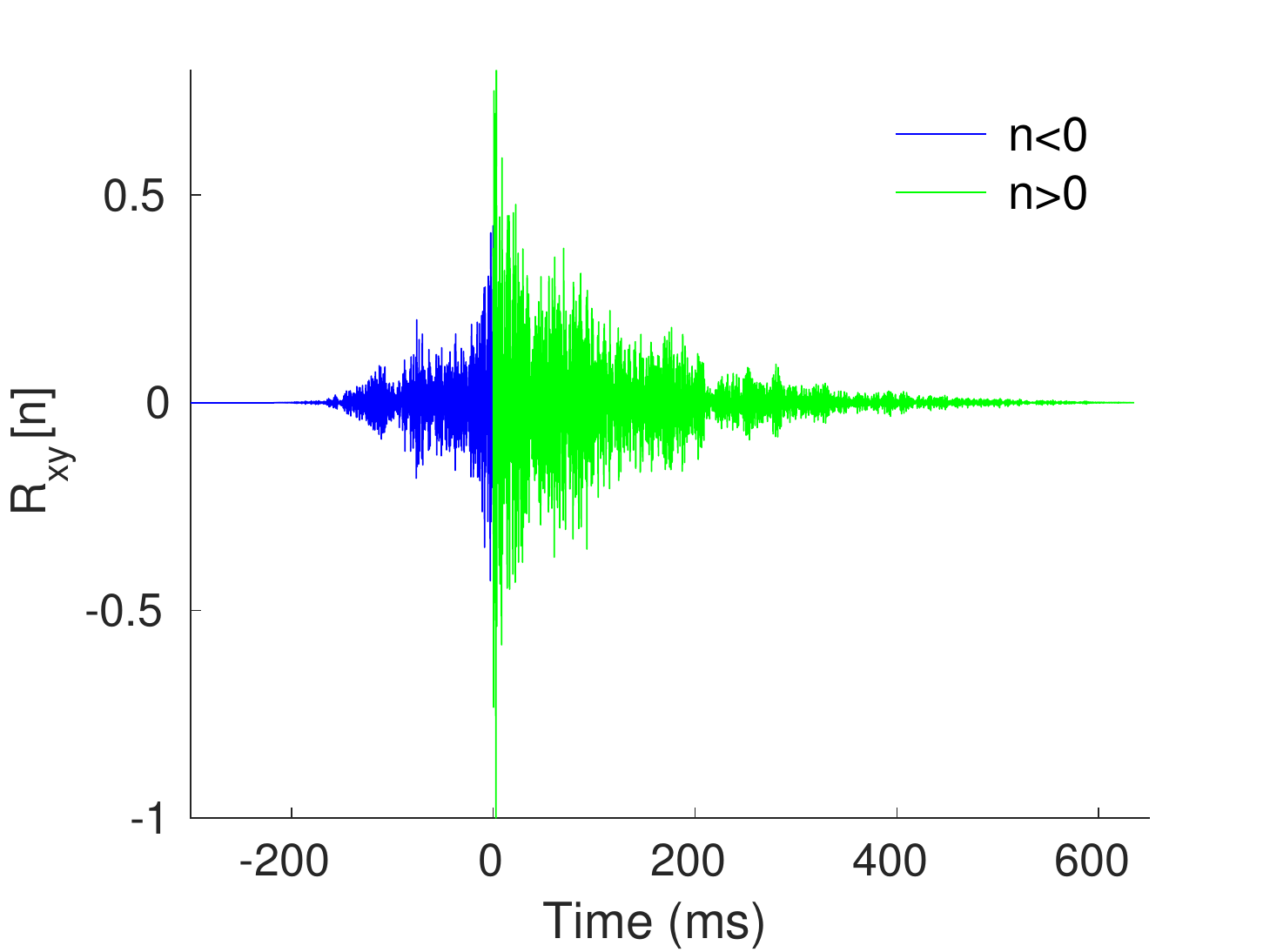}
\caption{Cross-correlation $R_{xy}[n]$}
\label{fig:a_crosscorr}
\end{subfigure}
\caption{Cross and autocorrelation correlation analysis for the vowel /a/. }
\label{fig:xcorr}
\end{figure}

\begin{align}
R_{xy}[n] & = \sum_{l=0}^{L-1} x[l]\cdot y[l+n] \\
& = \sum_{l=0}^{L-1} x[l]\cdot \bigg( \sum_{m=0}^{M-1} x[l+n-m]\cdot h[m] \bigg) \nonumber \\
& = \sum_{l=0}^{L-1} \sum_{m=0}^{M-1} x[l]\cdot x[l+n-m]\cdot h[m] \nonumber \\
& = \sum_{m=0}^{M-1} h[m] \underbrace{\sum_{l=0}^{L-1} x[l]\cdot x[l+n-m]}_{R_{xx}[n-m]} \nonumber \\
& =  \sum_{m=0}^{M-1} h[m] \cdot R_{xx}[n-m] \nonumber
\end{align}

\begin{figure}[t!]
\begin{subfigure}{0.50\textwidth}
\includegraphics[scale=0.44,trim={0cm 0cm 0cm 0cm},clip]{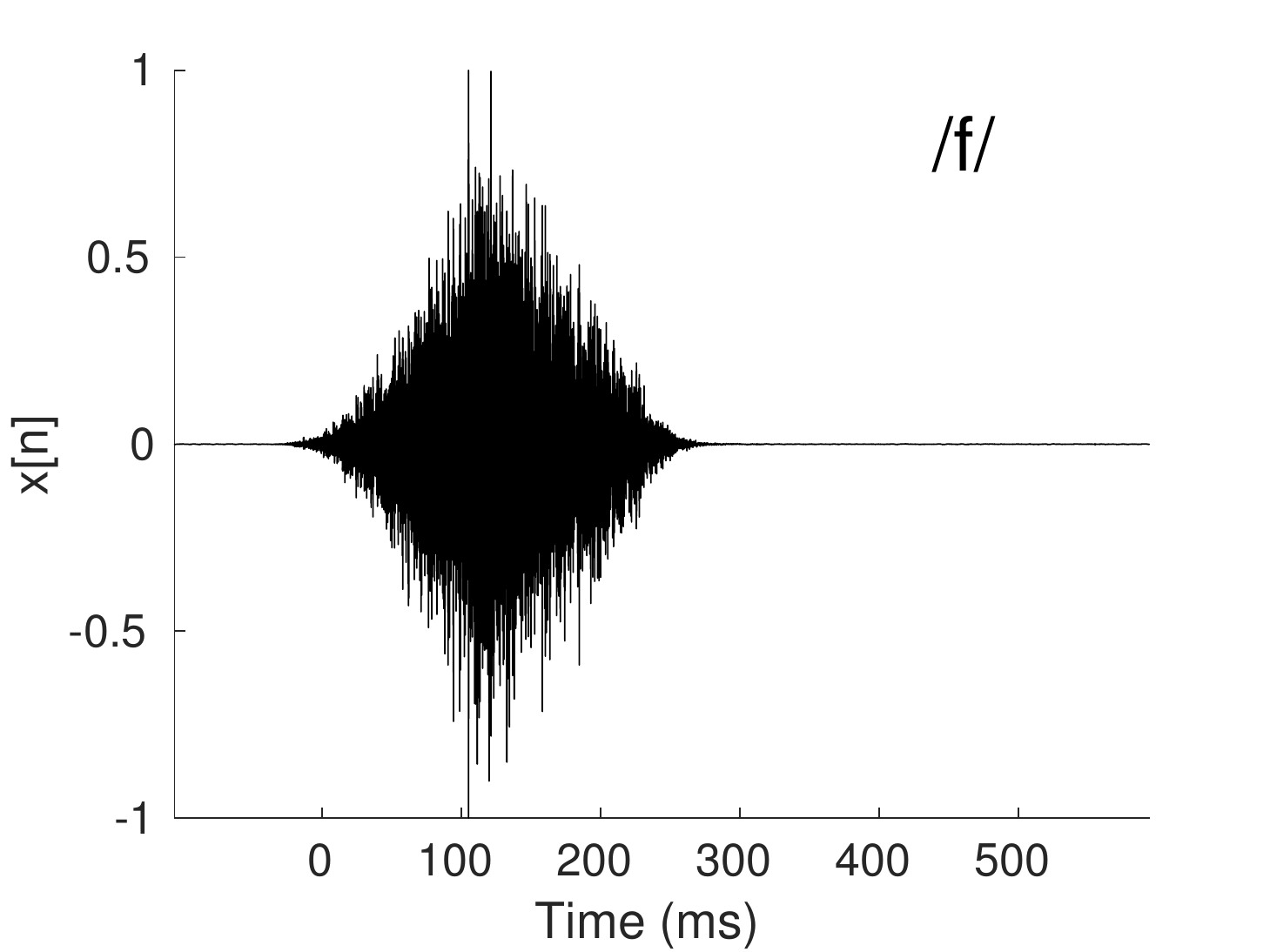}
\caption{Close-talking signal $x[n]$}
\label{fig:f}
\end{subfigure} \hspace{0.0\textwidth}
\begin{subfigure}{0.50\textwidth}
\includegraphics[scale=0.44,trim={0cm 0cm 0cm 0cm}]{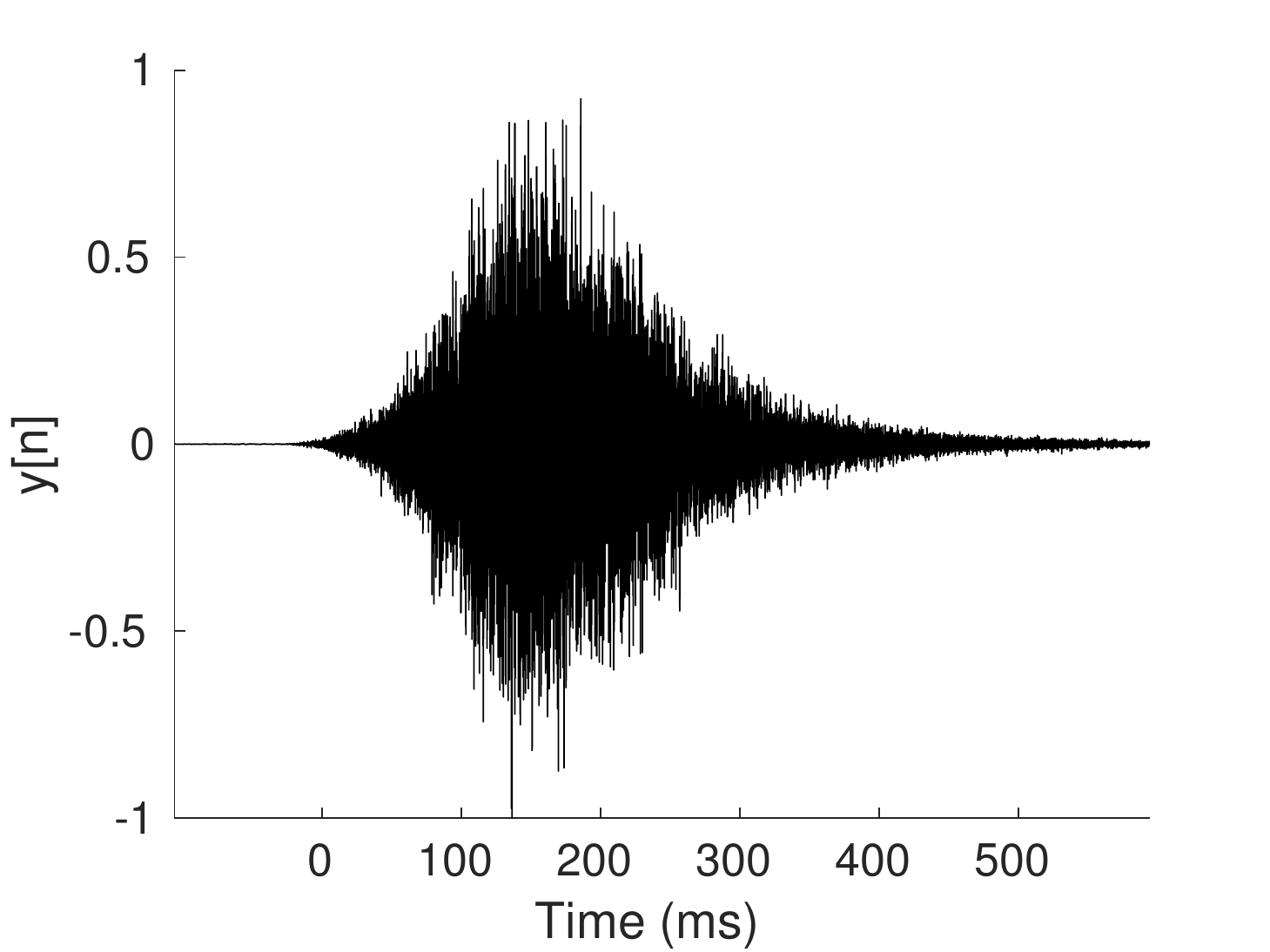}
\caption{Reverberated signal $y[n]$}
\label{fig:f_rev}
\end{subfigure}
\begin{subfigure}{0.50\textwidth}
\includegraphics[scale=0.44,trim={0cm 0cm 0cm 0cm}]{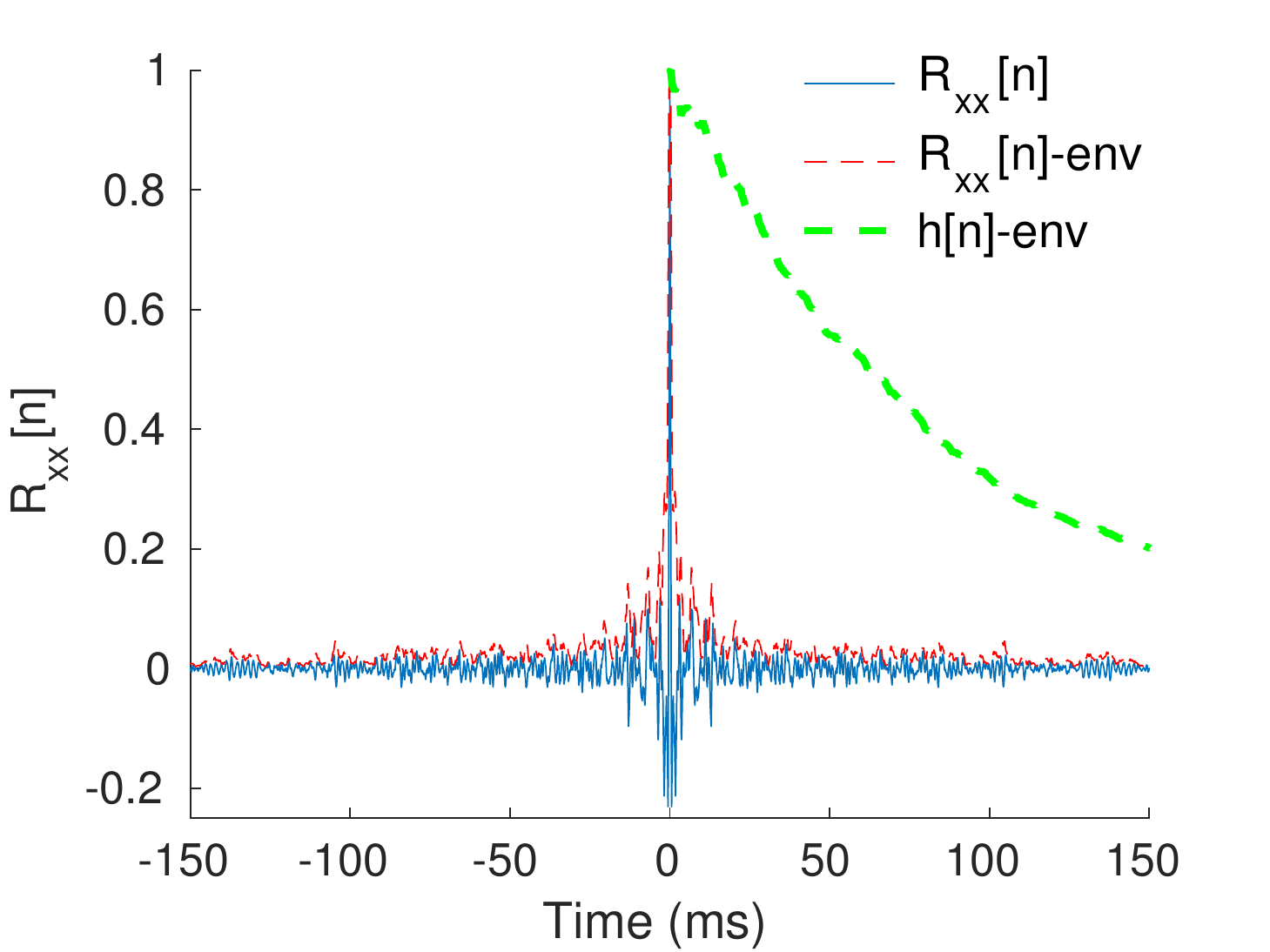}
\caption{Autocorrelation $R_{xx}[n]$}
\label{fig:f_corr}
\end{subfigure}
\begin{subfigure}{0.50\textwidth}
\includegraphics[scale=0.44,trim={0cm 0cm 0cm 0cm}]{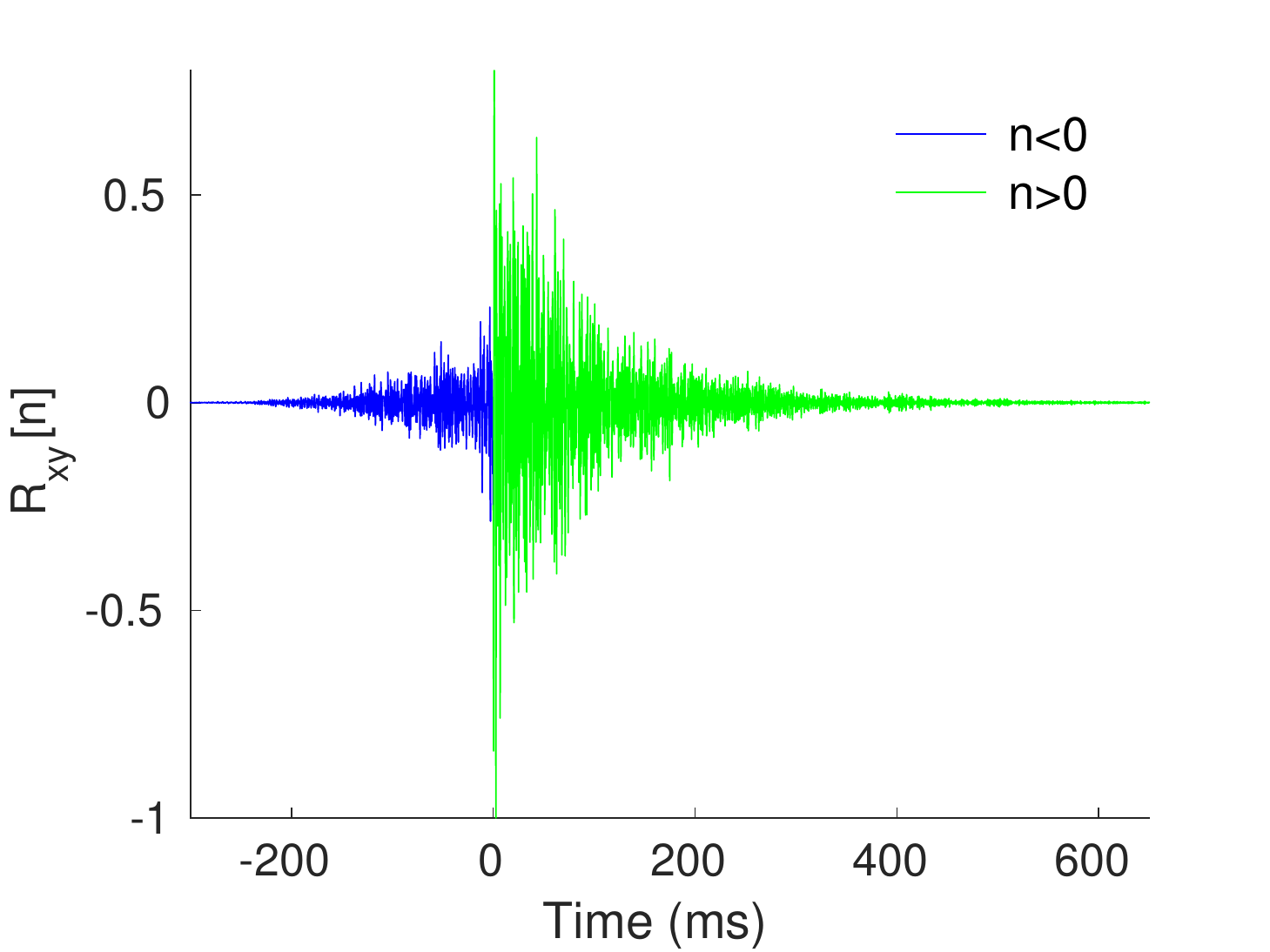}
\caption{Cross-correlation $R_{xy}[n]$}
\label{fig:f_crosscorr}
\end{subfigure}
\caption{Cross and autocorrelation correlation analysis for the fricative /f/. }
\label{fig:xcorr}
\end{figure}

The cross-correlation $R_{xy}[n]$ thus depends on both the impulse response $h[n]$ and the autocorrelation function $R_{xx}[n]$. 
The autocorrelation $R_{xx}[n]$ varies significantly according to the particular phoneme and the signal characteristics that are considered. Figure  \ref{fig:a_corr}, for instance, shows the autocorrelation $R_{xx}[n]$ of a vowel $/a/$, while Figure  \ref{fig:f_corr} illustrates  $R_{xx}[n]$ for a fricative $/f/$.

One can easily observe that different autocorrelation patterns are obtained: for the vowel sound $/a/$, $R_{xx}[n]$ is based on several peaks due to pitch and formants, while for $/f/$ a more impulse-like pattern is observed. The spread of the autocorrelation function around its center $t=0$ also depends on the specific phoneme. If we consider, for instance, the time instant where the energy of $R_{xx}[n]$ decays to 99.9\% of its initial value, the autocorrelation length is 104 ms with the vowel /a/, and about 25 ms with the fricative /f/. 
In both cases, however, the duration of the autocorrelation is significantly smaller than the length of the impulse response (see green dashed line of Figure   \ref{fig:a_corr} and Figure  \ref{fig:f_corr}). 
This characteristic, together with the causality of the impulse response ($h[n]=0 \quad \forall n<0$), originates an asymmetric trend in the cross-correlation $R_{xy}[n]$, which can be clearly appreciated from both Figure  \ref{fig:a_crosscorr} and Figure  \ref{fig:f_crosscorr}. 
As shown by the latter examples, with medium-high $T_{60}$ the right side of this function is influenced by the impulse response decay. 
The future samples ($n>0$) are thus, on average, more redundant than previous ones ($n<0$), and this effect is amplified when reverberation increases.

Based on these observations, we began to realize the possible benefits arising with an asymmetric context window that integrates more past than future frames. This solution, in fact, would be more appropriate than a traditional symmetric context since it results in a frame configuration less affected by the aforementioned forward correlation effects of reverberation. In other words,  with the asymmetric context we can feed the DSR system with information which is, on average, more complementary than that considered in a standard symmetric frame configuration, allowing the DNN to perform more robust predictions. 



To further validate this conjecture, we extended the previous signal-based results to the feature domain, using the Pearson correlation \cite{pearson,pearson2}. This coefficient, denoted as  $r_{x,y}(p)$, is computed between close-talking $\mathbf{x}$ and distant-talking  $\mathbf{y}$ feature sequences, and it is defined as follows:

\begin{equation}
r_{x,y}(p)=\dfrac{\mathlarger{\sum}_{k=1}^{N_{fr}} (\mathbf{x_{k}}-\mathbf{\bar{x}})\cdot (\mathbf{y_{k+p}}-\mathbf{\bar{y}})} {\sqrt{\sum_{k=1}^{N_{fr}} (\mathbf{x_{k}}-\mathbf{\bar{x}})^2} \cdot \sqrt{\sum_{k=1}^{N_{fr}} (\mathbf{y_{k+p}}-\mathbf{\bar{y}})^2} }
\label{eq:pearson}
\end{equation}
where $\mathbf{\bar{x}}=\sum_{k=1}^{N_{fr}}\mathbf{x_{k}}$ and $\mathbf{\bar{y}}=\sum_{k=1}^{N_{fr}}\mathbf{y_{k}}$.

The correlation $r_{x,y}(p)$, is here estimated for $-N_{p}\leq p \leq N_{f}$ integrating over all the frames $N_{fr}$. The argument $p$ denotes the frame lag between the input sequences. For instance, when $p=0$ the Pearson coefficient $r_{x,y}(0)$ is computed by considering the k-th frames for both close- and distant-talking features. When $p=m$, the correlation is computed between the $k$-th close-talking frames and $k+m$ frame of the distant talking feature sequence.

\begin{figure}[t!]
\begin{subfigure}{0.50\textwidth}
\includegraphics[scale=0.55]{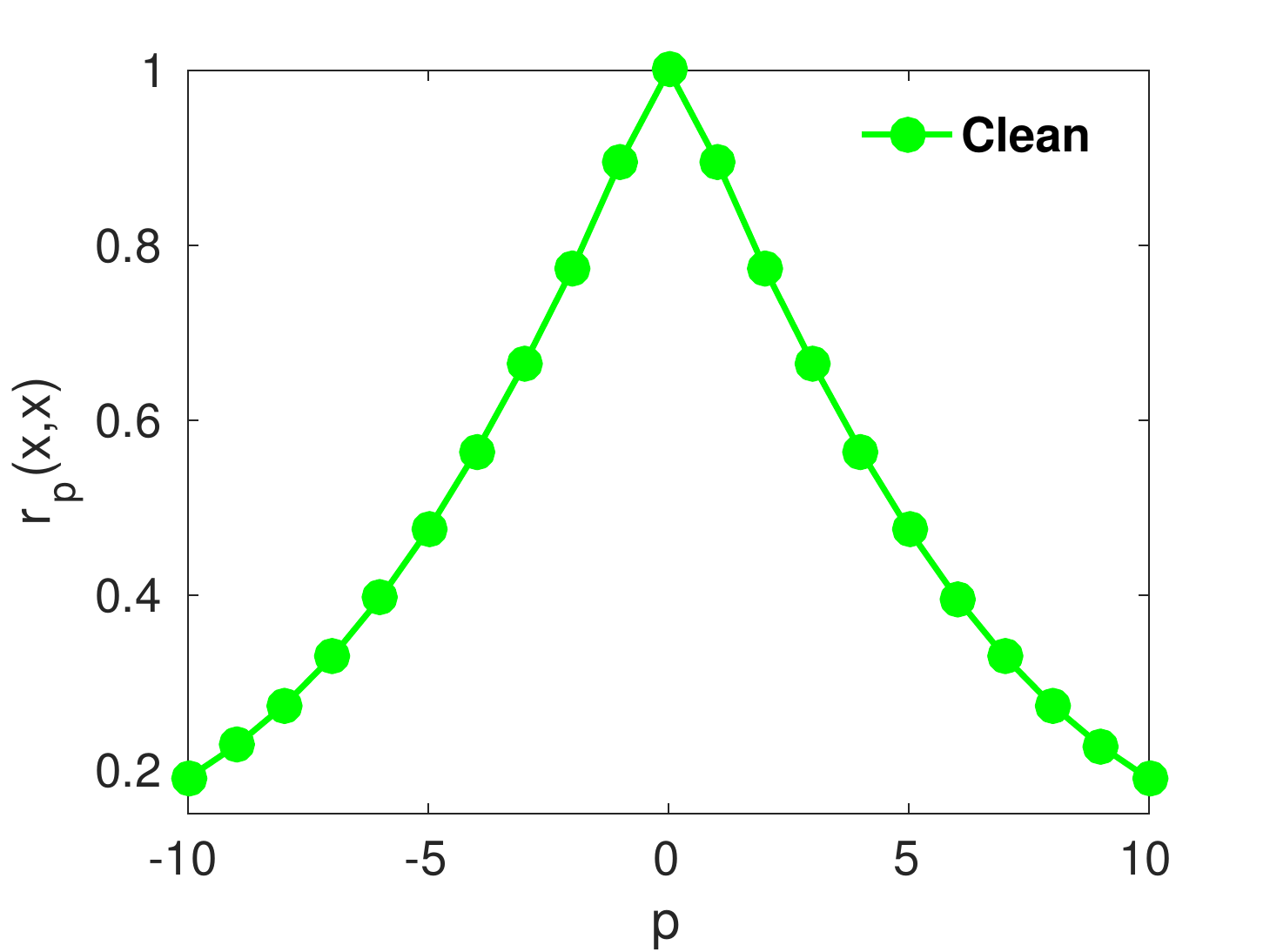}
\caption{Close-talking scenario (Clean)}
\label{fig:clean_fea}
\end{subfigure} \hspace{0.0\textwidth}
\begin{subfigure}{0.50\textwidth}
\includegraphics[scale=0.55]{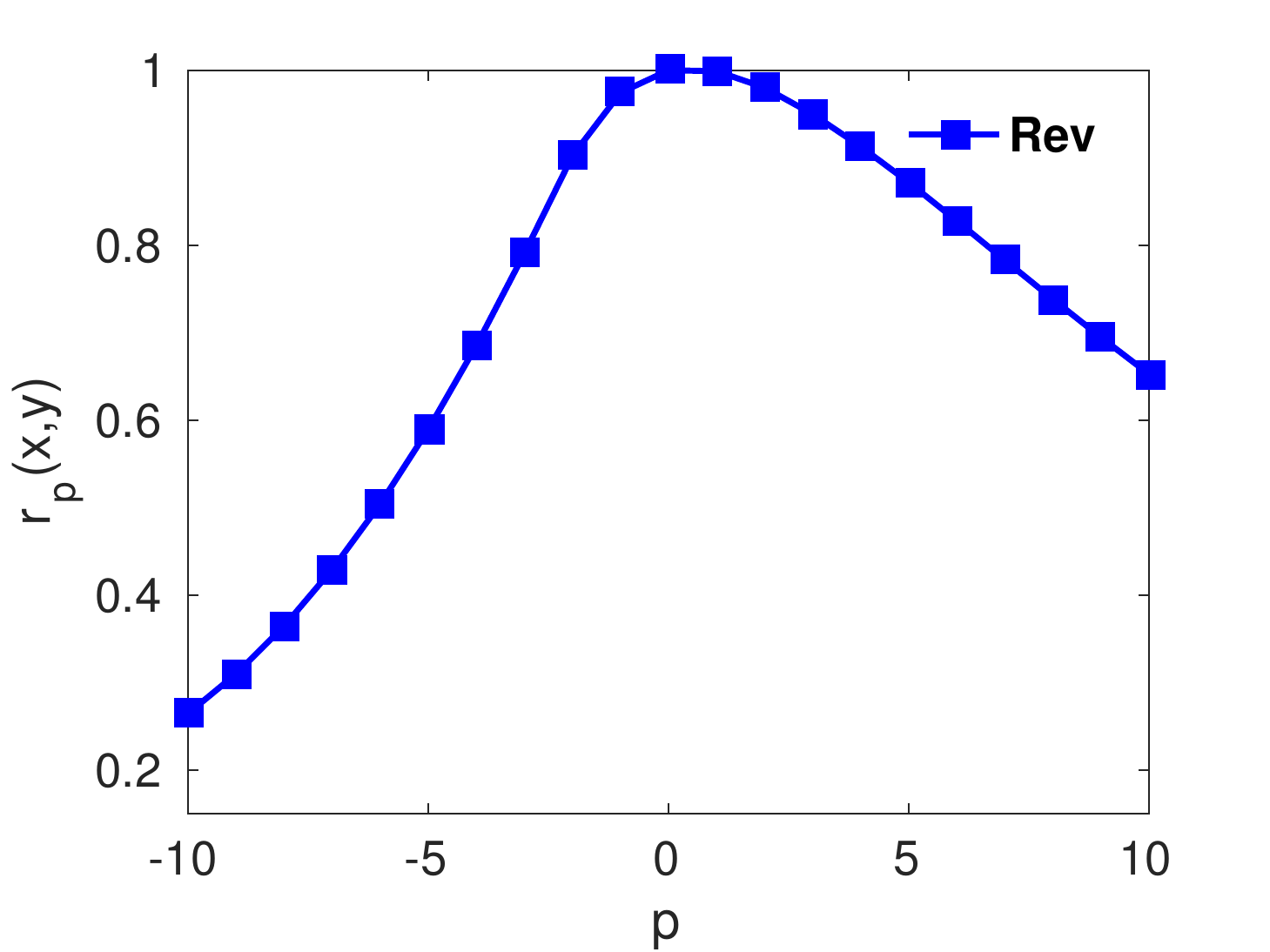}
\caption{Distant-talking scenario (Rev)}
\label{fig:rev_fea}
\end{subfigure}
\caption{Pearson Correlation $r_{p}$ across the various frames composing the context windows.}
\label{fig:fea_an}
\end{figure}

We computed the Pearson correlation using the DIRHA-English-WSJ5k dataset (simulated part of the set1 portion, see App. \ref{app:corpora}), considering 13 static MFCC features for both the close-talking recordings and the corresponding reverberated simulations. Note that simulations were generated by compensating time delays with the close-talking recordings.
The Pearson coefficient $r_{p}$, is reported in Figure  \ref{fig:fea_an} averaging the  correlation for all the sentences of this dataset. In particular, Figure  \ref{fig:clean_fea} reports the auto-correlation $r_{x,x}(p)$ obtained for the close-talking case. From the figure it is clear that in the close-talking case a symmetric trend is obtained, showing that past and future frames are equally correlated. Figure  \ref{fig:rev_fea} reports the Pearson cross-correlation coefficients $r_{x,y}(p)$ between close- and distant-talking signals.

\label{sec:fea_an}

 In this case, an asymmetric correlation trend is observed, highlighting that future frames ($p>0$) are consistently more correlated than past ones ($p<0$). 

Similarly to what observed for the signal cross-correlation $R_{xy}[n]$, the correlation unbalance of the Pearson coefficients further confirms the potential of the proposed asymmetric context window.

\subsection{DNN Weight Analysis} \label{sec:acw_inside_dnn}

Another interesting way to analyze the effect of reverberation is to inspect the DNN weights. In particular, we are interested in studying whether the network is able to automatically assign different importance to the different frames composing the context window.  With this regard, it is helpful to analyze the weights connecting the input frames with the first-layer neurons. If one frame is considered important by the network, we expect that, on average, the corresponding weights have a higher magnitude. On the other hand, if the network assigns little importance to a certain input, its weight connections should have a magnitude close to zero.  This directly stems from the role played by the gradient in the context of the SGD optimization. For a weight $w_{p,i,j}$ connecting the i-th features of the p-th frame with the j-th neurons (see Figure  \ref{fig:dnn_context}), the SGD updates are computed with the following equation:
\begin{equation}
w_{p,i,j}=w_{p,i,j} -\eta \frac{\partial C}{\partial w_{p,i,j}}
\end{equation}
where $\eta$ is the learning rate and C is the considered cost function (e.g., cross-entropy).
Large gradients $\frac{\partial C}{\partial w_{p,i,j}}$ are typically attributed to impactful weights, whose little perturbation produces larger effects on the final cost function C. 
On the contrary, weights with little impact on the cost function will have, on average, smaller gradients and their values will not deviate significantly from their original close-to-zero initialization.

Relying on this characteristic, we can define the importance $I_{p}$ for the p-th frame of the context window in the following way:
\begin{equation}
I_{p}=\sum_{i=1}^{N_{fea}}\sum_{j=1}^{N_{neu}} w_{p,i,j}^2
\label{eq:imp}
\end{equation}
where $N_{fea}$ is the number of features for each frame, and $N_{neu}$ is the number of neurons of the first hidden layer.

In this work we have first trained a DNN composed of six hidden layers with the WSJ dataset. Then, we computed the frame importance on the weights connecting the input features with the first hidden layer. This operation has been performed for both the standard close-talking version of the WSJ dataset, and for a corresponding reverberated version generated with impulse responses deriving from the DIRHA apartment (see App. \ref{app:mtc_s1} for more details).  

\begin{figure}[t!]
\begin{subfigure}{0.50\textwidth}
\includegraphics[scale=0.55]{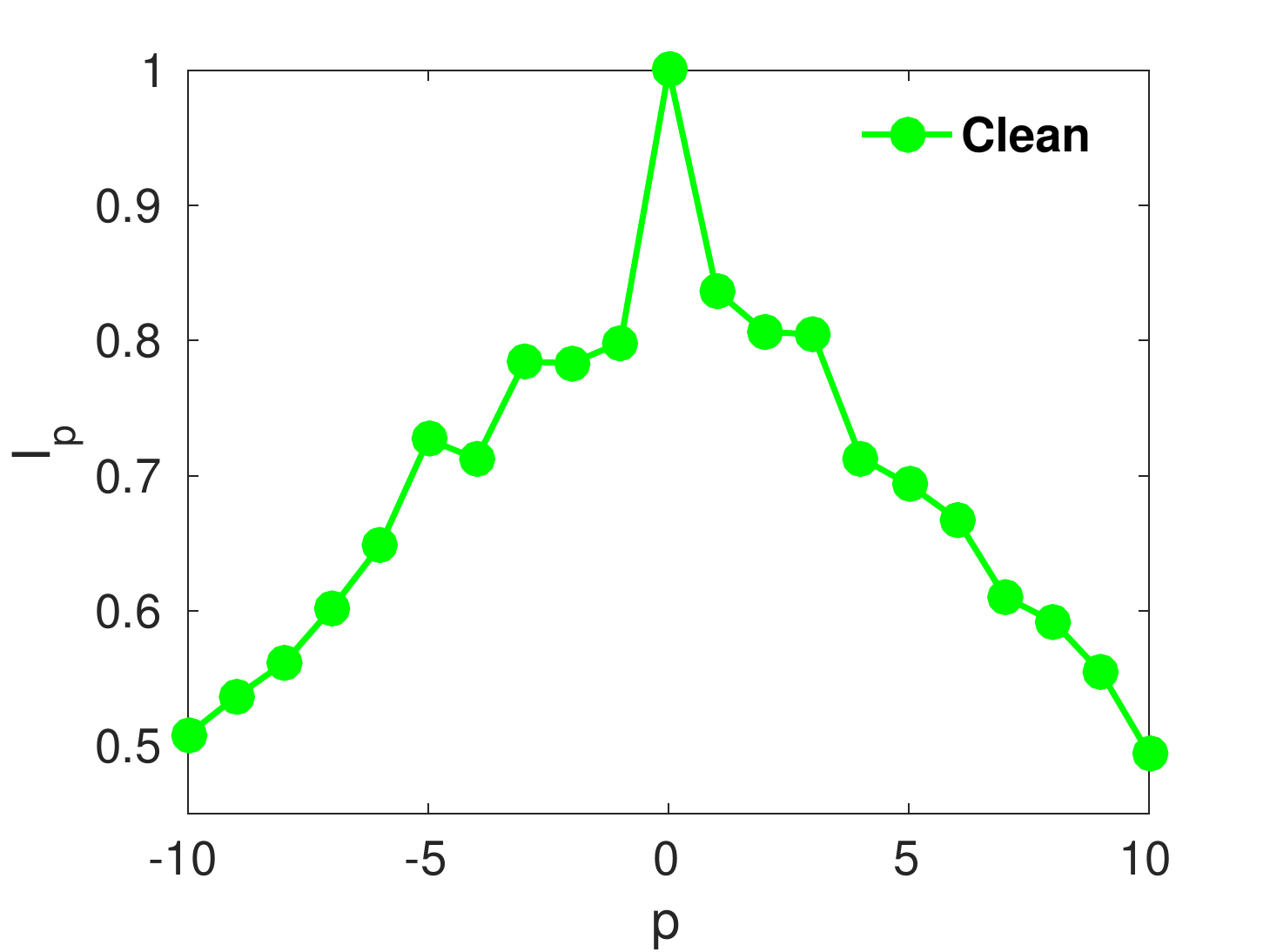}
\caption{Close-talking scenario (Clean)}
\label{fig:inside_dnn_clean}
\end{subfigure} \hspace{0.0\textwidth}
\begin{subfigure}{0.50\textwidth}
\includegraphics[scale=0.55]{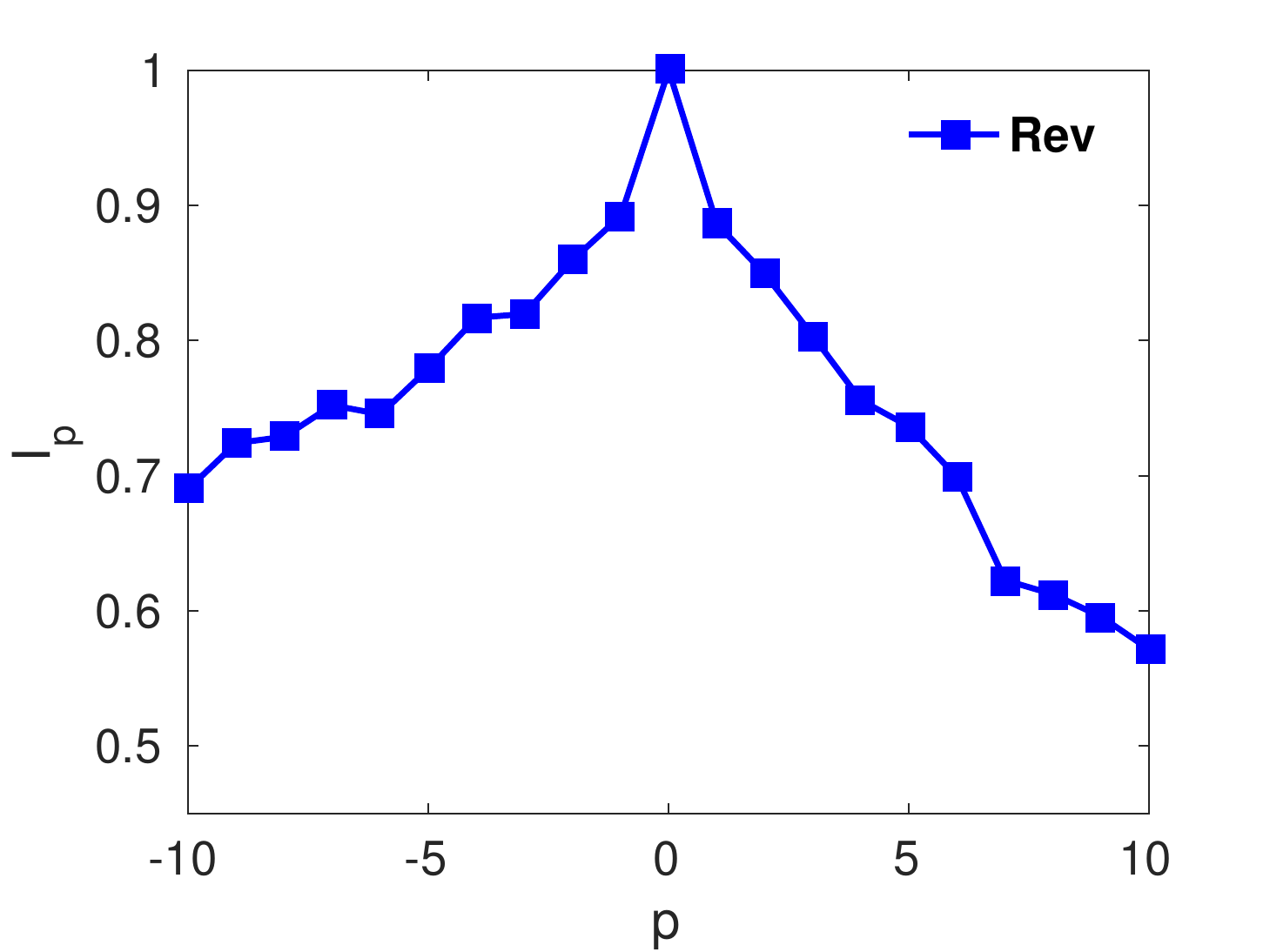}
\caption{Distant-talking scenario (Rev)}
\label{fig:inside_rev}
\end{subfigure}
\caption{Frame Importance $I_{p}$ obtained by analyzing the DNN weights connecting the various features frames of the context window.}
\label{fig:inside_dnn}
\end{figure}

The frame importance $I_{p}$ is reported in  Figure  \ref{fig:inside_dnn} for both the aforementioned close-talking (Figure  \ref{fig:inside_dnn_clean}) and distant-talking conditions (Figure  \ref{fig:inside_rev}). $I_{p}$ is normalized by its maximum value for representational convenience. The index $p=0$ represents the current frame, while negative and positive values of p refer to past and future frames, respectively. 

This experiment highlights that the network, as expected, is able to automatically assign more importance to the current frame (p=0). In both close and distant-talking conditions the importance $I_{p}$ clearly decreases when progressively moving far away from the current frame. Interestingly enough, in the reverberated case the network learns to place more importance to past frames ($p<0$) rather than to future frames ($p>0$). This can be readily appreciated from the asymmetric trend achieved in Figure  \ref{fig:inside_rev}, which is a further indication of the possible benefits deriving from using asymmetric time contexts. In fact, with a proper asymmetric context, we can directly feed the network with the most important frames, avoiding to overload it with useless information. Note that a very different trend is obtained with close-talking data (see Figure  \ref{fig:inside_dnn_clean}). In the latter case, indeed, the network shows once again no clear preference for past or future information.

\subsection{ASR experiments}  \label{sec:acw_dsr_performance}
In the previous section, we found that the distant-talking DNN tends to naturally attribute more importance to past rather than future frames. In this section, we take a step forward by verifying whether this fact is observed even in terms of recognition performance. 

The following part of this section reports the ASR experiments performed with different context window settings, features and architectures. We will also provide experimental evidence in different acoustic environments as well as under mismatching conditions.

The following experimental activity considered the WSJ dataset for training and the DIRHA-English-WSJ (set 1 portion) for test purposes. 
Close-talking, reverberated, and reverberated+noise versions of these corpora were considered. The experiments were based on a standard DNN composed of six hidden layers of 2048 neurons. Please, refer to the appendix for a more detailed description of the DIRHA-English-WSJ5k dataset
and for the related baselines (App. \ref{app:corpora}). For more information on the adopted experimental setup, please refer to App. \ref{app:mtc_s1}).

\subsubsection{ASR performance with different context settings}
\begin{figure}[t!]
\begin{subfigure}{0.50\textwidth}
\includegraphics[scale=0.55]{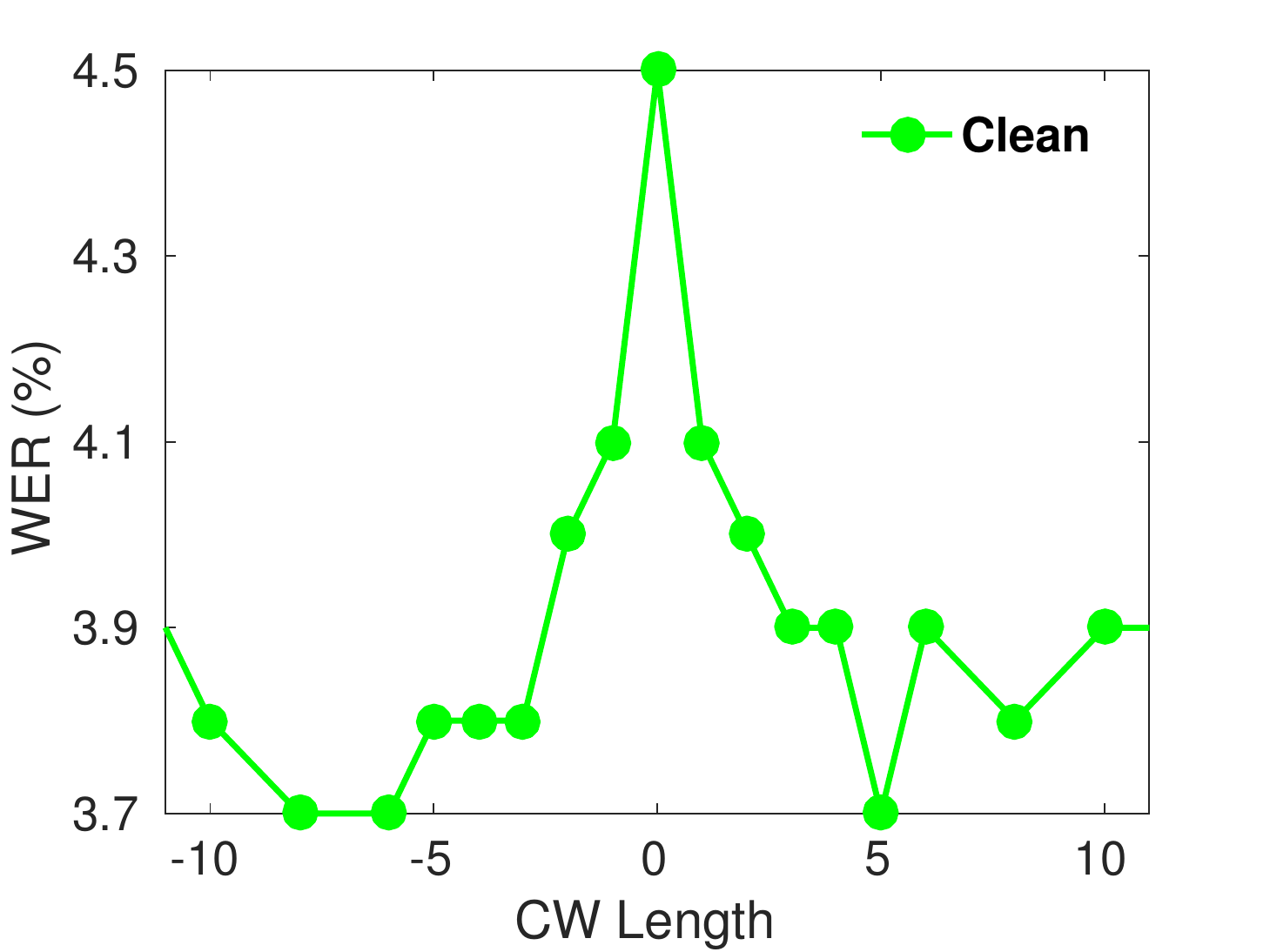}
\caption{Close-talking scenario (Clean)}
\label{fig:past_future_clean}
\end{subfigure} \hspace{0.0\textwidth}
\begin{subfigure}{0.50\textwidth}
\includegraphics[scale=0.55]{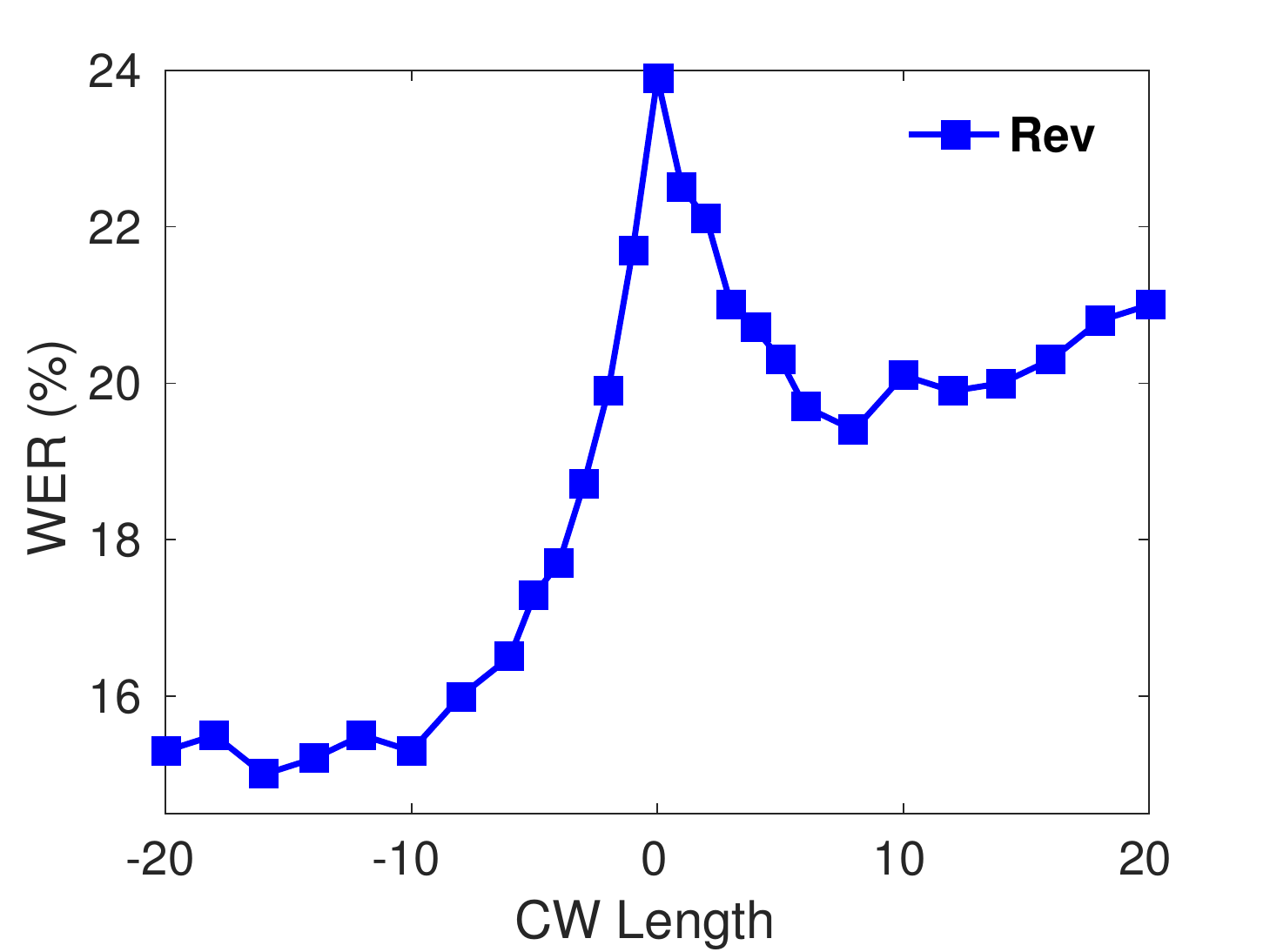}
\caption{Distant-talking scenario (Rev)}
\label{fig:past_future_rev}
\end{subfigure}
\caption{WER(\%) obtained by DNN context windows that progressively integrate only past or future frames. Training is performed with a contaminated version of the WSJ dataset (WSJ-rev), while test is based on the DIRHA-English-WSJ5k (set1, simulated part).}
\label{fig:past_future}
\end{figure}

As a first experiment, we examined the performance obtained in close-talking (Clean) and reverberated (Rev) conditions when fully asymmetric (i.e., single side) context windows are used. Figure  \ref{fig:past_future} shows the outcome of these experiments, reporting results in terms of  Word Error Rate (WER). Negative x-axis refers to the progressive integration of past frames only, while positive x-axis refers to future frames. In this set of experiments, fMLLR features are employed.

Results highlight that in a close-talking framework a rather symmetric behaviour is attained (Figure  \ref{fig:past_future_clean}), reiterating that in such contexts past and future information provide a similar contribution to improve the system performance. 
Differently, the role of past information is significantly more important in a distant-talking case, since a faster decrease of the WER(\%) is observed when past frames are progressively concatenated (Figure  \ref{fig:past_future_rev}). 
This result is in line with the findings emerged in the previous sections, and it confirms that an asymmetric context window is more suitable than a traditional symmetric one when reverberation arises.

\begin{figure*}[t]
\centering
  \includegraphics[scale=0.65]{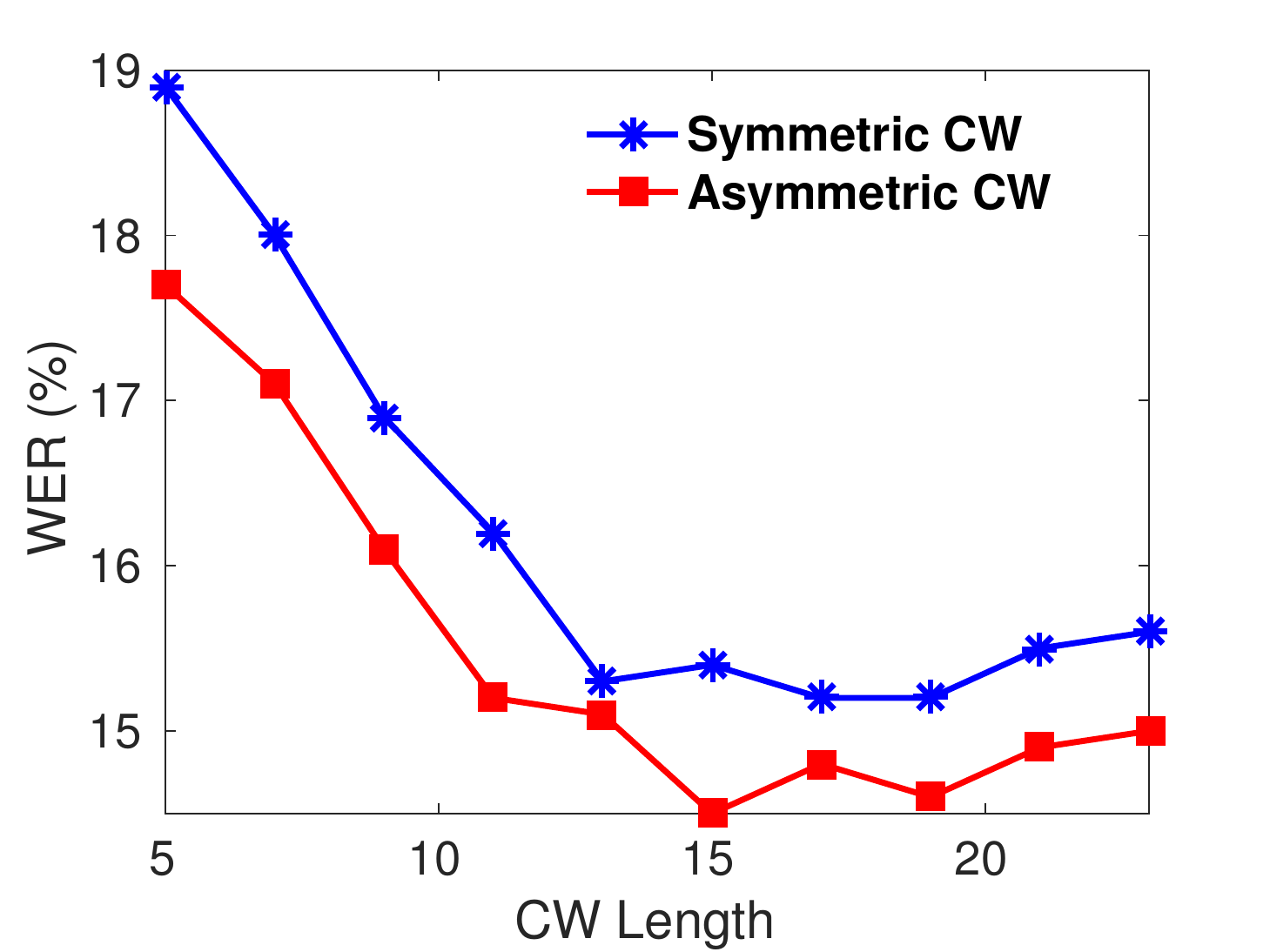}
\caption{Comparison between symmetric and asymmetric windows of different durations in a reverberated distant-talking scenario (using the set1-rev of the DIRHA-English-WSJ5k).}\label{fig:acw_scw}
\end{figure*}

In the latter experiment, we tested only fully asymmetric windows with $\rho_{cw}$=0\% (future frames) or $\rho_{cw}=100\%$ (past frames). However, it might be of interest to study more fuzzy configurations, where both past and future frames are considered. With this purpose, Figure  \ref{fig:acw_scw} compares this kind of asymmetric windows with standard symmetric contexts of various lengths. 

 
From this experiment, it emerges that the asymmetric window consistently outperforms the standard symmetric one for all the considered context durations. On average, about 5\%  relative improvement of the WER is obtained with essentially no computational cost.
This result refers to $\rho_{cw}=60\div65\%$, which was optimal for all the context durations.
Note that the optimal $\rho_{cw}$ could also be predicted by simply analyzing the local minima emerged in Figure  \ref{fig:past_future_rev}, i.e., the points where no longer benefits are observed when adding new frames ($N_{p}=11$, $N_{f}=7$, $\rho_{cw}=61\%$). The same trend is confirmed with other $\rho_{cw}$ ranging from $55\div90\%$.


\subsubsection{ASR performance with different features and architectures}
To further validate the benefits of the asymmetric context window, we performed additional experiments with also FBANK and MFCC coefficients. Moreover we also studied the proposed context window on a standard CNN, that was composed of two convolutional layers followed by four fully-connected layers (see App. \ref{app:mtc_s1} for more details).

\begin{table}[t!]
\centering
\tabcolsep=0.50cm
    \begin{tabular}{  | l | c | c | c | c | c |}
    \cline{1-4}
Architecture & Features & SCW (9-1-9) & ACW (11-1-7) \\ \hline
DNN & fMLLR & 15.2 & \textbf{14.8} \\ \hline
DNN & MFCC & 21.8 & \textbf{20.8} \\ \hline
DNN & FBANK & 20.7 & \textbf{20.2} \\ \hline
CNN & FBANK & 18.5 & \textbf{18.1 }\\ \hline

\end{tabular}
\caption{Comparison between the WER(\%) achieved with symmetric (SCW) and asymmetric context window (ACW) when different features and DNN architectures are used (using set1-rev of DIRHA-English-WSJ5k). The window configuration is reported in brackets with the following format: $N_{p}$-1-$N_{f}$.}
\label{tab:fea}
\end{table}
Results of Table  \ref{tab:fea} confirm that the asymmetric context window outperforms the symmetric one in all the considered settings. Although the best performance is achieved with fMLLR features, the computation of such coefficients is not compliant with real-time constraints, since an additional decoding step performed with a HMM-GMM acoustic model is required. For real-time applications, standard MFCCs or FBANKs thus remain the most viable choice. The last row of Tab \ref{tab:fea} highlights an interesting performance improvement achieved with CNNs. CNNs, which represent a valid alternative to fully-connected DNNs, are based on local connectivity, weight sharing, and pooling operations that allow them to exhibit some invariance to small feature shifts along the frequency axis, with well-known benefits against speaker and environment variations \cite{cnn1}.

\subsubsection{ASR performance under mismatching conditions}
\label{sec:mis}
In the previous experiments, training and test have been performed on the same environment with similar acoustic conditions.
In real applications, however, DSR systems often operate under mismatching conditions.
\begin{figure}[t!]
\begin{subfigure}{0.50\textwidth}
\includegraphics[scale=0.55]{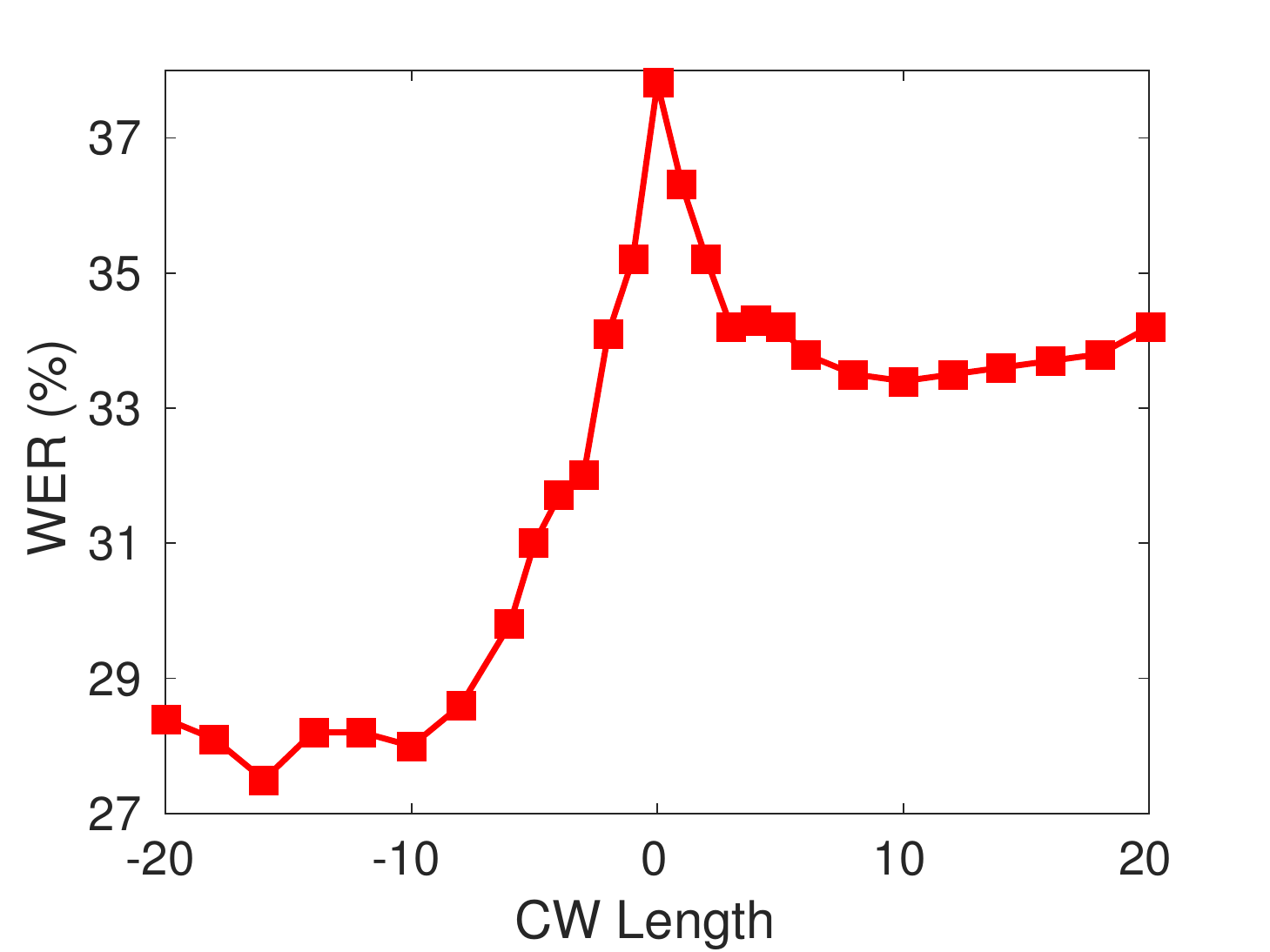}
\caption{Fully asymmetric context}
\label{fig:ab_revnoise}
\end{subfigure} \hspace{0.0\textwidth}
\begin{subfigure}{0.50\textwidth}
\includegraphics[scale=0.55]{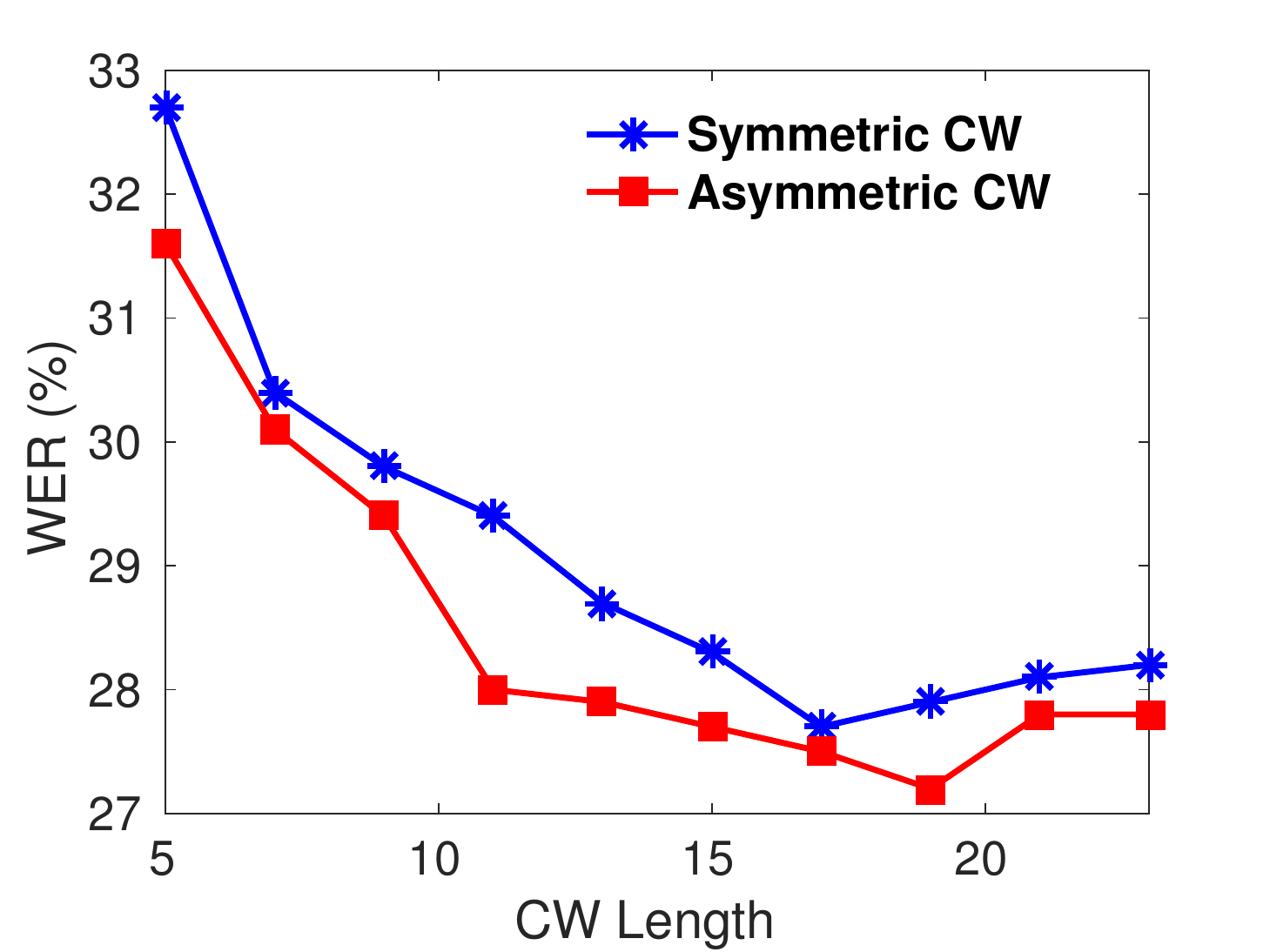}
\caption{Symmetric vs asymmetric window}
\label{fig:acw_revnoise}
\end{subfigure}
\caption{Comparison between symmetric and the proposed asymmetric context window under mismatching conditions. Training is performed with the WSJ-rev data, while test is performed on the set1-real (rev\&noise) part of the DIRHA-English Dataset.}
\label{fig:acw}
\end{figure}
As a first experiment to test the latter situation, we train the DNN with the reverberated data (Rev) so far considered, while test is performed with the DIRHA-English data corrupted by both noise and reverberation (DIRHA-English-rev\&noise).
Figure  \ref{fig:ab_revnoise} shows the results obtained when fully asymmetric context windows are adopted. Figure  \ref{fig:acw_revnoise}, instead, compares symmetric and asymmetric time contexts of different duration with the optimal $\rho_{cw}=60\div65$.  



Due to the more challenging conditions, the WER(\%) is significantly worse than that highlighted in Figure  \ref{fig:past_future_rev} and Figure  \ref{fig:acw_scw}. 
However, it is worth noting that the benefits deriving from using asymmetric contexts are maintained even under the addressed mismatching case.

Another mismatch typology might occur when training and test are performed in different acoustic environments.
In the experiment reported in Table  \ref{tab:mism}, training is carried out in the living-room of the DIRHA apartment so far considered, while test is performed in different environments, i.e., an office, a surgery room, as well as a room of another apartment. The test data were generated following our data simulation approach, but using impulse responses measured in the aforementioned environments. For these experiments, we inherit a context length of 19 frames, that resulted optimal in previous studies. 

\begin{table}[t!]
\centering
\tabcolsep=0.50cm
    \begin{tabular}{  | l | c | c | c | c | c |}
    \cline{1-4}
Environment & $T_{60}$ (ms) & SCW (9-1-9) & ACW (11-1-7) \\ \hline
Office  & 650 ms & 16.6 & \textbf{16.2} \\ \hline
Home  & 700 ms & 19.5 & \textbf{19.1} \\ \hline
Surgery Room  & 850 ms & 21.4 & \textbf{20.3} \\ \hline
\end{tabular}
\caption{WER(\%) obtained with symmetric (SCW) and asymmetric (ACW) context windows in different acoustic environments and under mismatching conditions.}
\label{tab:mism}
\end{table}
Results reveal that all the various testing environments take benefit from the use of the asymmetric contexts, even when the training is performed in a different acoustic environment. 

\subsubsection{ASR performance with different reverberation times}
\label{sec:t60}
As a last experiment, we further extend our validation by simulating acoustic environments with different reverberation times $T_{60}$. For this study, a set of impulse responses simulated with the directional image method described in Sec. \ref{sec:im_method} are used to contaminate both training (WSJ-clean) and testing corpora (DIRHA-WSJ-clean). Table  \ref{tab:time} summarizes the optimal results obtained with $T_{60}$ ranging from 250 ms to 1000 ms. 
\begin{table}[t!]
\centering
\tabcolsep=0.50cm
    \begin{tabular}{  | l | c | c | c | c | c |}
    \cline{1-5}
\multicolumn{2}{ | c | }{ $T_{60}$(ms)} & SCW & ACW & ACW*  \\ \hline

\multirow{2}{*}{250 ms} & CW Composition & \textit{ 5-1-5} & \textit{6-1-4} & \textit{4-1-6} \\  \cline{2-5}
                   & WER   & 5.5 & \textbf{5.1} & 5.7 \\ \hline
           
 \multirow{2}{*}{500 ms} & CW Composition & \textit{6-1-6} & \textit{7-1-5} & \textit{5-1-7} \\  \cline{2-5}
                    & WER  & 9.1 & \textbf{8.5} & 9.4 \\ \hline

 \multirow{2}{*}{750 ms} & CW Composition  & \textit{9-1-9} & \textit{13-1-5} & \textit{5-1-13} \\  \cline{2-5}
                    & WER  & 15.2 & \textbf{14.8} & 16.8 \\ \hline
                      
 \multirow{2}{*}{1000 ms} & CW Composition & \textit{12-1-12} & \textit{18-1-6} & \textit{6-1-18} \\  \cline{2-5}
                    & WER  & 20.5 & \textbf{20.1} & 23.2 \\ \hline
                      
    \end{tabular}
\caption{Comparison between WER(\%) obtained with symmetric (SCW) and asymmetric (ACW) context window under different reverberation conditions. The ACW* column reports the results obtained with an asymmetric window that integrates more future than past frames.}
\label{tab:time}
\end{table}

As expected, results show that the performance progressively degrades as $T_{60}$ increases. The proposed solution, however, is able to overtake standard symmetric windows in all the different reverberation conditions considered here. It is also worth noting that larger context windows are needed when passing from simple acoustic conditions to more challenging environments. This derives from additional optimization experiments that showed, for instance, that when $T_{60}$=250 ms the optimal window integrates only 11 frames, while 25 frames are necessary when $T_{60}$=1000 ms. For the sake of completeness, the last column of Table \ref{tab:time} ($ACW*$) reports the results obtained when, contrary to what proposed in this work, a context window embedding more future than past frames is considered. Results confirm once again that the latter choice is not optimal, since such a frame configuration integrates  redundant information when the reverberation time increases. This degree of redundancy can also be related to the relative performance loss from ACW to $ACW*$, which is ranging from 10\% to 15\%.

In this section we studied asymmetric context windows that turned out to be helpful for improving DSR performance thanks to better management of the redundancy introduced by reverberation. A noteworthy aspect is that these benefits are obtained without introducing additional computational costs.  Moreover, this approach minimizes the use of future frames,  making it very suitable for practical small-footprint and on-line ASR applications.
Although rather large contexts can be analyzed with this approach, one limitation lies in the relatively short-term analysis of speech modulations achievable with this methodology. The ASR performance, according to our experiments, starts to degrade when contexts larger than 150-200 ms are embedded. 

In the following sections we focus on architectures able to learn longer-term dependencies. We will initially recall the main state-of-the-art approaches and, finally, we  propose a novel RNN architecture.

\section{Managing Long-Term Dependencies} \label{sec:rnn_context}
To embed longer-term information, some DNN architectures have been proposed in the past. Remaining in the domain of feed-forward DNNs, an example are Time Delay Neural Networks (TDNNs) \cite{tdnn,tdnn2}, that consider a wide fixed-size context window to manage long-range temporal dependencies with a modular and incremental architecture. In standard DNNs, all the neurons of the first hidden layer are fed with the entire temporal context. Differently, in TDNNs the initial transformations are learnt on narrow contexts and the deeper layers process the hidden activations from a wider temporal context. 

Other popular approaches adopt a cascade of multiple DNNs for embedding larger contexts \cite{split_cw1,split_cw2,split_cw3}. TempoRAl PatternS (TRAPS), for instance, capture long-term speech modulations by progressively analyzing long temporal vectors of critical band energies \cite{IEEEexample:intro2}. More precisely, the TRAPS architecture consists of two MLPs: the first one performs a non-linear mapping from log critical band energy time trajectories to phonetic probabilities, while the second MLP combines these predictions to obtain the  overall  phonetic probabilities.
This method was able to successfully exploit contexts ranging from 500 ms to 1000 ms. A variant of TRAP is represented by the  Hidden Activation TRAPS (HAT) \cite{hat2,hat},  where the first stage outputs hidden activations rather than posterior probabilities. A paradigm that recently gained popularity is Hierarchical DNNs (H-DNN) \cite{IEEEexample:hbn2,IEEEexample:bn6,tb}, H-DNNs are based on a first DNN focusing  on short speech dependencies, while a second one works on a different time scale to capture long-term information. This approach was particularly useful in the context of the BABEL project\footnote{\url{https://www.iarpa.gov/index.php/research-programs/babel}},  where a bottleneck hidden layer in the first DNN was often used to derive language independent features \cite{IEEEexample:hbn3,IEEEexample:hbn2,IEEEexample:bn6}.

In the first part of this PhD study, some efforts were devoted to study possible variations of this hierarchical paradigm. One  contribution, described in \cite{tb}, was the study  of a novel H-DNN architecture, called multi-stream H-DNN. This variant, proposed as a joint work between FBK and ICSI-Berkeley, was tested and trained on spontaneous telephone speech conversations using the Cantonese IARPA Babel corpus  and turned out to be effective for improving the ASR performance. A second contribution, reported in \cite{ravanelli_eusipco}, was the extensions of this paradigm to an acoustic event classification task. To the best of our knowledge, our work, that was another result achieved in the context of the collaboration with ICSI, was the first exploring this paradigm in the latter domain. Please refer to  the referenced paper for more details.

The feed-forward DNNs discussed so far are still very popular and represent the most preferable choice when the computational power is limited or when there are real-time constraints. A typical application is thus low-latency/real-time speech recognition or small-footprint systems, as witnessed by the numerous studies in the literature \cite{small1,small2,small3,small4,small5,online2}. Nevertheless, when enough computational capabilities are available, the most suitable architecture to process long-term information is represented by Recurrent Neural Networks (RNNs). Thanks to their recurrent nature, RNNs can memorize and process very long time contexts, and are therefore potentially able to outperform standard feed-forward DNNs. RNNs have recently gained a lot of popularity in speech recognition and have been recently used in the context of both hybrid RNN-HMM and end-to-end speech recognizers. 

At the time of writing, the most popular RNN architecture able to learn long-term dependencies is LSTM, that exploits multiplicative gates to create gradient shortcuts.
General-purpose RNNs such as Long Short Term Memories (LSTMs) \cite{lstm} have been the subject of several studies and modifications over the past years \cite{peephole, lstm_odyssey, lstm_highway}, leading to some novel architectures, such as the recently-proposed  Gated Recurrent Units (GRUs) \cite{gru1}. Note that the asymmetric context window studied in the previous section, can also be used for modern GRU or LSTM architectures. However, some additional experiments not reported here, suggest that no benefits can be obtained with this approach. On the other hand, RNNs automatically learn how to exploit the time information through the recurrent connections, and the adoption of a context window is rather redundant and useless.

Instead, our work, that will be summarized in the next sections, attempts to improve the context management by further revising GRUs. Differently from previous efforts, our primary goal was not to derive a general-purpose RNN, but to modify the standard GRU design to better process time information for  speech recognition, in particular for noisy and reverberant speech inputs.


\section{Revising Gated Recurrent Units} \label{sec:li_gru}
LSTMs rely on memory cells that are controlled by forget, input, and output gates. Despite their effectiveness, such a sophisticated gating mechanism might result in an overly complex model. 

A noteworthy attempt to simplify LSTMs has recently led to a novel model called Gated Recurrent Unit (GRU) \cite{gru1,gru2}, that is based on just two multiplicative gates. 
In particular, the standard GRU architecture is defined by the following equations: 

\begin{subequations}
\begin{align}
z_{t}&=\sigma(W_{z}x_{t}+U_{z}h_{t-1}+b_{z}), \\
\label{eq:eq_2}r_{t}&=\sigma(W_{r}x_{t}+U_{r}h_{t-1}+b_{r}), \\
\label{eq:eq_3}\widetilde{h_{t}}&=\tanh(W_{h}x_{t}+U_{h}(h_{t-1} \odot r_{t})+b_{h}), \\
\label{eq:eq_4}h_{t}&=z_{t} \odot h_{t-1}+ (1-z_{t}) \odot \widetilde{h_{t}}.
\end{align}
\end{subequations}

where $z_{t}$ and $r_{t}$ are vectors corresponding to the update and reset gates, respectively, while $h_{t}$ represents the state vector for the current time frame $t$.
Computations denoted as $\odot$ represent element-wise multiplications.
The activations of both gates are element-wise logistic sigmoid functions $\sigma(\cdot)$, that constrain $z_{t}$ and $r_{t}$ to take values ranging from 0 and 1. The candidate state $\widetilde{h_{t}}$ is processed with a hyperbolic tangent. 
The network is fed by the current input vector $x_{t}$ (e.g., a vector of speech features), while the parameters of the model are the matrices $W_z$, $W_r$, $W_h$ (the feed-forward connections) and $U_z$, $U_r$, $U_h$ (the recurrent weights).
The architecture finally includes trainable bias vectors $b_z$, $b_r$ and $b_h$, that are added before the non-linearities are applied. 

As shown in Eq.~\ref{eq:eq_4}, the current state vector $h_{t}$ is a linear interpolation between the previous activation $h_{t-1}$ and the current candidate state $\widetilde{h_{t}}$. The weighting factors are set by the update gate $z_{t}$, that decides how much the units will update their activations. Note that this linear interpolation, which is similar to the gating in LSTMs \cite{lstm}, is the key component for learning long-term dependencies. For instance, if $z_{t}$ is close to one, the previous state is kept unaltered and can remain unchanged for an arbitrary number of time steps. On the other hand, if $z_{t}$ is close to zero, the network tends to favor the candidate state $\widetilde{h_{t}}$, that depends more heavily on the current input and on the closer hidden states. The candidate state $\widetilde{h_{t}}$ also depends on the reset gate $r_{t}$, that allows the model to possibly delete the past memory by forgetting the previously computed states. 

Despite the interesting performance achieved by GRUs that, according to several works in the literature   \cite{gru2,gru3,gru4}, is comparable to LSTMs in different machine learning tasks, in this thesis we tried to further simplify this standard model, deriving a GRU architecture able to better process long speech contexts. In particular, the main changes to the standard GRU model concern the reset gate, ReLU activations, and batch normalization, as outlined in the next sub-sections.

\subsection{Removing the reset gate}
From the previous introduction to GRUs, it follows that the reset gate can be useful when significant discontinuities occur in the sequence. For language modeling, this may happen when moving from one text to another that is not semantically related. In such situation, it is convenient to reset the stored memory to avoid taking a decision biased by an unrelated history. 
 \begin{figure}[t!]
 \centering
 \includegraphics[width=0.75\textwidth]{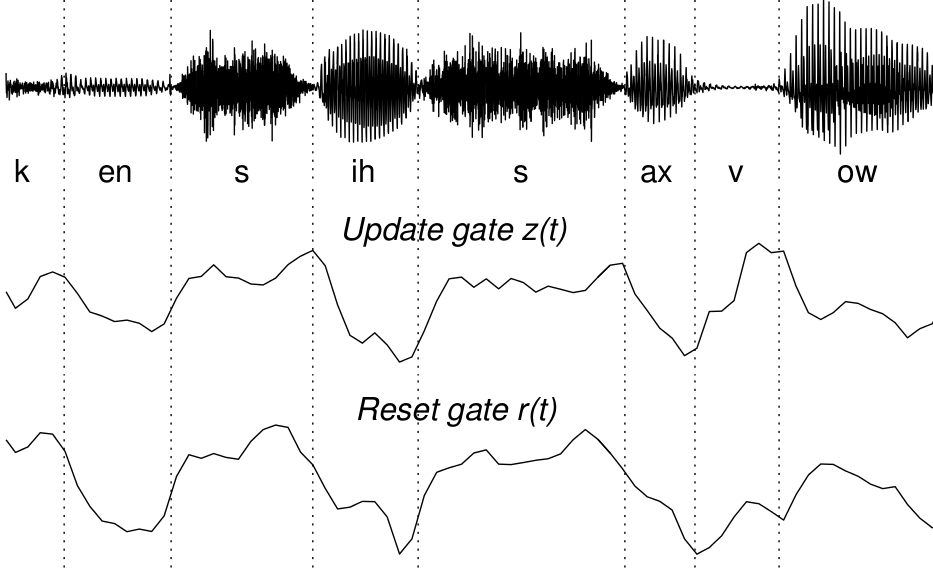}
 \caption{Average activations of the update and reset gates for a GRU trained on TIMIT in a chunk of the utterance ``\textit{sx403}" of speaker ``\textit{faks0}". The activations are within the rage 0.7$\div$0.3 for both gates.}
 \label{fig:im1}
 \end{figure}

Nevertheless, we believe that for some specific tasks like speech recognition this functionality might not be useful. 
In fact, a speech signal is a sequence that evolves rather slowly (the features are typically computed every 10 ms), in which the past history can virtually always be helpful.  
Even in the presence of strong discontinuities, for instance observable at the boundary between a vowel and a fricative, completely resetting the past memory can be harmful. On the other hand, it is helpful to memorize phonotactic features, since some phone transitions are more likely than others.

We also argue that a certain redundancy in the activations of reset and update gates might occur when processing speech sequences.  For instance, when it is necessary to give more importance to the current information,  the GRU model can set small values of $r_{t}$. A similar effect can be achieved with the update gate only, if small values are assigned to $z_{t}$. The latter solution tends to weight more the candidate state $\widetilde{h_{t}}$, that depends heavily on the current input. 
Similarly, 
a high value can be assigned either to $r_{t}$ or to $z_{t}$, in order to place more importance on past states. 
This redundancy is also highlighted in Figure  \ref{fig:im1}, where a temporal correlation in the average activations of update and reset gates 
can be readily appreciated for a GRU trained on TIMIT. 

Based on these reasons, the first variation to standard GRUs thus concerns the removal of the reset gate $r_{t}$. 
This change leads to the following modification of Eq. \ref{eq:eq_3}:

\begin{equation}
\widetilde{h_{t}}=\tanh(W_{h}x_{t}+U_{h} h_{t-1}+b_{h})
\end{equation}

The main benefits of this intervention are related to the improved computational efficiency, that is achieved thanks to a more compact single-gate model. Note that the removal of the reset gate does not prevent the architecture to learn long-term dependencies. The weights of the linear interpolation, in fact, are determined by the update gate only, and  even avoiding the reset gate, the aforementioned shortcut used to mitigate gradient vanishing problems is still preserved. 

\subsection{ReLU activations}
As outlined in Chapter \ref{cha:dl}, ReLU activations are very popular for feed-forward DNNs. 
We believe that extending the well-known benefits of ReLU to RNNs is of great importance, since it might help mitigating vanishing gradient issues and fostering a better learning of long-term dependencies.  

In this work, our second modification thus consists in replacing the standard hyperbolic tangent of standard GRU with ReLU activation.
In particular, we modify the computation of candidate state $\widetilde{h_{t}}$ (Eq.~\ref{eq:eq_3}), as follows:

\begin{equation}
\widetilde{h_{t}}=\mbox{ReLU}(W_{h}x_{t}+U_{h}h_{t-1}+b_{h})
\end{equation}

The adoption of ReLU-based neurons was not so common in the past for the RNN architectures. This was due to numerical instabilities originating from the unbounded ReLU functions applied over long time series. In the proposed architecture we were able to circumvent these numerical issues coupling it with batch normalization, as will be discussed in the next sub-section.




\subsection{Batch Normalization} \label{sec:bn}
Batch normalization \cite{batchnorm} has been recently proposed in the machine learning community and addresses the so-called \textit{internal covariate shift} problem by normalizing the mean and the variance of each layer's pre-activations for each training mini-batch. Several works have already shown that this technique is effective both to improve the system performance and to speed-up the training procedure \cite{cesar, baidu,ravanelli_SLT,ravanelli_icassp}. Batch normalization can be applied to RNNs in different ways. In \cite{cesar}, the authors suggest to apply it to feed-forward connections only, while in \cite{initbn} the normalization step is extended to recurrent connections, using separate statistics for each time-step.  In our work, we tried both approaches, but  we did not observe substantial benefits when extending batch normalization to recurrent parameters (i.e., $U_{h}$ and $U_{z}$).  For this reason, we applied this technique to feed-forward connections only (i.e., $W_{h}$ and $W_{z}$), obtaining a more compact model that is almost equally performing but significantly less computationally expensive. When batch normalization is limited to feed-forward connections, indeed, all the related computations become independent at each time step and they can be performed in parallel. This offers the possibility to apply it with reduced computational efforts.  As outlined in the previous sub-section, coupling the proposed model with batch-normalization \cite{batchnorm} could also help in limiting the numerical issues of ReLU RNNs. Batch normalization, in fact, rescales the neuron pre-activations, inherently bounding the values of the ReLU neurons. 
In this way, our model concurrently takes advantage of the well-known benefits of both ReLU activation and batch normalization.
In our experiments, we found that the latter technique helps against numerical issues also when it is limited to feed-forward connections only.

Formally, removing the reset gate, replacing the hyperbolic tangent function with the ReLU activation, and applying batch normalization, now leads to the following model:

\begin{subequations}
\begin{align}
\label{eq:eq_5a}&z_{t}=\sigma(BN(W_{z}x_{t})+U_{z}h_{t-1}), \\
\label{eq:eq_5b}&\widetilde{h_{t}}=\mbox{ReLU}(BN(W_{h}x_{t})+U_{h}h_{t-1}), \\
\label{eq:eq_5c}&h_{t}=z_{t} \odot h_{t-1}+ (1-z_{t}) \odot \widetilde{h_{t}}.
\end{align}
\end{subequations}
The batch normalization $BN(\cdot)$ works as described in \cite{batchnorm}, and is defined as follows:

\begin{equation}
BN(a)=\gamma \odot
\label{eq:bn} \frac{a-\mu_b}{\sqrt[]{\sigma_b^2+\epsilon}}+\beta
\end{equation}
where $\mu_b$ and $\sigma_b$ are the minibatch mean and variance, respectively. A small constant $\epsilon$ is added for numerical stability. The variables $\gamma$ and $\beta$ are trainable scaling and shifting parameters, introduced to restore the network capacity. Note that the presence of $\beta$ makes the biases $b_h$ and $b_z$ redundant. Therefore, they are omitted in Eq. \ref{eq:eq_5a} and \ref{eq:eq_5b}.

We called this architecture Light GRU (Li-GRU), to emphasize the simplification process conducted on a standard GRU.

\subsection{Related work} \label{sec:related_work}
A first attempt to remove $r_{t}$ from GRUs has recently led to a single-gate architecture called Minimal Gated Recurrent Unit (M-GRU) \cite{mgru}, that achieves a performance comparable to that obtained by standard GRUs in handwritten digit recognition as well as in a sentiment classification task.  To the best of our knowledge, our contribution is the first attempt that explores this architectural variation in speech recognition.  Recently, some attempts have also been done for embedding ReLU units in the RNN framework. For instance, in \cite{orth_init} authors replaced tanh activations with ReLU neurons in a vanilla RNN, showing the capability of this model to learn long-term dependencies when a proper orthogonal initialization is adopted. In this work, we extend the use of ReLU to a GRU architecture.

In summary, the novelty of our approach consists in the integration of three key design aspects (i.e, the removal of the reset gate, ReLU activations and batch normalization) in a single model, that resulted particularly suitable for speech recognition. 

In the following sub-sections, we describe the experimental activity conducted to assess the proposed model. Most of the experiments reported in the following are based on hybrid DNN-HMM speech recognizers, since the latter ASR paradigm typically reaches state-of-the-art performance. However, for the sake of comparison, we also extended the experimental validation to an end-to-end CTC model.
More precisely, in sub-section \ref{sec:corr}, we first  quantitatively analyze the correlations between the update and reset gates in a standard GRU. In sub-section \ref{sec:grad}, we extend our study with some analysis of gradient statistics. The role of batch normalization and the CTC experiments are described in sub-sections \ref{sec:bn_exp} and \ref{sec:cts}, respectively. The speech recognition performance will then be reported for TIMIT,  DIRHA-English-WSJ, CHiME as well as for the TED-talk corpus. 
The complete experimental setup is described in App. \ref{app:mtc_s2}, where more details about DNN and ASR setups are reported.

\subsection{Correlation analysis} \label{sec:corr}
The degree of redundancy between reset and update gates can be analyzed in a quantitative way using the cross-correlation metric $C(z,r)$:
\begin{equation}
C(z,r)=\overline{z}_t \star \overline{r}_t
\label{eq:cross}
\end{equation}
where $\overline{z}_t$ and $\overline{r}_t$ are the average activations (over the neurons) of update and reset gates, respectively, and $\star$ is the cross-correlation operator.

\begin{figure}[t]
\centering
  \includegraphics[scale=0.75]{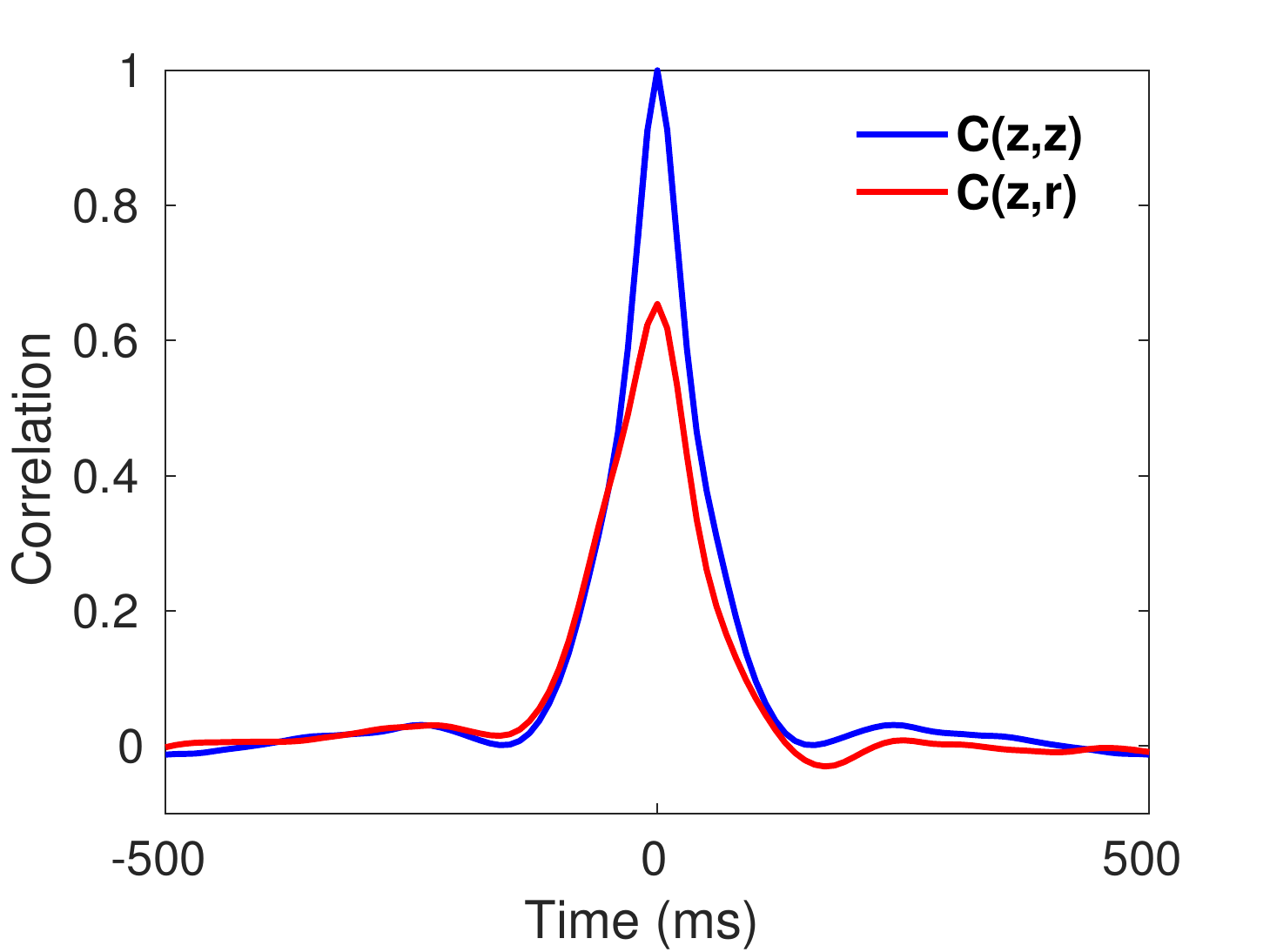}
\caption{Auto-correlation $C(z,z)$ and cross-correlation $C(z,r)$ between the average activations of the update (z) and reset (r) gates. Correlations are normalized by the maximum of $C(z,z)$ for graphical convenience.}\label{fig:corr}
\end{figure}

The cross-correlation $C(z,r)$ between the average activations of update $z$ and reset $r$ gates is shown in Figure ~\ref{fig:corr} for a standard GRU trained on the TIMIT dataset (see App. \ref{app:corpora}).
The gate activations are computed for all the input frames and, at each time step, an average over the hidden neurons is considered.
The cross-correlation $C(z,r)$ is displayed along with the auto-correlation $C(z,z)$, that represents the upper-bound limit of the former function.  

Figure  \ref{fig:corr} clearly shows a high peak of $C(z,r)$, revealing that update and reset gates end up being  redundant. This peak is about 66\% of the maximum of $C(z,z)$ and it is centered at $t=0$, indicating that almost no-delay occurs between gate activations.  This result is obtained with a single-layer GRU of 200 bidirectional neurons fed with MFCC features and trained with TIMIT. After the training-step, the cross-correlation is averaged over all the development sentences.

\begin{figure}[t]
\centering
  \includegraphics[scale=0.75]{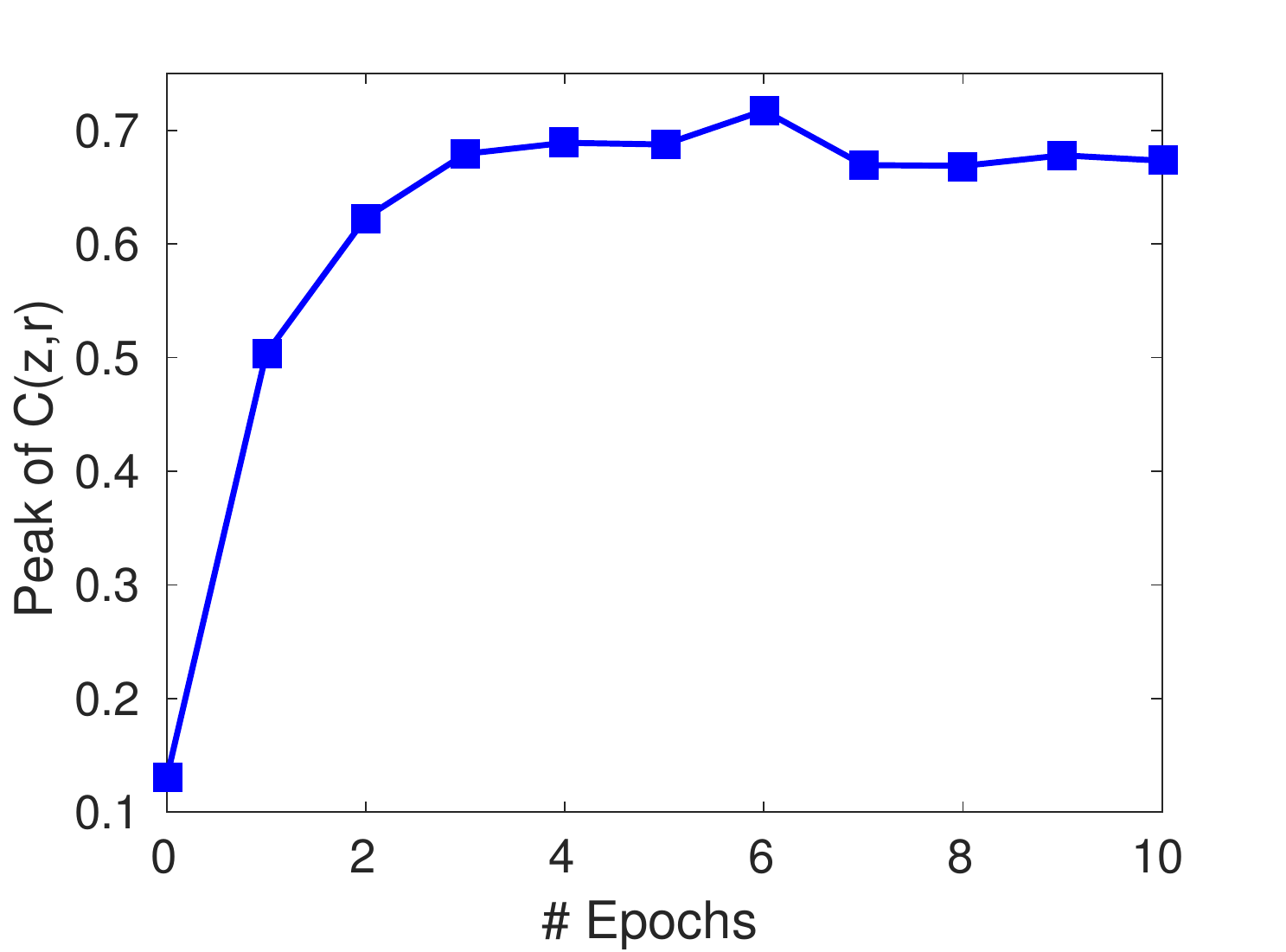}
\caption{Evolution of the  peak of the cross-correlation $C(z,r)$ over various training epochs.}\label{fig:corr2}
\end{figure}

It would be of interest to examine the evolution of this correlation over the epochs. With this regard, Figure  \ref{fig:corr2} reports the peak of $C(z,r)$ for some training epochs, showing that the GRU attributes rather quickly a similar role to update and reset gates. 
In fact, after 3-4 epochs, the correlation peak reaches its maximum value, that is almost maintained for all the subsequent training iterations.   

\subsection{Gradient analysis} \label{sec:grad}
The analysis of the main gradient statistics can give some preliminary  indications on the role played by the various parameters. 

With this goal, Table \ref{tab:grad} reports the L2 norm of the gradient for the main parameters of the GRU model considered in the previous section.  
\begin{table}[t!]
\centering
\tabcolsep=0.20cm
    \begin{tabular}{ | c | c | c | c | c | }
    \cline{1-4}
   {\backslashbox{\em{Param.}}{\em{Arch.}}} & GRU &  M-GRU & Li-GRU \\ \hline
$\|W_{h}\|$ & 4.80 & 4.84 & 4.86 \\ \hline
$\|W_{z}\|$ & 0.85 & 0.89 & 0.99 \\ \hline
$\|W_{r}\|$ & 0.22 & - & - \\ \hline
$\|U_{h}\|$ & 0.15 & 0.35 & 0.71 \\ \hline
$\|U_{z}\|$ & 0.05 & 0.11 & 0.13 \\ \hline
$\|U_{r}\|$ & 0.02 & - & - \\ \hline
\end{tabular}
\caption{$L_2$ norm of the gradient for the main parameters of the GRU models. The norm is averaged over all the training sentences and epochs.}
\label{tab:grad}
\end{table}
Results reveal that the reset gate weight matrices (i.e., $W_{r}$ and $U_{r}$) have a smaller gradient norm when compared to the other parameters. 
This result is somewhat expected, since the reset gate parameters are processed by two different non-linearities (i.e., the sigmoid of Eq. \ref{eq:eq_2} and the tanh of \ref{eq:eq_3}), that can attenuate their gradients.
This would anyway indicate that, on average, the reset gate has less impact on the final cost function, further supporting its removal from the GRU design. When avoiding it, the norm of the gradient tends to increase (see for instance the recurrent weights $U_{h}$ of M-GRU model). 
This suggests that the functionalities of the reset gate, can be performed by other model parameters.    
The norm further increases in the case of Li-GRUs, due to the adoption of ReLu units. This non-linearity, indeed, improves the back-propagation of the gradient over both time-steps and hidden layers, making long-term dependencies easier to learn.
Results are obtained with the same GRU used in subsection \ref{sec:corr}, considering M-GRU and Li-GRU models with the same number of hidden units.

\subsection{Impact of batch normalization} \label{sec:bn_exp}
After the preliminary analysis on correlation and gradients done in previous sub-sections, we now compare RNN models in terms of their final speech recognition performance on the TIMIT dataset. In particular, this section studies the effects of batch normalization on the proposed model. With this regard,  Table \ref{tab:bn} compares the Phone Error Rate (PER\%) achieved with and without this technique.

\begin{table}[t!]
\centering
\tabcolsep=0.20cm
    \begin{tabular}{ | l | c | c | c | c | }
    \cline{1-4}
   {\backslashbox{\em{Param.}}{\em{Arch.}}} & GRU &  M-GRU & Li-GRU \\ \hline
without batch norm & 18.4 & 18.6 & 20.4 \\ \hline
with batch norm & 17.1 & 17.2 & \textbf{16.7} \\ \hline
\end{tabular}
\caption{PER(\%) of the GRU models with and without batch normalization (TIMIT dataset, MFCC features).}
\label{tab:bn}
\end{table}

Results show that batch normalization is helpful to improve the ASR performance, leading to a relative improvement of about 7\% for GRU and M-GRU and 18\% for the proposed Li-GRU.
The latter improvement confirms that our model couples particularly well with this technique, due to the adopted ReLU activations. Without batch normalization, the ReLU activations of the Li-GRU are unbounded and tend to cause numerical instabilities.  According to our experience, the convergence of Li-GRU without batch normalization, can be achieved only by setting rather small learning rate values. The latter setting, however, can lead to a poor performance and, as clearly emerged from this experiment, coupling  Li-GRU  with this technique is strongly recommended. 

\subsection{CTC experiments} \label{sec:cts}
The proposed Li-GRU model has also been evaluated in the context of end-to-end speech recognition, considering the CTC technique introduced in Chapter \ref{cha:dsr}.
Table \ref{tab:resctc} summarizes the results of CTC on the TIMIT data set. In these experiments, the Li-GRU clearly outperforms the standard GRU, showing the effectiveness of the proposed model even in a end-to-end ASR framework. The improvement is obtained both with and without batch normalization and, similarly to what observed for hybrid systems, the latter technique leads to better performance when coupled with Li-GRU. However, a smaller performance gain is observed when batch normalization is applied to the CTC. This result could also be related to the different choice of the regularizer, as weight noise was used instead of recurrent dropout. 

In general, PERs are higher than those of the hybrid systems. End-to-end methods, in fact, are relatively young models, that are still less competitive than more complex (and mature) DNN-HMM approaches. We believe that the gap between CTC and hybrid speech recognizers could be partially reduced in our experiments with a more careful setting of the hyperparameters and with the introduction of an external phone-based language model. The main focus of this work, however, is to show the effectiveness of the proposed Li-GRU model, and a fair comparison between CTC and hybrid systems is out of the scope of this work. 

\begin{table}[t!]
\centering
\tabcolsep=0.25cm
    \begin{tabular}{ | l | c | c | }
    \cline{1-3}
   {\backslashbox{\em{Arch.}}{\em{Batch-norm.}}} & False &  True \\ \hline
GRU & 22.1 & 22.0 \\ \hline
Li-GRU & 21.1 & \textbf{20.9} \\ \hline
    \end{tabular}
\caption{PER(\%) obtained for the test set of TIMIT with various CTC RNN architectures.}
\label{tab:resctc}
\end{table}

\subsection{DNN-HMM Experiments} \label{sec:timit}
The results of  Table \ref{tab:bn} and \ref{tab:resctc} highlighted that the proposed Li-GRU model outperforms other GRU architectures. 
In this sub-section, we extend this study by performing a more detailed comparison with the most popular RNN architectures. To provide a fair comparison, batch normalization is hereinafter applied to all the considered RNN models. Moreover, at least five experiments varying the initialization seeds were conducted for each RNN architecture. The results are thus reported as the average error rates with their corresponding standard deviation. In the following sub-sections, the results obtained for TIMIT, DIRHA-English-WSJ, CHIME and TED-Talk datasets are reported.

\subsubsection{Recognition performance on TIMIT}

Table \ref{tab:res1_ligru} presents a comparison of the performance achieved with the most popular RNN models on the standard TIMIT dataset. 
\begin{table}[t!]
\centering
\tabcolsep=0.20cm
    \begin{tabular}{ | l | c | c | c | c | }
    \cline{1-4}
   {\backslashbox{\em{Arch.}}{\em{Feat.}}} & MFCC &  FBANK & fMLLR \\ \hline
relu-RNN & 18.7 $\pm$ 0.18 & 18.3 $\pm$ 0.23 & 16.3  $\pm$ 0.11 \\ \hline
LSTM & 18.1 $\pm$ 0.33 & 17.1 $\pm$ 0.36 & 15.7  $\pm$ 0.32 \\ \hline
GRU & 17.1 $\pm$ 0.20 & 16.7 $\pm$ 0.36 & 15.3  $\pm$ 0.28 \\ \hline
M-GRU & 17.2 $\pm$ 0.11 & 16.7 $\pm$ 0.19 & 15.2  $\pm$ 0.10 \\ \hline
Li-GRU & \textbf{16.7} $\pm$ 0.26 & \textbf{15.8} $\pm$ 0.10 & \textbf{14.9}  $\pm$ 0.27
\\ \hline  
    \end{tabular}
\caption{PER(\%) obtained for the test set of TIMIT with various RNN architectures.}
\label{tab:res1_ligru}
\end{table}
The first row reports the results achieved with a simple RNN with ReLU activations (no gating mechanisms are used here). Although this architecture has recently shown promising results in some machine learning tasks \cite{orth_init}, our results confirm that gated recurrent networks (rows 2-5) outperform traditional RNNs.
We also observe that GRUs tend to slightly outperform the LSTM model.
As expected, M-GRU (i.e., the architecture without reset gate) achieves a performance very similar to that obtained with standard GRUs, further supporting our speculation on the redundant role played by the reset gate in a speech recognition application. 
The last row reports the performance achieved with the proposed model, in which, besides removing the reset gate, ReLU activations are used. 
The Li-GRU performance indicates that our architecture consistently outperforms the other RNNs over all the considered input features. A remarkable achievement is the average PER(\%) of $14.9$\% obtained with fMLLR features. To the best of our knowledge, this result yields the best published performance on the TIMIT test-set.

In Table ~\ref{tab:res_ph} the PER(\%) performance is split into five different phonetic categories (vowels, liquids, nasals, fricatives and stops), showing that Li-GRU exhibits the best results for all the considered classes.
 
\begin{table}[t!]
\centering
\tabcolsep=0.20cm
    \begin{tabular}{  | l | c | c | c | c | c |}
    \cline{1-4}
Phonetic Cat. & Phone Lists & GRU & Li-GRU  \\ \hline
Vowels & \{\textit{iy,ih,eh,ae,...,oy,aw,ow,er}\} & 23.2 & \textbf{23.0}  \\ \hline
Liquids & \{\textit{l,r,y,w,el}\} & 20.1 & \textbf{19.0}  \\ \hline
Nasals & \{\textit{en,m,n,ng}\} & 16.8 & \textbf{15.9} \\ \hline
Fricatives & \{\textit{ch,jh,dh,z,v,f,th,s,sh,hh,zh}\} & 17.6 & \textbf{17.0} \\ \hline
Stops & \{\textit{b,d,g,p,t,k,dx,cl,vcl,epi}\} & 18.5 & \textbf{17.9} \\ \hline  
    \end{tabular}
\caption{PER(\%) of the TIMIT dataset (MFCC features) split into five different phonetic categories. Silences (\textit{sil}) are not considered here.}
\label{tab:res_ph}
\end{table}

Previous results are obtained after optimizing the main hyperparameters of the model on the development set. Table \ref{tab:opt} reports the outcome of this optimization process, with the corresponding best architectures obtained for each RNN architecture.
 \begin{table}[t!]
 \centering
 \tabcolsep=0.25cm
     \begin{tabular}{  | l | c | c | c | c | c |}
     \cline{1-4}
 Architecture & Layers & Neurons & \# Params  \\ \hline
 relu-RNN & 4 & 607 & 6.1 M   \\ \hline
 LSTM & 5 & 375 & 8.8 M   \\ \hline
 GRU & 5 & 465  & 10.3 M  \\ \hline
 M-GRU & 5 & 465 & 7.4 M  \\ \hline
 Li-GRU & 5 & 465 & 7.4 M  \\ \hline  
     \end{tabular}
 \caption{Optimal number of layers and neurons for each TIMIT RNN model. The outcome of the optimization process is similar for all considered features.}
 \label{tab:opt}
 \end{table}
For GRU models, the best performance is achieved with 5 hidden layers of 465 neurons. It is also worth noting that M-GRU and Li-GRU have about 30\% fewer parameters compared to the standard GRU.

\subsubsection{Recognition performance on DIRHA-English-WSJ} \label{sec:dirha}
After a first set of experiments on TIMIT, in this sub-section we assess our model on a more challenging and realistic distant-talking task, using the  DIRHA-English-WSJ corpus. A challenging aspect of this dataset is the acoustic mismatch between training and testing conditions. Training, in fact, is performed with a reverberated version of WSJ (WSJ-rev), while test is characterized by both non-stationary noises and reverberation. 

Tables \ref{tab:res2} and \ref{tab:res3} summarize the results obtained with the simulated and real parts of this dataset. 

\begin{table}[t!]
\centering
\tabcolsep=0.30cm
    \begin{tabular}{ | l | c | c | c | c | }
    \cline{1-4}
   {\backslashbox{\em{Arch.}}{\em{Feat.}}} & MFCC &  FBANK & fMLLR \\ \hline
relu-RNN & 29.7 $\pm$ 0.31  & 30.0 $\pm$ 0.38 & 24.7 $\pm$ 0.28  \\ \hline
LSTM & 29.5 $\pm$ 0.41  & 29.1 $\pm$ 0.42 & 24.6 $\pm$ 0.35 \\ \hline
GRU & 28.5 $\pm$ 0.37  & 28.4 $\pm$ 0.21 & 24.0 $\pm$ 0.27 \\ \hline
M-GRU & 28.4 $\pm$ 0.34  & 28.1 $\pm$ 0.30 & 23.6 $\pm$ 0.21 \\ \hline
Li-GRU & \textbf{27.8} $\pm$ 0.38  & \textbf{27.6} $\pm$ 0.36 & \textbf{22.8} $\pm$ 0.26 \\ \hline  
\end{tabular}
\caption{Word Error Rate (\%) obtained with the DIRHA English WSJ dataset (simulated part of the set2 rev\&noise portion) for various RNN architectures.}
\label{tab:res2}
\end{table}

\begin{table}[t!]
\centering
\tabcolsep=0.30cm
    \begin{tabular}{ | l | c | c | c | c | }
    \cline{1-4}
   {\backslashbox{\em{Arch.}}{\em{Feat.}}} & MFCC &  FBANK & fMLLR \\ \hline
relu-RNN & 23.7  $\pm$ 0.21  & 23.5 $\pm$ 0.30 & 18.9 $\pm$ 0.26 \\ \hline
LSTM & 23.2  $\pm$ 0.46  & 23.2 $\pm$ 0.42  & 18.9 $\pm$ 0.24 \\ \hline
GRU & 22.3  $\pm$ 0.39   & 22.5 $\pm$ 0.38 & 18.6 $\pm$ 0.23 \\ \hline
M-GRU & 21.5  $\pm$ 0.43   & 22.0 $\pm$ 0.37  & 18.0 $\pm$ 0.21 \\ \hline
Li-GRU & \textbf{21.3}  $\pm$ 0.38  & \textbf{21.4} $\pm$ 0.32 & \textbf{17.6} $\pm$ 0.20 \\ \hline  
\end{tabular}
\caption{Word Error Rate (\%) obtained with the DIRHA-English-WSJ dataset (real part of the set2 rev\&noise portion) for various RNN architectures.}
\label{tab:res3}
\end{table}

\begin{figure}[t]
\centering
  \includegraphics[scale=0.75]{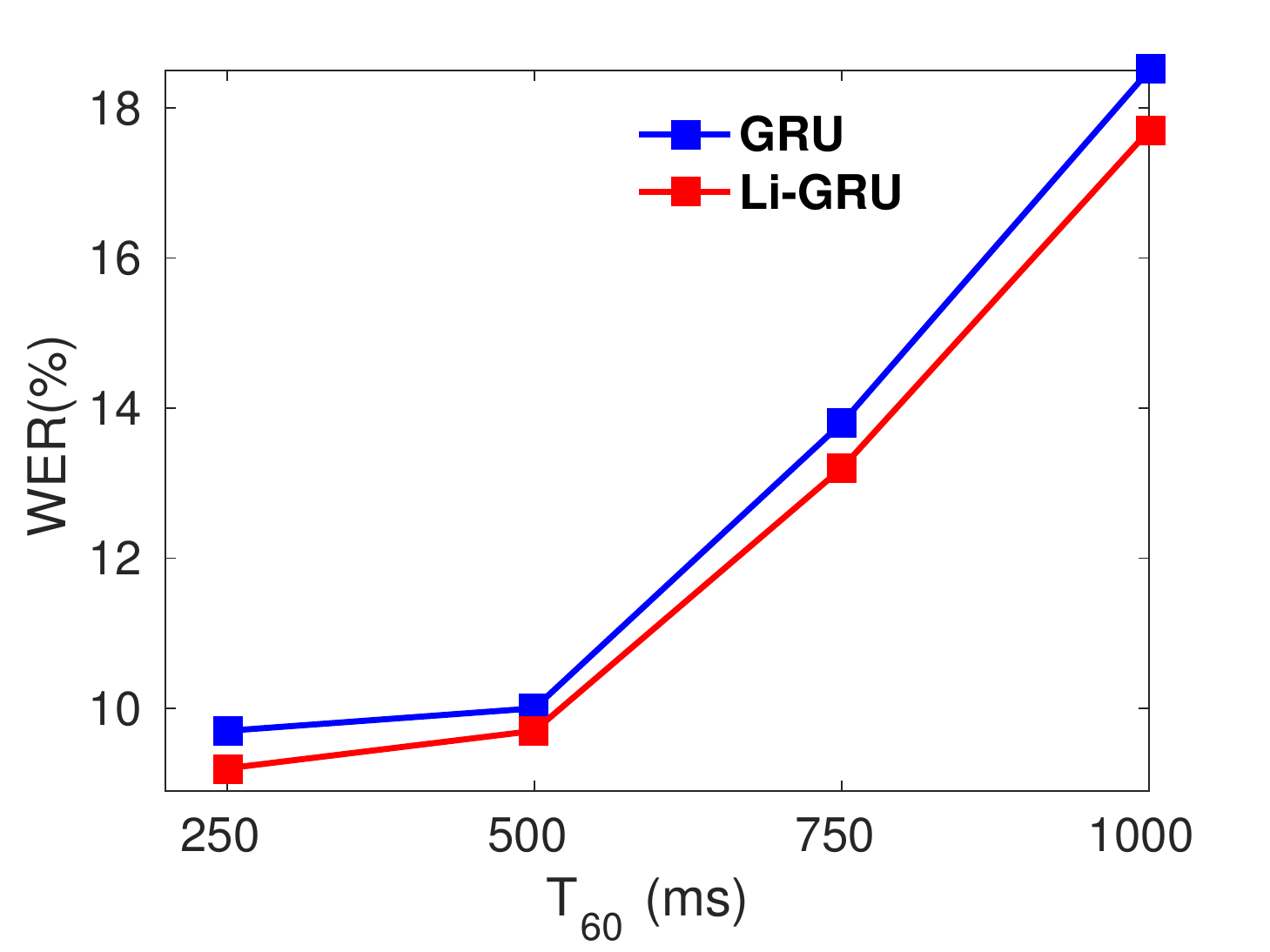}
\caption{Evolution of the WER(\%) for the DIRHA WSJ simulated data over different reverberation times $T_{60}$. }\label{fig:t60}
\end{figure}
These results exhibit a trend comparable to that observed for TIMIT, confirming that Li-GRU still outperforms GRU even in a more challenging scenario. The results are consistent over both real and simulated data as well as across the different features considered in this study.

The reset gate removal seems to play a more crucial role in the addressed distant-talking scenario.  If the close-talking performance reported in Table \ref{tab:res1_ligru} highlights comparable error rates between standard GRU and M-GRU, in the distant-talking case we even observe a  small performance gain when removing the reset gate. We suppose that this behaviour is due to reverberation, that implicitly introduces redundancy in the signal, due to the multiple delayed replicas of each sample. This results in a forward memory effect, that can make reset gate ineffective. 

In Figure ~\ref{fig:t60}, we extend our previous experiments by generating simulated data with different reverberation times $T_{60}$ ranging from 250 to 1000 ms, as outlined in Sec. \ref{sec:im_method}. In order to simulate a more realistic situation, different impulse responses have been used for training and testing purposes. No additive noise is considered for these experiments.

As expected, the performance degrades as the reverberation time increases. Similarly to previous achievements, we still observe that Li-GRU outperforms GRU under all the considered reverberation conditions. 

\subsubsection{Recognition performance on CHiME} \label{sec:chime}
In this sub-section we extend the results to the CHiME corpus, that is an important benchmark in the ASR field, thanks to the success of CHiME challenges \cite{chime,chime3}. In Table ~\ref{tab:chime} a comparison across the various GRU architectures is presented. For the sake of comparison, the results obtained with the official CHiME 4 are also reported in the first two rows\footnote{The results obtained in this section are not directly comparable with the best systems of the CHiME 4 competition. Due to the purpose of this work, indeed, techniques such as multi-microphone processing, data-augmentation, system combination as well as lattice rescoring are not used here.}. 

\begin{table}[t!]
\centering
\tabcolsep=0.10cm
    \begin{tabular}{ | l | c | c | c | c | }
    \cline{1-5}
   {\backslashbox{\em{Arch.}}{\em{Dataset}}} & DT-sim & DT-real &  ET-sim & ET-real \\ \hline
DNN & 17.8 $\pm$ 0.38 & 16.1 $\pm$ 0.31  & 26.1 $\pm$ 0.45 & 30.0 $\pm$ 0.48 \\ \hline   
DNN+sMBR & 14.7 $\pm$ 0.25  & 15.7 $\pm$ 0.23  & 24.0 $\pm$ 0.31 & 27.0 $\pm$ 0.35 \\ \hline
GRU & 15.8 $\pm$ 0.30  & 14.8 $\pm$ 0.25  & 23.0 $\pm$ 0.38 & 23.3 $\pm$ 0.35 \\ \hline
M-GRU & 15.9 $\pm$ 0.32 & 14.1 $\pm$ 0.31  & 22.8 $\pm$ 0.39 & 23.0 $\pm$ 0.41 \\ \hline
Li-GRU & \textbf{13.5} $\pm$ 0.25  & \textbf{12.5} $\pm$ 0.22  & \textbf{20.3} $\pm$ 0.31 & \textbf{20.0} $\pm$ 0.33 \\ \hline
 
\end{tabular}
\caption{Speech recognition performance on the CHiME dataset (single channel, fMLLR features).}
\label{tab:chime}
\end{table}

Results confirm the trend previously observed, highlighting a significant relative improvement of about 14\%  achieved when passing from GRU to the proposed Li-GRU. Similarly to our findings of the previous section, some small benefits can be observed when removing the reset gate. The largest performance gap, however, is reached when adopting ReLU units (see M-GRU and Li-GRU columns), confirming the effectiveness of this architectural variation. Note also that the GRU systems significantly outperform the DNN baseline, even when the latter is based on sequence discriminative training (DNN+sMBR)\cite{sequence_training}.  


Table ~\ref{tab:chime2} splits the ASR performance of the real test set into the four noisy categories.
\begin{table}[t!]
\centering
\tabcolsep=0.30cm
    \begin{tabular}{ | l | c | c | c | c | }
    \cline{1-5}
   {\backslashbox{\em{Arch.}}{\em{Env.}}} & BUS &  CAF & PED & STR \\ \hline
DNN & 44.1 & 32.0 & 26.2 & 17.7 \\ \hline
DNN+sMBR & 40.5 & 28.3 & 22.9 & 16.3 \\ \hline
GRU & 33.5 & 25.6 & 19.5 & 14.6 \\ \hline
M-GRU & 33.1 & 24.9 & 19.2 & 14.9 \\ \hline
Li-GRU & \textbf{28.0} & \textbf{22.1} & \textbf{16.9} & \textbf{13.2} \\ \hline
 
\end{tabular}
\caption{Comparison between GRU and Li-GRU for the four different noisy conditions considered in CHiME on the real evaluation set (ET-real).}
\label{tab:chime2}
\end{table}
Li-GRU outperforms GRU in all the considered environments, with a  performance gain that is higher when more challenging acoustic conditions are met. For instance, we obtain a relative improvement of 16\% in the bus (BUS) environment (the noisiest), against the relative improvement of 9.5\% observed in the street (STR) recordings.

\subsubsection{Recognition performance on TED-talks} 
Table  \ref{tab:ted_talks} reports a comparison between GRU and Li-GRU on the TED-talks corpus. The experiments are performed with standard MFCC features, and a four-gram language model is considered in the decoding step (see \cite{ted_lm} for more details).

Results on both test sets consistently shows the performance gain achieved with the proposed architecture. This further confirms the effectiveness of Li-GRU, even for a larger scale ASR task. In particular, a relative improvement of about 14-17\% is achieved. This improvement resulted statistically significant according to the matched-pair test \cite{statistical_significance} (conducted with NIST sclite $sc\_sta$t tool).

\label{sec:ted_talks}
\begin{table}[t!]
\centering
\tabcolsep=0.30cm
    \begin{tabular}{ | l | c | c | c | c | }
    \cline{1-3}
   {\backslashbox{\em{Arch.}}{\em{Dataset.}}} & TST-2011 &  TST-2012  \\ \hline
GRU & 16.3 & 17.0  \\ \hline
Li-GRU & \textbf{14.0} & \textbf{14.6} \\ \hline
 
\end{tabular}
\caption{Comparison between GRU and Li-GRU with the TED-talks corpus.}
\label{tab:ted_talks}
\end{table}

\subsubsection{Training time comparison} \label{sec:tr_time}
In the previous subsections, we reported several speech recognition results, showing that Li-GRU outperforms other RNNs. In this sub-section, we finally focus on another key aspect of the proposed architecture, namely its improved computational efficiency. 
In Table \ref{tab:tr_time}, we compare the per-epoch wall-clock training time of GRU and Li-GRU models. 

\begin{table}[t!]
\centering
\tabcolsep=0.20cm
    \begin{tabular}{ | l | c | c | c | c | }
    \cline{1-5}
   {\backslashbox{\em{Arch.}}{\em{Dataset.}}} & TIMIT &  DIRHA & CHiME & TED \\ \hline
GRU & 9.6 min & 40 min & 312 min & 590 min\\ \hline
Li-GRU & \textbf{6.5 min}   & \textbf{25 min} & \textbf{205 min} & \textbf{447 min} \\ \hline  
\end{tabular}
\caption{Per-epoch training time (in minutes) of GRU and Li-GRU models for the various datasets on an NVIDIA K40 GPU.}
\label{tab:tr_time}
\end{table}

The training time reduction achieved with the proposed architecture is about 30\% for all the datasets. This reduction reflects the amount of parameters saved by Li-GRU, that is also around 30\%. The reduction of the computational complexity, originated by a more compact model, also arises for testing purposes, making our model potentially suitable for small-footprint ASR, \cite{small1,small2,small3,small4,small5,online2}, which studies DNNs designed for portable devices with small computational capabilities.

\section{Summary and Future Challenges} \label{sec:discussion_ch5}
This chapter has confirmed the considerable importance of time contexts for improving distant speech recognition. In our studies we considered both feed-forward and recurrent recurrent neural networks. For the former architecture we proposed asymmetric context windows, that turned out to be very precious to mitigate the effects of acoustic reverberation. Our efforts have then be devoted to revise standard GRUs, proposing a novel architecture, called Light GRU, that improves the ASR performance while reducing the training time of more that 30\%.

Despite our contribution, we believe that there is still room to improve current techniques for managing time contexts. 
As discussed in this Chapter, the analysis of long-term dependencies is tightly connected with the problem of vanishing gradients over long computational chains, and the dominant approach consists in devising architectures with proper gradient shortcuts. Although this solution currently leads to interesting performance levels, it does not necessarily mean that this is the only possible methodology to embed long-term information. We indeed believe that, in the future, alternative architectures or novel training algorithms could play an important role for improving current techniques.

Moreover, there are other open challenges that state-of-the-art architectures are not able to address properly. For instance, a feature that is currently missing even in modern RNN implementations is the ability to recover the deleted memory. When processing long time sequences, the network can, at a certain point,  erroneously erase the past memory, influencing the processing of the following elements of the sequence, that cannot benefit anymore from longer context. Solutions able to mitigate this issue could have a remarkable impact on distant speech recognition.

\chapter{Cooperative Networks of Deep Neural Networks} \label{cha:ndnn}
The current development of deep learning  draws inspiration from multiple sources \cite{Goodfellow-et-al-2016-Book}. Among the others,  the most important ones are the biological brain, game theory and natural evolution.  According to Darwin's theories, a key role for the evolution of living forms is played by competition. Actually, competition across  deep neural networks has been recently explored in the context of Generative Adversarial Networks (GAN) \cite{gan}, where a generator and a discriminator continuously evolve themselves to improve their performance. Competition was also explored in deep reinforcement learning.  Super-human performance, for instance, has been achieved in automated game playing, by forcing a competition between agents \cite{alpha_go}. With this approach, DeepMind stunned the world when the AlphaGo system was able to defeat the world champion of Go, an ancient Chinese game with a huge number of possible moves (enormously larger than the combinations of chess).
 
Beyond competition, we believe that another key aspect to consider is cooperation \cite{cooperation_evolution}. Building neural networks able to automatically learn how to cooperate and communicate will represent a fundamental step towards artificial intelligence, that promises to bridge the gap between current neural networks and human brain. 

Inspired by this vision we developed a novel deep learning paradigm, called network of deep neural network. This paradigm can be exploited to solve challenging problems, where the cooperation across different DNNs can be helpful to counteract uncertainty. Distant speech recognition represents the natural application field for this approach: DSR, in fact, is not only a challenging multi-disciplinary problem, but it also requires a proper matching, communication and cooperation across the various modules involved in the speech recognition process. In the previous chapters of this thesis we tried to counteract the uncertainty originated by noise and reverberation using contaminated data or exploiting time contexts. Conversely, in the following we summarize our efforts to mitigate the harmful effects of these disturbances with another strategy: cooperation across DNNs.

The remaining part of this Chapter is organized as follows: the general paradigm will be described in Sec. \ref{sec:ndnn_gp}, while its application to DSR is discussed  in Sec. \ref{sec:ndnn_dsr}.
The first experiments on joint training will be presented in Sec. \ref{sec:joint}, while the experimental evidence emerged with the final network of DNNs is  reported in Sec. \ref{sec:ndnn}.

\section{General Paradigm} \label{sec:ndnn_gp}
The human ability to solve complex problems is a distinctive trait of our species that contributes more than other abilities to the development of human life on  earth. The typical approach to solve complex problems is to break them down into a series of simpler sub-problems. From an engineering point of view, this means that systems able to solve challenging tasks are often based on complex architectures. These architectures are composed of an ecosystem of smaller sub-modules, each one with its own functionality. 
A noteworthy example is our brain, in which different areas specialized on different tasks communicate together to achieve complex goals \cite{brain_communication}.

The ability to solve multiple complex tasks through cooperation represents one of the most significant gaps between biological and artificial neural networks. Current DNNs, in fact, are able to achieve interesting performance when addressing very specific problems (sometimes also reaching super-human abilities), but largely fail to properly address multiple tasks.
Multi-task learning was actually object of several research in the past years \cite{multi_task}. The typical approach consists in sharing the first one or two hidden layers of the DNN, in order to derive general and more robust feature representations. We believe that a better way to foster cooperation is to exchange higher-level information at multiple levels  rather than just sharing part of the architecture. 

\begin{figure}[t!]
\begin{subfigure}{0.50\textwidth}
\includegraphics[scale=0.63,trim={0cm 0cm 0cm 0cm},clip]{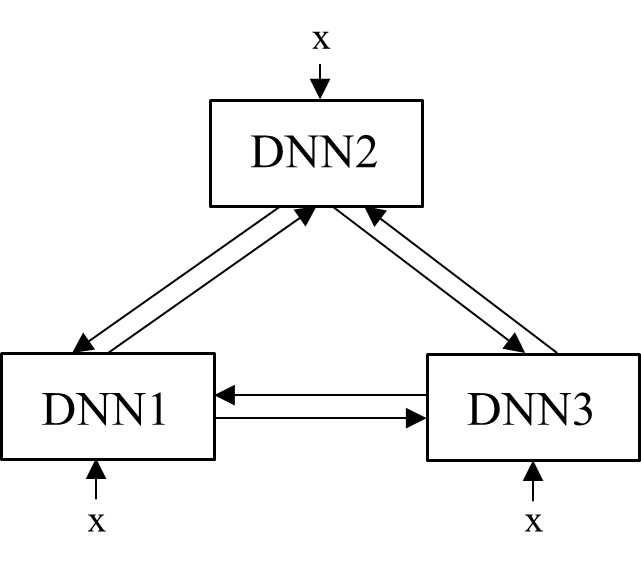}
\caption{A network of DNNs.}
\label{fig:ndnn1}
\end{subfigure} \hspace{0.0\textwidth}
\begin{subfigure}{0.50\textwidth}
\includegraphics[scale=0.63,trim={0cm 0cm 0cm 0cm},clip]{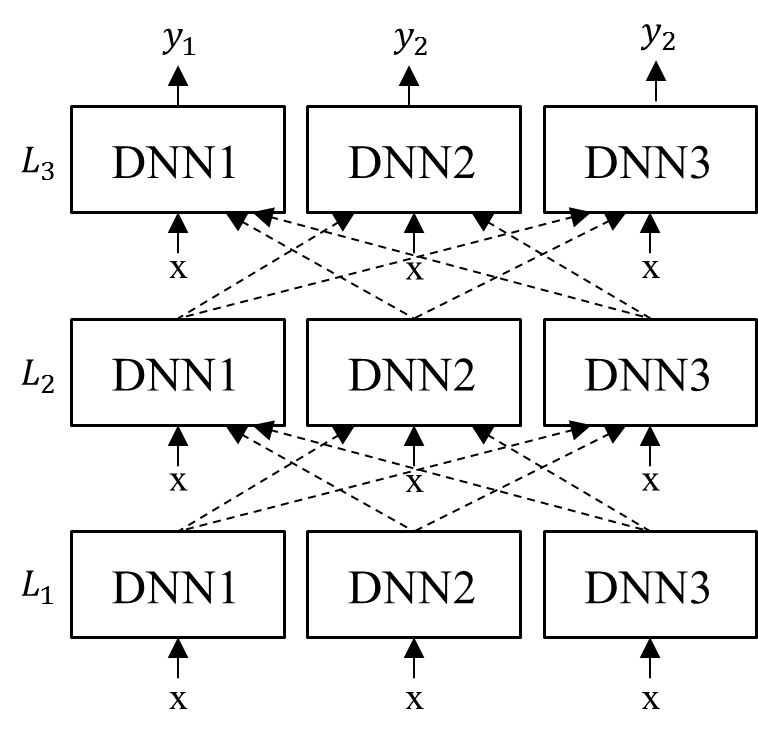}
\caption{Unfolding a network of DNN.}
\label{fig:ndnn_unfold}
\end{subfigure} \hspace{0.0\textwidth}
\caption{An example of network of deep neural networks.}
\label{fig:ndnn}

\end{figure}

The network of deep neural network, depicted in Figure  \ref{fig:ndnn1}, follows this philosophy. The proposed paradigm is not based on independent DNNs, but all the systems are organized in a network where a full-communication across the elements arises. Each element solves a different task and optimizes a different cost function. The cooperation with the other modules is realized by exchanging DNN's outputs.  In the example of  Figure  \ref{fig:ndnn1}, for instance, $DNN1$ is fed by the input features $x$ and by the output of the other DNNs. This network, instead of estimating $P(y_{1}/x)$, thus estimates posterior probabilities $P(y_{1}/x,y_{2},y_{3})$ that are enriched by additional information processed by other DNNs. This information can be regarded as a sort of prior knowledge, that might help to mitigate DNN uncertainty. The contribution of the other systems can be helpful or not.  However, instead of planning by hand the system communication, this paradigm leaves the network to freely decide which communication channels are more helpful, minimizing (ideally) human efforts in the architecture design.

Such a communication modality inherently entails a \emph{chicken-and-egg problem}, as clearly highlighted in the following recurrent equations:

\begin{eqnarray*}
y_{1}&=f_{1}(x,y_{2},y_{3},\theta_{1}) \\
y_{2}&=f_{2}(x,y_{1},y_{3},\theta_{2}) \\
y_{3}&=f_{3}(x,y_{1},y_{3},\theta_{3}) 
\label{eq:ndnn}
\end{eqnarray*}

The chicken-and-egg problem is caused by the fact that each DNN is fully connected with the others through a full bidirectional communication channel. 
To circumvent this issue we suggest to unfold the network of deep neural networks using an approach similar to that adopted for recurrent neural networks. Figure  \ref{fig:ndnn_unfold} shows an example of unfolded network of deep neural networks: at the first level the networks are independent and are fed only by the low-level input features $x$. A full communication is then established and continuously refined by progressively adding more levels to the architecture. 

\begin{algorithm}[t!]
\caption{back-propagation through network algorithm}
\label{alg_ndnn}
\begin{algorithmic}[1]
 \State Unfold the network of DNN until a communication level $L$
 \State \textbf{Forward Pass (from $l=1$ to $l=N$):} 
 \State Starting from input $x$ do a forward pass through all the DNNs.
  \State \textbf{Compute Cost Functions:}. 
 \State For each DNN $i$ of the level $l$ compute $C_{i,l}$
 
  \State \textbf{Gradient back-propagation (from $l=N$ to $l=1$):}
    \State For each DNN $i$ of the level $l$ compute the gradients and back-propagate them through all the connected nodes.
     \State The gradient of the parameters $\theta_{i,l}$ will be given by:
     
     $g_{\theta_{i,l}}=\frac{\partial C_{i,l}}{\theta_{i,l}}+ \sum_{m=l+1}^L \sum_{n=1}^N  \frac{\partial C_{n,m}}{\theta_{i,l}} $
     
  \State \textbf{Parameter Updates:}
  \State  Given $g_{\theta_{i,l}}$ use an optimizer to compute the new parameters $\theta_{i,l}$ 
\end{algorithmic}
\end{algorithm}

After the unfolding procedure, the overall computational graph can be considered as a single very deep neural network where all the DNNs can be jointly trained.  In particular, training can be carried out with the  \textit{back-propagation through network}, that is a variation of the standard back-propagation. The proposed algorithm is described in Alg.~\ref{alg_ndnn}, where $x$ are the input features, $L$ the number of communication levels, $N$ the number of DNNs, $g$ the gradients and $\theta$ the DNN parameters. This algorithm is repeated for all the minibatches and iterated for several epochs until convergence.
The basic idea is to perform a forward pass, compute the loss functions at the output of each DNN, compute the corresponding gradients, and back-propagate them. The gradient, that is back-propagated through all the connected DNNs, is given by the following equation:
\begin{equation}
g_{\theta_{i,l}}=\underbrace{\frac{\partial C_{i,l}}{\theta_{i,l}}}_{\text{local gradient}}+ \underbrace{\sum_{m=l+1}^L \sum_{n=1}^N  \frac{\partial C_{n,m}}{\theta_{i,l}}}_{\text{other contributions}}
\label{eq:eq1}
\end{equation}
The updates of each DNN not only depend on their local cost, but also on the higher-level losses. In this way the training of the DNNs  is in part driven by the contribution of the others, that  would  hopefully be helpful to improve the system performance. As will be discussed in the following sections, the integration of different gradients coming from the higher levels produces a regularization effect, that significantly helps the training of the system.

Note that no direct connections between DNNs solving the same task are considered in the unrolled architecture depicted in Fig. \ref{fig:ndnn_unfold} (for instance,  DNN1 $l=2$ does not feed  DNN1 $l=3$).  Results, not reported here, have shown that such communication channel degrades the overall system performance, since the higher level DNN tends just to copy at the output the information provided by the lower-level DNN.


\section{Application to DSR} \label{sec:ndnn_dsr}
\begin{figure}[t!]
\begin{subfigure}{0.62\textwidth}
\includegraphics[scale=0.65,trim={0cm 0cm 0cm 0cm},clip]{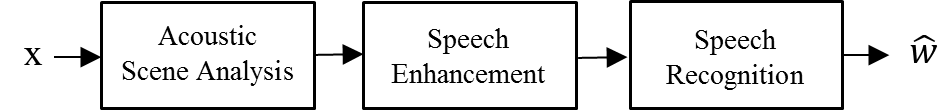}
\caption{State-of-the-art DSR system.}
\label{fig:ndnn_DSR1}
\end{subfigure} \hspace{0.0\textwidth}
\begin{subfigure}{0.35\textwidth}
\includegraphics[scale=0.65,trim={0cm 0cm 0cm 0cm},clip]{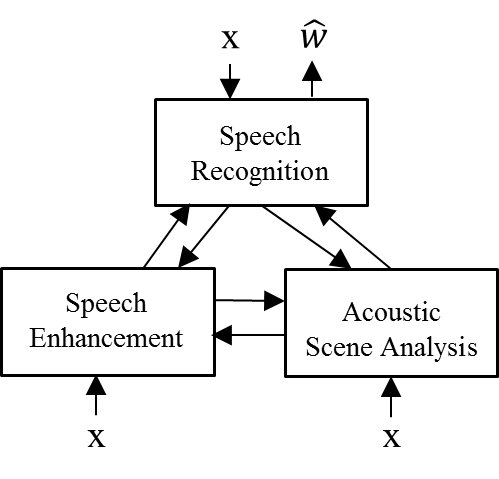}
\caption{Proposed architecture.}
\label{fig:ndnn_DSR2}
\end{subfigure} \hspace{0.0\textwidth}
\caption{Network of DNNs for distant speech recognition.}
\label{fig:ndnn}
\end{figure}

The most advanced state-of-the-art DSR systems are based on rather complex architectures, that are often composed of multiple modules working together \cite{nakatani}. An example is depicted in Figure \ref{fig:ndnn_DSR1}, where a pipeline of acoustic scene analysis (that might be, for instance, a module for environment classification or for acoustic event detection), speech enhancement and speech recognition technologies is shown. 
Although this pipeline sounds perfectly reasonable and  organized in a logical fashion, we believe that current systems suffer from the following prominent limitations:

\begin{itemize}
\item \textbf{Lack of matching}.  Even in modern DSR systems \cite{nakatani}, speech enhancement and speech recognition are often developed independently.  Moreover, in several cases, the enhancement part is tuned according to metrics which are not directly correlated with the final speech recognition performance. This weakness is due to the historical evolution of DSR technologies: the community working on speech enhancement and speech recognition were, indeed, largely independent for several years. Speech enhancement was mostly based on signal processing techniques, while ASR technology was based on pure machine learning approaches.
We strongly believe that deep learning can help break down the wall between these different disciplines.
The recent success of DNNs, in fact, has not only largely contributed to a substantial improvement of speech recognition  \cite{pawel2,hain,dnn_rev,dnn_rev2,dnn3,rav_in14,ravanelli15}, but has also enabled the development of competitive DNN-based speech enhancement solutions \cite{dnn_se1,dnn_se2,dnn_se3}, making an effective integration between these technologies easier.

\item \textbf{Lack of communication}.
As shown in Figure \ref{fig:ndnn_DSR1}, state-of-the-art system are based on a flat pipeline, based on a unidirectional communication flow. The speech enhancement, for instance, normally helps the speech recognizer, but the output of the latter is not commonly used, in turn, to improve the speech enhancement.
We argue that establishing this missing link can nevertheless be very useful, since a hint on the recognized phone sequence might help the speech enhancement in performing its task. This can be particularly helpful under noisy and reverberant conditions: when a significant uncertainty arises, a continuous interaction between the systems can guide them towards a better decision. 
\end{itemize}

We propose to mitigate the latter issues with the network of deep neural network approach described in the previous section. Figure \ref{fig:ndnn_DSR2} reports the proposed architecture, in which all the basic modules of the DSR system are implemented with DNNs. The lack of matching is mitigated using a joint-training strategy, while the lack of communication is improved through a full communication across 
DNNs.

To validate our architecture, in the next section we start from a simplified scenario where a flat pipeline of a speech enhancement and speech recognition DNNs is jointly trained with batch normalization.
Even though different types of DNNs can be embedded within the network of deep neural networks framework, the studies conducted in this thesis and discussed in the following are focused on standard feed-forward DNNs.


\section{Batch Normalized Joint Training} \label{sec:joint}


Within the DNN framework, one way to achieve a fruitful integration of the various components is joint training.  
The core idea is to pipeline a speech enhancement and a speech recognition deep neural network and to jointly update their parameters as if they were within a single bigger network. Although joint training for speech recognition is still an under-explored research direction, such a paradigm is progressively gaining more attention and some interesting works in the field have been recently published \cite{joint2,joint1,joint3,joint6,joint7,joint4,joint5}.

In this thesis, we contributed to this line of research by proposing an approach based on batch normalization, that turned out to be very precious to improve the network convergence, since it makes one network less sensitive to changes of the other. 
Differently to previous works \cite{joint1,joint3}, using this approach we were able to jointly train a cascade between speech enhancement and speech recognition DNNs without any pre-training step.

\begin{figure}[t!]
\centering
\includegraphics[width=0.7\textwidth]{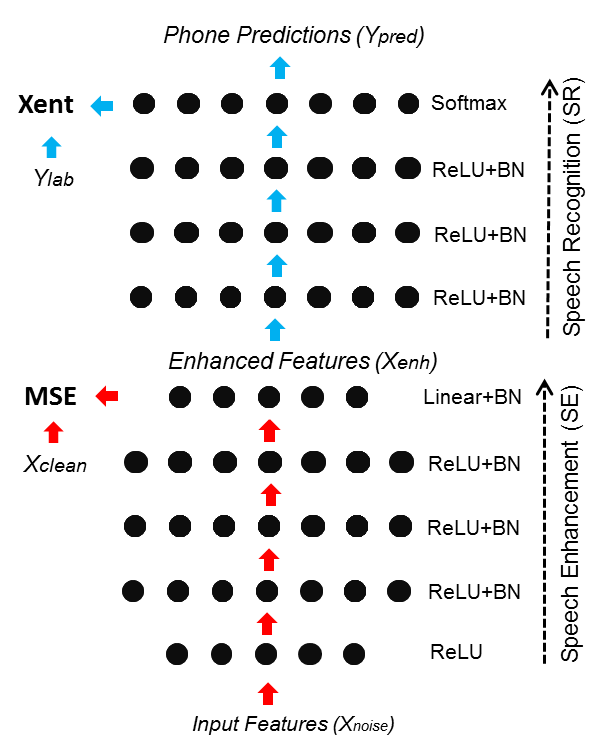}
\caption{The FF-DNN architecture proposed for joint training.}
\label{fig:arch}
\end{figure}


The proposed architecture is depicted in Figure ~\ref{fig:arch}. A bigger joint DNN is built by concatenating a speech enhancement and a speech recognition MLP. The speech enhancement DNN is fed with the noisy features $x_{noise}$ gathered within a context window and tries to reconstruct at the output the original clean speech (regression task). 
The speech recognition DNN is fed by the enhanced features $x_{enh}$ estimated at the previous layer and performs phone predictions $y_{pred}$ at each frame (classification task). The architecture of Figure  \ref{fig:arch} is trained with the algorithm described in Alg. \ref{alg}.

The basic idea is to perform a forward pass, compute the loss functions at the output of each DNN (mean-squared error for speech enhancement and negative multinomial  log-likelihood for speech recognition), compute and weight the corresponding gradients, and back-propagate them.
In the joint training framework, the speech recognition gradient is also back-propagated through the speech enhancement DNN. Therefore, at the speech enhancement level, the parameter updates not only depend on the speech enhancement cost function but also on the speech recognition loss, as shown by the following equation:

 \begin{equation}
 \theta_{SE} \gets \theta_{SE}- lr * (g_{SE}+\lambda g_{SR}) \,.
 \label{eq:updates}
 \end{equation}
Where $\theta_{SE}$ are the parameters of the speech enhancement DNN, $g_{SE}$ is the gradient of such parameters computed from the speech enhancement cost function (mean squared error), while $g_{SR}$ is the gradient of $\theta_{SE}$ computed from the speech recognition cost function (multinomial log-likelihood). Finally, $\lambda$ is a hyperparameter for weighting $g_{SR}$ and $lr$ is the learning rate.

The key intuition behind joint training is that since the enhancement process is in part guided by the speech recognition cost function, the front-end would hopefully be  able to provide enhanced speech which is more suitable and discriminative for the subsequent speech recognition task. 
From a machine learning perspective, this solution can also be considered as a way of injecting a useful task-specific prior knowledge into a deep neural network.
On the other hand, it is well known that training deep architectures is easier when some hints are given about the targeted function \cite{know_matter}. 
As shown previously \cite{know_matter}, such prior knowledge becomes progressively more precious as the complexity of the problem increases and can thus be very helpful for a distant speech recognition task. Similarly to the current work, in \cite{know_matter,Romero-et-al-ICLR2015-small} a task-specific prior knowledge has been injected into an intermediate layer of a DNN for better addressing an image classification problem.
In our case, we exploit the prior assumption that to solve our specific problem, it is reasonable to first enhance the features and, only after that, perform the phone classification.
Note that this is certainly not the only way of solving the problem, but among all the possible functions able to fit the training data, we force the system to choose from a more restricted subset, potentially making training easier. 
On the other hand, good prior knowledge is helpful to defeat the curse of dimensionality, and 
a complementary view is thus to consider the proposed joint training as a regularizer. 
According to this vision, the weighting parameter $\lambda$ of Eq. \ref{eq:updates} can be regarded as a regularization hyperparameter, as will be better discussed in Sec. \ref{sec:gw}. 
\begin{algorithm}[t!]
\caption{Pseudo-code for joint training}
\label{alg}
\begin{algorithmic}[1]
\State \textbf{DNN initialization} 
\For {i in minibatches}
 \State \textbf{Forward Pass:} 
 \State Starting from the input layer do a forward pass
 \State (with batch normalization) through the networks.
  \State \textbf{Compute SE Cost Function:} 
  \State $MSE_i=\frac{1}{N}\sum_{n=1}^{N}(x_{enh}^i-x_{clean}^i)^2$
  \State \textbf{Compute SR Cost Function:}
  \State $NLL_i=-\frac{1}{N}\sum_{n=1}^{N}y_{lab}^i log(y_{pred}^i)$ 
  \State \textbf{Backward Pass:}
  \State Compute the grad. $g_{SE}^i$ of $MSE_i$ and backpropagate it.
  \State Compute the grad. $g_{SR}^i$ of $NLL_i$ and backpropagate it.
  \State \textbf{Parameters Updates:}
   \State  $\theta_{SE}^i \gets \theta_{SE}^i - lr * (g_{SE}^i+\lambda g_{SR}^i)$
    \State  $\theta_{SR}^i \gets \theta_{SR}^i - lr * g_{SR}^i$
\EndFor
\State Compute NLL on the development dataset
\If {$NLL_{dev} < NLL_{dev}^{prev}$}
  \State Train for another epoch (go to 2) 
\Else
 \State Stop Training
\EndIf
\end{algorithmic}
\end{algorithm}



This joint training of the network of DNNs, however, can be complicated by the fact that the output distribution of the DNNs may change substantially during the optimization procedure.
Each DNN would have to deal with an input distribution that is non-stationary and unnormalized, possibly causing convergence issues. To mitigate this issue, known as  internal covariate shift, we suggest to couple the proposed architecture with batch normalization, that turned out to be crucial for achieving a better performance, to improve DNN convergence, and to avoid any time-consuming pre-training steps.
Particular attention should anyway be devoted to the initialization of the $\gamma$ parameter. Contrary to \cite{batchnorm}, where it was initialized to unit variance ($\gamma=1$), in this work we have observed better performance and convergence properties with a smaller variance initialization ($\gamma=0.1$).
A similar outcome has been found in \cite{initbn}, where fewer vanishing gradient problems are empirically observed with small values of $\gamma$ in the case of recurrent neural networks.

\subsection{Relation to prior work}
Similarly to our work, a joint training framework between speech enhanancement and speech recognition has been explored in \cite{joint2,joint1,joint3,joint6,joint7,joint4,joint5}.
In \cite{joint1,joint3}, for instance, the joint training was actually performed as a fine-tuning procedure, which was carried out only after training the two networks independently. This approach can be justified by the fact that jointly train all the DNNs from scratch is actually very challenging with standard deep learning techniques, due to possible convergence issues. 
However, pre-training significantly slows down the training time, making the overall learning procedure particularly heavy and computational demanding.  
Another critical aspect of such an approach is that the learning rate adopted in the fine-tuning step has to be properly selected in order to really take advantage of pre-training. 

A key difference with previous efforts is that we propose to combine joint training with batch normalization.
With this technique we are  not only able to significantly improve the performance of the DSR system, but also to perform joint training from scratch, skipping any pre-training phase without convergence issues. Note also that our approach naturally entails a regularization effect: the speech recognition, in fact, will be fed with poor enhanced features at the beginning of training, introducing a sort of noise that could improve DNN generalization.



\subsection{Joint Training Performance}
The experiments reported in the following summarize the main findings reported in \cite{ravanelli_SLT}.
The experimental validation has been conducted using different datasets, acoustic conditions, and tasks. A first set of experiments was conducted with the TIMIT corpus, using a phone-loop task. Training and test has been performed with contaminated versions of this dataset.
The impulse responses used for contamination were measured in the living-room of the DIRHA apartment.

The experiments have then be extended to a more realistic WSJ task. Training was performed with a reverberated version of the standard WSJ corpus, while evaluation considered the real recordings of the DIRHA-English-WSJ corpus under both $rev$ and $rev\&noise$ conditions.
MFCC features are used for all the experiments. See \ref{app:ndnn_s1} for more detailed description of the experimental setup.

As a reference baseline, we first report the close-talking performance achieved with a single DNN. The Phoneme Error Rate (PER\%) obtained by decoding the original test sentences of TIMIT is $19.5\%$ (using DNN models trained with the original dataset). The Word Error Rate (WER\%) obtained by decoding the close-talking DIRHA-English-WSJ sentences is $3.3\%$.  It is worth noting that, under such favorable acoustic conditions, the DNN model leads to a very accurate sentence transcription, especially when coupled with a language model. 

In Table \ref{tab:res1_jt}, we reported the performance obtained with distant-talking experiments where the proposed joint training approach is compared with other competitive strategies. 
\begin{table}[t!]
\centering
\tabcolsep=0.28cm
    \begin{tabular}{ | l | c | c | c | c | }
    \cline{1-4}
    \multirow{2}{*}{\backslashbox{\em{System}}{\em{Dataset}}} & \multicolumn{1}{ | c |}{TIMIT}  & \multicolumn{1}{ | c |}{WSJ} & \multicolumn{1}{ | c |}{WSJ}  \\ \cline{2-4}
    & \textit{Rev} & \textit{Rev} & \textit{Rev+Noise}  \\ \hline
      Single big DNN & 31.9  & 8.1 & 14.3    \\ \hline
      SE + clean SR & 31.4  & 8.5 & 15.7    \\ \hline
      SE + matched SR & 30.1  & 8.0 & 13.7    \\ \hline
      SE + SR joint training & \textbf{29.1}  & \textbf{7.8} & \textbf{12.7}    \\ \hline  
    \end{tabular}
\caption{Performance of the proposed joint training approach compared with other competitive DNN-based systems. Training is performed with WSJ-rev, while test is performed with DIRHA-English-WSJ (set2, real part).}
\label{tab:res1_jt}
\end{table}
\label{sec:bn_exp}
\begin{table}[t!]
\centering
\tabcolsep=0.108cm
    \begin{tabular}{ | l | c | c | c | c | c |}
    \cline{1-5}
    \multirow{2}{*}{\backslashbox{\em{Dataset}}{\em{System}}} & \multicolumn{2}{ | c |}{Without Pre-Training}  & \multicolumn{2}{ | c |}{With Pre-Training}  \\ \cline{2-5}
    & \textit{no-BN} & \textit{with-BN} & \textit{no-BN} & \textit{with-BN}  \\ \hline
      TIMIT-Rev & 34.2  & \textbf{29.1} & 32.6  & 29.5   \\ \hline
      WSJ-Rev & 9.0  & \textbf{7.8} & 8.8  & 7.8   \\ \hline
      WSJ-Rev+Noise & 15.7 & \textbf{12.7} & 15.0  & 12.9  \\ \hline
    
    \end{tabular}
\caption{Analysis of the role played by batch normalization within the proposed joint training framework.}
\label{tab:test2_jt}
\end{table}
In particular, the first line shows the results obtained with a single neural network. The size of the network has been optimized on the development set (4 hidden layers of 1024 neurons for TIMIT, 6 hidden layers of 2048 neurons for WSJ cases). The second line shows the performance obtained when the speech enhancement neural network (4 hidden layers of 2048 neurons for TIMIT, 6 hidden layers of 2048 neurons for WSJ) is trained independently and later coupled with the aforementioned close-talking DNN. These results are particularly critical because, especially in adverse acoustic conditions, the speech enhancement model introduces significant distortions. The close-talking DNN, trained in the usual way, is thus not able to cope with this significant amount of mismatch. To partially recover such a critical mismatch, one approach is to first train the speech enhancement, then pass all the training features through the speech enhancement DNN, and, lastly, train the speech recognition DNN with the dataset processed by the speech enhancement. The third line shows results obtained with such a matched training approach. The last line reports the performance achieved with the proposed joint training approach. Batch normalization is adopted for all the systems considered in Table \ref{tab:res1_jt}.

Although joint training exhibits the best performance in all the cases, it is clear that such a technique is particularly helpful especially when challenging acoustic conditions are met. For instance, a relative improvement of about $8\%$ over the most competitive matched training system is obtained for the WSJ task in noisy and reverberant conditions.

 \begin{figure}[t!]
 \centering
 \includegraphics[width=0.80\textwidth]{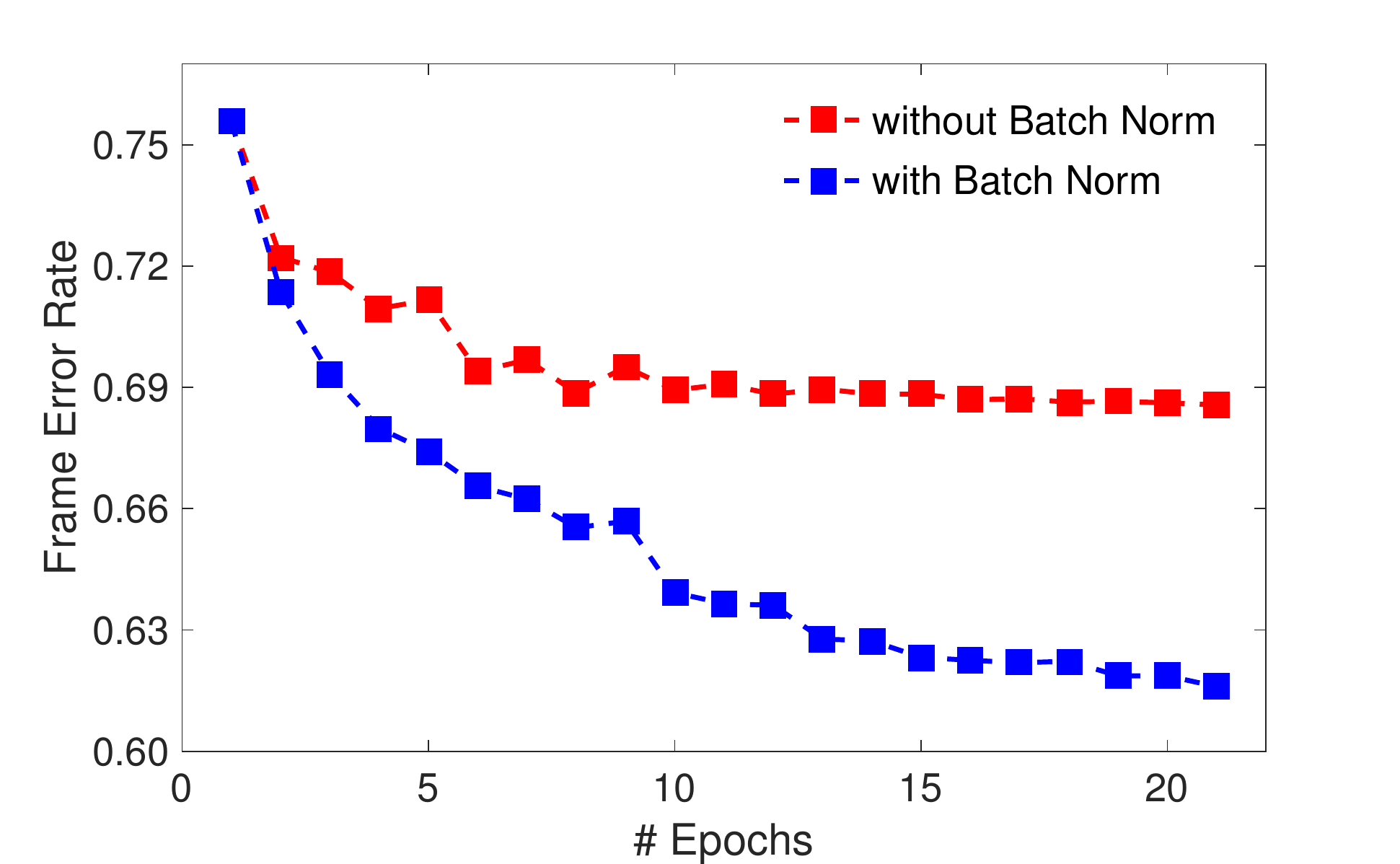}
 \caption{Evolution of the test frame error rate across various training epochs with and without batch normalization.}
 \label{fig:bn_frame}
 \end{figure}

\subsection{Role of batch normalization}
In Table \ref{tab:test2_jt}, the impact of batch normalization on the joint training framework is shown.
The first two columns report, respectively, the results obtained with and without batch normalization when no pre-training techniques are employed. The impact of pre-training is studied in the last two columns. The pre-training strategy considered here consists of initializing the two DNNs with the matched training system discussed in the previous section, and performing a fine-tuning phase with a reduced learning rate. The column corresponding to the pre-training without batch normalization represents a system that most closely matches the approaches followed in \cite{joint1,joint3}. 

Table~\ref{tab:test2_jt} clearly shows that batch normalization is particularly helpful. For instance, a relative improvement of about 23\% is achieved when batch normalization is adopted for the WSJ task in a noisy and reverberant scenario. The key importance of batch normalization is also highlighted in Figure ~\ref{fig:bn_frame}, where the evolution during training of the frame-level phone error rate (for the TIMIT-Rev dataset) is reported with and without batch normalization. From the figure it is clear that batch normalization, when applied to the considered deep joint architecture, ensures a faster convergence and a significantly better performance. Moreover, as shown in Table~\ref{tab:test2_jt}, batch normalization eliminates the need of DNN pre-training, since similar (or even slightly worse results) are obtained when pre-training and batch normalization are used simultaneously.




\subsection{Role of the gradient weighting}
\label{sec:gw}
\begin{figure}[t!]
 \centering
 \includegraphics[width=0.70\textwidth]{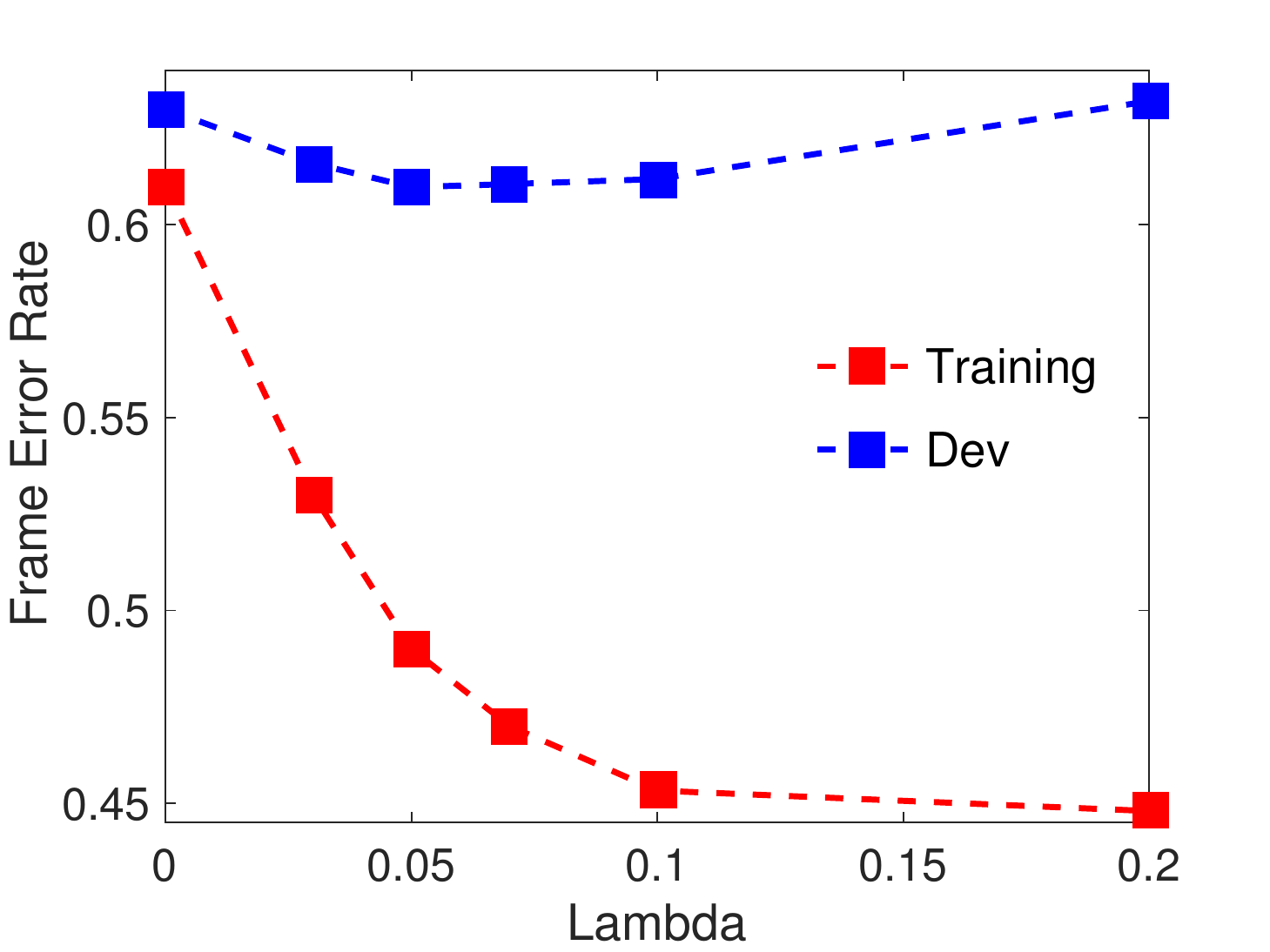}
 \caption{Training and development frame error rates obtained on the TIMIT-Rev dataset for different values of $\lambda$.}
 \label{fig:grad_w}
 \end{figure}

In Figure  \ref{fig:grad_w}, the role of the gradient weighting factor $\lambda $ is highlighted.
From the figure one can observe that small values of $\lambda$ lead to a situation close to underfitting, while higher values of $\lambda$ cause overfitting. The latter result is somewhat expected since, intuitively, with very large values of $\lambda$ the speech enhancement information tends to be neglected and training relies on the speech recognition gradient only.

In the present work, we have seen that values of $\lambda$ ranging from 0.03 to 0.1 provide the best performance. Note that these values are smaller than that considered in \cite{joint1,joint2}, where a pure gradient summation ($\lambda=1$) was adopted. We argue that this result is due to the fact that, as observed in \cite{initbn}, the norm of the gradient decays very slowly when adopting batch normalization with a proper initialization of $\gamma$, even after the gradient has passed through many hidden layers. This causes the gradient backpropagated through the speech recognition network and into the speech enhancement network to be very large.



\section{Cooperative Enhancement and Recognition} \label{sec:ndnn}
\begin{figure}[t!]
\centering
\includegraphics[width=0.80\textwidth]{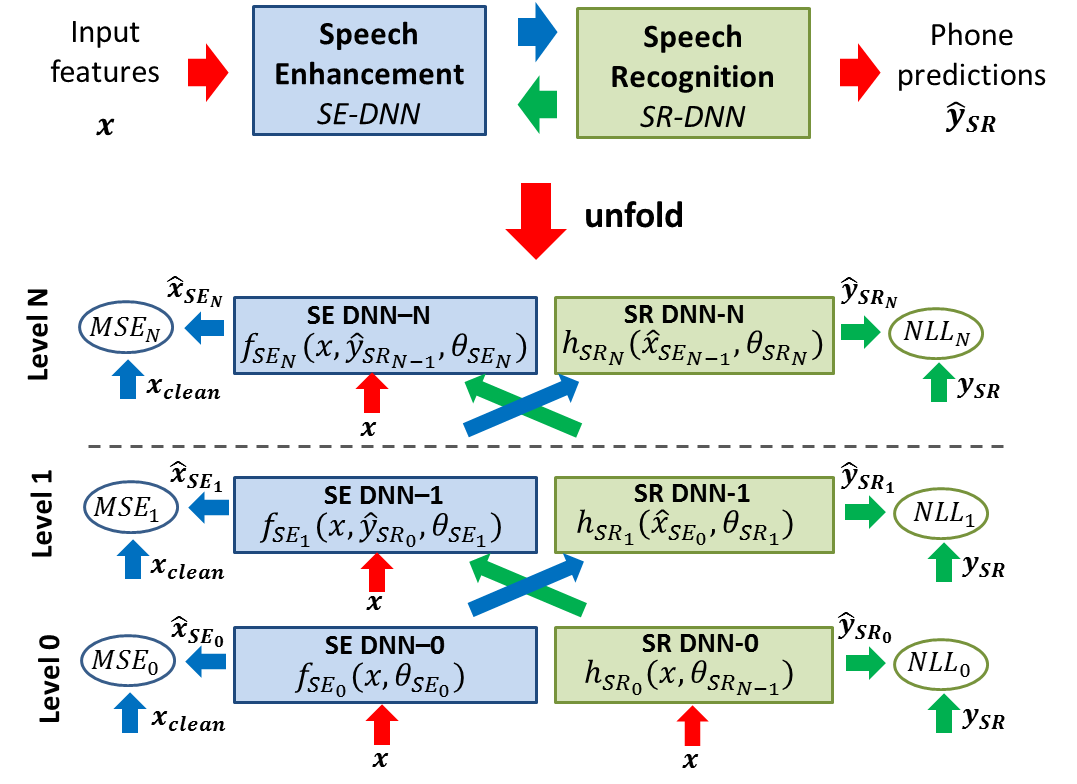}
\caption{The proposed network of deep neural networks for speech enhancement and speech recognition. Each element of the network is a FF-DNN.}
\label{fig:arch_last}
\end{figure}

In this section we evolve the flat joint trained pipeline experimented so far towards a cooperative network of DNNs. As reported in Figure  \ref{fig:arch_last}, this work considers a full communication between the speech enhancement and speech recognition, which are the most critical components of a DSR system. 
The proposed architecture is unfolded and trained with the backpropagation through network algorithm as described in Sec. \ref{sec:ndnn_gp}. As discussed in the previous section, all the components are jointly trained with a single learning procedure, and batch normalization is adopted to improve the network convergence.

In this specific application, an effective system communication can be impaired by the high dimensionality of the speech recognizer output $\hat{y}_{SR_\ell}$. This dimensionality derives from the number of considered context-dependent states, which typically ranges from 1000 to 4000 (depending on the phonetic decision tree and on the dataset). To avoid feeding the speech enhancement with such a high dimensional input, we jointly estimate the monophone targets (which are only some dozens). Similarly to \cite{multitask}, this is realized by adding an additional softmax classifier on the top of the last hidden layer of each speech recognition DNN.

\subsection{Related work}


Similarly to this work, an iterative pipeline based on feeding the speech recognition output into a speech enhancement DNN has recently been proposed in \cite{joint6,ndnn1}. The main difference with our approach is that the latter circumvents the chicken-and-egg problem by simply feeding the speech enhancement with the speech recognition alignments generated at the previous iteration, while our solution faces this issue by adopting the  unrolling procedure over different interaction levels previously discussed.

Our paradigm has also some similarities with traditional multi-tasking techniques \cite{multi_ow}. The main difference is that the latter are based on sharing some hidden layers across the tasks, while our method relies on  exchanging DNN outputs at various interaction levels. 
Finally, the proposed training algorithm has some aspects in common with the back-propagation through structure originally proposed for the parsing problem \cite{goller}.  The main difference is that the latter back-propagates the gradient through a tree structure, while the proposed variation back-propagates it on a less constrained network of components. Another difference is that in the original algorithm the same neural network is used across all the levels of the tree, while in this work different types of DNNs (i.e., speech enhancement and speech recognition) are involved.

\subsection{Network of DNNs performance}
The experiments reported in the following summarize the main achievements emerged in \cite{ravanelli_icassp}. The experiments were based on the same datasets and with the same experimental setup considered in the previous section. More details can be found in App. \ref{app:ndnn_s2}.



\begin{figure}[t!]
\centering
\includegraphics[width=0.80\textwidth]{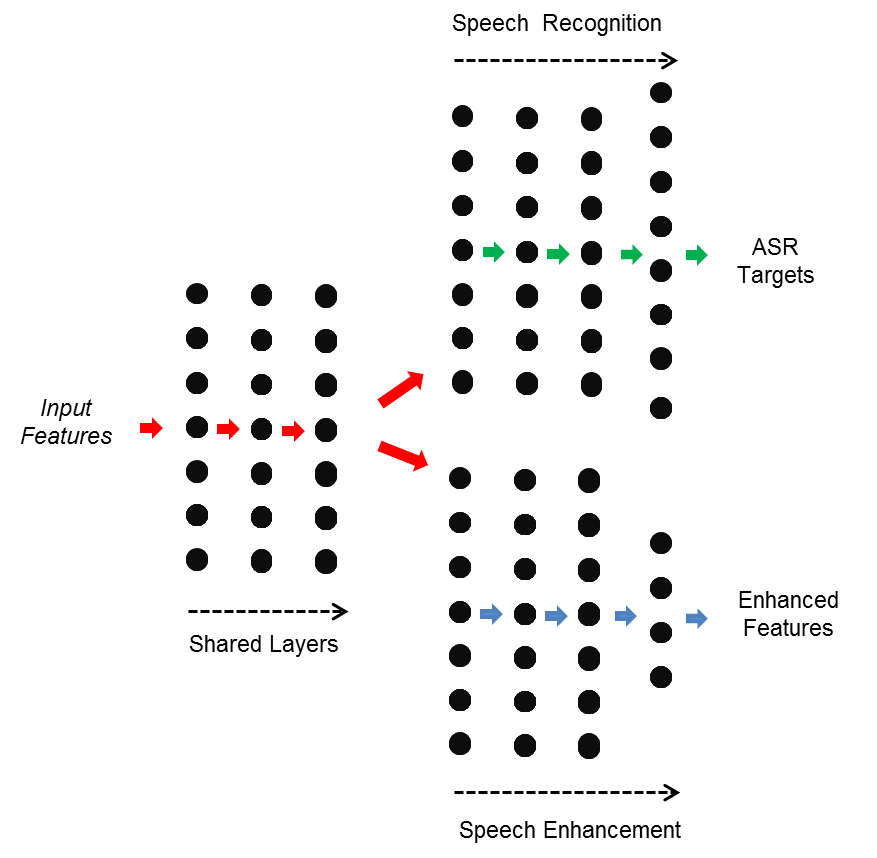}
\caption{Multi-task DNN for speech enhancement and recognition.}
\label{fig:multi_ndnn}
\end{figure}

The proposed network of DNNs approach is compared in Table \ref{tab:res1} with other competitive systems. 
The first line reports the results obtained with a single neural network. In this case, only the speech recognition labels are used and the DNN is not forced to perform any speech enhancement task.
The second line shows the performance obtained when the single DNN is coupled with a traditional multi-task learning, in which speech enhancement and speech recognition tasks are simultaneously considered, as shown in Figure  \ref{fig:multi_ndnn}. This multi-task architecture shares the first half of the hidden layers across the tasks, while the second half of the architecture is task-dependent. This approach aims to discover (within the shared layers) more general and robust features which can be exploited to better solve both correlated tasks.  
The third line reports the performance achieved with the joint training approach described in  the previous section. In this case a bigger DNN composed of a cascade of a speech enhancement and a speech recognition DNNs is jointly trained by back-propagating the speech recognition gradient also into the speech enhancement DNN. 
The last line finally shows the performance achieved with the proposed network of deep neural network approach (unfolded up to level 2). To allow a fair comparison, batch normalization is adopted for all the considered systems.

\begin{table}[t!]
\centering
\tabcolsep=0.28cm
    \begin{tabular}{ | l | c | c | c | c | }
    \cline{1-4}
    \multirow{2}{*}{\backslashbox{\em{System}}{\em{Dataset}}} & \multicolumn{1}{ | c |}{TIMIT}  & \multicolumn{1}{ | c |}{WSJ} & \multicolumn{1}{ | c |}{WSJ}  \\ \cline{2-4}
    & \textit{rev} & \textit{rev} & \textit{rev+noise}  \\ \hline
      Single DNN & 31.9  & 8.1 & 14.3    \\ \hline
      Single DNN +multitask & 31.4  & 8.1 & 13.8    \\ \hline
      Joint SE-SR training & 29.1  & 7.8 & 12.7    \\ \hline
      Network of DNNs & \textbf{28.7}  & \textbf{7.6} &  \textbf{12.3}    \\ \hline  
    \end{tabular}
\caption{Performance of the proposed network of DNN approach compared with other competitive DNN-based systems (PER\% for TIMIT, WER\% for WSJ). For WSJ experiments, test is performed with DIRHA-English-WSJ (set2, real-part).}
\label{tab:res1}
\end{table}
\label{sec:bn_exp}

\begin{table}[t!]
\centering
\tabcolsep=0.28cm
    \begin{tabular}{ | l | c | c | c | c | }
    \cline{1-4}
      Dataset & Level 0  & Level 1 & Level 2    \\ \hline
      TIMIT rev & 31.4 & 29.1 & \textbf{28.7}  \\ \hline
      WSJ rev & 8.0 & 7.7 & \textbf{7.6}  \\ \hline
      WSJ rev+noise & 14.3 & 12.7 & \textbf{12.3}  \\ \hline 
    \end{tabular}
\caption{Performance of the proposed network of DNN achieved at various levels of the architecture.}
\label{tab:res_ndnn}
\end{table}
\label{sec:bn_exp}

Table \ref{tab:res1} highlights that the proposed approach significantly outperforms all the single DNN systems. For instance, a relative improvement of about 14\%  over the single DNN baseline is obtained for the \textit{WSJ rev+noise} case. The network of deep neural networks also outperforms the considered joint training method. This result suggests that the improved cooperation between the networks achieved with our full communication scheme can overtake the standard DSR pipeline based on a partial and unidirectional information flow (which is still considered  in  the context of joint training approaches).     

Table \ref{tab:res_ndnn} shows the results obtained by decoding the speech recognition output at the various levels of the proposed architecture (denoted as $\hat{y}_{SR_0}$, $\hat{y}_{SR_1}$,  $\hat{y}_{SR_2}$ in Figure  \ref{fig:arch_last}). One can note that the performance become progressively better as the level of the network of DNNs increases till level 2. As expected, the first level speech recognizer (\textit{SR DNN-0}) performs similarly to the single DNN baseline. The second level (\textit{SR DNN-1}) is based on a simple cascade between a speech enhancement and a speech recognition DNNs,  and thus provides results similar to that obtained with standard joint training. The third level (\textit{SR DNN-2}) achieves the best performance,  confirming that the progressive interaction of the DNNs involved in the DSR process helps in improving the system performance. No additional benefits have been observed for the considered tasks by adding more than 3 levels. The number of communication levels, however, can be regarded as an hyperparameter of the architecture and its best value might be task-dependent. In general, one should anyway expect  a diminishing return when increasing $N$.

\section{Summary and Future Challenges} \label{sec:disc_ch6}
This Chapter discussed the network of deep neural networks paradigm and its application to DSR. The experimental part started with a simplified scenario, where a flat pipeline between speech enhancement and speech recognition was jointly trained with batch normalization. From these experiments emerged the great importance of this learning modality, that turned out  to improve matching between the various components. Our efforts were then focused on a network of DNNs between speech recognition and speech enhancement. The experiments have shown that cooperation and full communication across these modules is of key importance for counteracting the uncertainty originated by noise and reverberation. 

The proposed paradigm, however, is not yet fully mature and the real potential of this framework, according to us, is still far from being reached. 
Several efforts will be devoted to derive proper architectures able to foster the cooperation across the various DNNs. For instance, some preliminary results, reported in \cite{ravanelli_icassp}, show that the adoption of Residual Neural Networks (ResNet) is a very promising architectural variation, which can significantly improve gradient back-propagation.  

Moreover, inspired by the gating mechanisms used in the context of recurrent networks, we are studying the adoption of learnable communication gates to manage the information flow through the unfolded computation graph. The network of DNNs proposed in this thesis was limited to speech enhancement and speech recognition. In the future we will add other components, possibly including systems for acoustic scene analysis, such as environmental classification, acoustic event detection, unsupervised estimation of the reverberation time, speaker identification and speaker verification. 

We will also explore the extension of this paradigm to RNNs. Although this is rather natural and straightforward, a possible issue is the considerable memory required for unfolding the computational graph over both time and communication levels, making an efficient multi-gpu implementation strictly necessary.   

Finally, we would like to highlight that the proposed framework is a general paradigm that can be used in other fields. Cooperative DNNs can be considered, for instance, in application such as robotics, where multiple components have to properly work together to achieve a common global goal.


\chapter{Conclusion} \label{cha:conclusion}
The results reported in this PhD thesis summarize the main efforts undertaken over the last four years. During that time, the research community has experienced a revolution, thanks to the popularization and maturation of deep learning techniques. When this doctorate began in November 2013, the actual impact of this technology was still not clear. Today, after only few years of intense research, deep learning is a consolidated state-of-the-art technology for ASR and we can consider this paradigm as a major breakthrough, that can be compared, in term of impact to the research community, only to the widespread diffusion of HMMs, that were proposed more than thirty years ago. 

Since the beginning of this PhD, we were totally convinced of the great potential of this framework, especially to face challenging acoustic conditions characterized by significant levels of noise and reverberation. In the latter scenarios, previous HMM-GMM technology was, indeed, very far from achieving a satisfactory robustness, and we were confident that the significant performance gain originally observed  with deep learning in the context of standard close-talking ASR could also be extended to DSR. 
This thesis, and the related papers, represent one of the first attempts to revise standard deep learning algorithms, architectures ,and techniques for better addressing the specific problem of distant speech recognition. The dominant approach consisted, in most of the cases, of developing proper solutions for close-talking ASR and later inherit them for DSR. Conversely, we directly addressed the DSR application, since we believe that the peculiarities and challenges arising with distant speech deserve specific studies and methodologies.

In particular, we focused our efforts on the main flaws and weaknesses of DSR technology.  First, we concentrated on realistic data contamination. Data, in fact, are a crucial factor for the success of deep learning and a proper methodology for realistic data simulation in reverberant environment has been proposed and extensively validated. The realistic simulations have been used for generating high-quality multi-microphone simulated datasets, that have been released at international level. Moreover, some methodologies were proposed to better exploit contaminated data for DNN training. 

Another contribution of the thesis was the development of techniques for managing large time contexts. Our research has initially considered feed-forward neural networks, studying the role of asymmetric context windows to counteract the harmful effects of reverberation. We then focused on RNNs, proposing a novel architecture called Light GRU, that is able to better process time contexts, improve system performance, and significantly speed up DNN training. 

Inspired by the idea that uncertainty can also be counteracted with cooperation, we finally developed a novel deep learning paradigm called network of deep neural networks. The proposed framework is based on a full communication and interaction across the various DNN modules composing a DSR system and turned out to be very effective to improve the system performance.

The different results, presented in this dissertation, support the positive accomplishment of our main goal: contributing to reduce the gap between the possibility offered by current technology and user's expectations. In particular, the choice of specifically studying deep learning  for distant speech recognition was not only a forward-looking view, but allowed us to establish a niche in which our research was internationally appreciated.

Despite our contribution and the remarkable efforts of the research community working in the field, the road towards human or super-human distant speech recognition performance is still long. The future development of ASR and DSR technology will be closely influenced by the advances in deep learning and, more in general, by the evolution of AI. As for many other fields, we believe that a crucial role will be played by unsupervised learning. The vast majority of speech data, in fact, are available without a corresponding transcription and the development of techniques for better exploiting unlabelled data will be of crucial importance. With this regard, the adoption of generative adversarial training in the context of speech recognition represents a promising research direction.

Moreover, current deep learning models are discovering representations that are still working on low-levels of abstraction, thus focusing on rather superficial aspects of the world. It will be necessary to develop new algorithms and architectures that are able to better disentangle the underlying factors of variability. For DSR, proper solutions for deeply understanding and modeling the actual complexity of acoustic environments will have a major impact.

We also believe that long-life learning will play an important role in the next generation of speech recognizers. The study of suitable never-ending learning paradigms, that might operate in a distributed crowd-sourcing scenario, represents a promising direction towards human or super-human ASR. Reinforcement learning is also a very interesting and under-explored research direction for speech recognition. The study of automatic strategies for exploiting user feed-backs will be crucial for improving the system performance. 

We finally think that a major accomplishment, which can potentially have an important impact on several applications,  would be the development of a unified learning framework, where supervised, unsupervised, reinforced and long-life learning are jointly exploited within a network of deep neural network paradigm.

\clearemptydoublepage


\thispagestyle{empty}
\makeatletter
\addcontentsline{toc}{chapter}{Bibliography}
\bibliographystyle{plain}
\bibliography{mybibfile}

\clearemptydoublepage

\begin{appendices}

\chapter{Multi-Microphone Setups}
Many experiments conducted in this thesis are performed in a domestic scenario. The reference environment is a real apartment  (referred to as DIRHA apartment) that was available for experiments under the \textit{DIRHA project}\footnote{\url{https://dirha.fbk.eu/}}. This apartment have been equipped with various microphone configurations, that will be described in the following sections:

\begin{figure}[t!]
\centering
\includegraphics[width=1.0\textwidth]{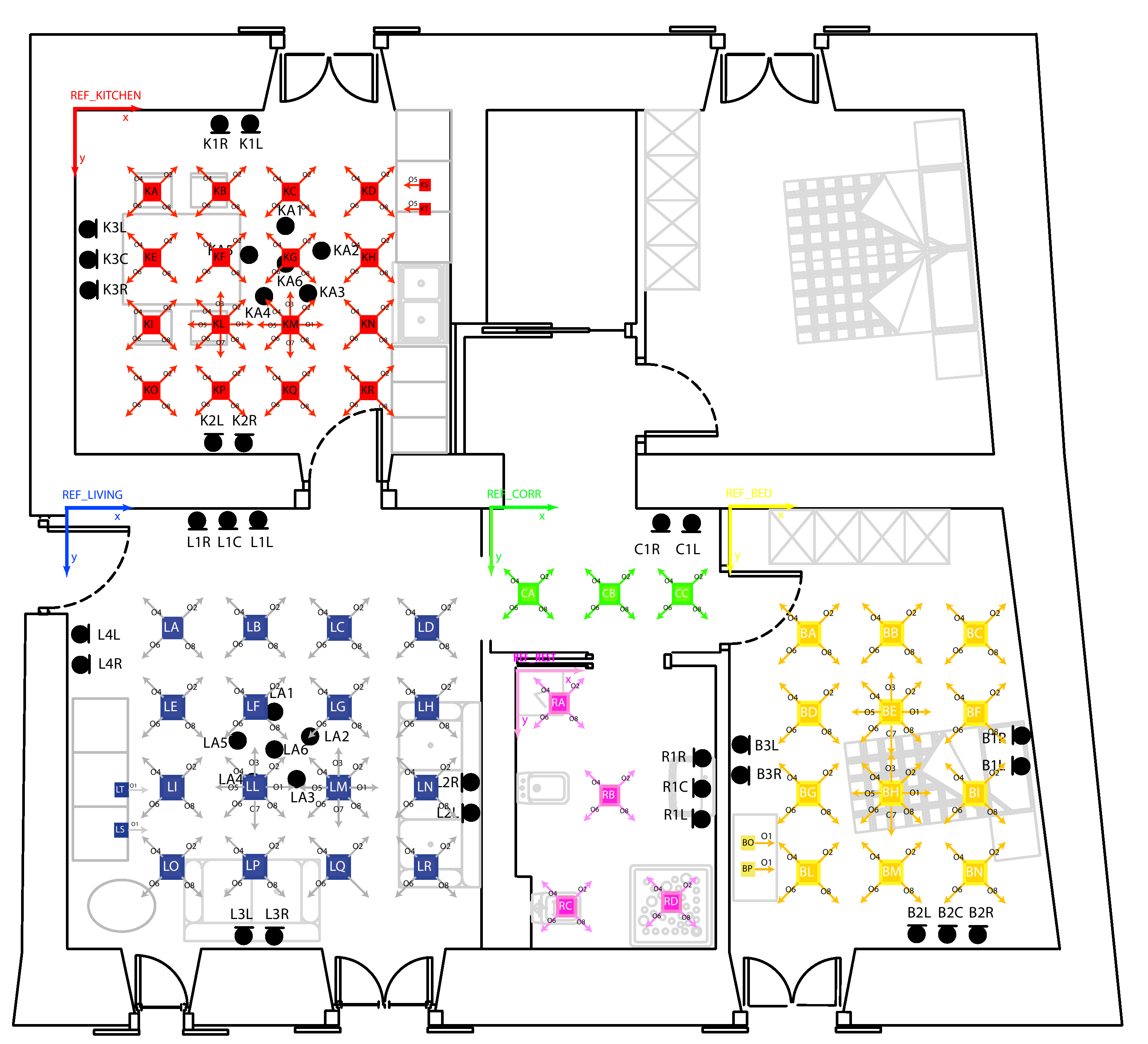}
\caption{Floor-plan of the DIRHA apartment equipped with a multi-room microphone network.}
\label{fig:dirhaflat}
\end{figure}

\section{Microphone Configuration 1 (MC-1)} \label{app:mc-1}
The first setup was based on 40 microphones distributed in five rooms of the DIRHA apartment (i.e., living-room, kitchen, corridor,bathroom, and bedroom).
The sensors  were high-quality omnidirectional microphones (Shure MX391/O), connected to multichannel clocked pre-amp and A/D boards (RME Octamic II), which allowed a perfectly synchronous sampling at 48 kHz, with 24 bit resolution.
Bathroom, corridor and bedroom were equipped with a limited number of microphone pairs and triplets (i.e., overall 12 microphones), while the living-room and the kitchen comprise the largest concentration of sensors and devices. As shown in Figure \ref{fig:dirhaflat}, the living-room includes three microphone pairs, a microphone triplet, and 6-microphone ceiling arrays. This setup was adopted for real recordings as well as for the extensive multi-room multi-microphone IR measurement champaign (see DIRHA-IRs Collection) used for the development of the DIRHA-Sim Corpora \cite{lrec} described in the following.

\begin{figure}[t!]
\centering
\includegraphics[width=0.8\textwidth]{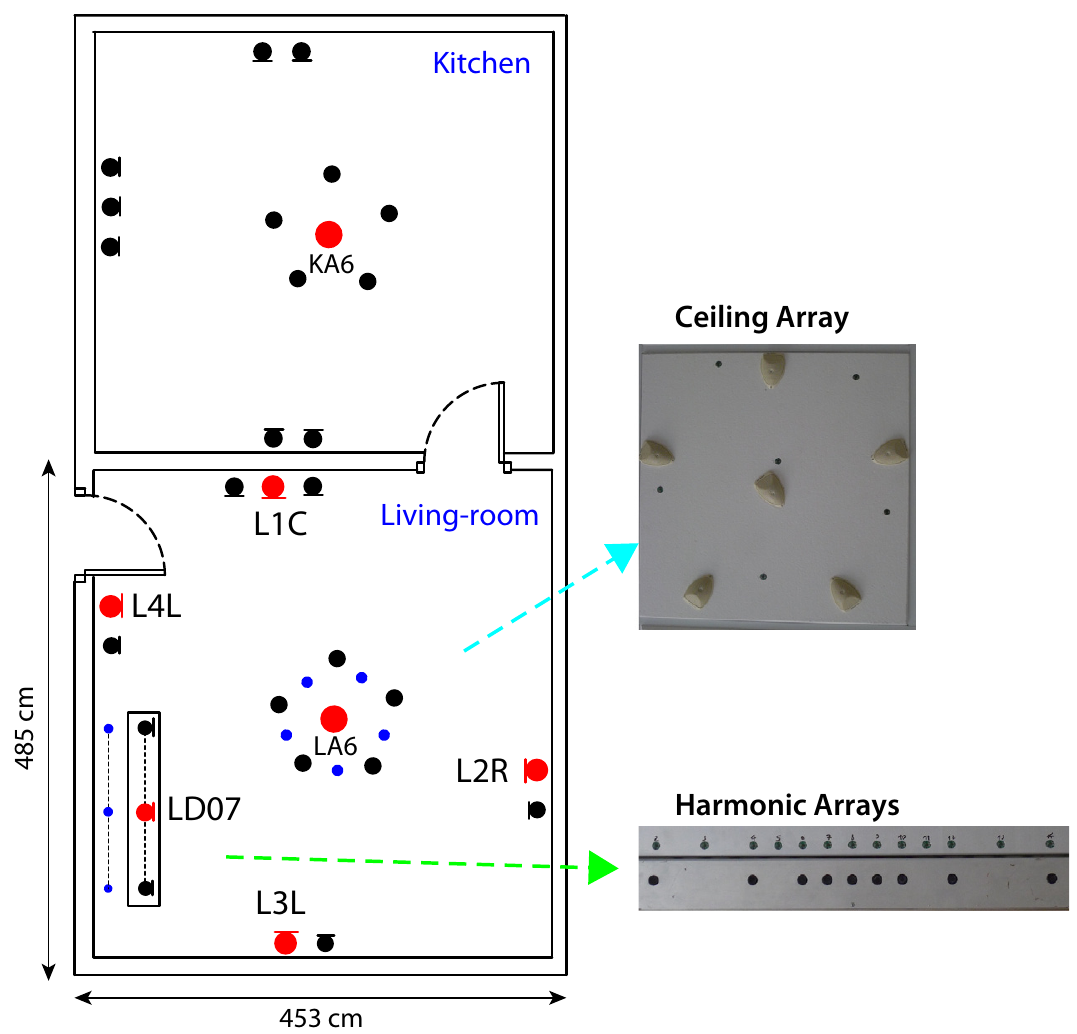}
\caption{An outline of the microphone set-up adopted for the DIRHA-ENGLISH corpus. Blue small dots represent digital MEMS microphones, red ones refers to the channels used for most of the experiments, while black ones represent the other available microphones. The right pictures show the ceiling array and the two linear harmonic arrays installed in the living-room.}
\label{fig:dirhaflat2}
\end{figure}

\section{Microphone Configuration 2 (MC-2)} \label{app:mc-2}
A second multi-microphone setup have been considered to explore a different microphone configuration (see Figure  \ref{fig:dirhaflat2}), which comprises 60 microphones distributed in the living-room (47) and the kitchen (13) of the DIRHA apartment.  In particular, the living-room was equipped with:

\begin{itemize}
\item a linear harmonic array composed of 13 Electrect microphones
\item a linear harmonic array composed of 13 MEMS microphones
\item a circular array composed of 6 Shure MX391/0 microphones
\item a circular array composed of 6 MEMS microphones
\item 9 Shure MX391/0 (3 pairs and 1 triplet) distributed in the four walls of the livingroom 
\end{itemize}

Shure and Electrect microphones were connected to multichannel clocked pre-amp and A/D boards (RME Octamic II), which allowed a perfectly synchronous sampling at 48 kHz, with 24 bit resolution. The MEMS microphones were developed by ST-Microelectronics and the  recordings have been conducted with their original acquisition board.

This setup was adopted for the real recordings  of the DIRHA-English Dataset. Moreover an IR measurement champaign (see DIRHA-IRs Collection) was conducted, and the measured IRs were used for generating the simulated part of the DIRHA-English Dataset. With this microphone configuration several hours of various noise sequences typical of a domestic environment were also recorded. The noise sequences were used to both add realistic noise in the simulations of the DIRHA-English Dataset and to contaminate training datasets (such as the WSJ corpus) with also additive noise.

This setup, that is limited to the kitchen and the livingroom, is depicted in Figure  \ref{fig:dirhaflat2}. The living-room includes three microphone pairs, a microphone triplet, and 6-microphone ceiling arrays

It also considers harmonic arrays and MEMS microphones which were unavailable in the previous setup. This setup was adopted for the development of the DIRHA-English Corpus
\cite{dirha_asru}.

\chapter{Corpora} \label{app:corpora}
The experimental evaluations of the  techniques proposed in this thesis have performed on different datasets, that are listed and briefly described in the following.

\begin{itemize}
\item \textbf{APASCI}:
It is an Italian high-quality speech database recorded in a recording studio under close-talking conditions \cite{apasci}. It includes 5,290 phonetically rich sentences and about 350 minutes of speech.
The speech material was read by 100 Italian speakers (50 male and 50 female). Each of them uttered 1 calibration sentence, 4 sentences with a wide phonetic coverage, 15 or 20 sentences with a wide diphonic coverage. Reverberated version of APASCI (denoted as \textit{APASCI-rev}) have been generated with the approach described in Chapter \ref{sec:cont} to train acoustic models of a speech recognizer. The versions denoted as \textit{APASCI-rev\&noise} also includes the contribution of additive noise that was recorded in the DIRHA apartment. 


\item \textbf{CHiME 4}: This dataset \cite{chime4_paper} is based on both real and simulated data recorded in four noisy environments (on a bus, cafe, pedestrian area, and street junction). The training set is composed of 43690 noisy WSJ sentences (75 hours) recorded by five microphones (arranged on a tablet) and uttered by a total of 87 speakers. The development set (DT) is based on 3280 WSJ sentences uttered by four speakers (1640 are real utterances referred to as DT-real, and 1640 are simulated denoted as DT-sim). 
The test set (ET) is based on  1320 real utterances (ET-real) and 1320 simulated sentences (DT-real) from other four speakers. The experiments reported in this thesis are based on the single channel setting, in which the test phase is carried out with a single microphone (randomly selected from the considered microphone setup). 
More information on CHiME data can be found in \cite{chime3}.

\item \textbf{DIRHA-IRs Collection}:
During the DIRHA project several efforts have been devoted to measure realistic impulse responses. 
A first IR measurement campaign has been carried out with the network of 40 microphones described in App. \ref{app:mc-1}. The IRs measurement process explored 57 different positions with at least four orientations for each position. In total, more than 9000 IRs have been measured. Cross-room impulse responses are also included in order to simulate cross-room audio propagation effects.
The IRs measurements were based on a professional studio monitor (Genelec 8030A) with the methodology described in Chapter \ref{sec:cont}.
The Time of Flight (TOF) information, crucial to applications such as acoustic event localization, beam-forming, and multi-microphone signal processing in general, has been preserved by means of six sample-synchronized multi-channel audio cards.
The IRs were measured at 48kHz sampling frequency with 16-bit accuracy.
A second IR measurement campaign was conducted in the same apartment with the same methodology, but with the microphone configuration described in App. \ref{app:mc-2}.

\item \textbf{DIRHA-English}: This corpus is multi-microphone dataset \cite{dirha_asru}, recently realized under the EC DIRHA project. The reference scenario is the DIRHA apartment equipped with the microphone configuration described in App. \ref{app:mc-2}. The overall  dataset is composed of 1-minute sequences, comprising phonetically-rich, conversational speech, keywords, commands, and wsj sentences. The corpus includes both real and simulated data. Simulations have been generated with the approach discussed in Chapter \ref{sec:cont} by combining high-quality close-talking recordings with impulse responses measured in the targeted environment. Background noise recorded in the apartment was also added to the simulations. Together with simulated data, real recording in different positions of the living-room are performed.  The corpus  includes 12 US and 12 UK English native speakers. Different subsets have been derived from the overall corpus and used in this thesis. For instance, the part denoted in the manuscript as \textit{DIRHA-English-WSJ5k} contains a collection of the wsj-5k sentences uttered by the aforementioned US speakers. Six speakers are part of the set1 subpart (409 sentences), while the other speakers belong to the set2 (330 sentence). 
The sentences are available in various scenarios of increasing complexity, i.e. close-talking \textit{DIRHA-English-WSJ5k (close-talk)}, reverberated \textit{DIRHA-English-WSJ5k (rev)}, reverb+noise conditions \textit{DIRHA-English-WSJ5k (rev\&noise)}. The WSJ part of the dataset in being distributed by the Linguistic Data Consortium (LDC), while the phonetically-rich sentences can be downloaded from the DIRHA website.
The DNN baselines obtained with Kaldi (using the standard Karel's recipe) in the aforementioned conditions are reported in Table \ref{tab:kaldi_baselines} for reference purposes (fMLLR features).
The ASR performance is reported in terms of Word Error Rate (\%) and the reference microphone is $LA6$\footnote{See our github page \url{https://github.com/SHINE-FBK/DIRHA_English_wsj} for a complete overview on the ASR baselines.}. 

\begin{table}[]
\centering
\tabcolsep=0.30cm
    \begin{tabular}{ | l | c | c | c | c | }
    \cline{1-5}
  \multirow{ 2}{*}{\backslashbox{\em{Condition}}{\em{Dataset.}}} & \multicolumn{2}{|c|}{set1} &  \multicolumn{2}{|c|}{set2}  \\ \cline{2-5}
& Sim & Real & Sim & Real \\ \hline
 
Close-Talking & - & 3.7 & - & 2.2 \\ \hline
Distant-Talking (Rev) & 16.2 & 13.6 & 9.8 & 9.4 \\ \hline
Distant-Talking (Rev\&Noise) & 22.8 & 30.0 & 25.0 & 16.6 \\ \hline
 
\end{tabular}
\caption{DNN baselines for the DIRHA-English-WSJ obtained with kaldi (karel's recipe).}
\label{tab:kaldi_baselines}
\end{table}

\item \textbf{DIRHA-SimCorpora}:
The database was collected with the microphone setup described in App. \ref{app:mc-1}.
The DIRHA SimCorpus is a multi-microphone, multi-language (Italian, German, Portuguese, Greek) and multi-room database containing simulated acoustic sequences derived from the DIRHA apartment. For each language, the corpus contains a set of acoustic sequences of duration 60
seconds, at 48kHz sampling frequency and 16-bit accuracy.
Each sequence consists of real background noise with superimposed various localized acoustic events. Acoustic events occur randomly (and rather uniformly) in time and in space (within predefined positions) with various dynamics. The acoustic wave propagation from the sound source to each single microphone is simulated by convoluting the clean signals with the respective impulse response (IR). Acoustic events are divided into two main categories, i.e., speech and non-speech. Speech events include different types of sentences (i.e., phonetically-rich sentences, read, commands, spontaneous speech and commands, keywords) uttered by different speakers in the four languages. Non-speech events have been selected from a collection of high-quality sounds typically occurring within a home environment (e.g., radio, TV, appliances, knocking, ringing, creaking, etc.).
For cross-language comparison purposes, each simulated acoustic sequence has been replicated in the four languages while preserving the same background noises and non-speech sources. Gender and timing of the active speakers have been preserved across the different languages, in order to further ensure homogeneity.
Data were generated by means of a multi-microphone simulation framework developed at FBK.
A subset of this corpus has been delivered in the contest of HSCMA 2014 \cite{hscma} and EVALITA 2014 \cite{evalita}.

\item \textbf{DIRHA-GRID}:
This corpus \cite{dirha_grid} is a multi-microphone and multi-room simulated database containing a set of acoustic scenes of 1-minute duration (at 16kHz sampling frequency and 16-bit accuracy) which are observed by the microphone network described in App. \ref{app:mc-1}. Each acoustic scene is composed of both speech, i.e., short English commands from the GRID database \cite{gridnews}, and non-speech acoustic
sources, i.e., typical home noises. Each acoustic event (both speech and non-speech) occurs randomly in time and in space and
can take place in any of the microphone-equipped rooms. In particular, a variable number of short commands (ranging from 4
to 7) arise in each 1-minute long acoustic scene. An overlap in time between speech and non-speech sources is possible, while an overlap between speech sources cannot occur. The corpus is divided into 3 chunks (dev 1, test1, test2) containing 75 acoustic sequences each with 12 different speakers (6 male and 6 female) involved for each dataset.
Data were generated by means of a multi-microphone simulation framework developed at FBK and can be downloaded from the DIRHA website.

\item \textbf{DIRHA-AEC}:
The database was collected in the  DIRHA apartment using the microphone setup described in App. \ref{app:mc-1}. In contrast to the other real databases recorded under the DIRHA project, the DIRHA AEC corpus is specifically designed for Acoustic Echo Cancellation (AEC) and is currently available in Italian. In particular, it is based on real recordings involving 13 speakers (6 males and 7 females) that read a list of 50 commands in five different
positions/orientations in the living- room. A set of simulated sequences were also generated, in which  all the commands were repeated with no overlapping sources (Real-S0), an overlap with a prompt (Real-S1), an overlap with the television (Real-S2) and with both overlaps (Real-
S3). In total, 2600 speech commands were recorded.

\item \textbf{Euronews}:
Data come from the portal Euronews and were acquired both from the Web and from TV \cite{gretter}. The overall corpus includes data in 10 languages (Arabic, English, French, German, Italian, Polish, Portuguese, Russian, Spanish and Turkish) and was designed both to train AMs and to evaluate ASR
performance. For each language, the corpus is composed of about 100 hours of speech for training and about 4 hours,
manually transcribed, for testing. Training data include the audio, some reference text, the ASR output and their alignment. The portion used in this thesis is the Italian part of the dataset. Reverberated version of Euronews (denoted as \textit{Euronews-rev}) have been generated with the approach described in Chapter \ref{sec:cont} to train acoustic models of a speech recognizer.

\item \textbf{TED-talks corpus}: This dataset was released in the context of the IWSLT evaluation campaigns \cite{iwslt_2011}. The training set is composed of 820 talks with a total of about 166 hours of speech. The development test is composed of 81 talks (16 hours), while the test sets (TST 2011 and TST 2012) are based on 8 talks (1.5 hours) and 32 talks (6.5 hours), respectively.

\item \textbf{TIMIT}:
This corpus contains  16kHz speech recording of 630 American English speakers, each reading 10 phonetically-rich sentences.  TIMIT is based on about 5 hours of speech recordings. Reverberated version of TIMIT (denoted as \textit{TIMIT-rev}) have been generated with the approach described in Chapter \ref{sec:cont} to train acoustic models of a speech recognizer. The versions denoted as \textit{TIMIT-rev\&noise} also includes the contribution of additive noise that was recorded in the DIRHA apartment.

\item \textbf{Wall Street Journal (WSJ)}:
The Wall Street Journal dataset (WSJ) consists of news text read by US native speakers. It is composed of two parts, often known as WSJ0 and WSJ1. The part used for this thesis is the WSJ0-5k, that is composed of 37416 sentences uttered by 283 speakers for a total of about 15 hours of speech material. Reverberated version of WSJ (denoted as \textit{WSJ-rev}) have been generated with the approach described in Chapter \ref{sec:cont} to train acoustic models of a speech recognizer. The versions denoted as \textit{WSJ-rev\&noise} also include the contribution of additive noise that was recorded in the DIRHA apartment.

\end{itemize}

\chapter{Experimental Setups} \label{app:setup}
We report now a detailed description of all the experimental setups adopted in the various experiments of this thesis.

\section{Data Contamination - Setup 1} \label{app:es1}
These experiments were obtained using contaminated data (for training purposes) and real data collected  in the DIRHA apartment equipped with the microphone network described in App \ref{app:mc-1}. The reference language of these experiments was Italian. In particular, different recording sessions were performed in the given environment, to collect both speech material and audio signals useful to estimate impulse responses.

Speakers and microphones were both located in the living-room. The sound source was located in position $LB$ with with frontal orientation towards the reference microphone $L3R$, that located at a distance of about 4 m (see Figure \ref{fig:dirhaflat}). 

For impulse response measurement purposes, different categories of loudspeakers were considered. In particular, the following results are based on the use of a professional studio monitor, i.e., \emph{Genelec 8030}. In order to estimate impulse responses, MLS, LSS, and ESS excitation sequences were diffused by each loudspeaker in the environment, with varying length and amplifying settings. Each excitation signal was preceded and followed by fade-in and fade-out sequences of 50 ms duration, in order to avoid the introduction of any possible numerical clicks.
Speech data collection was then conducted with all the real speakers located in the same position where the loudspeaker was previously placed. For comparison purposes, the speech uttered by each speaker was also recorded with a professional close-talking \emph{Countryman E6} microphone.

The speech material used to train the distant-speech recognizer consists in contaminated versions (one for each IR measurements settings) of APASCI.

In order to evaluate speech recognition performance, the utterances pronounced by 11 Italian speakers (which were not in the APASCI corpus) in the above-mentioned apartment. The speaker read a list of 125  command-and-control sentences.

The speech recognition system investigated in this work is based on a standard front-end processing consisting of a pre-emphasis step followed by feature extraction.
The pre-emphasized signal is blocked and Hamming windowed into frames of 20 ms duration (with 50\% overlapping). For each frame, 12 Mel-frequency Cepstral Coefficients (MFCCs) and the log-energy are extracted. MFCCs are normalized by subtracting the means, while the log-energy is normalized with respect to the maximum value on the whole utterance.
The resulting normalized MFCCs and log-energy, together with their first and second order derivatives, are then arranged into a single observation vector of 39 components.
Acoustic modeling is based on a GMM-HMM that operates at context-independent phone-like unit level, and is derived by applying the \emph{Baum-Welch} algorithm, while the recognition step is accomplished by using \emph{Viterbi} algorithm.

The Word Loop (WL) task is based on small vocabulary  of 233 words used to create any of the above-mentioned command-and-control sentences. Although the introduction of a grammar or language modeling would increase the system performance, in this work a word loop task was prefered to better emphasize any experimental evidence at acoustic level, which would otherwise be partially missed. It is worth noting that the vocabulary includes a quite large number of short and confusable words.

\section{Data Contamination - Setup 2} \label{app:es2}
The reference scenario for these experiments is the microphone configuration described in App. \ref{app:mc-2}. The considered task is the Wall Street Journal (WSJ-5k), in agreement with the task addressed in the CHiME 3 challenge. While CHiME 3 was pretty focused on robustness against noise, in this work the main source of disturbance is reverberation.

The training phase is based on the WSJ0 database (LDC catalog number LDC93S6A), which was contaminated with three  impulse responses corresponding to different positions of the speaker in the targeted living-room. Note that, according to our past experience, this small  number of high-quality IRs is sufficient to properly derive reverberant-robust acoustic models, as shown in our previous work \cite{Ravanelli-14}. The considered IRs were measured, or simulated with the image method, depending on the specific experiment. 

For test purposes we employed both real and simulated data of the DIRHA English dataset (set1 portion) \cite{dirha_asru}. We actually considered only 5 of the 6 available US speakers since one of the speakers ($spk01$) was not available for the recordings in the DIRHA apartment. For the experiments reported in this work, we thus removed it from the considered dataset. 

Beside real recordings, simulated data have also been created combining high-quality close-talking recordings with the measured impulse responses. With this purpose, in order to allow a fair comparison between real and simulated data, we asked the same speakers to utter again the sentences in the FBK recording studio, which led to a very high-quality clean speech material. Moreover, for each position/orientation of the speaker in the real recordings, a corresponding IR was measured, allowing us to derive a simulated corpus well-matching in terms of speaker positions, orientations, and other signal characteristics, the real data collected in the apartment. 

The experimental part of this work is based on the Kaldi toolkit \cite{kaldi}. The recipe considered for training and test the DSR system is similar to the s5 recipe proposed in the Kaldi release for WSJ data. In short, the speech recognizer is based on standard MFCCs and acoustic models of increasing complexity. In the following section, ``\textit{mono}'' is the simplest system based on 48 context-independent phones of the English language, each modeled by a three state left-to-right HMM (overall using 1000 gaussians). A set of context-dependent models are then derived. In ``\textit{tri1}'' 2.5k tied states with 15k gaussians are trained by exploiting a binary regression tree.``\textit{Tri2}''is an evolution of the standard context-dependent model in which a Linear Discriminant Analysis (LDA) is applied.
In both `\textit{`tri3}'' and ``\textit{tri4}'' models Speaker Adaptive Training (SAT) is also performed. The difference is that  ``\textit{tri4}'' is bootstrapped by the previously computed `\textit{`tri3}'' model.
The considered ``\textit{DNN}'', based on the Karel's recipe \cite{kaldi}, is composed of six hidden layers of 2048 neurons,  with a context window of 11 consecutive frames (5 before and 5 after the current frame) and an initial learning rate of 0.008. The weights are initialized via RBM pre-training, while the fine tuning is performed with stochastic gradient descent optimizing cross-entropy loss function. The sampling frequency for ASR experiments is 16 kHz.

\section{Data Contamination - Setup 3} \label{app:es3}
In this work, we use a Context-Dependent DNN-HMM speech recognizer, where every unit is modeled by a three state left-to-right HMM, and the tied-state observation probabilities are estimated through a DNN. 

Feature extraction is based on blocking the signal into frames of 25 ms with 10 ms overlapping. For each frame, 13 MFCCs plus pitch and Probability of Voicing (PoV) are extracted. The pitch and PoV are estimated through the normalized autocorrelation method discussed in \cite{kpitch}. The resulting features, together with their first and second order derivatives, are then arranged into a single observation vector of 45 components. 
Finally, a context window gathering several consecutive frames followed by a mean and variance normalization of the feature space are applied before feeding the DNN. 
The DNN, trained with the Kaldi toolkit (Karel's recipe) \cite{kaldi}, is composed of sigmoid-based hidden neurons, while the output layer is based on softmax activation functions.
The pre-training phase is carried out by stacking Restricted Boltzmann Machines (RBM) \cite{rbm1} to form a deep belief network, while 
the fine-tuning is performed by a stochastic gradient descent optimizing cross-entropy loss function.
In the latter phase, the initial learning rate is kept fixed as long as the increment of the frame accuracy on the dev-set is higher than 0.5$\%$. For the following epochs, the learning rate is halved until the increment of frame accuracy is less than the
stopping threshold of 0.1$\%$.
The decoding is performed 
by adopting a phone-loop based grammar.
As previously outlined, even though the use of more complex grammars or language models is certainly helpful in increasing the recognition performance, the adoption of a simple phone-loop is due to the need of an experimental evidence not biased by a LM.
In this study, as in \cite{rav_in14}, a set of 26 phone units of the Italian language was chosen for evaluation purposes

\section{Managing Time Contexts - Setup 1} \label{app:mtc_s1}
In this work the experiments were conducted in three different acoustic conditions of increasing complexity: close-talking (\textit{Clean}), distant-talking with reverberation (\textit{Rev}), and distant-talking with both noise and reverberation (\textit{Rev\&Noise}).

In the context of the close-talking experiments we considered the standard WSJ dataset for training, and the close-talking part of the DIRHA-English-WSJ Dataset (DIRHA-WSJ-close-talk) for test purposes (see App. \ref{app:corpora}).

The reference environment for most of the experiments conducted in this study was the living-room of the DIRHA apartment, that was equipped with the microphone setup described in App. \ref{app:mc-2}.

A set of experiments was carried out to study distant-talking conditions where only reverberation acts as a source of disturbance (Rev).  In these cases, training was performed using a contaminated dataset (WSJ-rev), which was generated by convolving the original WSJ data with a set of three impulse responses chosen from the aforementioned collection. 
Test data  were based on a contaminated version of the close-talking test data (DIRHA-Englsh-WSJ5k-clean, set1). In  order  to  simulate several speaker positions and orientations, a set of 36 impulse responses (different from those used for training) was used for the latter dataset. 

To explore more challenging conditions characterized by both noise and reverberation (Rev\&Noise), real recordings have also been performed.
The real recordings (DIRHA-English-WSJ, real set1 portion), are part of the recently-released DIRHA English WSJ corpus \cite{dirha_asru}.

To test our approach in different environments, other contaminated versions of the training and test data are generated with different impulse responses (either measured or computed with the image method \cite{image}), as will be discussed in Sec. \ref{sec:mis} and Sec. \ref{sec:t60}.
In this work, we use a context-dependent DNN-HMM speech recognizer, where every unit is modeled by a three state left-to-right HMM, and the tied-state observation probabilities are estimated through a DNN. 

Feature extraction was based on blocking the signal into frames of 25 ms with an overlap of 10 ms.  The experimental activity was conducted considering different acoustic features, i.e., 39 MFCCs (13 static+$\Delta$+$\Delta\Delta$), 40 log-mel filter-bank features (FBANKS), as well as 40 fMLLR features (extracted as reported in the s5 recipe of Kaldi \cite{kaldi}). Features of consecutive frames were gathered into both symmetric and asymmetric observation windows. Mean and variance normalization of the feature space is applied before feeding the DNN. 

The DNNs, trained with the Kaldi toolkit \cite{kaldi} (Karel's recipe), was composed of six sigmoid-based hidden layers of 2048 neurons, while the output was based on a softmax classifier. Weights were initialized with close-to-zero values with small variance.  
Training was performed with SGD optimizing the cross-entropy loss function, and its evolution was monitored using a small validation set (10\% of the training data) that was randomly extracted from the training corpus. The performance on the validation set was monitored after each epoch to perform learning rate annealing  as well as for checking the stopping condition.   
In particular, the initial learning rate was kept fixed at 0.008 as long as the increment of the frame accuracy on the validation was higher than 0.5$\%$. For the following epochs, the learning rate is halved until the increment of frame accuracy was less than the stopping threshold of 0.1$\%$. The labels for DNN training were derived from an alignment on the tied states, which was performed with a previously-trained HMM-GMM acoustic model \cite{kaldi} with the same input features and adopting the same corpus. 
For Convolutional Neural Network (CNNs) experiments, we replaced the first two fully-connected layers of the above-mentioned DNN with two convolutional layers based on 128 and 256 filters, respectively.
The decoding phase was performed with standard WSJ language model.

\section{Managing Time Contexts - Setup 2} \label{app:mtc_s2}

The architecture adopted for these experiments consisted of multiple recurrent layers, that were stacked together prior to the final softmax classifier. These recurrent layers were bidirectional RNNs \cite{graves}, which were obtained by concatenating the forward hidden states (collected by processing the sequence from the beginning to the end) with backward hidden states (gathered by scanning the speech in the reverse time order).
Recurrent dropout was used as regularization technique. Since extending standard dropout to recurrent connections hinders learning long-term dependencies, we followed the approach introduced in \cite{drop_asru,Gal2016}, that tackles this issue by sharing the same dropout mask across all the time steps. Moreover,  batch normalization was adopted exploiting the method suggested in \cite{cesar}.
The feed-forward connections of the architecture were initialized according to the \textit{Glorot}'s scheme \cite{xavier}, while recurrent weights were initialized with orthogonal matrices \cite{orth_init}. Similarly to \cite{ravanelli_SLT}, the gain factor $\gamma$ of batch normalization was initialized to 0.1 and the shift parameter $\beta$ was initialized to 0.01

Before training, the sentences were sorted in ascending order according to their lengths and, starting from the shortest utterance, minibatches of 8 sentences were progressively processed by the training algorithm.
This sorting approach minimizes the need of zero-paddings when forming mini-batches, resulting helpful to avoid possible biases on batch normalization statistics. 
Moreover, the sorting approach exploits a curriculum learning strategy \cite{curriculum} that has been shown to slightly improve the performance and to ensure numerical stability of gradients. The optimization was done using the Adaptive Moment Estimation (Adam) algorithm \cite{adam} running for 22 epochs (with $\beta_1=0.9$, $\beta_2=0.999$, $\epsilon=10^{-8}$). The performance on the development set was monitored after each epoch, while the learning rate was halved when the performance improvement went below a certain threshold ($th=0.001$). Gradient truncation was not applied, allowing the system to learn arbitrarily long time dependencies.

The main hyperparameters of the model (i.e., learning rate, number of hidden layers, hidden neurons per layer, dropout factor) were optimized on the development data. 
In particular, we guessed some initial values according to our experience, and starting from them we performed a grid search to progressively explore better configurations. A total of 20-25 experiments were conducted for all the various RNN models.
As a result, an initial learning rate of 0.0013 and a dropout factor of 0.2 were chosen for all the experiments. The optimal numbers of hidden layers and hidden neurons, instead, depend on the considered dataset/model, and range from 4 to 5 hidden layers with 375-607 neurons.

For DNN-HMM experiments, the DNN is trained to predict context-dependent phone targets. The feature extraction is based on blocking the signal into frames of 25 ms with an overlap of 10 ms.  The experimental activity is conducted considering different acoustic features, i.e., 39 MFCCs (13 static+$\Delta$+$\Delta\Delta$), 40 log-mel filter-bank features (FBANKS), as well as 40 fMLLR features (extracted as reported in the s5 recipe of Kaldi \cite{kaldi}). The labels were derived by performing a forced alignment procedure on the original training datasets. See the standard s5 recipe of Kaldi for more details \cite{kaldi}. During test, the posterior probabilities generated  for each frame by the RNN are normalized by their prior probabilities. The obtained likelihoods are processed by an HMM-based decoder, that, after integrating the acoustic, lexicon and language model information in a single search graph, finally estimates the sequence of words uttered  by the speaker. The RNN part of the ASR system was implemented with Theano \cite{theano}, that was  coupled with the Kaldi decoder \cite{kaldi} to form a context-dependent RNN-HMM speech recognizer.

The models used for the CTC experiments consisted of 5 layers of bidirectional RNNs of either 250 or 465 units. Unlike in the other experiments, weight noise was used for regularization. The application of weight noise is a simplification of adaptive weight noise \cite{graves2011practical} and has been successfully used before to regularize CTC-LSTM models \cite{graves2013speech}. The weight noise was applied to all the weight matrices and sampled from a zero mean normal distribution with a standard deviation of $0.01$. Batch normalization was used with the same initialization settings as in the other experiments. Glorot's scheme was used to initialize all the weights (also the recurrent ones). The input features for these experiments were 123 dimensional FBANK features (40 + energy + $\Delta$+$\Delta\Delta$). These features were also used in the original work on CTC-LSTM models for speech recognition \cite{graves2013speech}.
The CTC layer itself was trained on the 61 label set. Decoding was done using the best-path method \cite{CTC_graves}, without adding any external phone-based language model.

Unlike in the other experiments, the utterances were not sorted by length. 

\section{Networks of DNNs - Setup 1} \label{app:ndnn_s1}
The features considered in this work are standard 39 Mel-Cepstral Coefficients (MFCCs) computed every 10 ms with a frame length of 25 ms. The speech enhancement DNN is fed with a context of 21 consecutive frames and predicts (every 10 ms) 11 consecutive frames of enhanced MFCC features. The idea of predicting multiple enhanced frames was also explored in \cite{joint3}. 
All the layers used Rectified Linear Units (ReLU), except for the output of the speech enhancement (linear) and the output of speech recognition (softmax).
Batch normalization \cite{batchnorm} is employed for all the hidden layers, while dropout \cite{dropout} is adopted in all part of the architecture, except for the output layers. 

The datasets used for joint training are obtained through a contamination of clean corpora (i.e., TIMIT and WSJ) with noise and reverberation. 
The labels for the speech enhancement DNN (denoted as $x_{clean}$ in Alg.1) are the MFCC features of the original clean datasets.
The labels for the speech recognition DNN (denoted as $y_{lab}$ in Alg.1) are derived by performing a forced alignment procedure on the original training datasets. See the standard s5 recipe of Kaldi for more details \cite{kaldi}.

The weights of the network are initialized according to the \textit{Glorot} initialization \cite{xavier}, while biases are initialized to zero.
Training is based on a standard Stochastic Gradient Descend (SGD) optimization with mini-batches of size 128. The performance on the development set is monitored after each epoch and the learning rate is halved when the performance improvement is below a certain threshold. The training ends when no significant improvements have been observed for more than four consecutive epochs. 
The main hyperparameters of the system (i.e., learning rate, number of hidden layers, hidden neurons per layer, dropout factor and $\lambda$) have been optimized on the development set. 

The proposed system, which has been implemented with Theano \cite{theano}, 
has been coupled with the Kaldi toolkit \cite{kaldi} to form a context-dependent DNN-HMM speech recognizer.

In order to provide an accurate evaluation of the proposed technique, the experimental validation has been conducted using different training datasets, different tasks and various environmental conditions\footnote{To allow reproducibility of the results reported in this section,  the code of our joint-training system will be available at \url{https://github.com/mravanelli}. In the same repository, all the scripts needed for the data contamination will be available. The public distribution of the DIRHA-English dataset is under discussion with the Linguistic Data Consortium (LDC).}. 

The experiments with TIMIT are based on a phoneme recognition task (aligned with the Kaldi s5 recipe). The original training dataset has been contaminated with a set of realistic impulse responses measured in a real apartment. The reverberation time ($T_{60}$) of the considered room is about 0.7 seconds. Development and test data have been simulated with the same approach. More details about the data contamination approach can be found in \cite{IRs_paper,lrec,rav_is16}.

The WSJ experiments are based on the popular wsj5k task (aligned with the CHiME 3 \cite{chime3} task) and are conducted under two different acoustic conditions. For the \textit{WSJ-Rev} case, the training set is contaminated with the same set of impulse responses adopted for TIMIT. For the \textit{WSJ-Rev+Noise} case, we also added non-stationary noises recorded in a domestic context (the average SNR is about 10 dB). The test phase is carried out with the DIRHA English Dataset (set 2 portion), consisting of 409 WSJ sentences uttered by six native American speakers in the above mentioned apartment. For more details see \cite{dirha_asru,rav_is16}.

\section{Networks of DNNs - Setup 2} \label{app:ndnn_s2}
To provide an accurate evaluation of the proposed technique, the experimental validation has been conducted using different training datasets, different tasks and various environmental conditions. 
In particular, a set of experiments with TIMIT has been performed to test the proposed paradigm in low-resources conditions. To validate on a more realistic task, the proposed technique has also been evaluated on a WSJ task.

The experiments with TIMIT are based on a phoneme recognition task (aligned with the Kaldi s5 recipe). The original training dataset has been contaminated with a set of impulse responses measured in a real apartment. The reverberation time ($T_{60}$) of the considered room is about 0.7 seconds. Development and test data have been simulated with the same approach, but considering a different set of impulse responses. 

The WSJ experiments are based on the popular wsj5k task (aligned with the CHiME 3 \cite{chime3} task) and are conducted under two different acoustic conditions. For the \textit{WSJ rev} case, the training set is contaminated with the same set of impulse responses adopted for TIMIT. For the \textit{WSJ rev+noise} case, we also added non-stationary noises recorded in a domestic context (the average SNR is about 10 dB). The test phase is carried out with the DIRHA-English corpus (real-data part), consisting of 409 WSJ sentences uttered by six native American speakers in the above mentioned apartment. More details on this corpus and on the impulse responses adopted in this work can be found in \cite{dirha_asru,rav_is16}.

The features considered in this work are standard 39 Mel-Cepstral Coefficients (MFCCs) computed every 10 ms with a frame length of 25 ms. The speech enhancement DNNs are fed with a context of 21 consecutive frames and predict (every 10 ms) 11 consecutive frames of enhanced MFCC features.
The speech recognition DNNs are fed by such 11 speech enhanced frames and predict both context-dependent and monophone targets at their output. 
All the layers used Rectified Linear Units (ReLU), except for the output of the speech enhancement DNNs (linear) and the output of  the speech recognition modules (softmax).
Batch normalization \cite{batchnorm} and dropout \cite{dropout} are employed for all the hidden layers. 
The labels for the speech enhancement DNN (denoted as $x_{clean}$ in Figure  ~\ref{fig:arch}) are the MFCC features of the original clean datasets.
The labels for the speech recognition DNN (denoted as $y_{SR}$ in Figure  ~\ref{fig:arch}) are derived by performing a forced alignment procedure on the original training datasets. See the standard s5 recipe of Kaldi for more details \cite{kaldi}.

The weights of the networks are initialized according to the \textit{Glorot} initialization \cite{xavier}, while biases are initialized to zero.
Training is based on a standard Stochastic Gradient Descend (SGD) optimization with mini-batches $N$ of size 128. The performance on the development set is monitored after each epoch and the learning rate $\eta$ is halved when the performance improvement is below a certain threshold. The training ends when no significant improvements have been observed for more than four consecutive epochs. 

The main hyperparameters of the system (i.e., learning rate $\eta$, number of hidden layers, hidden neurons per layer, dropout factor, gradient weighting factor $\lambda$ and number of unfolding levels $L$) have been optimized on the development set (DIRHA-English-WSJ5k set1 sim). As a result, speech enhancement and speech recognition DNNs with 4 hidden layers of 1024 neurons and DNNs with 6 hidden layers of 2048 neurons are employed for TIMIT and WSJ tasks, respectively. The initial learning rate is 0.08, the dropout factor is 0.2 and the considered number of levels is 3 ($l=0,..,2$).  Similarly to \cite{ravanelli_SLT}, $\lambda$ is fixed to 0.1.

The proposed system, which has been implemented with Theano \cite{theano}, 
has been coupled with the Kaldi toolkit \cite{kaldi} to form a context-dependent DNN-HMM speech recognizer.

\end{appendices}









\end{document}